 \newcommand{\mydriver}{pdflatex} 

\documentclass[12pt,\mydriver]{thesis}  

\usepackage{titlesec}
   \titleformat{\chapter}
      {\normalfont\large}{Chapter \thechapter:}{1em}{}
\usepackage{tikz}
\usepackage{times}
\usepackage{subcaption}
\usepackage{graphicx}
\usepackage{cite}
\usepackage{lscape}
\usepackage{indentfirst}
\usepackage{latexsym}
\usepackage{multirow}
\usepackage{epstopdf}
\usepackage{tabls}
\usepackage{wrapfig}
\usepackage{slashbox}
\usepackage{float}
\usepackage{caption}
\usepackage{supertabular}
\usepackage[onehalfspacing]{setspace} 
\usepackage{amsmath}
\usepackage{amsfonts}
\usepackage{booktabs}
\usepackage{comment}
\usepackage{algorithm}
\usepackage{algpseudocode}

\DeclareMathOperator*{\argmin}{arg\,min}

\usepackage{pifont}
\newcommand{\cmark}{\ding{51}}%
\newcommand{\xmark}{\ding{55}}%
\newcommand{\VS}[1]{#1}
\newcommand{\norm}[1]{\left\lVert#1\right\rVert_2}

\newcommand{\revone}[1]{{\color{black}{#1}}}

\usepackage[normalem]{ulem}

\usepackage{tablefootnote}
\usepackage{gensymb}

\usepackage{tocloft,calc}
\renewcommand{\cftlottitlefont}{\hspace*{\fill}\large}
\renewcommand{\cftafterlottitle}{\hspace*{\fill}}
\renewcommand{\cftloftitlefont}{\hspace*{\fill}\large}
\renewcommand{\cftafterloftitle}{\hspace*{\fill}}
\renewcommand{\cftchapaftersnum}{:\ }
\renewcommand{\cftchappresnum}{\chaptername\space}
\makeatletter
\g@addto@macro\appendix{%
  \addtocontents{toc}{%
    \protect%
  }%
}
\usepackage[colorlinks=true,urlcolor=black,linkcolor=blue,citecolor=blue]{hyperref}
\usepackage{nameref}

\renewcommand{\baselinestretch}{2}
\begin{document}
\pagestyle{empty}

\hbox{\ } \vspace{.7in}
\renewcommand{\baselinestretch}{1}
\small \normalsize

\begin{center}
\large{{ABSTRACT}}

\vspace{3em}

\end{center}
\hspace{-.15in}
\begin{tabular}{ll}
Title of Dissertation:    & {\large  PLANNING AND PERCEPTION FOR}\\
&                     {\large  UNMANNED AERIAL VEHICLES IN} \\
&                     {\large  OBJECT AND ENVIRONMENTAL MONITORING} \\
\ \\
&                          {\large  Harnaik Singh Dhami} \\
&                           {\large Doctor of Philosophy, 2024} \\
\ \\
Dissertation Directed by: & {\large  Professor Pratap Tokekar} \\
&               {\large  Department of Computer Science } \\
\end{tabular}

\vspace{3em}

\renewcommand{\baselinestretch}{2}
\large \normalsize

Unmanned Aerial vehicles (UAVs) equipped with high-resolution sensors are enabling data collection from previously inaccessible locations on a remarkable spatio-temporal scale. These systems hold immense promise for revolutionizing various fields such as precision agriculture and infrastructure inspection where access to data is important. To fully exploit their potential, the development of autonomy algorithms geared toward planning and perception is critical. In this dissertation, we develop planning and perception algorithms, specifically when UAVs are used for data collection in monitoring applications.

In the first part of this dissertation, we study problems of object monitoring and the planning challenges that arise with them. Object monitoring refers to the continuous observation, tracking, and analysis of specific objects within an environment. We start with the problem of visual reconstruction where the planner must maximize visual coverage of a specific object in an unknown environment while minimizing the time and cost. Our goal is to gain as much information about the object as quickly as possible. By utilizing shape prediction deep learning models, we leverage predicted geometry for efficient planning. We further extend this to a multi-UAV system. With a reconstructed 3D digital model, efficient paths around an object can be created for close-up inspection. However, the purpose of inspection is to detect changes in the object. The second problem we study is inspecting an object when it has changed or no prior information about it is known. We study this in the context of infrastructure inspection. We validate our planning algorithm through real-world experiments and high-fidelity simulations. Further, we integrate defect detection into the process.

In the second part, we study planning for monitoring entire environments rather than specific objects. Unlike object monitoring, we are interested in environmental monitoring of spatio-temporal processes. The goal of a planner for environmental monitoring is to maximize coverage of an area to understand the spatio-temporal changes in the environment. We study this problem in slow-changing and fast-changing environments. Specifically, we study it in the context of vegetative growth estimation and wildfire management. For the fast-changing wildfire environments, we utilize informative path planning for wildfire validation and localization. Our work also leverages long short-term memory (LSTM) networks for early fire detection. 

\thispagestyle{empty} \hbox{\ } \vspace{1.5in}
\renewcommand{\baselinestretch}{1}
\small\normalsize
\begin{center}

\large{{PLANNING AND PERCEPTION FOR UNMANNED AERIAL \\
VEHICLES IN OBJECT AND ENVIRONMENTAL MONITORING}}\\

\ \\
\ \\
\large{by} \\
\ \\
\large{Harnaik Singh Dhami}
\ \\
\ \\
\ \\
\ \\
\normalsize
Dissertation submitted to the Faculty of the Graduate School of the \\
University of Maryland, College Park in partial fulfillment \\
of the requirements for the degree of \\
Doctor of Philosophy \\
2024
\end{center}

\vspace{7.5em}

\noindent Advisory Committee: \\
\hbox{\ }\hspace{.5in}Professor Pratap Tokekar, Chair/Advisor \\
\hbox{\ }\hspace{.5in}Professor Nikhil Chopra, Dean's Representative \\
\hbox{\ }\hspace{.5in}Professor Dinesh Manocha \\
\hbox{\ }\hspace{.5in}Professor Nirupam Roy\ \\
\hbox{\ }\hspace{.5in}Professor Michael Otte

\thispagestyle{empty}
\hbox{\ }

\vfill
\renewcommand{\baselinestretch}{1}
\small\normalsize

\vspace{.5in}

\begin{center}
\large{\copyright \hbox{ }Copyright by\\
Harnaik Singh Dhami  
\\
2024}
\end{center}

\vfill

\newpage 

\pagestyle{plain} \pagenumbering{roman} \setcounter{page}{2}



\phantomsection 
\addcontentsline{toc}{chapter}{Dedication}

\renewcommand{\baselinestretch}{2}
\small\normalsize
\hbox{\ }
 
\vspace{.5in}

\begin{center}
\large{Dedication}
\end{center} 

To my Baba, Tarlochan Singh Dhami.

\phantomsection 
\addcontentsline{toc}{chapter}{Acknowledgements}

\renewcommand{\baselinestretch}{2}
\small\normalsize
\hbox{\ }
 
\vspace{.5in}

\begin{center}
\large{Acknowledgments} 
\end{center} 

\vspace{1ex}

First, I would like to start by thanking my advisor, Pratap Tokekar. I never imagined that working up the courage after class in my junior year of undergrad to join as a researcher in your lab would lead to this. Your guidance and mentorship through these past eight years have been instrumental in helping me grow as a researcher. There was not a single day during these eight years where I felt unsupported and always knew I could come to you for advice both related and unrelated to our research endeavors. You knew exactly how to push me to become a better researcher without adding any extra stress. I could not ask for a better advisor and mentor.

I am thankful to my committee members Prof. Dinesh Manocha, Prof. Nirupam Roy, Prof. Michael Otte, and Prof. Nikhil Chopra for taking the time to give valuable feedback for my dissertation. I would also like to thank Prof. Ryan Williams and Prof. Song Li for their support during my master's studies at Virginia Tech. Our work together helped convince me to continue with my research journey.

Next, I am grateful for all my collaborators and co-authors: Vishnu Dutt Sharma, Kevin Yu, Troi Williams, Charith Reddy, Dennis Trimarchi, Jun Liu, Murtaza Rangwala, Kulbir Singh Ahluwalia, Shayan Ghajar, Tianshu Xu, Vineeth Vajipey, Qian Zhu, Kshitiz Dhakal, James Friel, and Prof. Benjamin Tracy. This dissertation would not be possible without your help in all the work we did together. This would also not be possible without the financial support from National Science Foundation grants \#1840044 and \#1943368, National Institute of Food and Agriculture grants \#67007-28380 and \#2018-51181-28384, Office of Naval Research grant \#N00014-18-1-2829, and N5 Sensors.

I am also thankful for all my friends who helped me stay sane during all these years. Whether it was listening to me vent about my work or just having fun, your support helped re-energize me while away from my work. 

I am extremely grateful to my fiancée, Sandip Minhas, for all her support and encouragement. When we met, I remember being worried and anxious about where my research was headed after dealing with a few paper rejection cycles. Now, I am close to the end of my graduate studies journey and you played a big part in that. You have been my external motivation and I work every day to make you proud. Without your love and support (and pressure!), I would not be where I am right now.

Finally, I am forever thankful for my family. They are the reason I am who I am today. Throughout the years, their support and encouragement helped me grow as a person and strive to become better. Thank you to my parents, Parminder and Gagandeep Dhami, and my (always a baby to me) sister Happi. I would not have been able to pursue this dissertation without your continual love, support, and guidance through the years. Special thanks to my grandparents, my Dadi Joginder, my Nana Bakhshish, and my Nani Mohinder. My Baba Tarlochan sadly is not here to see the completion of this chapter of my life, but I know he would be the most proud. Also my cousins Jaspreet, Shappi, and Joey Hira as well as their parents, my Bhua Rupinder and my Fufar Davinder. I always thought of you more as siblings than cousins since we grew up together. With where we have all ended up, our parents must have done something right!  
    \cleardoublepage

\phantomsection 
    \addcontentsline{toc}{chapter}{Table of Contents}
    \renewcommand{\contentsname}{Table of Contents}
\renewcommand{\baselinestretch}{1}
\small\normalsize
\tableofcontents 
\newpage

\phantomsection 
\addcontentsline{toc}{chapter}{List of Tables}
    \renewcommand{\contentsname}{List of Tables}
\listoftables 
\newpage

\phantomsection 
\addcontentsline{toc}{chapter}{List of Figures}
    \renewcommand{\contentsname}{List of Figures}
\listoffigures 
\newpage


\newpage
\setlength{\parskip}{0em}
\renewcommand{\baselinestretch}{2}
\small\normalsize

\setcounter{page}{1}
\pagenumbering{arabic}

\renewcommand{\thechapter}{1}

\chapter{Introduction}\label{chap:intro}

Robots are autonomous machines designed to perform tasks that are dull, dirty, or dangerous. They rely on technologies such as planning and perception to navigate and interact with their environment effectively. Planning involves algorithms that enable robots to find optimal routes for the task at hand. Perception involves sensor processing, allowing robots to interpret and understand their surroundings. Both of these are fundamental enablers for robots. Effective planning and perception algorithms are crucial for improving efficiency, accuracy, and safety of operation of robots. 


The goal of this dissertation is to develop efficient planning and enhancing perception capabilities for Unmanned Aerial Vehicles (UAVs) when they are used for data collection in monitoring applications. UAVs are currently being used (or hold the promise for use) for data collection in applications such as precision agriculture~\cite{tsouros2019review}, air quality monitoring~\cite{villa2016overview}, surveillance~\cite{semsch2009autonomous}, infrastructure inspection~\cite{shakhatreh2019unmanned}, and first response~\cite{jin2020research}. In such applications, UAVs instrumented with appropriate sensors such as RGB cameras, infrared cameras, thermal sensors, LIDARs, etc. can be tasked to collect data from hard-to-reach places or for extended periods of time. While current practices involve manually piloting the UAVs, our vision is to be able to enable fully autonomous operations. This requires effective joint planning and perception algorithms that decide \emph{where, when, and how} to collect the data. In this dissertation, we develop planning and perception algorithms that address these fundamental questions in the context of object and environmental monitoring. With better planning and perception, we can significantly improve the accuracy, coverage, and effectiveness of UAV-based systems.

We begin by describing two tasks where planning and perception are needed, specific applications of these tasks, and some of the challenges that arise when solving these tasks.

\section{Tasks, Applications, and Challenges}

We are motivated by scenarios where the robot is used for monitoring. We may be interested in monitoring one specific object of interest in the environment or monitoring the entire environment as a whole. Depending on whether the task is to monitor a specific object or the entire environment will affect the planner and the perception module. For object monitoring, the planner must focus on directing the robot to specific targeted areas relative to the object. On the other hand, for environmental monitoring, the planner must balance covering the entire area while still gathering targeted information from interesting areas. Furthermore, the environment itself may change which requires reasoning about a spatio-temporal field. 
In this dissertation, we will address planning and perception challenges in both tasks. These tasks are discussed in more detail below.


\subsection{Object Monitoring}

Object monitoring refers to the continuous observation, tracking, and analysis of specific objects within an environment. 
Object monitoring has many applications where it enables real-time monitoring, event detection, and data collection for specific objects, allowing for timely responses and informed decision-making based on the gathered information.

\begin{figure}[ht!]
        \centering
        \includegraphics[width=0.7\linewidth]{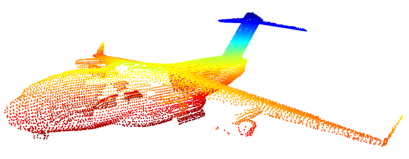}
        \caption{Motivating Application 1: Visual Reconstruction. Observations taken from UAVs over time should completely reconstruct the object of interest. These observations should be efficiently obtained. The image is from simulations conducted in AirSim where 3D pointcloud observations were captured to reconstruct a C17 airplane.}
        \label{fig:reconstruct}
\end{figure}

One of these applications is visual reconstruction. Visual reconstruction plays a pivotal role in both natural and built environments. This technique involves the creation of three-dimensional (3D) digital models of objects, structures, and landscapes, providing valuable insights and data for informed decision-making and resource management~\cite{DebevecPhd1996}. An example 3D reconstructed pointcloud model is shown in Figure~\ref{fig:reconstruct}. These reconstructed 3D models are used in many domains. In medical imaging, these 3D models allow for a non-invasive method for diagnosis~\cite{pichat2018survey}. In agriculture, farmers can use 3D models for crop monitoring, yield estimation, and management of resources~\cite{dong20174d}. For inspection, 3D models can help with defect detection, part inspection, and quality assurance~\cite{bitzidou2013multi, xu2021toward}. Solving visual reconstruction problems is challenging. The goal of reconstruction is to get enough images from all sides of the target object to maximize coverage. Naturally, the more images there are, the better the reconstruction is up to a point. To efficiently carry out reconstruction, we must carefully choose where to obtain the images from so that we maximize coverage while minimizing time and cost. The first problem that we study in the dissertation is: \textbf{Where should we obtain measurements from to quickly reconstruct an object of interest?}

One of the applications of building 3D digital models is inspecting them for defects. Inspection is crucial for ensuring the safety, functionality, and longevity of critical assets such as bridges, buildings, pipelines, and power lines~\cite{ariaratnam2001assessment, madanat1994optimal}. Regular inspections help detect and address structural defects, wear and tear, or potential hazards before they lead to accidents or costly failures~\cite{chang2003health, ellingwood2005risk}. By creating accurate 3D models, engineers and inspectors can identify defects, structural damage, or wear and tear, ensuring the safety and integrity of these structures.

UAVs have become prominent for inspection by offering a cost-effective and efficient alternative to traditional inspection methods~\cite{gillins2016cost, eschmann2013high}. UAVs equipped with high-resolution cameras, LiDAR sensors, thermal imaging, and other sensors can access difficult-to-reach or hazardous areas, providing detailed visual and data-based assessments~\cite{duque2018synthesis, ellenberg2015use}. UAVs can quickly identify structural damage, corrosion, cracks, or irregularities, enabling engineers and inspectors to make informed decisions about maintenance or repairs~\cite{gopalakrishnan2018crack, lemos2023automatic}.

\begin{figure}[ht!]
        \centering
        \includegraphics[width=0.7\linewidth]{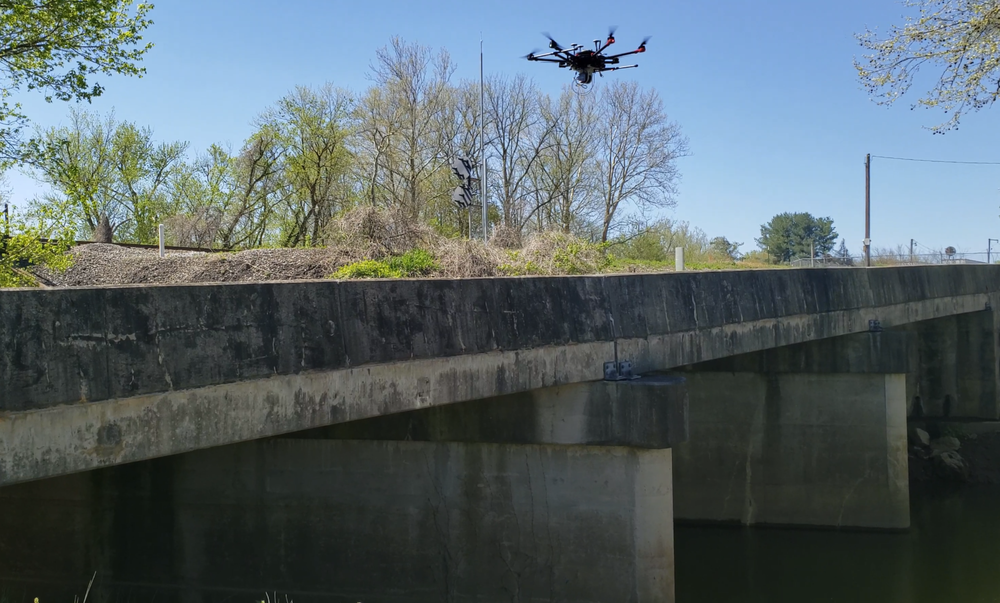}
        \caption{Motivating Application 2: Visual Inspection. UAVs are used to efficiently inspect the surface of a target infrastructure while avoiding obstacle collisions and having limited prior information. Image is from experiments at Whitehorne Road Bridge at Virginia Tech's Kentland Farm.}
        \label{fig:inspect}
\end{figure}

Inspection provides its own challenges for planning. Instead of focusing on the entirety of the object, the goal here is to get specific details about the object. This requires getting close-up imagery of the surface of the object so as not to miss critical details. For most current methods, it is assumed information about the inspection surface, such as a 3D digital model, is known apriori. However, the purpose of inspection is to detect changes implying that the 3D digital model could be inaccurate. There can also be environmental changes between the time the digital model was created and when the inspection is done. Another key question we address in this work is: \textbf{Where should we obtain measurements from to quickly inspect a surface of interest with partial or no prior information?} We consider this problem in the context of infrastructure inspection in this dissertation.

\subsection{Environmental Monitoring}

With object monitoring, the focus is on a specific object in the environment to obtain detailed information about that object. Environmental monitoring is the process of observing conditions within an entire geographical area. The objective of environmental monitoring is obtaining information on the state of the environment including any changes over time. It typically involves deploying a network of sensors or instruments, such as robots, to gather data over time. Environmental monitoring also has many applications allowing monitoring of these geographical areas through informed decision-making. 

\begin{figure}[ht!]
        \centering
        \includegraphics[width=0.7\linewidth]{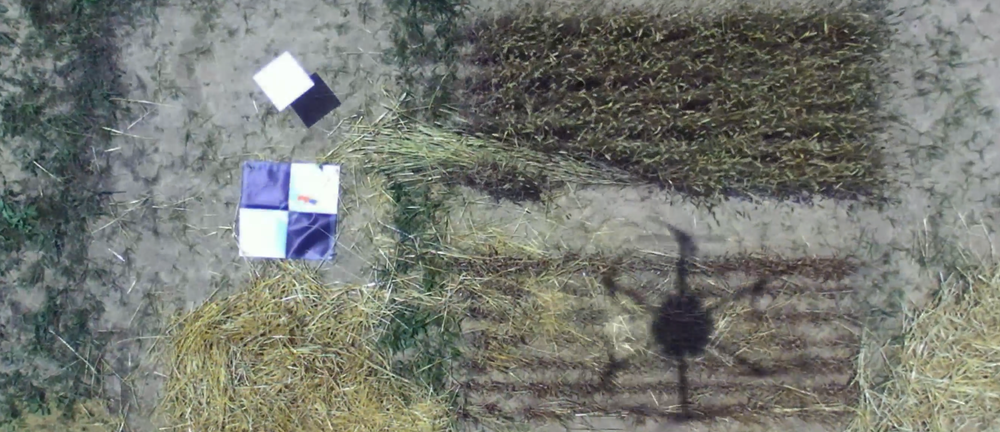}
        \caption{Motivating Application 3: Change Detection. UAV is used to detect changes that are occurring in a slow-changing environment, specifically in the context of precision agriculture and crop height estimation. Image is from experiments conducted over a wheat field at Virginia Tech's Kentland Farm.}
        \label{fig:changedetect}
\end{figure}

One application of environmental monitoring is detecting changes. A need to detect changes arises in various domains. In precision agriculture, collected information about environmental changes is used to help with efficient resource management, sustainability, improved crop quality, increased yields, and reduced costs in modern farming~\cite{maes2019perspectives, gokool2023crop}. In wildfire management, monitoring changes can help with early fire detection~\cite{mohapatra2022early}. Different methods are used for detecting changes such as stationary sensors~\cite{kumar2018impact}, satellites~\cite{murugan2017development}, and UAVs~\cite{casbeer2005forest}. Change detection is a challenging problem.
It is fundamentally different from visual reconstruction and inspection because the underlying environment changes over time. The goal of a planner is to maximize the coverage of a target area to understand the spatio-temporal changes in an environment. The next key question we address in this work is: \textbf{How can we detect changes in the environment from the measurements we have obtained?} This is another fundamental problem we consider in this dissertation in the context of vegetative growth estimation, as shown in Figure~\ref{fig:changedetect}. 

\begin{figure}[ht!]
        \centering
        \includegraphics[width=0.7\linewidth]{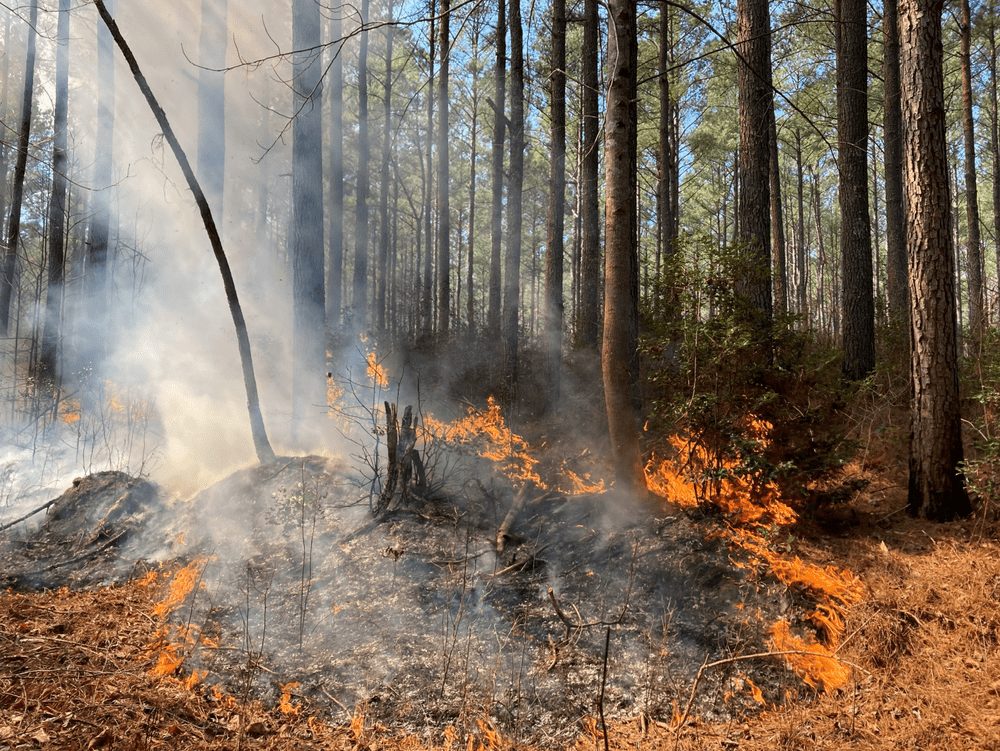}
        \caption{Motivating Application 4: Change Validation. In environments that change rapidly, such as when there are wildfires, detected changes need to be validated quickly due to high levels of noise and false positives. The image is from a planned burn that took place at Big Woods, VA where N5 deployed ChemNode sensor stations to capture data.}
        \label{fig:changevalid}
\end{figure}

Once changes have been detected, there is a need for validation. Stationary sensors might give off false positives and these need to be validated before resources go to waste. Low Earth orbit satellites cannot monitor the same area all the time, so the information from them could be out of date. Therefore, fast validation is needed after early detection. When using UAVs for this, informative path planning can lead to quick and efficient validation. Informative path planning generates paths that maximize information gain while minimizing cost. It has found use in domains such as search-and-rescue~\cite{Lim2015} and surveillance~\cite{bostrom2019informative}. 
There are many challenges in change validation. The type of changes can be very nuanced and complex, especially in environments where the changes happen fast. It is very similar to the challenge of visual reconstruction where a planner must maximize coverage while minimizing time and cost. The important difference here is that the environment is changing during planning. The last key question we address in this work is: \textbf{Where should we obtain measurements from to quickly validate changes in the environment?} We study this fundamental problem in the context of wildfires (Figure~\ref{fig:changevalid} in this dissertation).

In the sections that follow, this dissertation will delve into each of these tasks, presenting prior work as well as our contributions; each showcasing the versatility and effectiveness of UAVs in object and environmental monitoring. We emphasize the need for developing planning and perception solutions in each application.

\section{Prior Work}
In this section, we go over some of the previous work for object and environmental monitoring. For object monitoring, we highlight previous work in the applications of visual reconstruction and inspection. For environmental monitoring, the tasks we focus on are change detection in the context of precision agriculture and change validation in the context of wildfire management. 

\subsection{Object Monitoring: Visual Reconstruction}
As stated previously, visual reconstruction is an application of object monitoring. This is a well-studied problem, especially in the field of robotics. One common planning approach to reconstruct objects in unknown environments is utilizing next-best-views (NBV)~\cite{scott2003view}. Using previous observations, these NBV algorithms select the next observation location where information gain will be maximized. NBV algorithms generally fall under 2 categories, utilizing probabilistic~\cite{delmerico2018comparison} or geometric~\cite{tarabanis1995survey} models. Most robotic NBV works fall under the first category, using probabilistic models to solve exploration problems~\cite{kuipers1991robot, yamauchi1997frontier}. 

Recent methods employ prediction-based models to aid object reconstruction. They predict shapes in the environment for more efficient exploration~\cite{georgakis2022uncertainty}. Various works use these predictions to generate views for efficient object reconstruction~\cite{Yang_2019, yuan2018pcn, yu2021pointr}. Other works use 3D predictions and machine learning to accomplish planning and perception tasks. Reinforcement learning can be used to predict NBVs for the reconstruction of 3D house models~\cite{Peralta2020}. Another example is using supervised learning to predict NBVs using point clouds~\cite{zeng2020pc}. One issue with these approaches is that they are not easily translatable to other problems due to the nature of the machine learning methods they deploy. 

All of the above work focuses on single-robot systems. Deploying multi-robot systems could help improve performance; however, this must be done in a way where there is no significant overlap in observations between the robots in the system. There has already been significant work on creating exploration methods designed for multi-agent systems that find NBVs from probabilistic models~\cite{burgard2005coordinated, amanatiadis2013multi, hardouin2020next}. One work combines a classical and learning-based method to create a hybrid planner for selecting NBVs based on partial observations from the robots~\cite{almadhoun2021multi}. Another work uses a predictive point cloud model to improve plant phenotyping efficiency~\cite{wu2019plant}. These methods do not consider robotic control effort for their planning, which can be an issue for UAVs with limited battery life. They also use probabilistic models for NBV selection as opposed to geometric models. 

\subsection{Object Monitoring: Inspection}
Once a 3D digital model of an object is available, it can be used for inspection. There are also other methods to gain information about the target object. Another method of gaining this information is exploration. One common approach to solving exploration is utilizing frontiers~\cite{zhu20153d, da2020novel, niroui2017robot}. Frontiers are defined as the edge between known space and unknown space. Naturally, if you fly to a frontier, the unknown part of the environment becomes discovered and explored. Variants of frontier exploration algorithms have also been proposed~\cite{dai2020fast, shen2012autonomous}. Another approach for exploration is to generate paths that maximize information gain~\cite{corah2019communication, premkumar2020combining}. Some approaches use the previously discussed NBV method for exploration~\cite{pito1999solution}.

For solving inspection problems, the above exploration approaches are lacking since they provide no guarantees towards targeted inspection. However, solving inspection with robots is a well-studied problem. Some methods use a prior map of the environment to create inspection paths~\cite{peng2019adaptive, roberts2017submodular}. Along these paths, they obtain high-resolution observations of the target. However, they are not applicable in cases where prior information about the environment and target is not known. 

One method of solving this problem is to first explore the environment and then inspect it~\cite{bircher2018receding}. This method uses information gain in a receding horizon manner. For both exploration and inspection, paths are generated using rapidly-exploring random trees~\cite{LaValle1998RapidlyExploringRT} (RRT), and the path with the highest information gain is selected. While previous information is not needed for the exploration of the environment, their method uses the previously known inspection surface to determine information gain for the inspection planner.

Other methods relax the inspection surface assumption. One such method proposed a high-level planner for coverage and a low-level planner for inspection~\cite{song2020online}. The low-level planner considers viewpoint constraints and selects local paths that maximize information gain about the target infrastructure. One drawback is that a bounding box around the target infrastructure is given beforehand to give the planner constraints.

\subsection{Environmental Monitoring: Change Detection}
As stated before, some applications require that the entire environment needs to be monitored and not just a specific object. A specific emphasis of this is to detect changes within the environment. One application of environmental change detection is precision agriculture and crop health monitoring. A well-studied crop health monitoring problem is crop height estimation. One method equipped a UAV with a downward facing 2D LiDAR for corn height measurement~\cite{anthony_elbaum_lorenz_detweiler_2014}. This method uses LiDAR to estimate the ground plane and the top of the corn, the difference of which is the estimated crop height with 5 cm accuracy. The authors only focus on height estimation with this method.

Another method utilized a similar approach from the ground as opposed to the air. They deploy an unmanned ground vehicle (UGV) equipped with 3D LiDAR to make the ground plane and wheat height estimations~\cite{madec}. They also deploy a UAV with an RGB camera mounted and create a dense 3D point cloud using a structure-from-motion (SfM) method. Their ground-based method had a root-mean-square error (RMSE) of 3.5 cm while the air-based method had an RMSE between 2.6-6.8 cm. As opposed to algorithmically dividing the 3D point cloud for estimations, they arbitrarily break it up. 

Utilizing other sensors on UAVs as opposed to RGB cameras and 2D LiDARs is another well-studied approach. One method compares estimations from an ultrasonic sensor onboard a UAV and a UGV with 3D LiDAR~\cite{yuan_li_bhatta_shi_baenziger_ge_2018}. They found that their UGV-based method had an RMSE of 5 cm while the UAV ultrasonic approach had an RMSE of 9 cm. One assumption of their algorithm is the height of the 3D LiDAR before estimations. This is manually measured beforehand and then used for the estimations. 

Some work also uses fixed-wing UAVs as opposed to multi-rotors. One such method mounted an RGB camera on a fixed-wing UAV and deployed ground control points along the field for image calibration and height estimation~\cite{ziliani_parkes_hoteit_mccabe_2018}. They also deployed a 3D LiDAR on a ground vehicle that drove around the field and compared the data collected to the point cloud generated from the UAV RGB images using SfM. They found a correlation of 0.99 between the ground-based LiDAR method and the air-based RGB method. However, during the flowering of the crops, the variability increased and the correlation dropped down to 0.65. Another drawback of their method is that the ground vehicle only drives around the field and cannot make any measurements within the field.  


\subsection{Environmental Monitoring: Change Validation}

In environments with fast changes or noisy measurements, detected changes need to be validated. An example of this type of environment is wildfires. Fast-spreading fires cause the environment to change fast and wildfire management is needed. This incorporates fire detection, monitoring, and suppression~\cite{Ollero2006}. For detection, one method proposed using a pair of UAVs, one equipped with a visual camera and another equipped with an infrared camera~\cite{Merino2005}. Their algorithm utilizes segmentation to detect fires within the visual and infrared images along with UAV position information to localize detected fires. However, they assume that the fires are static and not spreading. 

This is addressed in further work by the authors in which they use a fleet of UAVs to map out a fire front~\cite{Merino2012}. Their method allows for the use of a heterogeneous UAV fleet as long as the vehicle can carry the necessary perception sensors (infrared or visual cameras), can localize itself, and can fly to specific waypoints. All information is sent to a centralized station for processing. Before fires, a UAV patrols the area for initial detection. This can become unrealistic as the size of the target monitoring area increases. 

Another area of focus is using deep learning frameworks for fire detection. One study tested the accuracy of various convolutional neural networks (CNNs)~\cite{Krizhevsky2012, Szegedy2014, simonyan2015deep} on making fire predictions from input images~\cite{Lee2017}. After training, the study evaluated that CNNs provide a highly accurate method of autonomous fire detection and proposed using such networks onboard UAVs.

\section{Research Contributions}

\begin{figure}
    \centering
    \begin{subfigure}[b]{.49\columnwidth}%
        \centering
        \includegraphics[width=\linewidth]{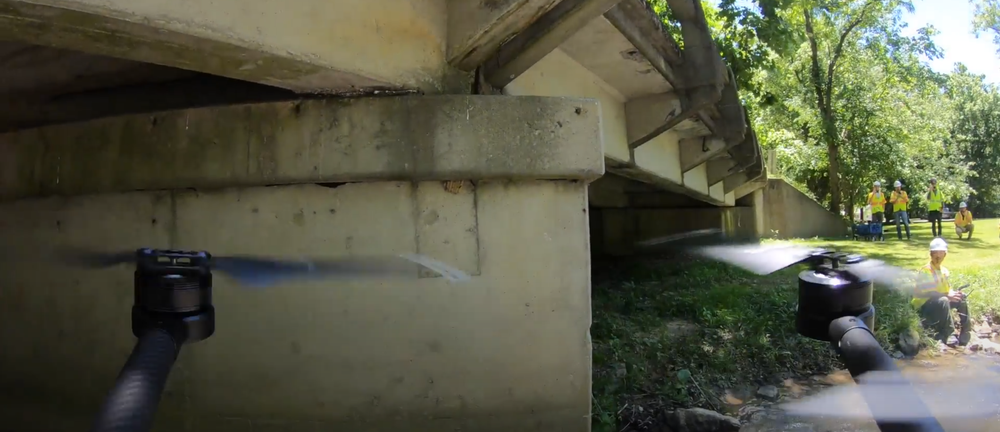}
        \caption{DJI M600 Pro during flight for bridge infrastructure inspection.}
    \end{subfigure}%
    \hfill%
    \begin{subfigure}[b]{.49\columnwidth}%
        \centering
        \includegraphics[width=\linewidth]{figs/kent_drone.png}
        \caption{DJI M600 Pro during flight capturing 3D pointcloud data at wheat farm.}
    \end{subfigure}%
    \hfill%
    \begin{subfigure}[b]{.49\columnwidth}%
        \centering
        \includegraphics[width=\linewidth]{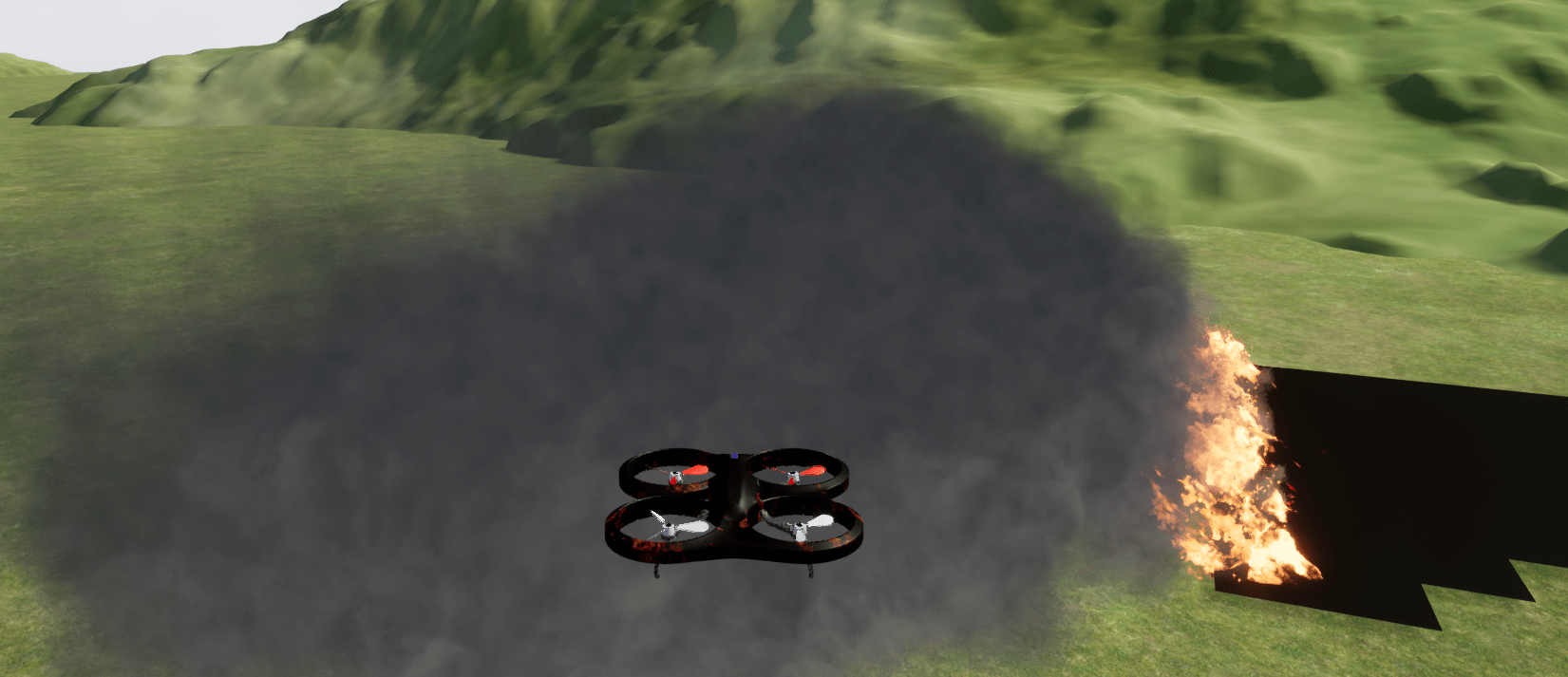}
        \caption{AirSim Multirotor during flight in wildfire simulation.}
    \end{subfigure}%
    \hfill%
    \begin{subfigure}[b]{.49\columnwidth}%
        \centering
        \includegraphics[width=\linewidth]{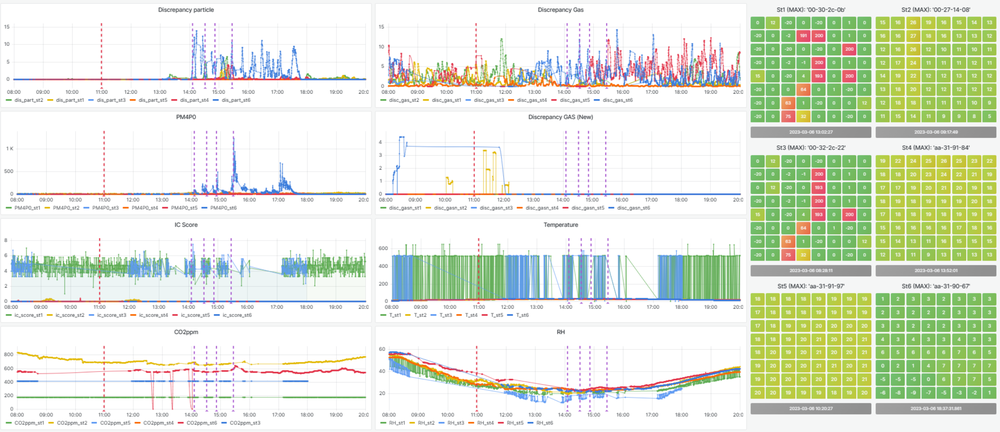}
        \caption{Dashboard highlighting multi-modal data captured from N5's ChemNode.}
    \end{subfigure}%
    \caption{Overview of the hardware (both real and simulated) that was used in this dissertation.}
    \label{fig:hardware_overview}
\end{figure}

In this section, we highlight our contributions to perception and planning for object and environmental monitoring. Our key contribution is planning algorithms that reason about \emph{where} to obtain measurements to efficiently carry out the monitoring task. The specific planning algorithms are based on the nature of the object or environment to be monitored. As such, the planner makes use of the structure in the underlying sensing models used for monitoring. We ground our work in the specific applications of infrastructure inspection, precision agriculture, and wildfire management, as shown in Figure~\ref{fig:hardware_overview}). However, our work can be used in other applications as well.

\subsection{Object Monitoring: Visual Reconstruction}
For visual reconstruction, planning needs to determine where to obtain measurements for quick object reconstruction. A balance between capturing enough measurements to reconstruct the object and minimizing time needs to be met. Due to this, there is an emphasis on efficient robot navigation. Prediction-based active perception has emerged as a promising approach to elevate both the efficiency and safety of robot navigation. This method accomplishes this by proactively anticipating and accounting for uncertainties in unfamiliar environments. However, current research in 3D shape prediction often assumes the availability of partial observations, making it unsuitable for practical real-world planning scenarios. Furthermore, these approaches often neglect the consideration of control efforts required for next-best-view planning.

In response to these limitations, we introduce \textit{Pred-NBV}~\cite{dhami2023prednbv}, an innovative object shape reconstruction framework. \textit{Pred-NBV} consists of two key components: PoinTr-C, an enhanced 3D prediction model trained on the ShapeNet~\cite{shapenet2015} dataset, and an information and control effort-based next-best-view strategy. This novel approach addresses the shortcomings of existing methods, promising practical applicability in object reconstruction.

\textit{Pred-NBV} demonstrates significant advancements, boasting a remarkable 25.46\% improvement in object coverage compared to traditional methods when evaluated within the AirSim~\cite{airsim2017fsr} simulator. Furthermore, in real-world scenarios using data acquired from a Velodyne 3D LiDAR mounted on DJI M600 Pro, \textit{Pred-NBV} outperforms PoinTr, the state-of-the-art shape completion model, showcasing its potential to excel in complex, real-world environments. This work was accepted at the \textbf{2023 IEEE/RSJ International Conference on Intelligent Robots and Systems (IROS 2023)}~\cite{dhami2023prednbv}\footnote{This work was done in collaboration with Vishnu Dutt Sharma.}. 

\begin{figure}[ht!]
    \centering
    \includegraphics[width=.75\linewidth]{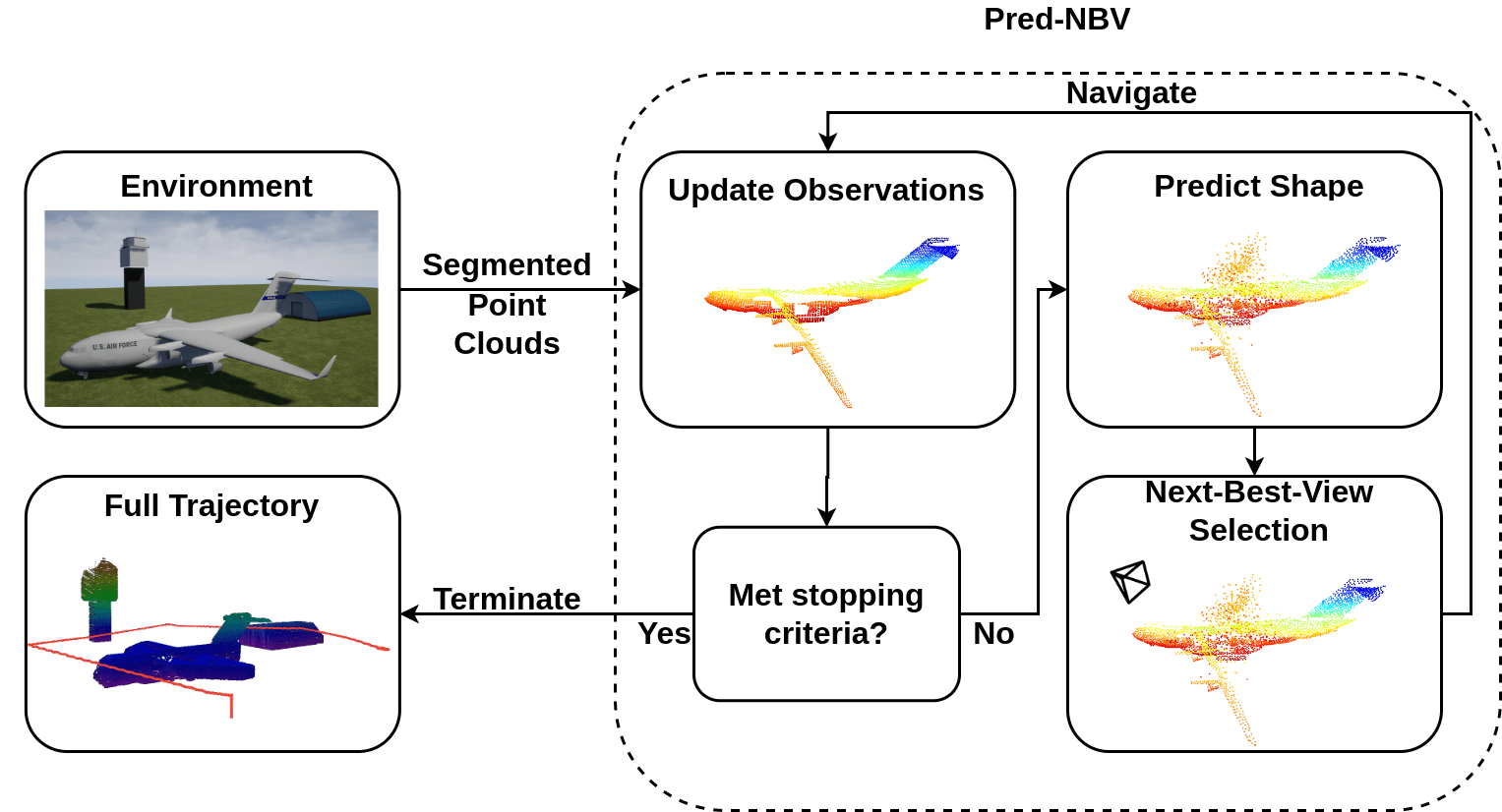}
    \caption{Overview of \textit{Pred-NBV}: An autonomous planner for 3D object reconstruction utilizing point cloud completion predictions. Pred-NBV predicts the complete point cloud using all previous observations and then selects the next view based on where the most predicted points can be seen targeting efficient visual reconstruction.}
    \label{fig:overview_pred}
\end{figure}

We also introduce an extension of \textit{Pred-NBV}, \textit{MAP-NBV}~\cite{dhami2023mapnbv}, an algorithm tailored for multi-agent systems. Traditional prediction-based approaches have made significant strides in enhancing active perception tasks by harnessing data-driven insights into object structures. However, these methods predominantly cater to single-agent systems, leaving a gap in collaborative multi-robot scenarios.

\textit{MAP-NBV} fills this void by pioneering a next-best-view strategy that leverages geometric metrics derived from predictive models. Our method takes a holistic approach, jointly optimizing information gain and control effort to facilitate efficient and collaborative 3D reconstruction of objects. Through extensive testing, \textit{MAP-NBV} is shown to improve reconstruction quality by 19\% compared to a non-predictive frontier-based NBV planner. Further, we show \textit{MAP-NBV} is 17\% better than a non-cooperative prediction-guided method while performing similarly in a centralized approach. This work was accepted at the \textbf{2024 IEEE/RSJ International Conference on Intelligent Robots and Systems (IROS 2024)}.~\cite{dhami2023mapnbv}\footnote{This work was done in collaboration with Vishnu Dutt Sharma.}.

\begin{figure}[ht!]
    \centering
    \includegraphics[width=\columnwidth]{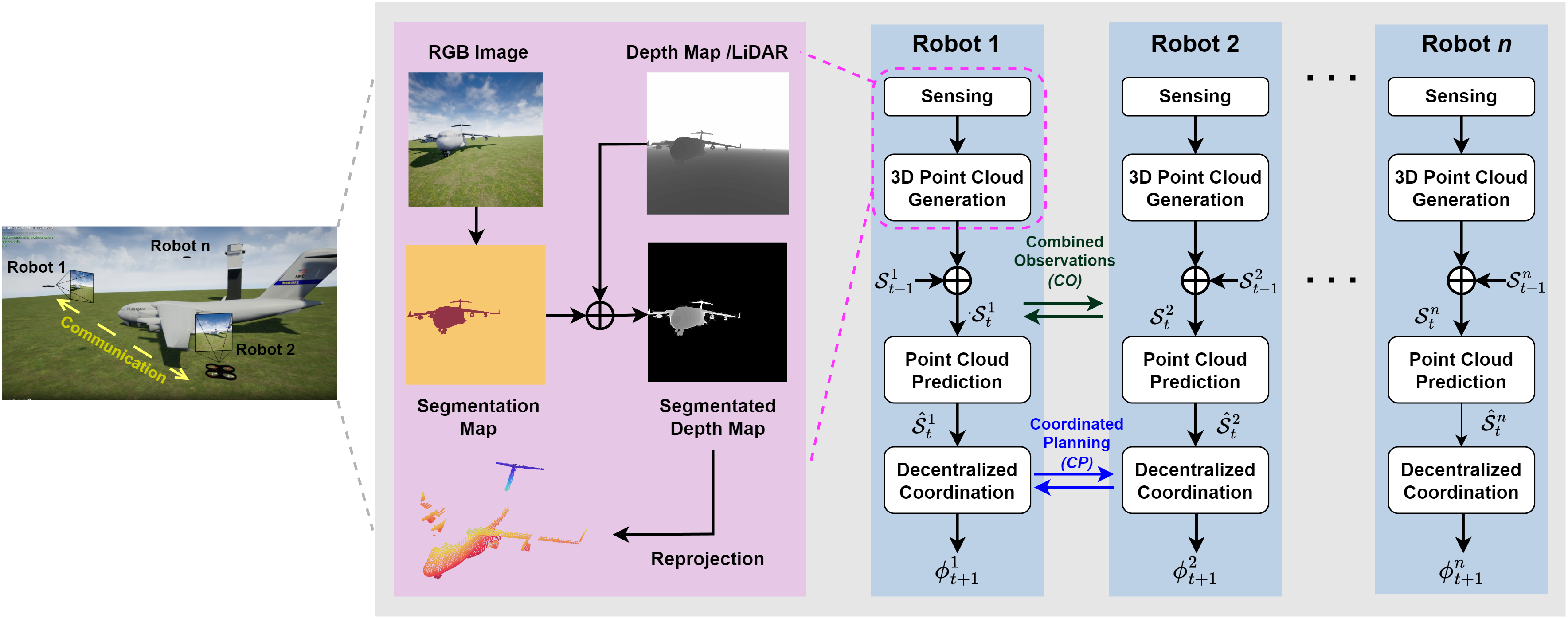}
    \caption{Overview of \textit{MAP-NBV}: A multi-agent extension of \textit{Pred-NBV}. It utilizes multiple UAVs along with predictions to target efficient visual reconstruction in a decentralized manner.}
    \label{fig:pointsObserved}
\end{figure}

\subsection{Object Monitoring: Inspection}
For inspection of an object given partial or no prior information, planning determines where to obtain measurements to quickly inspect the surface of interest. As opposed to visual reconstruction where the focus is on capturing the geometric information of an object, the focus of inspection is to observe finer details about specific parts of that object. Also, with visual reconstruction, we considered a single next-best-view to increase our reconstruction quality. However, what if we consider multiple views, i.e. a trajectory, to increase the inspection quality? We specifically tackle the challenging task of visually inspecting infrastructure surface for defects using a UAV, without assuming prior knowledge of the infrastructure's geometric model. Our novel planning system, \textit{GATSBI}~\cite{dhamiGATSBI}, operates in a receding horizon manner to systematically inspect every point on the infrastructure's surface. To achieve this, \textit{GATSBI} takes as input a real-time 3D occupancy map generated from LiDAR scans. Within this map, voxels representing the infrastructure are semantically identified and used to construct a specialized infrastructure-only occupancy map.

Inspecting each infrastructure voxel necessitates capturing images from specific viewing angles and distances using the UAV. To optimize this process, we formulate a Generalized Traveling Salesperson Problem (GTSP)~\cite{henry1969record, saskena1970mathematical, srivastava1969generalized} instance that groups potential viewpoints for inspecting the infrastructure voxels. Subsequently, we employ a readily available GTSP solver~\cite{Smith2017GLNS} to compute the optimal inspection path based on the given instance. As the algorithm progressively explores more of the environment, it dynamically recalculates the inspection path to cover new areas of the infrastructure while avoiding obstacles.

To validate the effectiveness of our approach, we conducted extensive evaluations through high-fidelity simulations using AirSim~\cite{airsim2017fsr}, as well as real-world experiments. Our comparative analysis pits \textit{GATSBI} against a conventional exploration algorithm as well as a baseline inspection algorithm, yielding insightful results. Notably, our findings underscore that focusing inspection efforts exclusively on segmented infrastructure voxels and meticulously planning with a GTSP solver yield a more efficient and comprehensive inspection process compared to the baseline algorithm. This work was accepted at the \textbf{2023 International Conference on Unmanned Aircraft Systems (ICUAS 2023)}~\cite{dhamiGATSBI} and the extended work is currently under review at a journal and a manuscript is available on arXiv~\cite{dhami2024gatsbijournal}.

\begin{figure}
    \centering
    \includegraphics[width = 0.6\columnwidth]{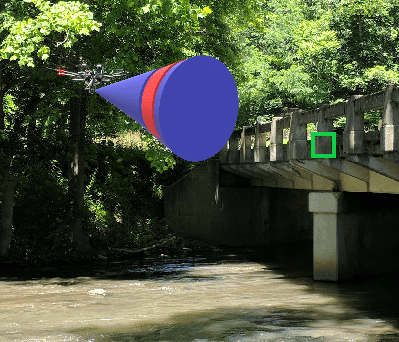}
    \caption{Example of UAV inspecting bridge during mid-flight of \textit{GATSBI}. \textit{GATSBI} targets complete surface inspection of a specific object in the environment while avoiding collision with obstacles.}
    \label{fig:vcvd}
\end{figure}

\subsection{Environmental Monitoring: Change Detection}
Next, we shift focus from monitoring a specific object in the environment to monitoring the entire environment. For environmental monitoring, there is a need to detect changes in obtained measurements over an entire geographical area. Detecting environmental changes can help one learn important spatiotemporal characteristics of the environment. We study this problem in the context of vegetative height estimation. Precision agriculture serves as a prime example of a slow-changing environment, where noticeable alterations within a single flight are typically absent. Nonetheless, over time, such environments do exhibit variations, particularly as plant life undergoes growth cycles. We introduce techniques for the precise measurement of crop heights utilizing a 3D LiDAR sensor mounted on a UAV~\cite{dhami2020crop}. The accurate assessment of plant heights holds importance in the monitoring of overall plant health and growth cycles, particularly in the context of high-throughput plant phenotyping.

Our study presents a methodology tailored to the extraction of plant heights from 3D LiDAR point clouds, with a specific focus on plot-based phenotyping environments. Additionally, we unveil a toolchain designed for the creation of phenotyping farms, which can be employed in Gazebo~\cite{Koenig2004Gazebo} simulations. This tool enables the generation of randomized farms featuring realistic 3D plant and terrain models.

To validate our approach, we conducted a series of simulations and hardware experiments under both controlled and natural conditions. Our algorithm demonstrated the capability to estimate plant heights within a field containing 112 plots, achieving an RMSE of 6.1 cm. Notably, this marks the first dataset of its kind, showcasing 3D LiDAR data collected from an airborne robot over a wheat field. This research holds promise in the fields of environmental monitoring and precision agriculture. This work was accepted at the \textbf{2020 IEEE/RSJ International Conference on Intelligent Robots and Systems (IROS 2020)}~\cite{dhami2020crop}.

\begin{figure}[htp]
    \centering
    \includegraphics[width=.75\linewidth]{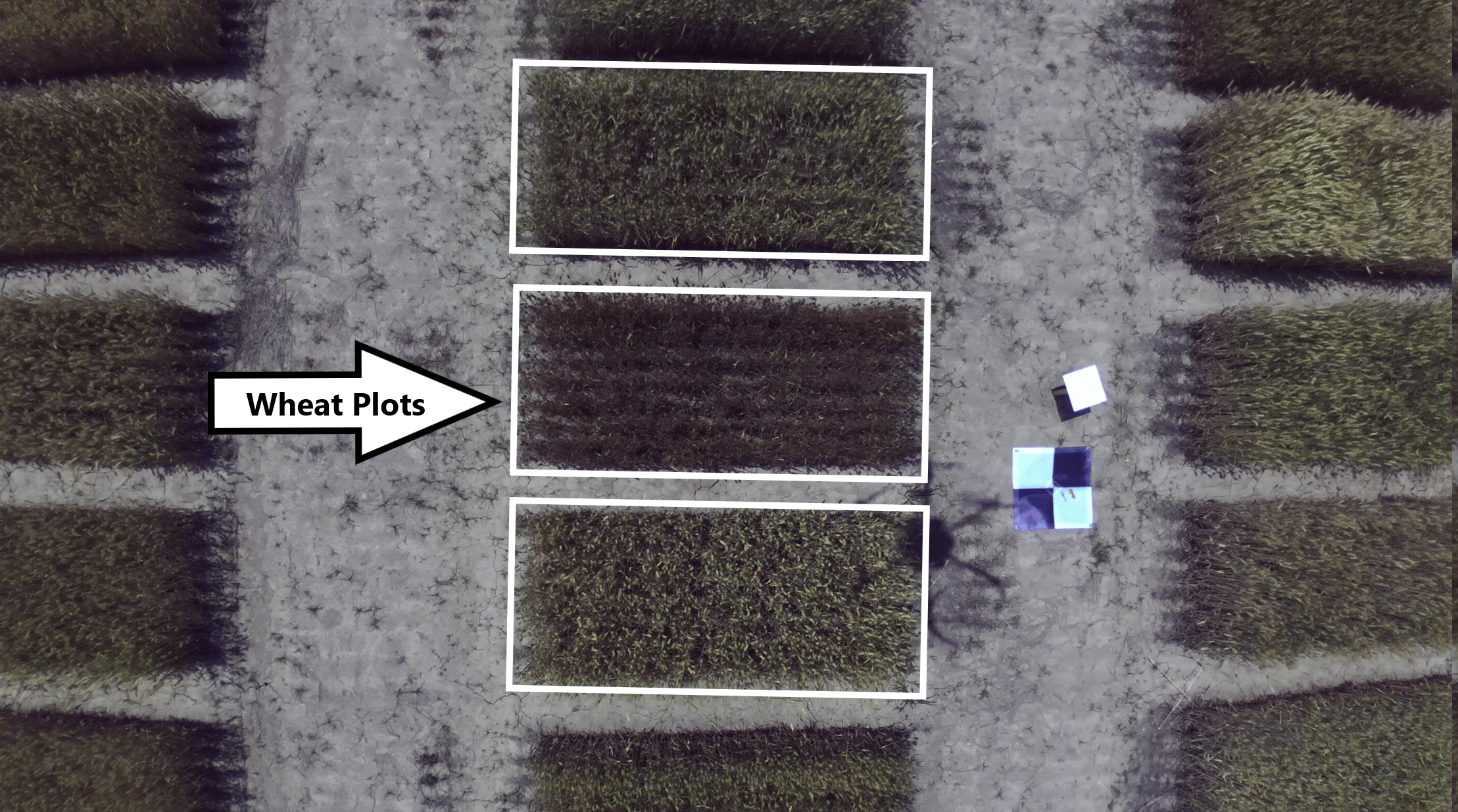}
    \caption{On-board image captured from UAV during crop height estimation flight. Data captured from 3D LiDAR is used to detect crop height changes in the wheat. Our technique also finds the plots and specifically estimates the height of the crops within a plot. These experiments were conducted at Virginia Tech's Kentland Farm.}
    \label{fig:wheat_rgb}
\end{figure}

Next, we used the same UAV and 3D LiDAR configuration to study the progression of growth within a pasture. Our fieldwork took place at Virginia Tech's Turfgrass Research Center, where we demarcated a specific plot of land, maintaining its boundaries while periodically mowing the grass. Over four weeks, we conducted UAV flights, mapping the plot and manually recording grass height at various points within it. Leveraging the boundary as a reference for ground-level estimation, we determined the plot's height within the 3D LiDAR point cloud map, enabling us to track its growth rate on a week-to-week basis. Our method exhibited a commendable accuracy level, with a margin of error of $\pm$7.91\% compared to ground truth measurements.

While our initial results showed promise, we recognized the need for further validation. Given the nature of these experiments, developing a growth model proved advantageous, as it facilitated the creation of diverse datasets for comprehensive testing. To this end, we adapted the simulation framework initially employed for crop height estimation. We designed a pasture generation model, which found application in subsequent research published in the \textbf{IEEE Transactions on Automation Science and Engineering (T-ASE 2022)}~\cite{Liu2022} and the \textbf{Journal of Agronomy (Agronomy 2021)}~\cite{Rangwala2021}.

\begin{figure}[ht!]
    \centering
    \includegraphics[width=.75\linewidth]{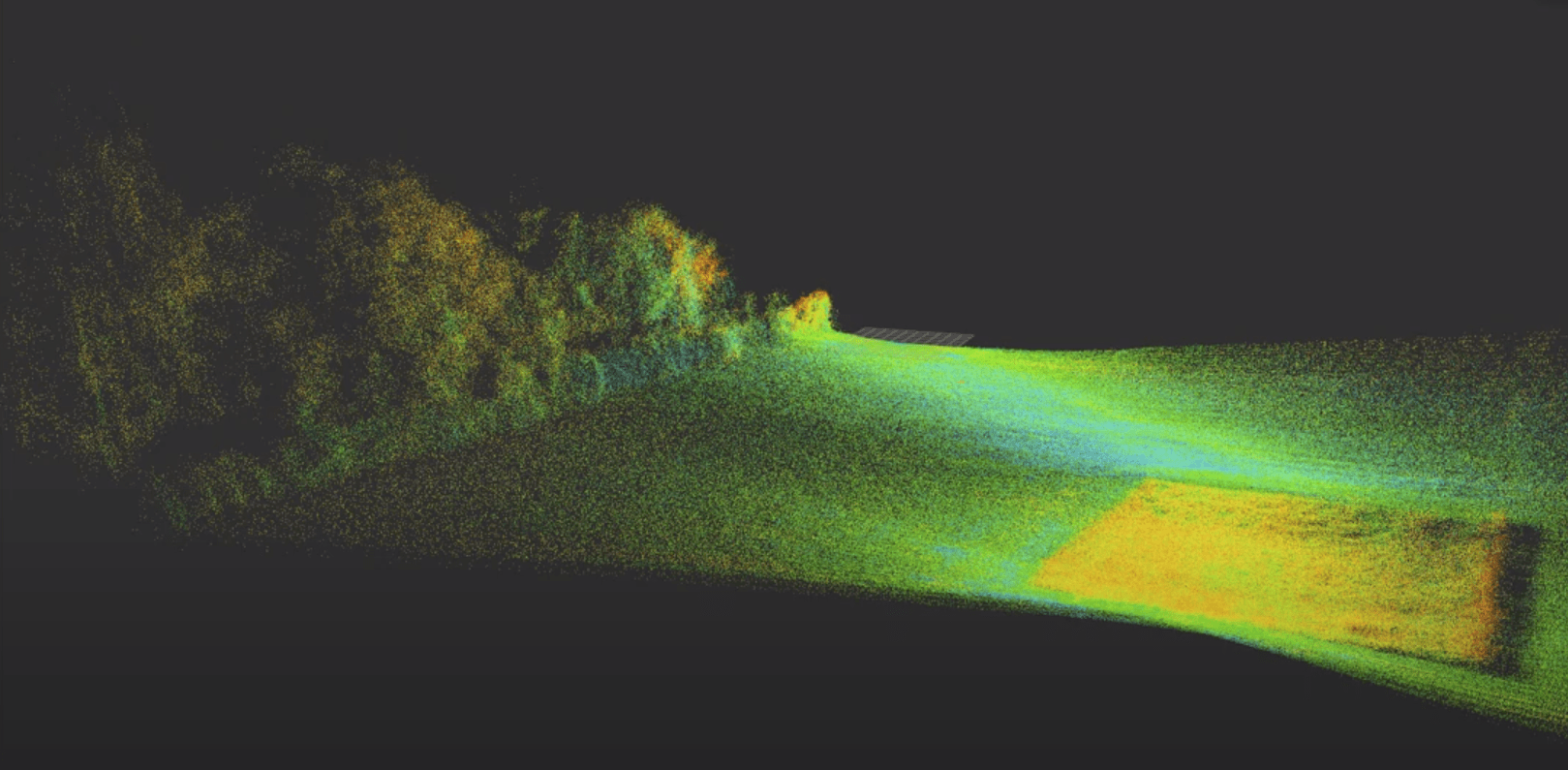}
    \caption{3D LiDAR data captured during flight used to build a map of the environment. This is used to detect growth changes in the plot on the bottom right of the image. These experiments were conducted at Virginia Tech's Turfgrass Research Center.}
    \label{fig:turfgrass}
\end{figure}

\subsection{Environmental Monitoring: Change Validation}
\begin{figure}
    \centering
    \begin{subfigure}[b]{.49\columnwidth}%
        \centering
        \includegraphics[height = 6cm]{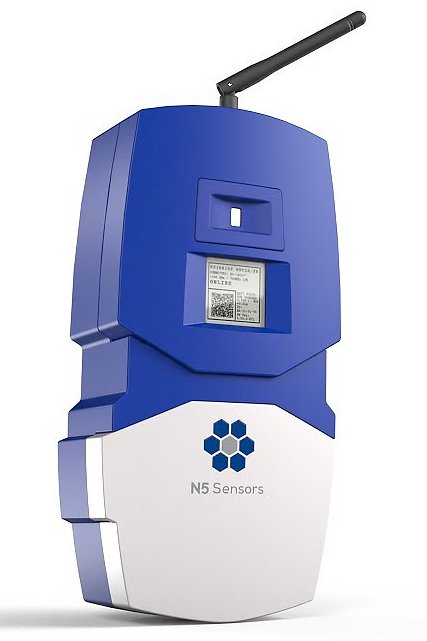}
        \caption{N5 Sensors~\cite{N5Sensors_2023} ChemNode: These sensor stations are used to detect environmental changes signaling the potential presence of a wildfire.}
        \label{fig:chemnode}
    \end{subfigure}%
    \hfill%
    \begin{subfigure}[b]{.49\columnwidth}%
        \centering
        \includegraphics[height = 4.5cm]{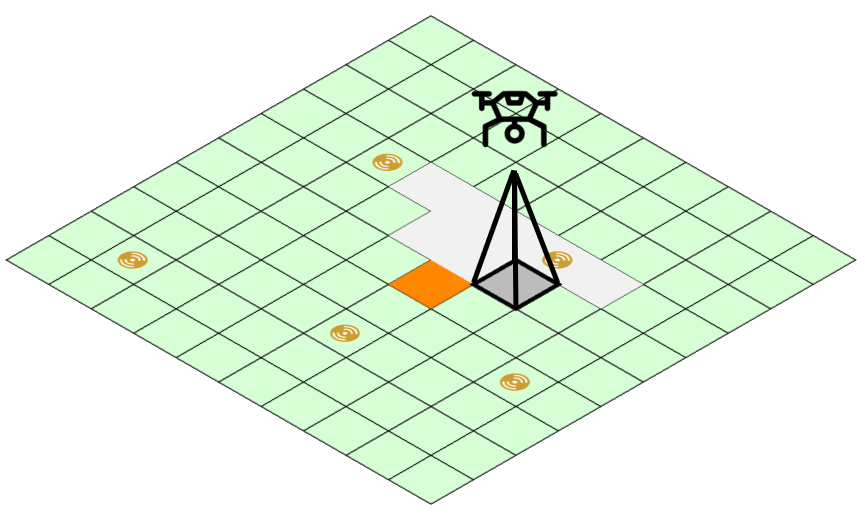}
        \caption{Once a wildfire is detected, the UAV uses ~\textit{Fire-GIPP}, an informative path planner for fast fire validation and localization. Due to false positives with wildfire detection, change validation is needed to confirm the presence of a wildfire.}
        \label{fig:overview_GIPP}
    \end{subfigure}%
    \caption{\textit{Fire-GIPP} is used to validate and locate changes that are detected by sensor stations indicating wildfires.}
    \label{fig:fire_gipp}
\end{figure}

Once changes have been detected in a fast-changing environment, there is a need to navigate the environment for fast validation. Due to complex environmental dynamics, many observations are noisy. This can lead to false positives where changes are detected but no underlying change occurs which is why validation is needed. We study this problem in the context of wildfire management. We present our work in informative path planning for improved wildfire validation and localization. In current work, we utilize datasets from cutting-edge multi-modality sensor systems~\cite{N5Sensors_2023}, shown in Figure~\ref{fig:chemnode}, for early fire detection. These sensors incorporate a comprehensive suite of particulate matter, gas, infrared (IR), carbon dioxide (CO2), temperature, and humidity sensors. These multi-modality sensors allow us to create a holistic understanding of the environmental conditions and potential fire risks. The data from these sensors is used to train a Long Short-Term Memory (LSTM), a type of Recurrent Neural Network (RNN) model that allows us to leverage both spatial and temporal data for wildfire detection. By continuously monitoring these parameters and utilizing these models, abnormal patterns can be identified that are indicative of a potential wildfire outbreak allowing for early wildfire detection. However, these are highly sensitive sensors and can lead to false positives. Therefore, a need for validation arises.
We assume a scenario where the multi-modality sensors are deployed in the field for early fire detection. Once a potential fire is detected, our planner must focus on accurately validating and localizing the fire's exact position. To address this issue, we have developed an informative path planning algorithm, \textit{Fire-GIPP}, that utilizes a probabilistic model for fire validation and localization. This algorithm optimizes the deployment of a UAV for validation and localization, focusing on areas with the highest likelihood of fire occurrence. In contrast, traditional baseline methods aim for complete coverage of the search area, which can be inefficient and time-consuming. \textit{Fire-GIPP} has demonstrated remarkable improvements over the baseline approach. It is 2.2x more effective at fire validation and 3.1x more effective at localization, significantly increasing the accuracy of identifying and pinpointing wildfire locations. Moreover, \textit{Fire-GIPP} achieves this while being 16x computationally more efficient. The research on realistic wildfire monitoring and informative path planning for fire validation and localization is currently under review at a conference.

\section{Overview of the Dissertation}

This dissertation is organized into 6 chapters, following this chapter.

In Chapters~\ref{chap:prednbv} and~\ref{chap:mapnbv}, we present our work on visual reconstruction. In Chapter~\ref{chap:prednbv}, we present \textit{Pred-NBV}\footnote{\url{https://raaslab.org/projects/PredNBV/}}, our prediction based next-best view planner for 3D object reconstruction. Chapter~\ref{chap:mapnbv} extends this to work with multiple agents in a decentralized manner\footnote{\url{https://raaslab.org/projects/MAPNBV/}}.

In Chapter~\ref{chap:insp}, we present our visual inspection planner ~\textit{GATSBI}\footnote{\url{https://raaslab.org/projects/GATSBI/}}. \textit{GATSBI} targets complete surface inspection of a specific object in the environment while avoiding obstacle collisions and having minimal prior information about the target object. 

Next, we shift focus from object monitoring where a specific object in the environment is targeted to environmental monitoring. Chapter~\ref{chap:agri} presents our work on change detection where we use observations to detect environmental changes that have occurred in the context of precision agriculture\footnote{\url{https://github.com/hsd1121/PointCloudProcessing}}. In environments where changes happen fast, high noise levels lead to the need for change validation. This is studied in Chapter~\ref{chap:fire} in the context of wildfire validation and localization~\footnote{\url{https://github.com/raaslab/Fire-GIPP}}.

Lastly, in Chapter~\ref{chap:con}, we conclude the dissertation by discussing our contributions and future research opportunities. 

\renewcommand{\thechapter}{2}

\chapter{Object Monitoring: Visual Reconstruction}\label{chap:prednbv}

\section{Introduction}

This chapter focuses on the visual reconstruction problem of object monitoring. As stated before, visual reconstruction is the creation of 3D digital models of objects of interest. In this chapter, we use a mobile robot to improve the efficiency of mapping and reconstructing an object of interest. This is a long-studied and fundamental problem in the field of robotics~\cite{bajcsy2018revisiting}. In particular, the commonly used approach is Next-Best-View (NBV) planning. In NBV planning, the robot seeks to find the best location to go to next and obtain sensory information that will aid in reconstructing the object of interest. Several approaches for NBV planning have been proposed over the years~\cite{delmerico2018comparison}. In this chapter, we show how to leverage the recent improvements in perception due to deep learning to improve the efficiency of 3D object reconstruction with NBV planning. In particular, we present a 3D shape prediction technique that can predict a full 3D model based on the partial views of the object seen so far by the robot to find the NBV. Notably, our method works ``in the wild'' by eschewing some common assumptions made in 3D shape prediction, namely, assuming that the partial views are still centered at the full object center.

There are several applications where robots are being used for visual data collection. Some examples include inspection for visual defect identification of civil infrastructure such as bridges~\cite{shanthakumar2018view,dhamiGATSBI}, ship hulls~\cite{kim_2009_IROS} and aeroplanes~\cite{ropek_2021}, digital mapping for real estate~\cite{46965,ramakrishnan2021hm3d}, and precision agriculture~\cite{dhami2020crop}. The key reasons why robots are used in such applications are that they can reach regions that are not easily accessible to humans and we can precisely control where the images are taken from. However, existing practices for the most part require humans to specify a nominal trajectory for the robots that will visually cover the object of interest. Our goal in this chapter is to automate this process.  

\begin{figure}[ht!]
    \centering
    \includegraphics[width=\linewidth]{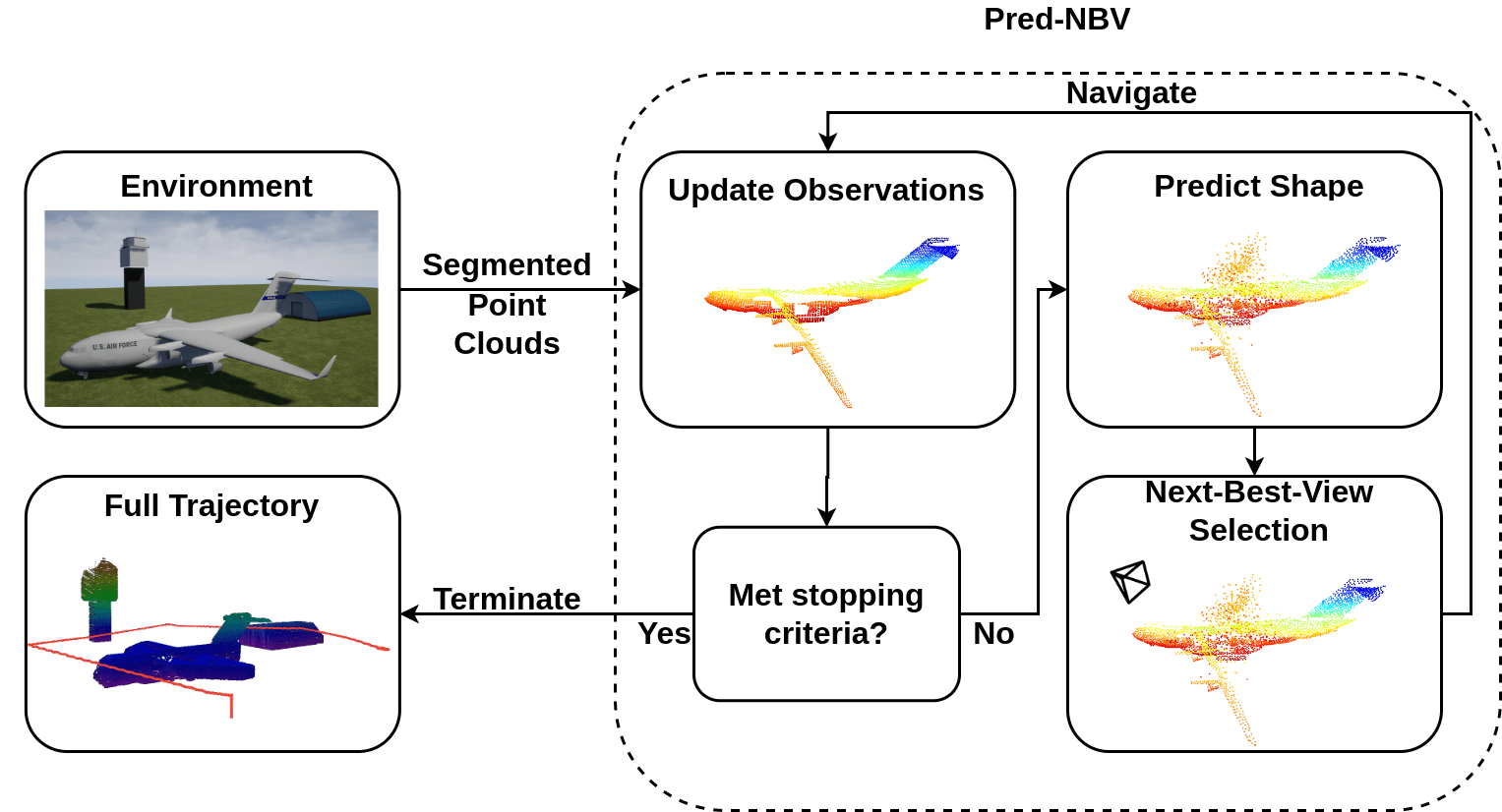}
    \caption{Overview of the proposed approach}
    \label{fig:overview}
\end{figure}

The NBV planning method is the commonly used approach to autonomously decide where to obtain the next measurement from. NBV planning typically uses geometric cues such as symmetry~\cite{debevec1996modeling} or prior information~\cite{breyer2022closed} for deciding the next best location. In this chapter, we do not rely on such assumptions but instead leverage the predictive power of deep neural networks for 3D shape reconstruction.

Recent works have explored predictions as a way of improving these systems by anticipating the unknowns with prediction and guiding robots' motion accordingly. This approach has been studied for robot navigation, exploration, and manipulation~\cite{ramakrishnan2020occupancy,georgakis2022uncertainty,Wei2021,yan2019data} with the help of neural network-based methods that learn from datasets. While 2D map representation works have shown the benefits of robotic tasks in simulation as well as the real world, similar methods for 3D prediction have been limited to simplistic simulations. The latter approaches generally rely on synthetic datasets due to the lack of realistic counterparts for learning.

Strong reliance on data results in the neural network learning implicit biases. Some models can make predictions only for specific objects~\cite{3drecgan}. Others require implicit knowledge of the object's center, despite being partially visible~\cite{yu2021pointr}. These situations are invalid in real world, mapless scenarios and may result in inaccurate shape estimation. Many shape prediction works assume the effortless motion of the robot~\cite{wu20153d}, whereas an optimal path for a robot should include the effort required to reach a position and the potential information gain due to time and power constraints. Monolithic neural networks replacing both the perception and planning components present an alternative, but they tend to be specific to the datasets used for training and may require extensive finetuning for real world deployment. Lack of transparency in such networks poses another challenge that could be critical for the safe operation of a robot when working alongside humans.

To make 3D predictive planning more realistic, efficient, and safe, we propose a method consisting of a 3D point cloud completion model, that relaxes the assumption about implicit knowledge of the object's center using a curriculum learning framework~\cite{bengio2009curriculum}, and an NBV framework, that maximizes the information gain from image rendering and minimizes the distance traveled by the robot. Furthermore, our approach is modular making it interpretable and easy to upgrade.

We make the following contributions in this chapter:
\begin{itemize}
    \item We use curriculum learning to build an improved 3D point cloud completion model, which does not require the partial point cloud to be centered at the full point cloud's center and is more robust to perturbations than earlier models. We show that this model, termed \textit{PoinTr-C}, outperforms the base model, PoinTr~\cite{yu2021pointr}, by at least $\textbf{23.06}\%$ and show qualitative comparison on the ShapeNet~\cite{shapenet2015} dataset, and a real point cloud obtained with a Velodyne 3D LiDAR mounted on DJI M600 Pro.
    \item We propose a next-best-view planning approach that performs object reconstruction without any prior information about the geometry, using predictions to optimize information gain and control effort over a range of objects in a model-agnostic fashion.
    \item We show that our method covers on average \textbf{25.46\%} more points on all models evaluated for object reconstruction in AirSim~\cite{airsim2017fsr} simulations compared to the non-predictive baseline approach, \textit{Basic-Next-Best-View}~\cite{aleotti2014global}, and performs even better for complex structures such as airplanes.
\end{itemize}

We share the qualitative results, project code, and visualization from our method on our project website\footnote{Project webpage: \url{http://raaslab.org/projects/PredNBV/}}.

\section{Related Work}\label{sec:rel_work}

~Active reconstruction in an unknown environment can be accomplished through next-best-view (NBV) planning, which has been studied by the robotics community for a long time~\cite{scott2003view}. In this approach, the robot builds a partial model of the environment based on observations and then moves to a new location to maximize the cumulative information gained. The NBV approaches can be broadly classified into information-theoretic and geometric methods. The former builds a probabilistic occupancy map from the observations and uses the information-theoretic measure~\cite{delmerico2018comparison} to select the NBV. The latter assumes the partial information to be exact and determines the NBV based on geometric measures~\cite{tarabanis1995survey}.

The existing works on NBV with robots focus heavily on information-theoretic approaches for exploration in 2D and 3D environments~\cite{kuipers1991robot,vasquez2014volumetric}. Subsequent development for NBV with frontier and tree-based approaches was also designed for exploration by moving the robot towards unknown regions~\cite{yamauchi1997frontier, gonzalez2002navigation, adler2014autonomous, bircher2018receding}. Prior works on NBV for object reconstruction also rely heavily on information-theoretic approaches to reduce uncertainty in pre-defined closed spaces~\cite{morooka1998next, vasquez2009view}. Geometric approaches require knowing the model of the object in some form and thus have not been explored to a similar extent.
Such existing works try to infer the object geometry from a database or as an unknown closed shape~\cite{banta2000next, kriegel2013combining} and thus may be limited in application.

In recent years, prediction-based approaches have emerged as another solution. One body of these approaches works in conjunction with other exploration techniques to improve exploration efficiency by learning to predict structures in the environment from a partial observation. This is accomplished by learning the common structures in the environment (buildings and furniture, for example) from large datasets. This approach has recently gained traction and has been shown to work well for mobile robot navigation with 2D occupancy map representations~\cite{ramakrishnan2020occupancy, Wei2021, sharma2023proxmap}, exploration~\cite{georgakis2022uncertainty}, high-speed maneuvers~\cite{Katyal2021}, and elevation mapping~\cite{yang2022real}.

Similar works on 3D representations have focused mainly on prediction modules. Works along this line have proposed generating 3D models from novel views using single RGB image input~\cite{Hani2020}, depth images~\cite{Yang_2019}, normalized digital surface models (nDSM)~\cite{Alidoost2019}, point clouds~\cite{yu2021pointr, xie2020grnet, yuan2018pcn}, etc. The focus of these works is solely on inferring shapes based on huge datasets of 3D point clouds~\cite{shapenet2015}. They do not discuss the downstream task of planning. A key gap in these works is that they assume a canonical representation of the object, such as the center of the whole object, to be provided either explicitly or implicitly. Relaxing this assumption does not work well in the real world where the center of the object may not be estimated accurately, discouraging the adoption of 3D prediction models for prediction-driven planning.

Another school of work using 3D predictions combines the perception and planning modules as a neural network. These works, aimed at predicting the NBV to guide the robot from partial observations, were developed for simple objects~\cite{POP2022160}, 3D house models~\cite{Peralta2020}, and a variety of objects~\cite{zeng2020pc} ranging from remotes to rockets. Peralta et al.~\cite{Peralta2020} propose a reinforcement-learning framework, which can be difficult to implement due to sampling complexity issues. The supervised-learning approach proposed by Zeng et al.~\cite{zeng2020pc} predicts the NBV using a partial point cloud, but the candidate locations must lie on a sphere around the object, restricting the robot's planning space. Moreover, monolithic neural networks suffer from a lack of transparency and real world deployment may require extensive fine-tuning of the hyperparameters. Prediction-based modular approaches solve these problems as the intermediate outputs are available for interpretation and the prediction model can be plugged in with the preferred planning method for a real environment.

A significant contribution of our method is to relax the implicit assumption used in many works that the center and the canonical orientation of the object under consideration are known beforehand, even if the 3D shape completion framework uses partial information as the input. A realistic inspection system may not know this information and thus the existing works may not be practically deployable.

\section{Problem Formulation}\label{sec:prob_form}

We are given a robot with a 3D sensor onboard that explores a closed object with volume $\mathcal{V} \in \mathbb{R}^3$. The set of points on the surface of the object is denoted by $\mathcal{S} \in \mathbb{R}^3$. The robot can move in free space around the object and observe its surface. The surface of the object $s_i \subset \mathcal{S}$ perceived by the 3D sensor from the pose $\phi_i \subset \Phi$ 

is represented as a voxel-filtered point cloud. We define the relationship between the set of points observed from a viewpoint $\phi_i$ with a function $f$, i.e., $s_i = f(\phi_i)$. The robot can traverse  a trajectory $\xi$ that consists of viewpoints $\{\phi_1, \phi_2, \ldots, \phi_m \}$. The surface observed over a trajectory is the union of surface points observed from the consisting viewpoints, i.e. $s_{\xi} = \bigcup_{\phi \in \xi} f(\phi)$. The distance traversed by the robot between two viewpoints $\phi_i$ and $\phi_j$ is denoted by $d(\phi_i, \phi_j)$.

Our objective is to find a trajectory $\xi_i$ from the set of all possible trajectories $\Xi$, such that it observes the whole voxel-filtered surface of the object while minimizing the distance traversed.  
\begin{align}
    \xi^* = \argmin_{\xi \in \Xi} \sum_{i=1}^{| {\xi}| - 1} d(\phi_i, \phi_{i+1}),~ 
    \textit{such that} \bigcup_{\phi_i \in \xi} f(\phi_i) = \mathcal{S}.
\end{align}

In unseen environments, $\mathcal{S}$ is not known apriori, hence the optimal trajectory can not be determined. We assume that the robot starts with a view of the object. If not, we can always first explore the environment until the object of interest becomes visible. 

\section{Proposed Approach}\label{sec:approach}
We propose \textit{Pred-NBV}, a prediction-guided NBV method for 3D object reconstruction highlighted in Fig.~\ref{fig:overview}. Our method consists of two key modules: (1) \textit{PoinTr-C}, a robust 3D prediction model that completes the point cloud using only partial observations, and (2) an NBV framework that uses prediction-based information gain to reduce the control effort for active object reconstruction. We provide the details in the following subsection.

\vspace{-1mm}
\subsection{\textit{PoinTr-C}: 3D Shape Completion Network}
\vspace{-0.5mm}
Given the current set of observations $v_o \in \mathcal{V}$, we predict the complete volume using a learning-based predictor $g$, i.e., $\hat{\mathcal{V}} = g(v_o)$.

To obtain $\hat{\mathcal{V}}$, we use PoinTr~\cite{yu2021pointr}, a transformer-based architecture that uses 3D point clouds as the input and output. 

PoinTr uses multiple types of machine learning methods to perform shape completion. It first identifies the geometric relationship in low resolution between points in the cloud by clustering. Then it generates features around the cluster centers, which are then fed to a transformer~\cite{vaswani2017attention} to capture the long-range relationships and predict the centers for the missing point cloud. Finally, a coarse-to-fine transformation over the predicted centers using a neural network outputs the missing point cloud. 

This model was trained on the ShapeNet~\cite{shapenet2015} dataset and outperforms the previous methods on a range of objects. However, PoinTr was trained with implicit knowledge of the center of the object. Moving the partially observed point cloud to its center results in incorrect prediction from PoinTr.

To improve predictions, we fine-tune PoinTr using a curriculum framework, which dictates training the network over easy to hard tasks by increasing the difficulty in steps during learning~\cite{bengio2009curriculum}. Specifically, we fine-tune PoinTr over increasing perturbations in rotation and translation to the canonical representation of the object to relax the assumption about implicit knowledge of the object's center. We use successive rotation-translation pairs of $(25^\circ, 0.0)$, $(25^\circ, 0.1)$, $(45^\circ, 0.1)$, $(45^\circ, 0.25)$, $(45^\circ, 0.5)$, $(90^\circ, 0.5)$, $(180^\circ, 0.5)$, and $(360^\circ, 0.5)$  for curriculum training.
We assume that the object point cloud is segmented well, which can be achieved using distance-based filters or segmentation networks.

\vspace{-1mm}
\subsection{Next-Best View Planner}
\vspace{-0.5mm}
Given the predicted point cloud $\mathcal{\hat{V}}$ for the robot after traversing the trajectory $\xi_{t}$, we generate a set of candidate poses $\mathcal{C} = \{ \phi_1, \phi_2, ..., \phi_m \}$ around the object observed so far. Given $v_o$, the observations so far, we define the objective to select the shortest path that results in observing at least $\tau \%$ of the maximum possible information gain over all the candidate poses. Considering $\mathcal{\hat{V}}$ as an exact model, we use a geometric measure to quantify the information gained from the candidate poses. Specifically, we define a projection function $I(\xi)$, over the trajectory $\xi$, which first identifies the predicted points distinct from the observed point cloud over the trajectory, then apply a hidden point removal operator on them~\cite{katz2007direct}, without reconstructing a surface or estimating a normal, and lastly, find the number of points that will be observed if we render an image on the robot's camera. Thus, we find the NBV from the candidate set $\mathcal{C}$ as follows:
\begin{align*}
    \phi_{t+1} = \argmin_{\phi \in \mathcal{C}} d(\phi, \phi_{t}),
    ~\textit{such that}~ \frac{I(\xi_t \cup \phi)}{\max_{\phi \in \mathcal{C}} I(\xi_t \cup \phi)} \geq \tau.
\end{align*}

We find the $d(\phi_i, \phi_j)$, using RRT-Connect~\cite{kuffner2000rrt}, which incrementally builds two rapidly-exploring random trees rooted at $\phi_i$ and $\phi_j$ through the observed space to provide a safe trajectory. After selecting the NBV, the robot follows this trajectory to reach the prescribed viewpoint. We repeat the prediction and planning process until the ratio of observations in the previous step and the current step is $0.95$ or higher.

To generate the candidate set $\mathcal{C}$, we first find the distance $d_{max}$ of the point farthest from the center of the predicted point cloud $\mathcal{\hat{V}}$ and z-range. Then, we generate candidate poses on three concentric circles: one centered at $\mathcal{\hat{V}}$ with radius $1.5 \times d_{max}$ at steps of $30^\circ$, and one $0.25 \times \text{z-range}$ above and below with radius $1.2 \times d_{max}$ at steps of $30^\circ$. We use $\tau = 0.95$ for all our experiments.

\section{Evaluation}\label{sec:eval}
In this section, we evaluate the \textit{Pred-NBV} pipeline. We start with a qualitative example followed by a comparison of the individual modules against respective baseline methods. The results show that \textit{Pred-NBV} can outperform the baselines significantly using large-scale models from the Shapenet~\cite{shapenet2015} dataset and with real world 3D LiDAR data.

\subsection{Qualitative Example}
\begin{figure}[ht!]
    \centering
    \begin{subfigure}[b]{\columnwidth}%
        \includegraphics[width = \textwidth]{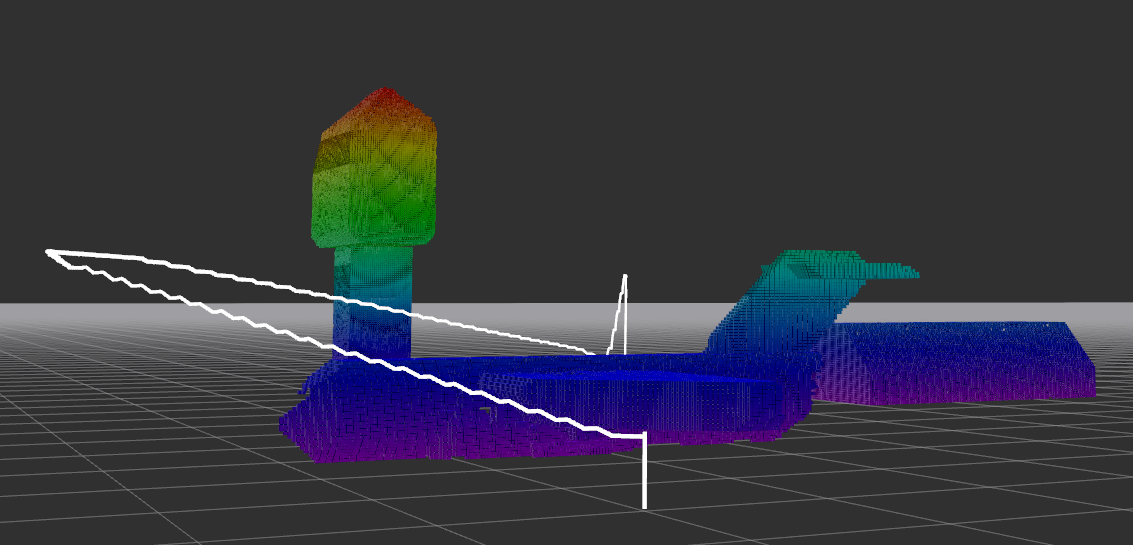}%
    \end{subfigure}%
    \hfill%
    \begin{subfigure}[b]{\columnwidth}%
        \includegraphics[width = \textwidth]{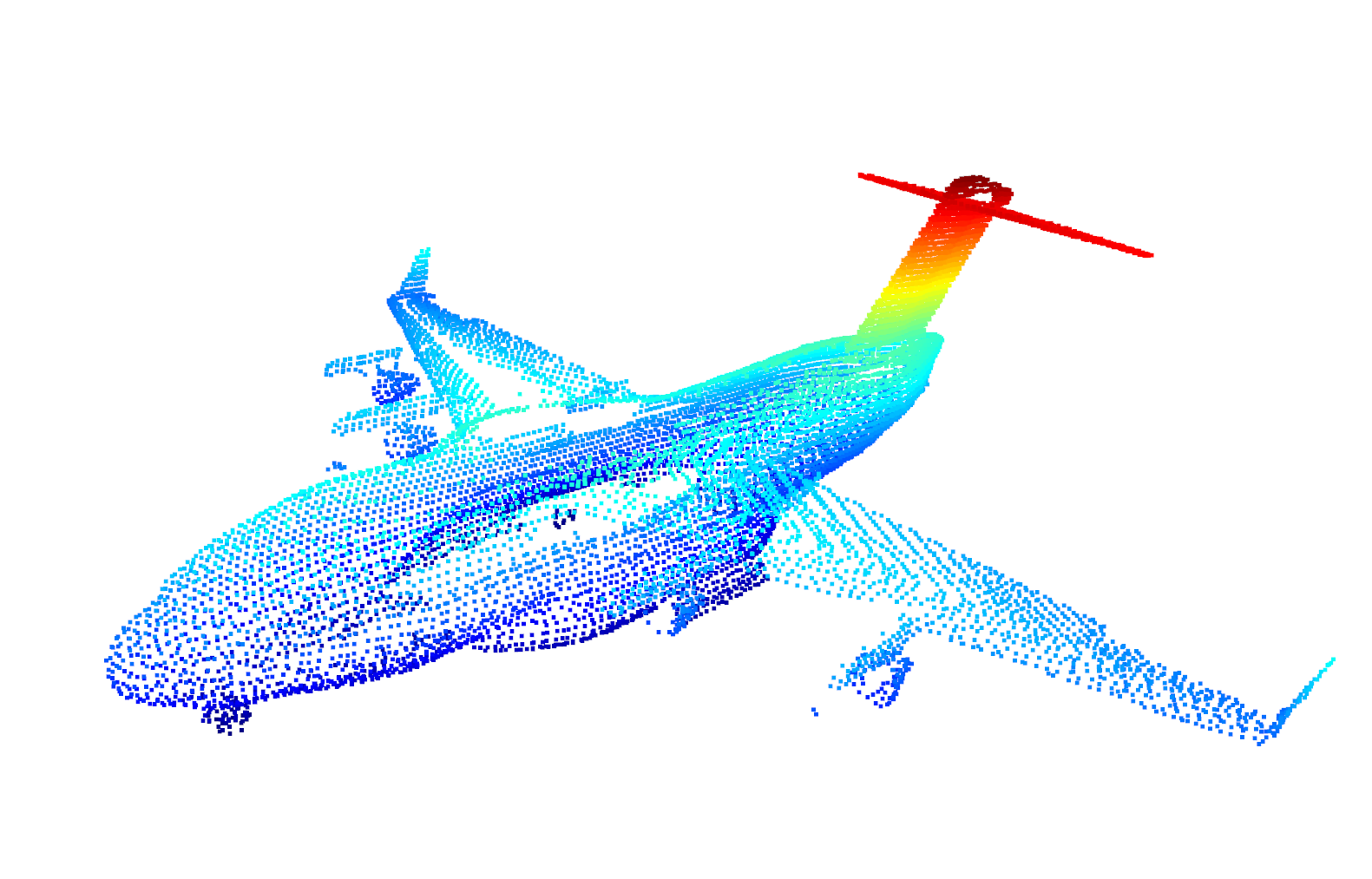}%
    \end{subfigure}%
    \caption{Flight path and total observations of C17 Airplane after running our NBV planner in AirSim simulation.}
    \label{fig:observations_airsim}
    \vspace{-5mm}
\end{figure}

\vspace{-0.5mm}
 Fig.~\ref{fig:observations_airsim} shows the reconstructed point cloud of a C17 airplane and the path followed by a UAV in AirSim~\cite{airsim2017fsr}. We create candidate poses on three concentric rings at different heights around the center of the partially observed point cloud. The candidate poses a change as more of the object is visible. 

 As shown in Fig.~\ref{fig:plane_res}, \textit{Pred-NBV} observes more points than the NBV planner without prediction in the same time budget. 

\vspace{-1mm}
\subsection{3D Shape Prediction}
\vspace{-0.5mm}
\subsubsection{Setup}

We train \textit{PoinTr-C} on a 32-core, 2.10Ghz Xeon Silver-4208 CPU and Nvidia GeForce RTX 2080Ti GPU with 11GB of memory. The network is fine-tuned over the ShapeNet\cite{shapenet2015} dataset, trained with perturbation as described in Section~\ref{sec:approach}. Similar to PoinTr~\cite{yu2021pointr}, we use Chamfer distance (CD) and Earth Mover's Distance (EMD), permutation-invariant metrics suggested by Fan et al.~\cite{fan2017point}, as the loss function for training \textit{PoinTr-C}. For evaluation we use two versions of Chamfer distance: CD-$l_1$ and CD-$l_2$, which use L1 and L2-norm, respectively, to calculate the distance between two sets of points, and F-score which quantifies the percentage of points reconstructed correctly.

\subsubsection{Results}
Table~\ref{tab:base_vs_best_mean} summarizes our findings regarding the effect of perturbations. \textit{PoinTr-C} outperforms the baseline in both scenarios. It only falters in CD-$l_2$ in the ideal condition, i.e., no augmentation. Furthermore, \textit{PoinTr-C} doesn't undergo large changes in the presence of augmentations, making it more robust than PoinTr. The relative improvement for \textit{PoinTr-C} is at least $23.05\%$ (F-Score).

\begin{table}[ht]
    \centering
    \vspace{-2.5mm}
    \caption{Comparison between the baseline model (PoinTr) and \textit{PoinTr-C} over test data with and without perturbation. Arrows show if a higher ($\uparrow$) or a lower ($\downarrow$) value is better.}
    \begin{tabular}{llrrr}
        \toprule
        Perturbation  & Approach  &  F-Score $\uparrow$   &  CD-$l_1$ $\downarrow$  & CD-$l_2$ $\downarrow$\\
        \midrule
        \multirow{2}*{\xmark} & PoinTr\cite{yu2021pointr} & 0.497 & 11.621 & \textbf{0.577}\\
        {} & \textit{PoinTr-C} & \textbf{0.550} & \textbf{10.024} & 0.651\\
        \midrule
        \multirow{2}*{\cmark} & PoinTr\cite{yu2021pointr} & 0.436 & 16.464 & 1.717\\
        {} & \textit{PoinTr-C} & \textbf{0.550} & \textbf{10.236} & \textbf{0.717}\\
        \bottomrule
        \end{tabular}
        \vspace{-2.5mm}
        \label{tab:base_vs_best_mean}
\end{table}

\begin{figure}
    \centering
    \vspace{0mm}
    \begin{subfigure}[b]{0.4\columnwidth}%
        \includegraphics[width=\linewidth]{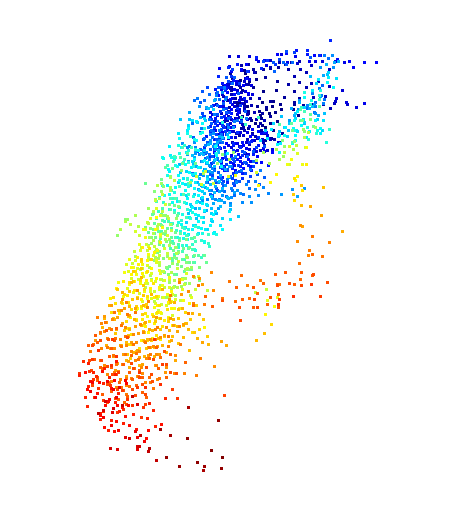}
        \subcaption{Input}
    \end{subfigure}%
    \hfill%
    \begin{subfigure}[b]{0.4\columnwidth}%
        \includegraphics[width=\linewidth]{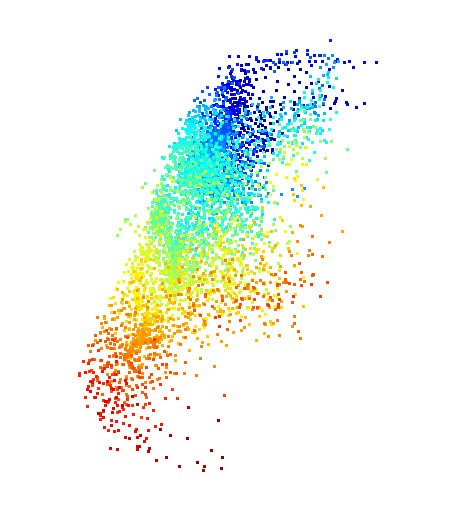}
        \subcaption{PoinTr~\cite{yu2021pointr}}
    \end{subfigure}%
    \hfill%
    \begin{subfigure}[b]{0.4\columnwidth}%
        \includegraphics[width=\linewidth]{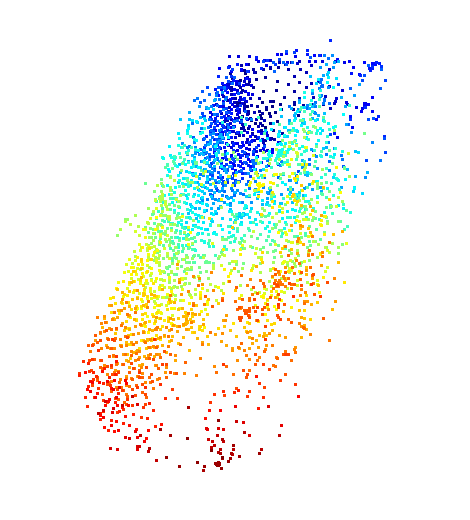}
        \subcaption{PoinTr-C}
    \end{subfigure}%
    \caption{Results over the real world point cloud of a car obtained using LiDAR (Interactive figure available on \href{http://raaslab.org/projects/PredNBV/}{our webpage}).}
    \label{fig:qualitative_real_colored}
\end{figure}

We provide a qualitative comparison of the predictions from the two models for various objects under perturbations on our \href{http://raaslab.org/projects/prednbv/}{project webpage}. Fig.~\ref{fig:qualitative_real_colored} shows the results for a real point cloud of a car (visualized from the camera above the left headlight) obtained with a Velodyne LiDAR sensor. The results show the predictions from PoinTr are scattered around the center of the visible point cloud, whereas \textit{PoinTr-C} makes more realistic predictions. Our webpage provides interactive visualizations of these point 
clouds for further inspection.

\vspace{-1mm}
\subsection{Next-Best-View Planning}
\vspace{-0.5mm}
\subsubsection{Setup}

We use Robot Operating System (ROS) Melodic and AirSim~\cite{airsim2017fsr} on Ubuntu 18.04 to carry out the simulations. We equipped the virtual UAV with a RGBD camera. AirSim's built-in image segmentation is used to segment out the target object from the rest of the environment. We created a ROS package to publish a segmented depth image containing pixels only belonging to the bridge based on the RGB camera segmentation. This segmented depth image was then converted to a point cloud. We use the MoveIt~\cite{coleman2014reducing} software package based on the work done by Köse~\cite{tahsinko86:online} to implement the RRT connect algorithm. MoveIt uses RRT connect and the environmental 3D occupancy grid to find collision-free paths for point-to-point navigation.

\subsubsection{Qualitative Example}

We evaluate \textit{Pred-NBV} on 20 objects from 5 ShapeNet classes:  airplane, rocket, tower, train, and watercraft. We selected these classes as they represent larger shapes suitable for inspection. Fig.~\ref{fig:overview} shows the path followed by the UAV using \textit{Pred-NBV} for the C-17 airplane simulation. There are non-target obstacles in the environment, such as a hangar and air traffic control tower. \textit{Pred-NBV} finds a collision-free path that selects viewpoints targeting the maximum coverage of the airplane.

\subsubsection{Comparison with Baseline}\label{sec:sim:baseline}
We compare the performance of \textit{Pred-NBV} with a baseline NBV method~\cite{aleotti2014global}. The baseline selects poses based on frontiers in the observed space using occupancy grids. We modified the baseline to improve it for our application and make it comparable to \textit{Pred-NBV}. The modifications include using our segmentation for the occupancy grid so that frontiers are weighted toward the target object. We also set the orientation of the selected poses towards the center of the target object similar to how \textit{Pred-NBV} works. The algorithms had the same stopping criteria as \textit{Pred-NBV}.

\begin{figure}
    \vspace{1.3mm}
    \centering
    \begin{subfigure}[b]{\columnwidth}%
        \hspace{0.8cm}
        \includegraphics[width = .75\textwidth]{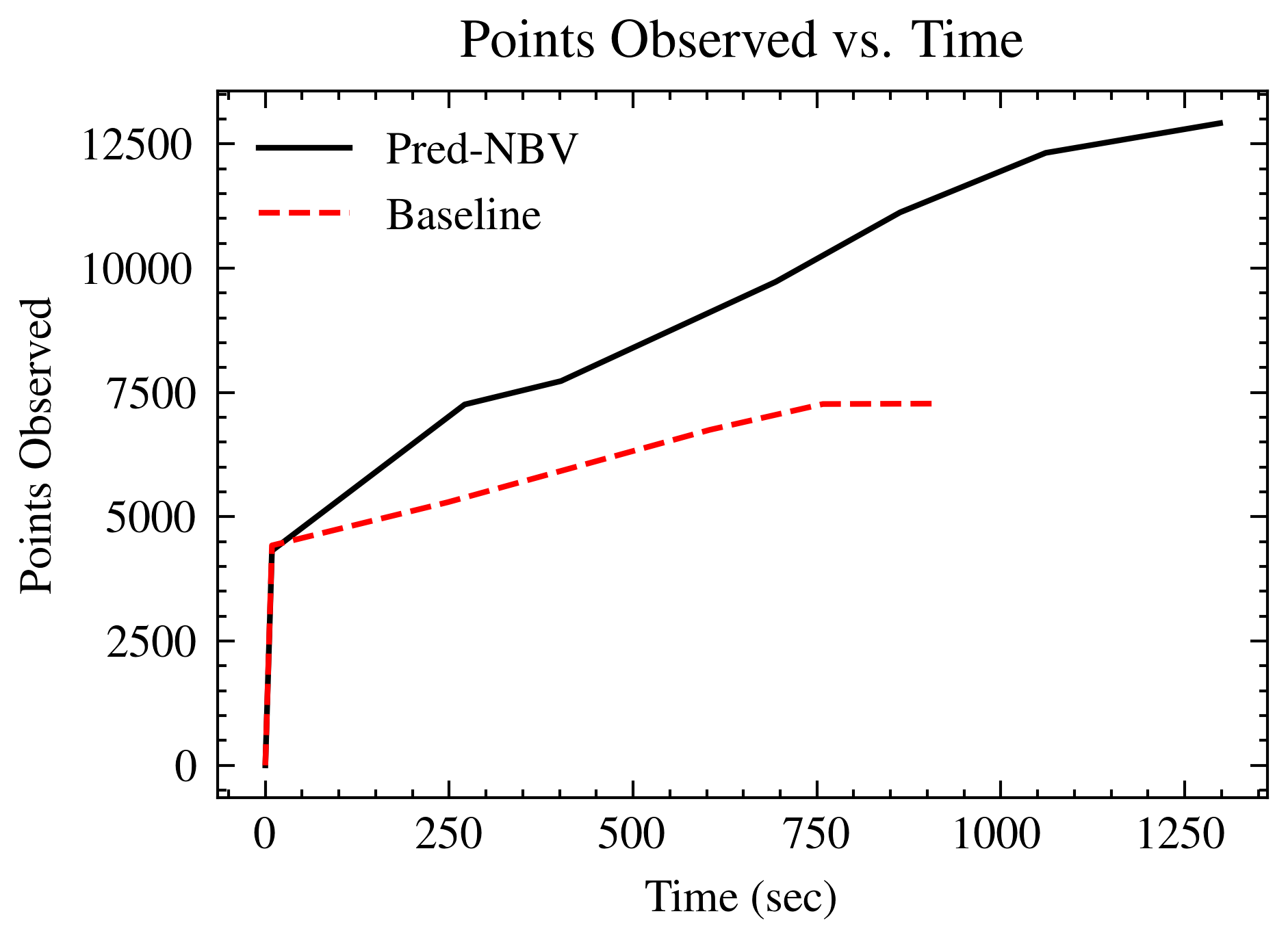}%
    \end{subfigure}%
    \hfill%
    \begin{subfigure}[b]{\columnwidth}%
        \hspace{1.0cm}
        \includegraphics[width = .725\textwidth]{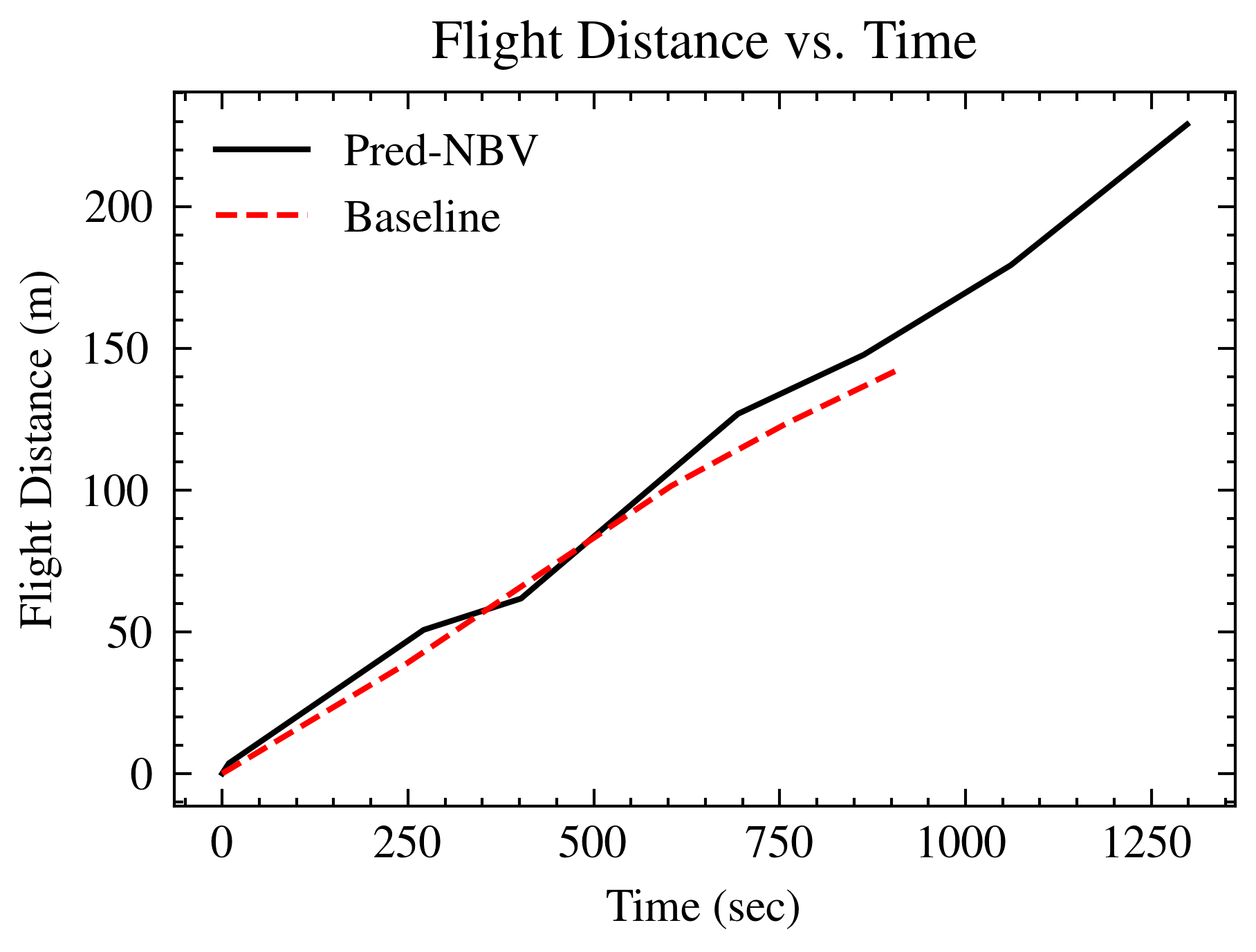}%
    \end{subfigure}%
    \caption{Comparison between \textcolor{black}{\textbf{Pred-NBV}} and the \textcolor{red}{\textbf{baseline NBV algorithm}}~\cite{aleotti2014global} for a C-17 airplane.}
    \label{fig:plane_res}
\end{figure}

We see in Table~\ref{tab:airsim_results} that our method observes on average 25.46\% more points than the baseline for object reconstruction across multiple models from various classes. In Fig.~\ref{fig:plane_res}, we show that \textit{Pred-NBV} observes more points per step than the baseline while not flying further per each step. 

\begin{table}[ht!]
    \centering
    \vspace{1.3mm}
    \caption{Points observed by \textit{Pred-NBV} and the baseline NBV method~\cite{aleotti2014global} for all models in AirSim.}
    \begin{tabular}{llrrr}
    
        \toprule
        \multirow{2}{*}{ Class} & \multirow{2}{*}{Model} & Points Seen & Points Seen & \multirow{2}{*}{Improvement} \\
        & & \textit{Pred-NBV} & Baseline & \\
        \midrule
        \multirow{5}{*}{Airplane}  
            & 747 & \textbf{11922} & 9657 & 20.99\% \\
            & A340 & \textbf{8603} & 5238 &  48.62\% \\
            & C-17 & \textbf{12916} & 7277 & 55.85\% \\
            & C-130 & \textbf{9900} & 7929 & 22.11\% \\
            & Fokker 100 & \textbf{10192} & 9100 & 11.32\%\\
        \midrule
        \multirow{5}{*}{Rocket} 
            & Atlas & \textbf{1822} & 1722 & 5.64\% \\
            & Maverick & \textbf{2873} & 2643 & 8.34\% \\
            & Saturn V & \textbf{1111} & 807 & 31.70\% \\
            & Sparrow & \textbf{1785} & 1639 & 8.53\% \\
            & V2 & \textbf{1264} & 1086 & 15.15\% \\
        \midrule
        \multirow{5}{*}{Tower} 
            & Big Ben & \textbf{4119} & 3340 & 20.89\% \\
            & Church & \textbf{2965} & 2588 & 13.58\% \\
            & Clock & \textbf{2660} & 1947 & 30.95\% \\
            & Pylon & \textbf{3181} & 2479 & 24.80\% \\
            & Silo & \textbf{5674} & 3459 & 48.51\% \\
        \midrule
        \multirow{2}{*}{Train}
            & Diesel & \textbf{3421} & 3161 & 7.90\% \\
            & Mountain & \textbf{4545} & 4222 & 7.37\% \\
        \midrule
        \multirow{3}{*}{Watercraft} 
            & Cruise & \textbf{4733} & 3522 & 29.34\% \\
            & Patrol & \textbf{3957} & 2306 & 52.72\% \\
            & Yacht & \textbf{9499} & 6016 & 44.90\% \\
        \bottomrule
    \end{tabular}
    \label{tab:airsim_results}
\end{table}

\vspace{-5mm}
\section{Conclusion}\label{sec:con}
We propose a realistic and efficient planning approach for robotic inspection using learning-based predictions. Our approach fills the gap between the existing works and the realistic setting by proposing a curriculum-learning-based point cloud prediction model, and a distance and information gain aware inspection planner for efficient operation. Our analysis shows that our approach can outperform the baseline approach in observing the object surface by 25.46\%. Furthermore, we show that our predictive model can provide satisfactory results for real world point cloud data. We believe the modular design paves the path to further improvement by enhancement of the constituents. 

In this chapter, we use noise-free observations but show that \textit{Pred-NBV} has the potential to work well on real, noisy inputs with pre-processing. In future work, we will explore making the prediction network robust to noisy inputs and with implicit filtering capabilities. We used a geometric measure for NBV in this chapter and will extend it to information-theoretic measures using an ensemble of predictions and uncertainty extraction techniques in future work.


\renewcommand{\thechapter}{3}

\chapter{Object Monitoring: Multi-agent Visual Reconstruction}\label{chap:mapnbv}

\section{Introduction}

In this chapter, we extend our object reconstruction method to work with multiple robots. Employing a team of UAVs for this problem is an intuitive solution as multiple UAVs can simultaneously cover multiple viewpoints. NBV planning can be used to move the UAVs, gather new observations, and fill the gaps in the object representation. An efficient reconstruction, however, requires not only a correct estimation of the missing information but also coordination among the UAVs to minimize redundancies in the observations.

\begin{figure}[ht!]
\centering
\begin{subfigure}{.47\columnwidth}
    \includegraphics[width=1.0\linewidth]{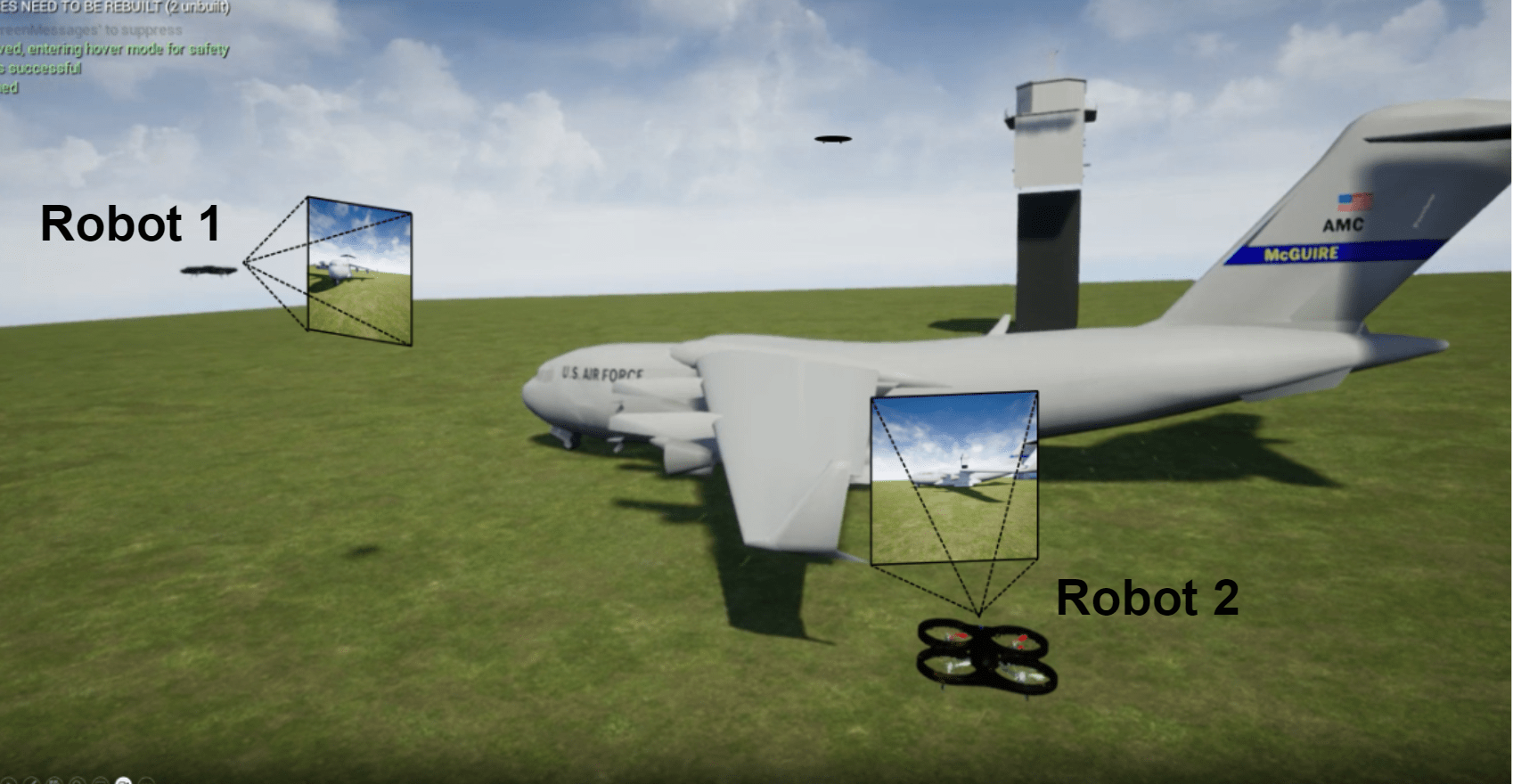}
        \caption{C-17 and the robots in AirSim simulation}
        \label{fig:airsimScreengrab}
\end{subfigure}%
\hfill
\begin{subfigure}{.47\columnwidth}
        \includegraphics[width=0.7\linewidth]{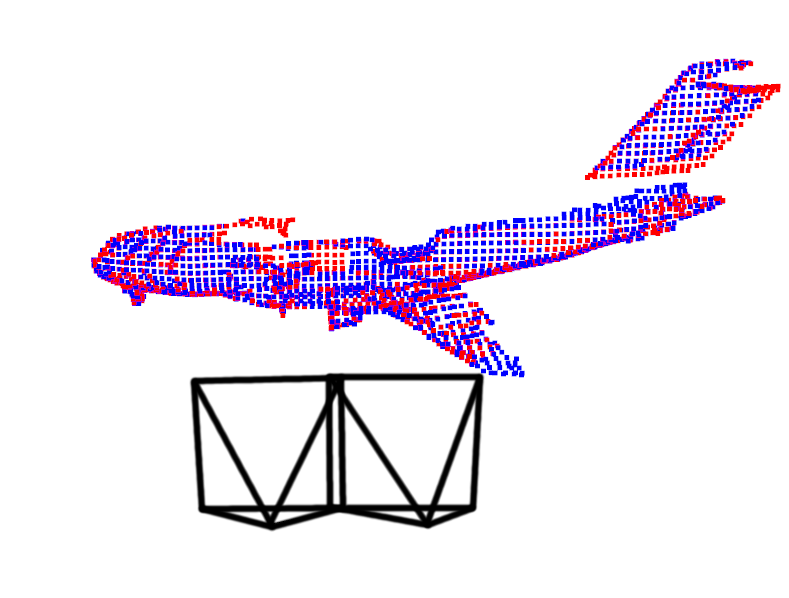}
        \caption{Initial observations by \textcolor{red}{robot 1} and \textcolor{blue}{robot 2}}
        \label{fig:c17Observed}
\end{subfigure}
\medskip
\begin{subfigure}{.47\columnwidth}
        \includegraphics[width=1.0\linewidth]{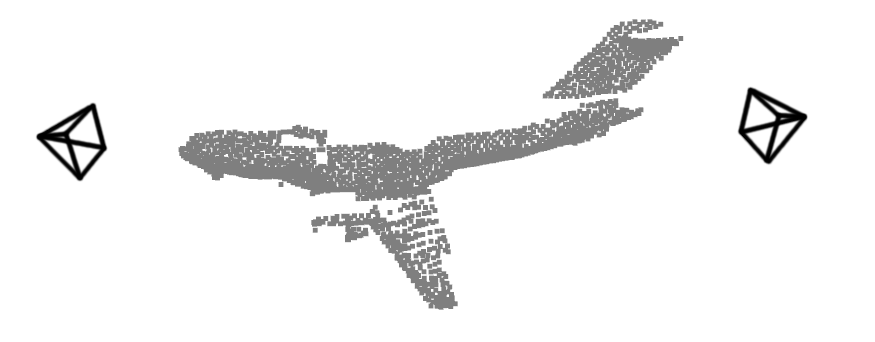}
        \caption{Poses selected by frontiers baseline based on \textcolor{gray}{observations}}
        \label{fig:c17BaselineNBV}
\end{subfigure}%
\hfill
\begin{subfigure}{.47\columnwidth}
        \includegraphics[width=0.97\linewidth]{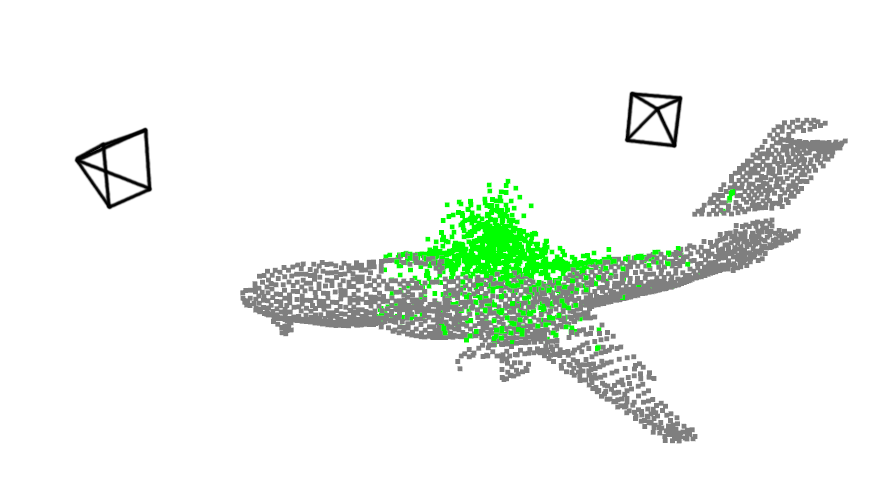}
        \caption{Poses selected by \textit{MAP-NBV} based on \textcolor{gray}{observations}}
        \label{fig:c17MAPNBV}
\end{subfigure}

\medskip
\begin{subfigure}{.47\columnwidth}
        \includegraphics[width=1.0\linewidth]{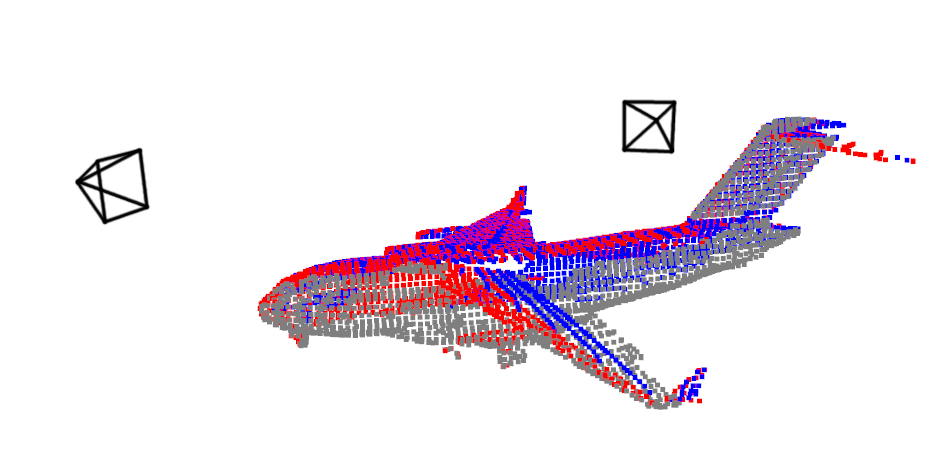}
        \caption{Observations after the first \textit{MAP-NBV} iteration}
        \label{fig:c17MAPNBV_Observations}
\end{subfigure}%
\hfill
\begin{subfigure}{.47\columnwidth}
        \includegraphics[width=0.97\linewidth]{figs/C17FullObserved.png}
        \caption{Full point cloud observed by \textit{MAP-NBV} after termination}
        \label{fig:c17MAPNBV_Observations_Final}
\end{subfigure}

\caption{MAP-NBV uses \textcolor{green}{predictions} to select better NBVs for a team of robots compared to the non-predictive baseline approach.}
\label{fig:c17predicted}
\label{fig:pointsObserved_mapnbv}
\vspace{-3mm}
\end{figure}

As shown in the previous chapter, \textit{Pred-NBV}~\cite{dhami2023prednbv}, estimating the unseen parts of the objects with point cloud completion networks can improve NBV planning, and hence the reconstruction efficiency, for a single UAV. These findings lead us to ask: \textbf{Can prediction improve the efficiency of multi-agent object reconstruction, given that multiple robots can themselves provide good coverage of the object?} 
Assuming predictions can augment the perception of multi-agent systems as well, a naive extension of methods designed for a single agent may result in significant overlaps in the observations by the team, necessitating coordination among all the robots. Prior works have shown that for target coverage problems (which object reconstruction is) \textit{explicit coordination} plays an important role in developing an efficient solution~\cite{corah2019communication}. However, this prior work only focused on scenarios where the coordination used past observation and not predictions. This begs the question: \textbf{How does coordination, in perception and planning, affect multi-agent object reconstruction when each robot has access to predictions?}

\VS{To answer these questions, we make the following contributions in this chapter}:
\begin{enumerate}
    \item We propose a decentralized, multi-agent, prediction-based NBV planning approach, named\textit{ MAP-NBV}, for active 3D reconstruction of various objects with a novel objective combining visual information gain and control effort.\looseness=-1
    
    \textit{MAP-NBV} uses partial point clouds and predicts what the rest of the point cloud would be (Figure~\ref{fig:c17predicted}) and exploits the submodular nature of the objective to coordinate in a decentralized fashion.
    \item We show that predictions effectively improve the performance by \textbf{19.41\%} over non-predictive baselines that use frontier-based NBV planning~\cite{aleotti2014global} in AirSim~\cite{airsim2017fsr} simulations.\looseness=-1
    \item We also show that \textit{MAP-NBV} results in at least \textbf{17.12\%} better reconstruction than non-cooperative prediction-guided method with experiments using the ShapeNet~\cite{shapenet2015} dataset and performs comparable to a centralized approach.\looseness=-1

\end{enumerate}
We share the qualitative results and release the project code from our method on our project website.\footnote{\url{http://raaslab.org/projects/MAPNBV/}}

\begin{figure*}[ht!]{}
    \vspace{1mm}
    \centering
    \includegraphics[width = 0.98\columnwidth]{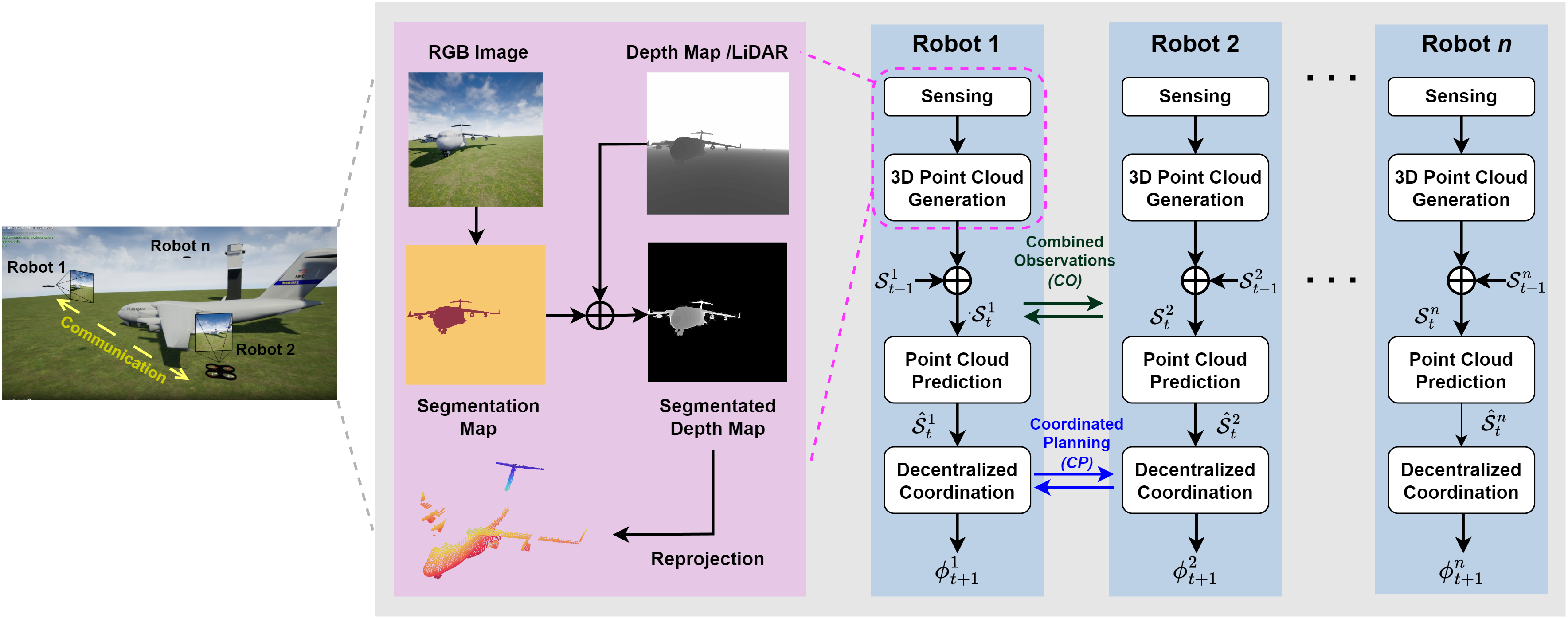}
    \caption{Algorithm Overview: Each robot runs the same algorithm including perception, prediction, and planning steps. The robots that communicate with each other can share observations and coordinate planning, whereas robots in isolation (e.g., Robot n) perform individual greedy planning.}
    \label{fig:algoOverview}
    \vspace{-5mm}
\end{figure*}

\section{Related Work}\label{sec:rel_work_mapnbv}

The use of robots for data acquisition purposes is an extensively studied topic for various domains. Their usage range from infrastructure inspection~\cite{ozaslaninspection} and environment monitoring~\cite{dunbabin2012environmental,sung2019competitive} for real-world application to the real-world digitization for research datasets and simulations~\cite{46965,ramakrishnan2021hm3d,ammirato2017dataset}. When the environment is unknown, active methods such as NBV~\cite{connolly1985determination} are used to construct an object model on the fly by capturing additional observations. A majority of the works on NBV planning use information-theoretic measures~\cite{delmerico2018comparison} for selection to account for uncertainty in observations~\cite{delmerico2018comparison,kuipers1991robot,vasquez2014volumetric}. The widely used frontier and tree-based exploration approaches also utilize uncertainty about the environment for guiding the robot motion~\cite{yamauchi1997frontier, gonzalez2002navigation, adler2014autonomous, bircher2018receding}. Some work devise geometric methods that make inferences about the exact shape of the object of interest and try to align the observations with the inferred model~\cite{tarabanis1995survey, banta2000next, kriegel2013combining}. Prediction-based NBV approaches have emerged as another alternative in recent years, where a neural network takes the robot and/or the environment state as the input and NBV pose or velocity as the output~\cite{johns2016pairwise, mendoza2020supervised, zeng2020pc, dhami2023prednbv}.

Directly extending single-robot NBV approaches to multi-robot systems may result in sub-optimal performance due to significant overlap in observations. This issue led to the development of exploration algorithms specifically for multi-robot systems~\cite{burgard2005coordinated, amanatiadis2013multi, hardouin2020next} with information-theoretic measures for determining NBV. 

Some recent works on multi-robot systems have explored the use of predictions for improvement in task efficiency. Almadhoun et al.~\cite{ almadhoun2021multi} designed a hybrid planner that switches between a classical NBV approach and a learning-based predictor for NBV selection but uses the partial model obtained by robot observations only.
Wu et al.~\cite{wu2019plant} use a point cloud prediction model for plants to use the predicted point cloud as an oracle leading to better results than the traditional approaches. This method uses entropy-based information gain measures for NBV and is designed for plant phenotyping with robotic arms. These methods do not consider the control effort required which is important for UAVs with energy constraints when deployed for observing large objects such as airplanes and ships. Also, these works employ information theoretic NBV approaches. We aim to explore a prediction-based approach for geometric NBV selection.\looseness=-1

In this chapter, we used point cloud predictions similar to Pred-NBV~\cite{dhami2023prednbv} and built a decentralized multi-robot NBV planner. The prediction on the point cloud makes the pipeline modular and interpretable, allowing for improvements by enhancing individual modules. We select NBV based on information gain, as well as control effort, making our approach more grounded in the real world.

\section{Problem Formulation}\label{sec:prob_form_mapnbv}
We are given a team of \textit{n} robots, each equipped with an RGB-D camera or LiDAR sensor. 
This team is tasked with navigating around a closed object of volume $\mathcal{V} \in \mathbb{R}^3$ and observes its surface as a set of 3D points, $\mathcal{S}$.
At any given time $t$, the set of surface points $S^i_t$ observed by the robot $r_i$ from the view-point $\phi^i_t \in \Phi$ at time $t$ is represented as a voxel-filtered point cloud and the relationship between them is defined as $S^i_t = f(r_i, \phi^i_t)$. Each robot $r_i$ follows a trajectory $\xi_{i}$, which consists of a sequence of viewpoints
aimed at maximizing the coverage of $\mathcal{S}$ while minimizing redundancy among observed points. 

The distance traveled by a robot between two poses $\phi_i$ and $\phi_j$ is represented by $d(\phi_i, \phi_j)$. The point cloud observed by the team of robots is the union of the surface points observed by the individual robots over their respective trajectories, i.e., $S_{\Bar{\xi}} =  \bigcup_{i=1}^n \bigcup_{\phi \in \xi_{i}} f(r_i, \phi)$ and $\Bar{\xi}$ represents the set of trajectories for each robot, i.e., $\Bar{\xi} = \{\xi_{1}, \xi_{2},..., \xi_{n}\}$.

The objective is to find a set of feasible trajectories $\Bar{\xi}^* = \{ \xi_{1}^*, \xi_{2}^*, ..., \xi_{n}^*\}$, such that the team observes the whole voxel-filtered surface $\mathcal{S}$, while also minimizing the total distance traveled by the robots on their respective trajectories.
\begin{align}
    \Bar{\xi}^* = \argmin_{\Bar{\xi}} \sum_{i=1}^n \sum_{j=1}^{| {\xi_{i}}-1|} d(\phi_j^i, \phi_{j+1}^i)\\ 
    \textit{such that}~~ \bigcup_{i=1}^n \bigcup_{\phi \in \xi_{i}} f(r_i, \phi) = \mathcal{S} 
\end{align}

Given a finite set of trajectories, if the object model, $\mathcal{S}$, is known, we can find the optimal set of trajectories through an exhaustive search. As the object model is not known apriori in an unknown environment, the optimal solution can not be found beforehand. Thus, each robot needs to determine the NBV based on the partial observations of the team to reconstruct the object's surface. Here we assume that each robot can observe the object at the start of the mission, which can be accomplished by moving the robots till they see the object. While a centralized server can help find an optimal assignment solution, a limited communication range can make a centralized solution infeasible. Thus, we define this problem as a decentralized one; each robot solves this objective but the communicating robots can collaborate and coordinate with their neighbors.

\section{Proposed Approach}\label{sec:approach_mapnbv}
In this section, we present \textit{Multi-Agent Pred-NBV (MAP-NBV)}, a prediction-guided NBV approach for a team of robots. Figure~\ref{fig:algoOverview} shows the overview of our process, which consists of two parts: (1) \textit{3D Model Prediction}, where we combine the observations from the neighboring robots to build a partial model of the object and use PoinTr-C~\cite{dhami2023prednbv}, a 3D point cloud completion network, to predict the full shape of the objects, 
and (2) \textit{Decentralized Coordination} which combines the observations from the communicating robots and solves an NBV objective to maximize information gain and minimize the control effort with sequential greedy assignment. Robots that do not communicate with anyone effectively run an individual greedy algorithm. This approach is detailed in Algorithm~\ref{algo:mapnbv}. Apart from being feasible and scalable, this approach also reduces the computation complexity resulting in fast runtime.

\begin{algorithm}
\caption{MAP-NBV Algorithm (for robot $r_i$)}
\begin{algorithmic}[1]
    \State \textbf{Inputs:} Initial positions $\phi^i_0$; Stopping threshold $\tau$; Information gain threshold $\lambda$
    \State \textbf{Output:} 3D Point Cloud $\mathcal{S}$
    \State \textbf{Initialization}: $novelty$ $\gets 0$; $t \gets 0$; $S^i_{-1} \gets \emptyset$, $\xi_i \gets \{\phi^i_0\}$
    \While{$novelty \le \tau$} \label{algo:line:stopping_condition}
        \State $\mathcal{S}^i_t \gets \bigcup_{k \in Neighbors} f(r_i, \phi^i_t)$ $\bigcup \mathcal{S}^i_{t-1}$  \label{algo:line:combined_observations}
        \State $\hat{\mathcal{S}}^i_t \gets$ getPredictionFromPoinTr-C($\mathcal{S}^i_t$) \label{algo:line:pc_completion}
        \State $\mathcal{C} \gets$ generateCandidatePoses($\hat{\mathcal{S}}^i_t, \mathcal{S}^i_t$) \label{algo:line:cand_pose_generation}

        \State $\mathcal{I}_{seen} \gets \bigcup_{k \in Neighbors; k < i} I (\xi_k)$ \label{algo:line:infogather}

        \State $\hat{\mathcal{I}} \gets \{ \mathcal{I}(\xi_i \cup \phi) - \mathcal{I}_{seen}; \forall  \phi \in \mathcal{C} \}$  \label{algo:line:potential_I_calculation}

        \State $\phi^i_{t+1} \gets \argmin_{\phi \in \mathcal{C}} d(\phi, \phi^i_{t+1})$, s.t. $\frac{\hat{\mathcal{I}}(\phi)}{\max \hat{\mathcal{I}}} \ge \lambda$ \label{algo:line:objective}
        
        \State $\xi_i \gets \xi_i \cup \phi^i_{t+1}$ \label{algo:line:traj_update}
        
        \State Broadcast($\xi_i$) \label{algo:line:broadcast}
        
        \State $novelty \gets \frac{| \mathcal{S}^i_t |}{| \mathcal{S}^i_{t-1}|}$ \label{algo:line:novelty_update}
        \State $t \gets t+1$ \label{algo:line:time_update}
    \EndWhile
    \State $\mathcal{S} \gets \bigcup_{i=1}^n \mathcal{S}^i_{t-1}$  \label{algo:line:novelty_update_1}\label{algo:line:final_pcd}
\end{algorithmic}
\label{algo:mapnbv}
\end{algorithm}

\subsection{3D Model Prediction (Line~\ref{algo:line:combined_observations}-\ref{algo:line:pc_completion})} 
\VS{To start, we use the RGB images to segment out the object and the depth sensors to generate the point cloud from the current observations for each robot, giving us segmented point clouds. This allows the algorithm to focus on only the target infrastructure as opposed to also including other obstacles. For the robots that can communicate with each other (e.g., Robot 1 and Robot 2 in Fig~\ref{fig:algoOverview}), each segmented point cloud per robot is transformed into a global reference frame and concatenated together into a single point cloud (Line~\ref{algo:line:combined_observations}). This point cloud represents the entire communication subgraph's observations of the target object at the current timestamp.} The point cloud concatenation can be replaced with a registration algorithm~\cite{huang2021comprehensive}, but we use concatenation due to its ease of use. Lastly, this current timestamp's point cloud is then concatenated with previous observations to get an up-to-date observation point cloud.

\VS{In order to get an approximation $\mathcal{\hat{S}}$ of the full model $\mathcal{S}$, we use PoinTr-C~\cite{dhami2023prednbv}. The combined observed point cloud of the object at time $t$, $\mathcal{S}_t$ goes as input to PoinTr-C and it predicts the full object point cloud $\mathcal{\hat{S}}_t$ (Line~\ref{algo:line:pc_completion}). PoinTr-C requires isolating the object point clouds from the scene. This can be realized with the help of distance-based filters and state-of-the-art segmentation networks\cite{kirillov2023segment} without any fine-tuning. An example of an input point cloud and a predicted point cloud, both from individual observations and combined observations, is shown in Figure~\ref{fig:c17predicted}}.

\subsection{Decentralized Coordination (Line~\ref{algo:line:cand_pose_generation}-\ref{algo:line:broadcast})}
\textbf{Next Best View Planning}. We use the predicted point cloud $\mathcal{S}_t$ as an approximation of the ground truth point cloud for NBV planning. For this, we first generate a set of candidate poses around the partially observed object (Line~\ref{algo:line:cand_pose_generation}). From these, we select a set of poses as NBVs for each robot, based on information gain and control effort. For a set of $k$ candidate viewpoints, we define the information gain, $\hat{\mathcal{I}}$, as the expected number of new, unique points the robots will observe after moving to these viewpoints. The control effort is defined as the total distance traversed by the robots to reach the viewpoints.

The number of new points varies with each iteration as robots move to new locations, observing more of the object's surface. While PoinTr-C predicts the point cloud for the whole object, the robots can observe only points on the surface close to it. Hence, before counting the number of new points, we apply hidden point removal~\cite{katz2007direct} to the predicted point cloud. We represent this relationship between the number of points observed and the trajectories traversed till time for a robot $r_i$ as $I(\xi_i)$, where $\xi = \{\phi^i_0, \phi^i_t, ..., \phi^i_t\}$ represents the trajectory for the robot $r_i$ till time $t$ consisting of the set of viewpoints the robot has traversed. To balance the information gain and control effort, we use a hyperparameter $\lambda$. First, we find the estimated information gain for each candidate pose $\hat{\mathcal{I}}$ (Line~\ref{algo:line:potential_I_calculation}) by treating the $\hat{\mathcal{S}}$ as the actual full object model. Each robot selects the candidate pose closest to the current pose where it achieves at least $\lambda \%$ of the maximum possible information gain over all the candidate poses (Line~\ref{algo:line:objective}).\looseness=-1

To find the control effort, we use RRT-Connect~\cite{kuffner2000rrt} to find the path between the robot $r_i$'s current location $\phi^i_t$ to each candidate pose $\phi$ and use its length as $d(\phi, \phi^i_t)$. 
The candidate poses are generated similarly to Pred-NBV~\cite{dhami2023prednbv}, i.e., on circles at different heights around the center of the predicted object point cloud (Line~\ref{algo:line:cand_pose_generation}).
One circle is at the same height as the predicted object center with radius $1.5 \times d_{max}$, where $d_{max}$ is the maximum distance of a point from the center of the predicted point cloud. The other two circles are located above and below this circle $0.25 \times \text{z-range}$ away, with a radius of $1.2 \times d_{max}$. The viewpoints are located at steps of $30^\circ$ on each circle.\looseness=-1

\begin{figure}[ht!]
    \centering
    \includegraphics[width=\linewidth]{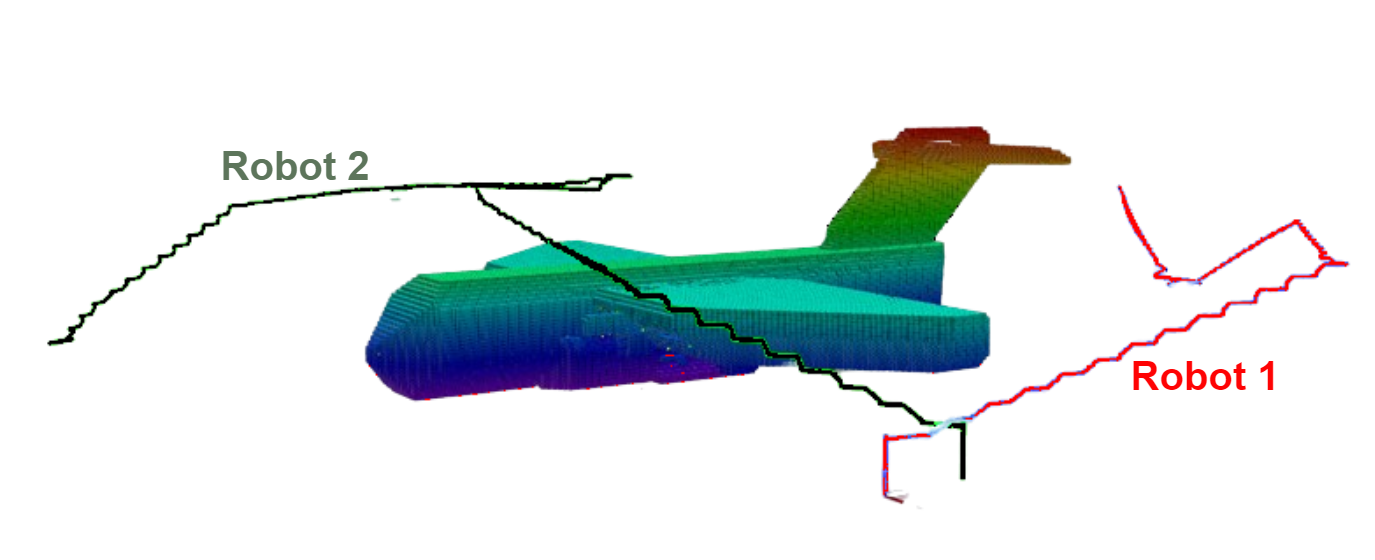}
    \caption{Flight paths of the two robots during C-17 simulation.}
    \label{fig:flightPath}
\end{figure}

\textbf{Sequential Greedy Assignment}. To effectively and efficiently use multiple robots, we coordinate among them for NBV assignment. We devise a decentralized coordination strategy to accommodate the dynamic communication graph effectively. We also leverage the submodular nature of the problem and use a sequential greedy assignment which results in a near-optimal solution~\cite{calinescu2011maximizing}. With each communications subgraph, the robots select their NBVs in the order determined by their IDs (lower to higher). Each robot first considers the trajectories of its neighbors with lower IDs to find the information gain attained by their movement $\mathcal{I}_{seen}$ (Line~\ref{algo:line:infogather}). For the robot with the lowest ID, $\mathcal{I}_{seen}$ is empty. Then we greedily choose the candidate position that would result in the information gain above a threshold $\lambda$ while minimizing the distance traveled. This candidate's position for the next iteration 
$\phi_{t+1}^i$ is the NBV for $r_i$. We add this pose to the robot's trajectory (Line~\ref{algo:line:traj_update}) and broadcast this information with the neighbors (Line~\ref{algo:line:broadcast}) to minimize overlaps.

In our experiments, we consider multi-hop communication, thus each robot can coordinate with every robot on its communication subgraph. Some robots may not be within any other robot's communication range (e.g., Robot \textit{n} in Fig~\ref{fig:algoOverview}). For such robots, this strategy effectively turns into a greedy prediction-guided assignment similar to Pred-NBV~\cite{dhami2023prednbv}. At each iteration, we calculate $novelty$, i.e., the ratio of the number of points observed over subsequent iterations (Line~\ref{algo:line:novelty_update}. The robot stops if $novelty$ is above a predefined threshold $\tau$ (we set $\tau = 0.95$ in our experiments).
In the end, all the robots assemble at the same location and combine their observations as the object model $\mathcal{S}$ (Line~\ref{algo:line:final_pcd}).\looseness=-1

\section{Experiments and Evaluation}\label{sec:eval_mapnbv}
We design experiments to answer the two key research questions: (1) can point cloud prediction improve multi-agent object reconstruction? and (2) how does coordination between agents affect the reconstruction? To answer the first question, we compare \textit{MAP-NBV} with a non-predictive frontier-based baseline. For the second question, we compare \textit{MAP-NBV} with a centralized and a non-coordinated variation of \textit{MAP-NBV}. We first describe the experiment setups used to answer these questions and then discuss the results obtained.

\subsection{Setup}
\label{subsec:setup}
\textbf{AirSim}: We use AirSim~\cite{airsim2017fsr} simulator as it allows us to load desired object models and get photo-realistic inputs while also supporting multiple robots. We use Robot Operating System (ROS) Melodic to run the simulations on Ubuntu 18.04. We spawn multiple UAVs close to each other looking towards the object. We equip each UAV with a depth camera and an RGB camera. Each UAV publishes a segmented image using AirSim's built-in segmentation. This segmented image is used along with a depth map to remove the background and isolate the depth map for the object. We then convert these segmented depth images into 3D point clouds.  For collision-free point-to-point planning, we use the MoveIt~\cite{coleman2014reducing} package implementing the work done by Köse~\cite{tahsinko86:online}. 


\begin{figure}[htp]
\begin{minipage}[b]{0.43\linewidth}
    \centering
    \begin{subfigure}[b]{\linewidth}
        \centering
        \includegraphics[width=\textwidth]{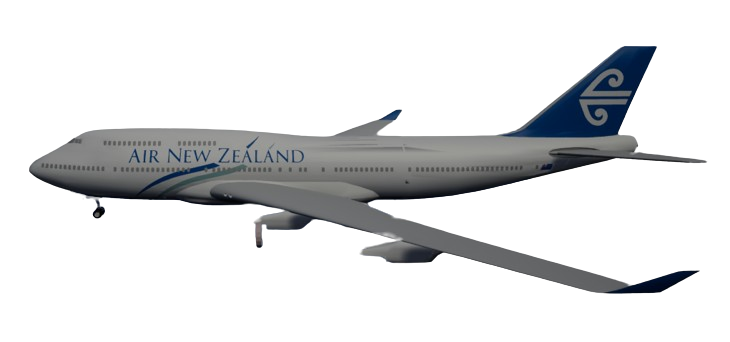}
        \caption{Airplane}
        \label{fig:airplane}
    \end{subfigure}
    

    \begin{subfigure}[b]{\linewidth}
        \centering
        \includegraphics[width=\linewidth]{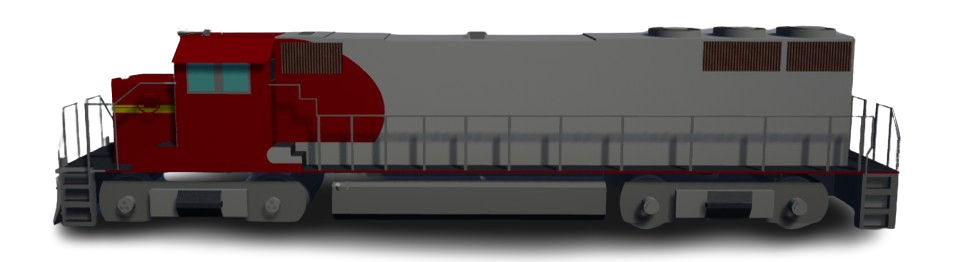}
        \caption{Train}
        \label{fig:train}
    \end{subfigure}


    \begin{subfigure}[b]{\linewidth}
        \centering
        \includegraphics[width=\linewidth]{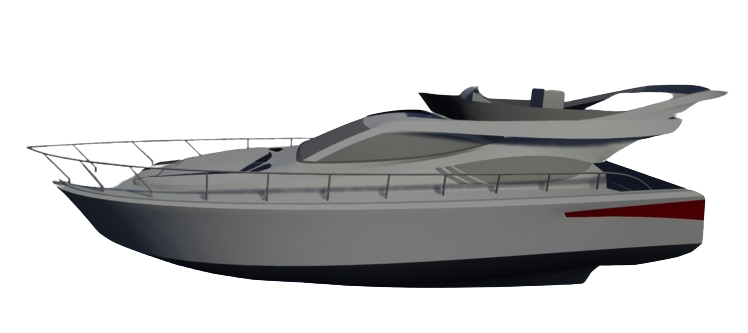}
        \caption{Boat}
        \label{fig:boat}
    \end{subfigure}
\end{minipage}
\begin{minipage}[b]{0.56\linewidth}
    \centering
    \begin{subfigure}[b]{0.41\linewidth}
        \centering
        \includegraphics[width=0.80\textwidth]{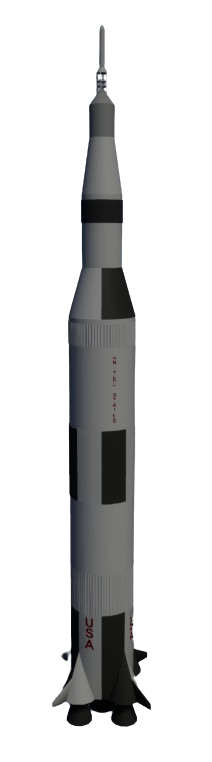}
        \caption{Rocket}
        \label{fig:rocket}
    \end{subfigure}
    \begin{subfigure}[b]{0.55\linewidth}
        \centering
        \includegraphics[width=0.90\textwidth]{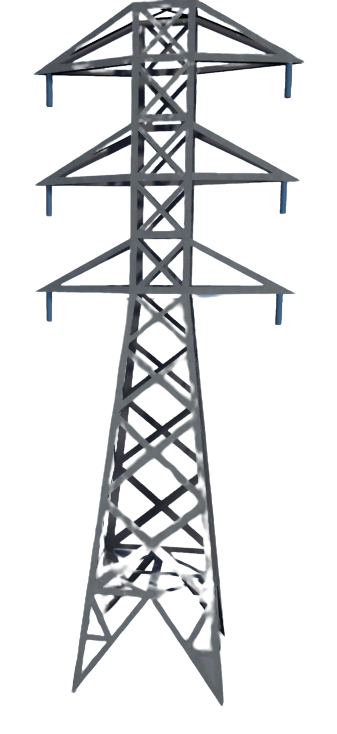}
        \caption{Tower}
        \label{fig:tower}
    \end{subfigure}
\end{minipage}

\caption{Examples of the 5 simulation model classes.}
\label{fig:simModels}
\end{figure}

\textbf{ShapeNet}: The AirSim setup helps us evaluate the coverage in a realistic setting, but it lacks a ground truth point cloud to compare the reconstruction quality. Hence, we run simulations with the ShapeNet~\cite{shapenet2015} dataset to obtain a ground truth and compare the reconstruction quality across different methods. We simulate the UAVs as free-floating cameras which can teleport to any location. These cameras do not have a roll and pitch and can only change their yaw. We use the Open3D~\cite{Zhou2018} library to load an object mesh (normalized) and select a random starting position around it. The UAVs are placed close to each other at this location. We render the depth image from the object mesh and reproject it to the 3D point cloud. Unlike the previous setup, we do not need segmentation as there is no background in this setup. 
To evaluate the reconstruction quality, we convert the mesh to a point cloud by uniformly sampling points on it and use it as the ground truth to quantify the reconstruction quality with Directional Chamfer Distance with $\ell_2$-norm (CD-$\ell_2$) (from ground truth point cloud to the observed point cloud). We use Euclidean distance as the distance metric $d(.,.)$ in this setup.\looseness=-1

\subsection{Qualitative Example}
We evaluate \textit{MAP-NBV} on the same 20 objects that were used in Pred-NBV to allow a direct comparison. 
The 20 objects consist of five different ShapeNet classes:  airplane, rocket, tower, train, and watercraft. Examples of each class are shown in Figure~\ref{fig:simModels}. These classes represent diverse shapes and infrastructures that are regularly inspected. Figure~\ref{fig:flightPath} shows the path followed by two UAVs as given by \textit{MAP-NBV} in the C-17 airplane simulation. This environment includes other obstacles that are not of interest but still need to be accounted for in collision-free path planning. \textit{MAP-NBV} finds a collision-free path for both UAVs while targeting the maximum coverage of the C-17 airplane.

\subsection{Can Point Cloud Prediction Improve Reconstruction?}\label{sec:sim:MAbaseline}
We compared the performance of \textit{MAP-NBV} with a modified baseline NBV method~\cite{aleotti2014global} designed for multi-agent use on 20 objects, as listed in Table~\ref{tab:airsim_multiagent_results}. The baseline method employs frontiers to select the next-best views. Frontiers are points located at the edge of the observed space near unknown areas. We utilized the same modifications described in Pred-NBV~\cite{dhami2023prednbv}. Specifically, we used our segmented point cloud to choose frontiers near the target object. To ensure that the UAVs always face the target object, the orientation of all poses selected by the baseline aligns with the center of the observed target object point clouds.

We further adapted this baseline method to function in a multi-agent setting. The pose for the first UAV is selected in the same manner as in the single-agent baseline. For each subsequent UAV, the remaining best pose is chosen, as long as it does not fall within a certain distance threshold compared to the previously selected poses in the current iteration of the algorithm.

Both \textit{MAP-NBV} and the baseline algorithm employ the same stopping criteria. The algorithm terminates if the total points observed in the previous step exceed 95\% of the total points observed in the current step. We run both the algorithms in the AirSim setup. Our evaluation, presented in Table~\ref{tab:airsim_multiagent_results}, demonstrates that \textit{MAP-NBV} observes, on average, 19.41\% more points than the multi-agent baseline for object reconstruction across all 20 objects from the five different model classes. In our simulations, we utilized 2 UAVs for both algorithms.\looseness=-1

Furthermore, the \textit{MAP-NBV} algorithm can be readily extended to accommodate more than just 2 robots. By incorporating additional UAVs, the algorithm can effectively leverage the collaborative efforts of a larger multi-agent system to improve object reconstruction performance and exploration efficiency. However, in our current evaluation, we utilized 2 UAVs for both algorithms due to limited computational resources. The simulations were computationally intensive, and our computer experienced significant slowdowns with just 2 robots in the simulation. Despite this limitation, the promising results obtained with 2 UAVs suggest that scaling up the algorithm to include more robots has the potential to yield even more significant performance improvements.

\begin{figure}[ht!]
    \vspace{-3mm}
    \centering
    \includegraphics[width=0.93\linewidth]{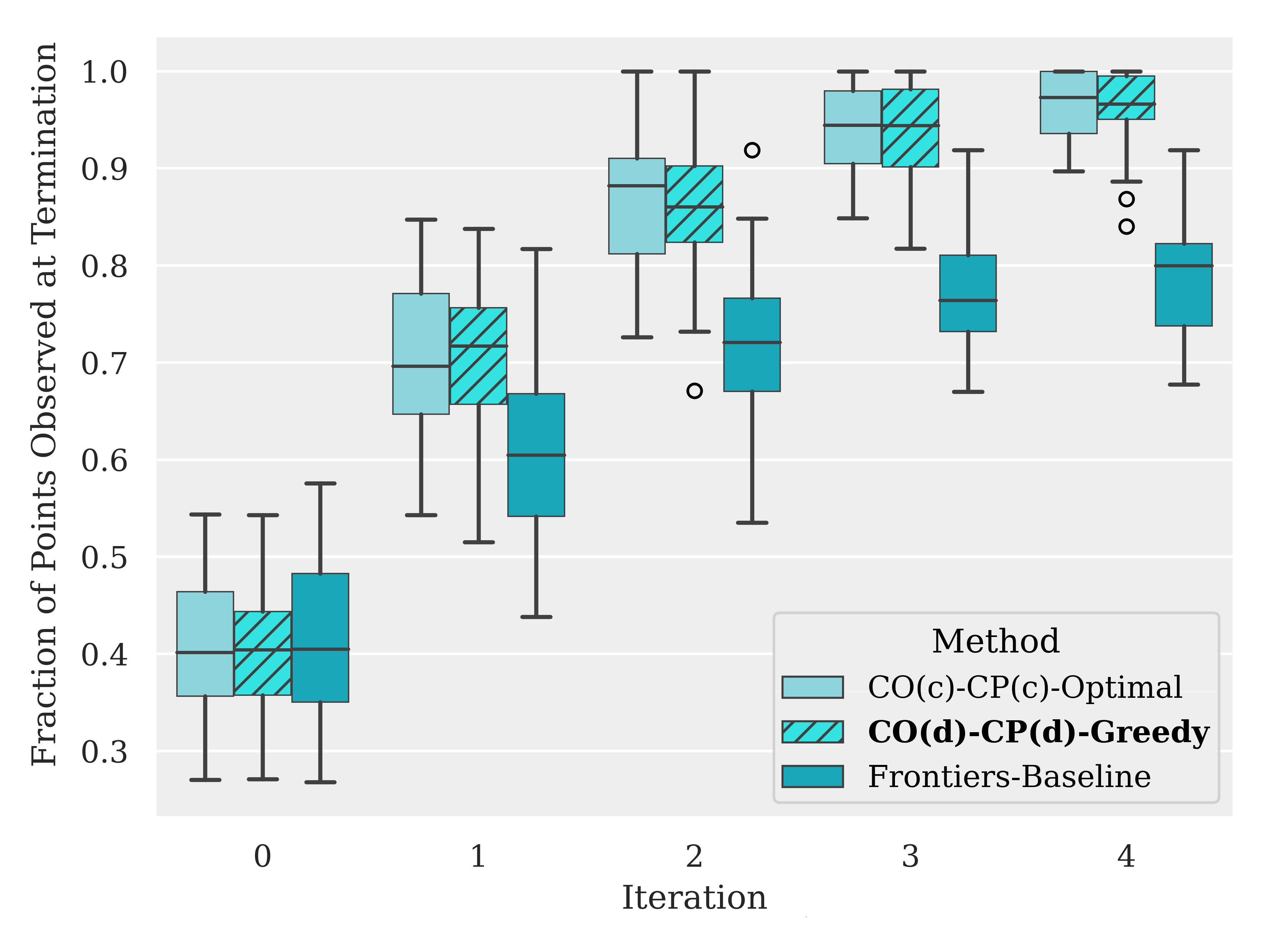}
    \caption{MAP-NBV (\texttt{CO(d)-CP(d)-Greedy}) performs comparably to the optimal solution (\texttt{CO(c)-CP(c)-Optimal}; Section~\ref{sec:COCPdesc}),  and much better than the frontiers-baseline in AirSim experiments.\looseness=-5 }
    \label{fig:plane_res_mapnbv}
\end{figure}

Additionally, Figure~\ref{fig:plane_res_mapnbv} illustrates that \textbf{\textit{MAP-NBV} observes more points per step than the multi-agent baseline}.

\begin{table}[ht!]
\vspace{1.0mm}
    \centering

    \caption{\textit{MAP-NBV} results in a better coverage compared to the multi-agent baseline NBV method~\cite{aleotti2014global} for all models in AirSim upon algorithm termination.}
    
    \begin{tabular}{p{2cm}p{2cm}rrr}
    \toprule
    \multirow{2}{*}{Class} & \multirow{2}{*}{Model} & \multicolumn{2}{c}{Points Seen} & \multirow{2}{*}{Improvement} \\
    \cmidrule(lr){3-4}
    & & \textit{MAP-NBV} & MA Baseline & \\
        \midrule
        \multirow{5}{*}{Airplane}  
            & 747 & \textbf{16140} & 13305 & 19.26\% \\
            & A340 & \textbf{10210} & 8156 &  22.37\% \\
            & C-17 & \textbf{13278} & 10150 & 26.70\% \\
            & C-130 & \textbf{6573} & 5961 & 9.77\% \\
            & Fokker 100 & \textbf{14986} & 13158& 12.99\%\\
        \midrule
        \multirow{5}{*}{Rocket} 
            & Atlas & \textbf{2085} & 1747 & 17.64\% \\
            & Maverick & \textbf{3625} & 2693 & 29.50\% \\
            & Saturn V & \textbf{1041} & 877 & 17.10\% \\
            & Sparrow & \textbf{1893} & 1664 & 12.88\% \\
            & V2 & \textbf{1255} & 919 & 30.91\% \\
        \midrule
        \multirow{5}{*}{Tower} 
            & Big Ben & \textbf{4294} & 3493 & 20.57\% \\
            & Church & \textbf{7884} & 6890 & 13.46\% \\
            & Clock & \textbf{3163} & 2382 & 28.17\% \\
            & Pylon & \textbf{2986} & 2870 & 3.96\% \\
            & Silo & \textbf{5810} & 4296 & 29.96\% \\
        \midrule
        \multirow{2}{*}{Train}
            & Diesel & \textbf{4013} & 3233 & 21.53\% \\
            & Mountain & \textbf{5067} & 4215 & 18.36\% \\
        \midrule
        \multirow{3}{*}{Watercraft} 
            & Cruise & \textbf{5021} & 3685 & 30.69\% \\
            & Patrol & \textbf{4078} & 3683 & 10.18\% \\
            & Yacht & \textbf{11678} & 10341 & 12.14\% \\
        \bottomrule
    \end{tabular}
    
    \label{tab:airsim_multiagent_results}
\end{table}

\subsection{How does coordination affect reconstruction?}\label{subsec:ablation}
\VS{For studying the effect of coordination we randomly select 25 objects, 5 from each of the ShapeNet classes mentioned in Section~\ref{sec:sim:MAbaseline} for the ShapeNet setup (Section~\ref{subsec:setup}).}

\label{sec:COCPdesc}
In a multi-agent setting, the communicating robots can combine their observation (\textbf{CO}) or they may choose to rely on their observations (\textbf{IO}) for point cloud predictions. After the prediction, they may choose to collaborate for planning (\textbf{CP}) or make individual decisions without relying on others (\textbf{IP}). To highlight the effect of these coordination strategies on reconstruction quality, for a team of $n$ robots and $k$ candidate viewpoints, we compare MAP-NBV with two contrasting coordination approaches: 
\begin{itemize}
    \item \textbf{\texttt{CO(c)-CP(c)-Optimal}}: This approach relies on combined observation and coordinated planning in a centralized manner. Here, a central server aggregates the observations, performs prediction, and finds the NBVs for each robot. For selecting NBVs, the algorithm evaluates all robot-candidate pose assignments and chooses the optimal setting, i.e.,  which results in maximum joint information gain. Among all possible permutations of robots that result in the optimal setting, we select the one that minimizes the maximum displacement for any robot. This process has a runtime complexity of $\mathcal{O}(k^{n})$.
    \item \textbf{\texttt{IO-IP}}: In this approach, each robot operates individually and does not share observations or coordinate with others regardless of the communication range. This is effectively a naive extension of Pred-NBV to a multi-agent setup. The runtime complexity here is $\mathcal{O}({k})$.
\end{itemize}

Following the notations above, \textbf{\uline{MAP-NBV is {CO(d)-CP(d)-Greedy}}}, i.e., combined observations (decentralized) and coordinated planning (decentralized) with the greedy assignment. \textit{Decentralized} indicates that the observations and planning are shared among only those robots that form a communication subgraph. \textit{MAP-NBV} thus has a runtime complexity of $\mathcal{O}(n \cdot k)$ in the worst case only, i.e., when all the robots are connected. We study these algorithms for teams of 2, 4, and 6 robots. 

\begin{figure}[ht!]
    \centering
    \begin{subfigure}[b]{0.51\columnwidth}%
        \includegraphics[width = \textwidth]{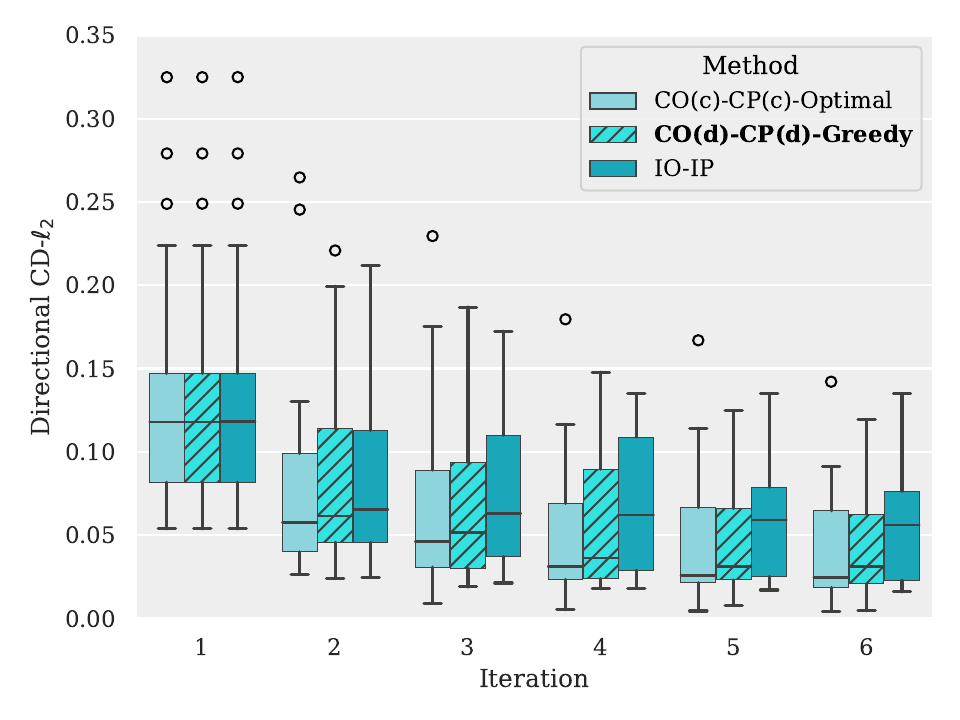}%
        \subcaption{2 Robots}
    \end{subfigure}%
    \hfill%
    \begin{subfigure}[b]{0.51\columnwidth}%
        \includegraphics[width = \textwidth]{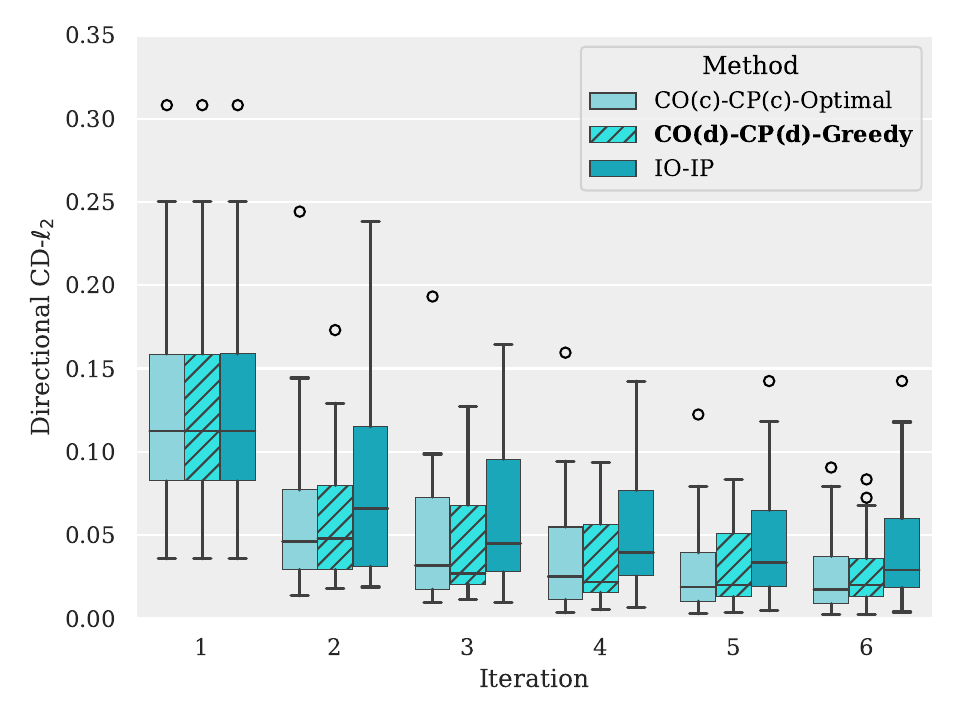}%
        \subcaption{4 Robots}
        
    \end{subfigure}%
    \hfill%
    \begin{subfigure}[b]{0.51\columnwidth}%
        \includegraphics[width = \textwidth]{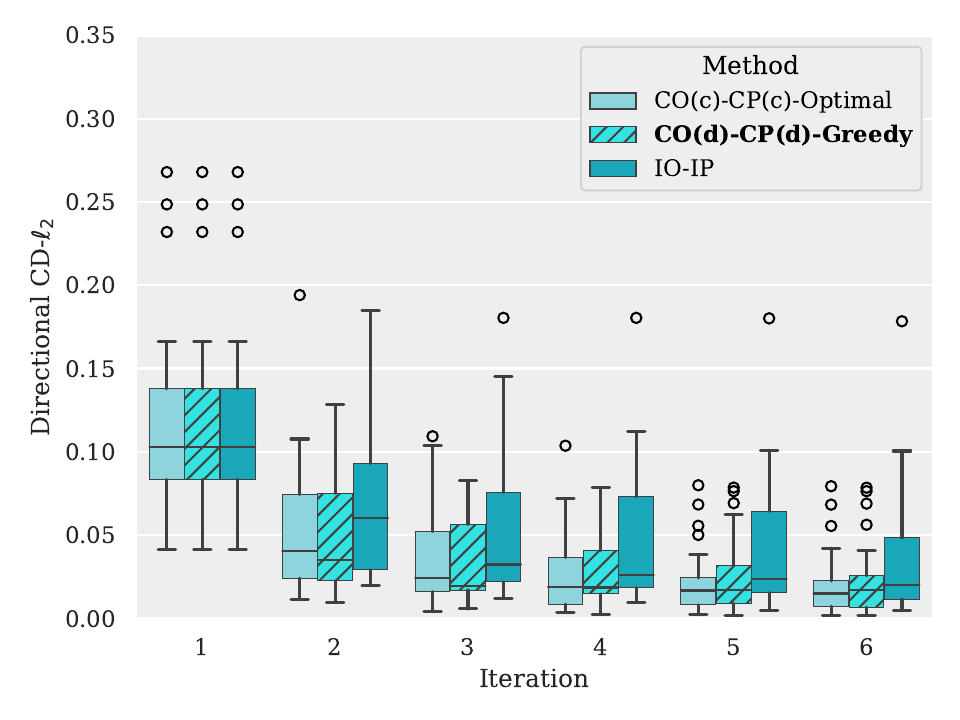}%
        \subcaption{6 Robots}
    \end{subfigure}%
    \caption{Directional CD-$\ell_2$ for teams of 2, 4, and 6 robots on ShapeNet models~\cite{shapenet2015} with different coordination strategies.}
    \label{fig:ablation}
\end{figure}

At each iteration, we combine the observations from all robots and compute Direction CD-$\ell_2$ over it. This thus represents the reconstruction quality if the mission terminated at that iteration. We found that 6 iterations were enough for all robots to reach the stopping criteria in each setting. Our findings are shown in Figure~\ref{fig:ablation}.

We observed that coordination and information sharing play a crucial role in improving the reconstruction quality. \texttt{CO(c)-CP(c)-Optimal} exhibits the best reconstruction over time, as expected, owing to shared observations which lead to better estimation of the partial point cloud, and coordinated planning, which minimizes overlap in information gain. In early iterations, the limited observations may lead to an imprecise prediction, which may result in inefficient NBV assignments. Still, over time the observation coverage increases leading to better predictions and performance over the other algorithms. We observed similar trends for the AirSim setup as well as shown in Figure~\ref{fig:plane_res_mapnbv}.

However, \texttt{CO(c)-CP(c)-Optimal} requires large computation time and does not scale well in comparison with \textit{MAP-NBV} \texttt{(CO(d)-CP(d)-Greedy)}. On a Ubuntu 20.04 system with 32-core, 2.10Ghz Xeon Silver-4208 CPU and Nvidia GeForce RTX 2080Ti GPU, \textit{MAP-NBV} was 2x faster for 4 robots and 10x faster for 6 robots compared to \texttt{CO(c)-CP(c)-Optimal}. Even with faster execution, \textit{MAP-NBV} performs comparably to \texttt{CO(c)-CP(c)-Optimal}, making it more attractive than the former. Additionally, the decentralized nature makes \textit{MAP-NBV} a more feasible algorithm than \texttt{CO(c)-CP(c)-Optimal} which requires centralization. \texttt{IO-IP}, where the robots do not share observations or coordinate plans, exhibits the worst improvement over time, highlighting that \textbf{the coordination plays a crucial role in multi-agent object reconstruction}. In fact, \textit{MAP-NBV} achieves a relative improvement of \textbf{17-22\%} in directional CD-$\ell_2$ over \texttt{IO-IP} after termination.

Interestingly, we observe that as the number of robots increases, the gap between these algorithms decreases. This is expected as more robots lead to more coverage at any time. Since the robots start the mission at close distances in our experiments, we found that \texttt{IO-IP} still exhibits worse improvement as the robots suffer significant overlaps in observations. Thus coordination still provides benefits and faster runtime of \textit{MAP-NBV} makes it a more suitable choice than the centralized alternative, even if centralization is feasible. 
We also performed experiments with a centralized, greedy selection variant of \textit{MAP-NBV} and found it performs only marginally better than \textit{MAP-NBV}'s performance. We share these findings on our \href{https://raaslab.org/projects/MAPNBV/}{project webpage} due to lack of space.\looseness=-1

\subsection{Qualitative Real-World Experiment}
\begin{figure*}[ht!]
    \centering
    \begin{subfigure}[c]{.51\textwidth}%
        \includegraphics[width = \textwidth]{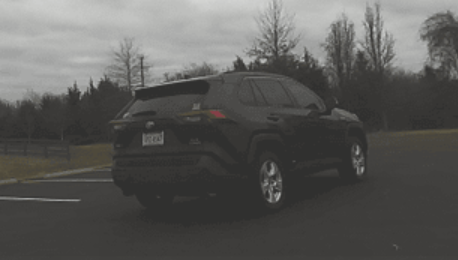}%
        \caption{}
    \end{subfigure}%
    \hfill
    \begin{subfigure}[c]{.51\textwidth}%
        \includegraphics[width = \textwidth]{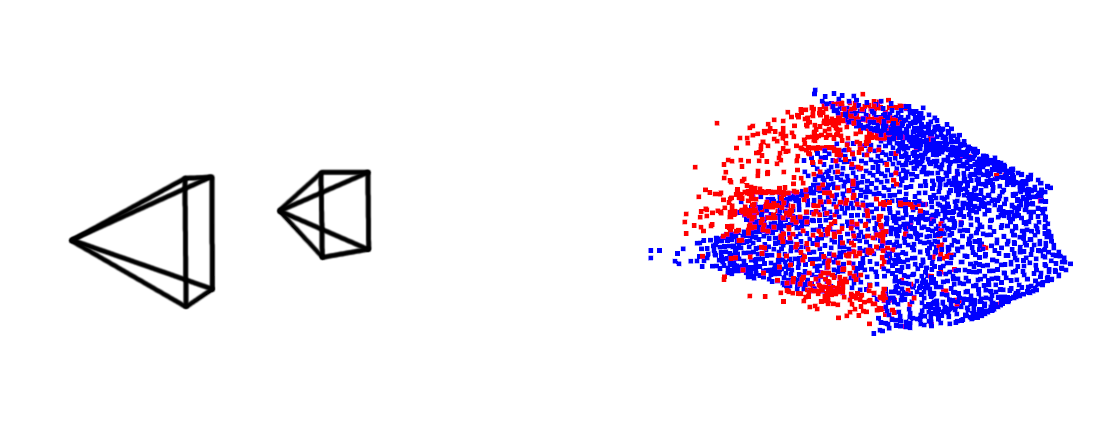}%
        \caption{}
    \end{subfigure}%
    \hfill
    \begin{subfigure}[c]{.51\textwidth}%
        \includegraphics[width = \textwidth]{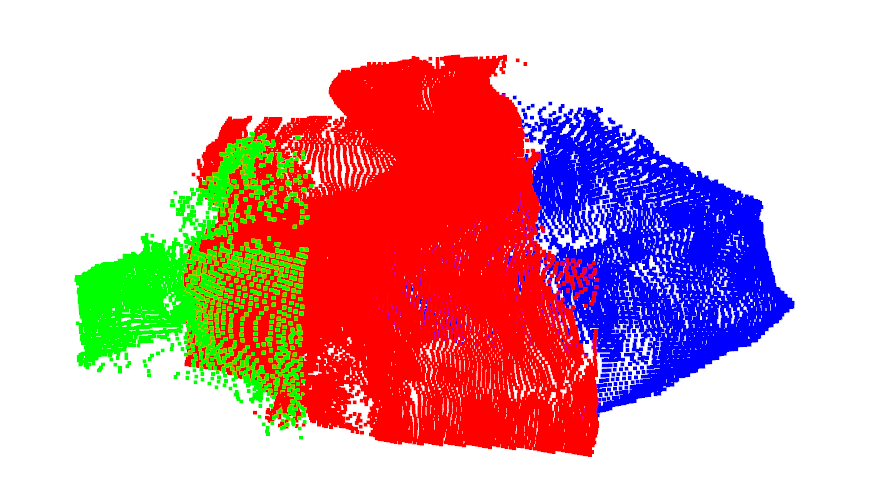}%
        \caption{}
    \end{subfigure}%
    \hfill
    \begin{subfigure}[c]{.51\textwidth}%
        \includegraphics[width = \textwidth]{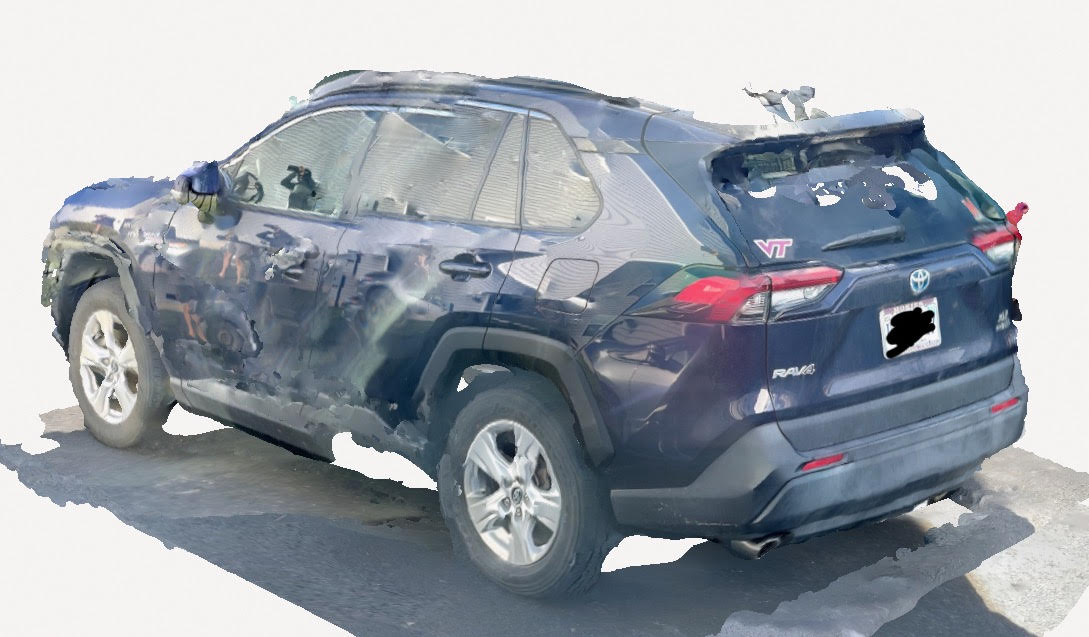}%
        \caption{}
    \end{subfigure}
    \caption{Real-World MAP-NBV experiment. (a) RGB Image. (b) \textcolor{blue}{Observations}, \textcolor{red}{Predictions}, and MAP-NBV poses. (c) \textcolor{blue}{Initial}, \textcolor{red}{Drone 1}, and \textcolor{green}{Drone 2} points after MAP-NBV iteration. (d) Reconstruction.}
    \label{fig:zed_experiment}
    \vspace{-5mm}
\end{figure*}

We also conducted a real-world experiment to evaluate the feasibility of implementing MAP-NBV on hardware. A single iteration of the MAP-NBV pipeline for 2 robots was run using a ZED camera. Depth images and point clouds were captured from the ZED of a Toyota RAV4 car. The car was extracted from the point clouds and used as input for the MAP-NBV pipeline. Then, depth images and point clouds were captured from the poses MAP-NBV outputted. Due to the lower quality of the ZED camera, an iPhone 12 Pro Max with built-in LiDAR was used to capture data to create a high-resolution reconstruction shown in Figure~\ref{fig:zed_experiment}d.

\section{Conclusions}\label{sec:con_mapnbv}
We present a multi-agent, decentralized, prediction-guided NBV planning approach for active 3D reconstruction. This method can be helpful in a variety of applications including civil infrastructure inspection. We show that our method can faithfully reconstruct the object point clouds efficiently compared to non-predictive multi-agent methods and other prediction-guided approaches. Our NBV planning objective considers both information gain and control effort, making it more suitable for real-world deployment given the flight time limit imposed on UAVs by their battery capacity. 

We are currently working on a bandwidth-aware extension of \textit{MAP-NBV} and the preliminary studies show encouraging results (shared on our \href{https://raaslab.org/projects/MAPNBV/}{webpage}). In this chapter, we focus solely on geometric measures for information gain. Many existing works on NBV have developed sophisticated information theoretic measures. We will explore combining both types of measures in our future work.


\renewcommand{\thechapter}{4}

\chapter{Object Monitoring: Visual Inspection}\label{chap:insp}

\section{Introduction}

In the previous chapters, we were only interested in the next-best-view. However, would it be better if we planned for several next views to take into account the potential overlap in views? With that in mind, how can we find a trajectory instead of just a single next-best-view? Instead of focusing on predictions like the previous chapters, we only focus on the observed portion of the object. We do that in the context of visual inspection with the additional constraint that views need to be within a certain distance and angle from the part of the object we are trying to inspect. Our motivation is the application of infrastructure inspection. Currently, infrastructure inspection is performed manually either by an inspector being suspended across the infrastructure's surface or by the inspector piloting a UAV. The first is dangerous whereas the second prevents the inspector from completely focusing on detecting defects, such as cracks.  Developing an autonomous inspection planner addresses these concerns.

Recently, several commercial solutions such as the ones from Skydio~\cite{skydio} and Exyn~\cite{Exyn_Technologies_undated-gq} and ongoing work in academia provide robust autonomy including SLAM and low-level planning (how to navigate from point A to point B). Our work on high-level planning (determining what the next waypoint B should be) is complementary to these works. Current forms of planning mostly consist of someone clicking on waypoints for the UAVs to fly to. As a result, we develop tools that autonomously solve the more general problem of inspecting infrastructure with no prior information about its geometry.

Inspection is closely related to coverage and exploration, which are problems that have been well-studied in the literature. However, coverage and exploration are not necessarily the best approaches for inspection. Given a 3D model of the environment (including the infrastructure), we can find a coverage path that covers all points on the infrastructure using an offline planner~\cite{9048979}. In practice, we often do not have any prior model of the layout of the infrastructure. Even if a prior 3D model is available, it may be inaccurate due to changes in the environment surrounding the infrastructure as well as structural changes made to the infrastructure. In this chapter, we address the problem of designing targeted inspection plans as the 3D model of the environment is built online. 

\begin{figure}
    \centering
    \includegraphics[width = \columnwidth]{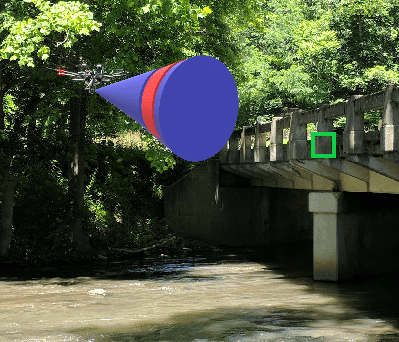}
    \caption{An example of $View$. Our goal is to ensure that each point on the surface is viewed by the onboard camera within the \emph{View} region. The green box depicts the face of an infrastructure voxel, the blue cone depicts the viewing cone, and the red band on the cone depicts the viewing distance.}
    \label{fig:vcvd_gatsbi}
\end{figure}

Frontier-based strategies~\cite{yamauchi1997frontier} are typically used for exploring an initially unknown environment. A frontier is a boundary between explored and unexplored regions. The strategy chooses which of the frontiers to visit (and which path to follow to get to the chosen frontier) to help speed up exploration. The algorithm terminates when there are no more accessible unexplored regions. A bounding box placed around the infrastructure can restrict exploration when operating in an open environment. Exploring the infrastructure does not necessarily mean that the UAV will get inspection-quality images. Instead, an inspection planner that can take into account viewing and distance constraints may be more efficient. We present such a planner, termed \emph{GTSP-Based Algorithm for Targeted Surface Bridge Inspection (GATSBI)}, and show that it outperforms previous inspection strategies in efficiently inspecting the infrastructure (Section~\ref{sec:eval_gatsbi}).

GATSBI consists of six modules: semantic segmentation to find 3D points from a pointcloud that corresponds to the surface of the infrastructure, simultaneous localization and mapping (SLAM) to align each incoming 3D pointcloud, 3D occupancy grid mapping using the 3D pointcloud, a high-level planner for finding inspection paths for the UAV, a navigation algorithm for executing the planned path, and a crack detection neural network to detect defects. We use off-the-shelf modules for SLAM (LIO-SAM~\cite{liosam2020shan}), occupancy grid mapping (OctoMap~\cite{hornung2013octomap}), solving GTSP (GLNS~\cite{Smith2016GLNS}), point-to-point navigation (MoveIt~\cite{coleman2014reducing}), and crack detection (YOLO-World~\cite{Cheng_2024_CVPR}). The specific point-to-point planner we use is RRT*. Since the distances generated by RRT* might be different than the Euclidean distance, we perform a lazy evaluation before navigation described in Section~\ref{sec:GATSBI}. Our key algorithmic contribution is the high-level planner module. Specifically, we show how to show \emph{how} to reduce the inspection problem to a GTSP instance and the full pipeline that outperforms baseline strategies. Our technique takes into account overlapping viewpoints that can view the same parts of the infrastructure and simultaneously selects \emph{where} to take images and \emph{what order} to visit those viewpoints. In summary, we make the following contributions:

\begin{itemize}
    \item Present our receding horizon algorithm, GATSBI~\cite{dhamiGATSBI}, an efficient infrastructure inspection planner with no prior structural information;
    \item Demonstrate that GATSBI outperforms a baseline inspection method that knows the infrastructure surface a priori by \textbf{38\%} through numerous simulations in AirSim using 3D models of infrastructure;
    \item Validate the practical feasibility in experiments with a practical version of GATSBI;
    \item Provide ROS packages that integrate MoveIt with AirSim for simulations and DJI's SDK for real-world experiments.\footnote{\url{https://github.com/raaslab/GATSBI}}
\end{itemize}

\section{Related Work}\label{sec:related}

In this section, we review some of the previous work in this domain. We first describe some of the recent work on the planning side and then describe recent work on crack detection and localization.
\subsection{Planning}

As described earlier, frontier-based exploration is a widely-used method for 3D exploration of unknown environments~\cite{zhu20153d,da2020novel,niroui2017robot}. Other works proposed variants of frontier exploration focused on choosing the next frontier to visit~\cite{dai2020fast,shen2012autonomous}. Another popular approach is to model the exploration problem as one of information gathering and choose a path (or a frontier) that maximizes the information gain~\cite{corah2019communication,premkumar2020combining}. Additionally, there are Next-Best-View approaches~\cite{pito1999solution, dhami2023prednbv, dhami2023mapnbv} that, as the name suggests, plan the next-best location to take an image from to explore the environment. We refer the reader to a recent, comprehensive survey on multi-robot exploration by Li~\cite{li2020exploration} that covers a variety of exploration strategies. There is also work that uses GTSP for view planning~\cite{tokekar2016algorithms}, however, it focuses on only 2D space and does not consider practical issues such as obstacle avoidance. 

As shown in our conference paper~\cite{dhamiGATSBI}, generic exploration strategies are inefficient when performing targeted infrastructure inspection. There has been work on designing inspection algorithms that plan paths that take into account the viewpoint considerations~\cite{peng2019adaptive,hollinger2013active,roberts2017submodular,song2020online}. When prior information is available, one can plan inspection paths carefully by considering the geometric model of the environment. Typically, algorithms use prior information, such as a low-resolution version of the environment, to create an inspection path and obtain high-resolution measurements of the environment~\cite{peng2019adaptive,roberts2017submodular}. Unlike these works, we consider a scenario where the robot has no prior environmental information and must plan using incrementally revealed data.

Bircher et al.~\cite{bircher2018receding} presented a receding horizon planner for exploration and inspection. Both algorithms use a Rapidly-Exploring Random Tree to generate a set of candidate paths in the known, free space of the environment. Then the algorithm selects a path based on a criterion that values how much information a path gains about the environment. The planner uses a receding horizon algorithm repeatedly invoked with new information. We follow a similar approach; however, their algorithm knows the inspection surface a priori, \revone{while our work only requires the UAV to see a portion of the infrastructure at the start of inspection}. The environment is not known for their exploration algorithm but it is known for their inspection algorithm. GATSBI has no prior knowledge about the environment and \revone{minimal knowledge of the} target inspection surface. Furthermore, we cluster potential viewpoints using GTSP which leads to further efficiency. Their inspection planner, Structural Inspection Planner (SIP)~\cite{BABOOMS_ICRA_15}, is the baseline inspection strategy that GATSBI is compared against in Section~\ref{sec:eval_gatsbi}.

Song et al.~\cite{song2020online} recently proposed an online algorithm that consists of a high-level coverage planner and a low-level inspection planner. The low-level planner takes into account the viewpoint constraints and chooses a local path that gains additional information about the structure under inspection. Our work differentiates by guaranteeing the quality of inspection, not requiring a bounding box around the target infrastructure, and segmenting the infrastructure of interest from the environment which guarantees inspection of only the target infrastructure.

\subsection{Crack Detection and Localization}
Detecting cracks and defects in the infrastructure with artificial intelligence (AI)  has been a subject of interest in recent years. Onboard deployment of such defect detection systems on aerial robots requires consideration of accuracy, inference speed, and lack of data. The earlier versions of defect detection methods use CNN-based approaches such as YOLO~\cite{alfarrarjeh2018deep, faramarzi2020road, kuang2020computer}, Faster R-CNN~\cite{cha2017deep}, and SSD~\cite{maeda2018road}. Subsequently, the development of more sophisticated networks led to more diverse approaches using architectures based on Graph Neural Networks~\cite{shang2022superpixel} and Transformers~\cite{liu2021crackformer}. While these approaches can provide good detection and localization accuracy, they also require huge training datasets, which may not be infeasible for specific defects such as cracks in bridges. Also, more complex and larger models result in increased inference time, making them less attractive for real-time applications.

The existing datasets for crack detection such as CrackTree260~\cite{zou2012cracktree},  CrackLS315~\cite{zou2018deepcrack}, Stone331~\cite{konig2021optimized}, etc., contain only a few hundred images. Data augmentation and transfer learning~\cite{nath2021s2d2net} are often used to tackle the lack of data transfer learning. Most of these networks for defect inspection predict a binary segmentation map~\cite{zavrtanik2021draem, bovzivc2021mixed, li2021cutpaste, bovzivc2021end, tabernik2020segmentation, dougan2022new, li2021automatic, yang2019feature, liu2021crackformer, inoue2021crack}. Such pixel-to-pixel prediction networks can be computationally intensive and detecting localization may need post-processing. Bounding box predictors provide a good alternative for defect detection and localization~\cite{sun2022new, kuang2020computer, joshi2022automatic}. However, the small size of the training data remains a challenge for these tasks as well. To address these issues, we use a real-time open-set detector, pre-trained on large-scale data with text-based prompts. Specifically, we use the YOLO-World~\cite{Cheng_2024_CVPR} model with `crack' as the prompt for crack detection and localization. Unlike the previous works, we do not fine-tune this network and show that the resulting model provides appreciable detection and localization over real-world images, without needing any training or fine-tuning on task-specific datasets.

\section{Problem Formulation}\label{sec:prob}

The goal of our unified system is to find infrastructure defects, specifically cracks that can be detected visually, using a UAV. Assuming we have a UAV with a 3D pointcloud sensor and RGB camera, our goal is to find a path to inspect every point on the infrastructure surface while minimizing total flight distance.

We consider the scenario where the geometric model of the infrastructure may be unknown a priori. We assume that the UAV starts the algorithm at a location where at least some part of the infrastructure is visible. If this is not the case, we can run a frontier exploration strategy until the infrastructure is visible. We then plan an inspection path for the part of the infrastructure that is visible. As the UAV sees more of the infrastructure, we replan to find a better tour in a receding horizon fashion. 

We use a 3D semantic, occupancy grid built using localized pointcloud data to represent the model of the infrastructure built online. GATSBI assigns each voxel in the occupancy grid a semantic label. The label indicates whether the voxel is free space $v_{F} \in V_{F}$; is occupied space, part of the infrastructure, and previously inspected $v_{BI} \in V_{BI}$; is occupied, part of the infrastructure, but not yet inspected $v_{BN} \in V_{BN}$; and occupied but not an infrastructure voxel (i.e., obstacles)  $v_{O} \in V_{O}$. Our goal is to inspect all the voxels that correspond to the infrastructure surface, i.e., to ensure that $V_{BN} = \emptyset$. 

A voxel $v_{BN} \in V_{BN}$ is inspected if we inspect at least one of its six faces. A face is inspected if the center of that face falls within a cone given an apex angle centered at the UAV camera and within a minimum and a maximum range of the UAV camera. The apex angle represents the field of view of the camera that is rigidly attached to the UAV. The viewing distance is a minimum and maximum distance range that the UAV should inspect an infrastructure voxel to ensure quality images for inspection. Figure~\ref{fig:vcvd_gatsbi} shows an example of these viewing constraints. 
For the rest of the chapter, we refer to the viewing cone and a distance as $View$. The RGB camera is used to take pictures of the bridge once the UAV has reached a target $View$ point. The problem we are trying to solve can be defined as follows:

\paragraph*{Problem 1} \label{prob:1}
Given a 3D occupancy map consisting of four sets of voxels ($V_{F}, V_{BI}, V_{BN}, V_{O}$), find a minimum length path that inspects every voxel in $V_{BN}$.

We repeatedly solve~\nameref{prob:1} as we gain new information until $V_{BN} = \emptyset$. In Section~\ref{sec:GATSBI}, we show how to model this problem as a GTSP instance.

Given a collection of RGB images captured at each inspection point, we seek to localize defects along the visible surfaces of the infrastructure via two phases offline. The first phase uses a machine learning pipeline to propose potential defects and their locations within each image. In the next phase, a human expert accepts or rejects each proposed defect and determines if the defect requires repairs. We assume the existence, number, and locations of defects are unknown \textit{a priori}.

\section{System Overview} \label{sec:overview}
\begin{figure}
    \centering
    \includegraphics[width = \columnwidth]{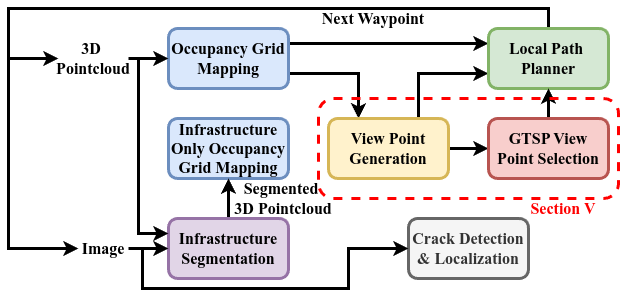}
    \caption{Flow diagram of GATSBI. \revone{The algorithm creates an occupancy map of the environment using incoming LiDAR scans. Then, it segments the points corresponding to the bridge into another point cloud using the RGB camera images. It then makes another occupancy map of only the bridge using the segmented point cloud. GATSBI uses both the environment and bridge occupancy maps to generate viewpoints, points in free space where the UAV can inspect the bridge. It sends these to the GTSP instance to make a tour and then a local path planner to get the flight path.}}
    \label{fig:flowDiagram}
\end{figure}

In this section, we give an overview of the GATSBI algorithm. We show the full pipeline (Fig.~\ref{fig:flowDiagram}) which broadly consists of three perception modules (segmentation, SLAM, and occupancy grid mapping) and two planning modules (high-level GTSP inspection planning and low-level point-to-point planning). Data captured during the flight is fed into the 6th module, crack detection. The main algorithmic contribution, the high-level planner, is described in the next section~\ref{sec:GATSBI}. Here we describe all other off-the-shelf components that form the full system.

\subsection{Perception} \label{sec:perception}

\begin{figure}
    \centering
    \includegraphics[width = \columnwidth]{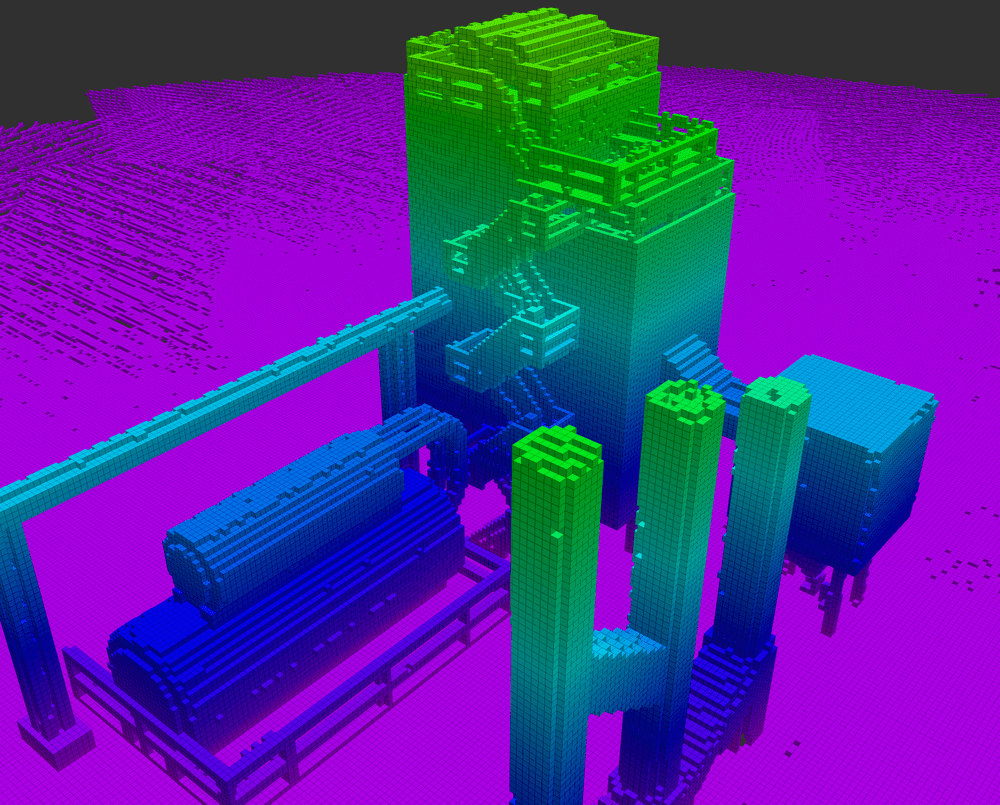}%
    \caption{Full voxel map containing $V_{BI}$ (inspected infrastructure voxels), $V_{BN}$ (uninspected infrastructure voxels), and $V_O$ (obstacle voxels).}%
    \label{fig:fullAndSeg}%
\end{figure}

\subsubsection{Simultaneous Localization and Mapping}
To start, the raw pointcloud data is localized and mapped. Localizing the pointcloud allows for the alignment of each successive pointcloud to all previous ones which is subsequently used to build a map of the environment. As the UAV flies to new areas, more and more of the environment gets mapped including the target infrastructure. 

\subsubsection{Segmentation}
We then use the localized 3D points to segment the infrastructure from the environment. Segmenting out the infrastructure allows the algorithm to differentiate between the infrastructure and obstacles in the environment. This allows only the infrastructure to be inspected as opposed to every object in the environment. The 3D points that lie on the segmented infrastructure are classified as infrastructure points. 

\subsubsection{Occupancy Grid Mapping}
These 3D localized infrastructure points are then used to create a 3D occupancy grid. The map produced by the SLAM module lies in continuous space whereas utilizing occupancy grid mapping allows us to discretize the 3D space. In parallel, the complete point cloud (segmented infrastructure and non-segmented obstacle points) is used to generate an environmental 3D occupancy grid. Together, these two occupancy grids output a set of voxels: free $V_F$, bridge $V_{BI}$, bridge $V_{BN}$, and obstacle $V_O$. The algorithm uses the segmented voxels ($V_{BI}, V_{BN}$) to plan inspection paths. The algorithm uses the other voxels ($V_{O}$, $V_{F}$) to plan collision-free paths and take into account viewing constraints. An example of a 3D occupancy grid is shown in Fig.~\ref{fig:fullAndSeg}.

\subsection{Planner} \label{sec:planner}

\subsubsection{High-Level GTSP Inspection Planning}
To inspect infrastructure, we need to inspect all voxels in $V_{BN}$ (as described in Section~\ref{sec:prob}). GATSBI works in a receding horizon fashion. The $V_{BI}$ set keeps track of inspected voxels. This avoids unnecessarily inspecting the same voxel more than once. Specifically, the UAV must view each voxel in $V_{BN}$ from some point on its path within $View$. We formulate this problem as a GTSP instance. The specific details on how we formulate it as a GTSP instance are described in Section~\ref{sec:GATSBI}. 

\subsubsection{Low-Level Point-to-Point Planning}
Once a GTSP tour is received from the high-level planner, point-to-point planning is needed. The GTSP tour does not guarantee collision-free paths, it only gives us points where all of the uninspected infrastructure surface can be inspected. Using our 3D environmental occupancy grid, we can generate collision-free paths between each point in the GTSP tour with a point-to-point planner. 

\subsection{Crack Detection}
While the UAV is flying on its inspection path, images from the flight are captured. We can use these images to detect defects in the infrastructure. For our purposes, we focus on crack detection specifically. The raw images can be input into a crack detection neural network where crack locations are output. Since we have a localization and mapping module, the exact location of these cracks can also be tagged in our environment map. 

\section{The GATSBI Planner} \label{sec:GATSBI}
We describe our high-level GTSP inspection planner. As shown in Figure~\ref{fig:flowDiagram}, this is done in two steps: viewpoint generation and GTSP-based viewpoint selection. GTSP generalizes the Traveling Salesperson Problem and is NP-Hard~\cite{Smith2016GLNS}. The input to GTSP consists of a weighted graph, $G$, where vertices are clustered into sets.  The edges of the graph are the distances between the vertices. The objective is a minimum weight tour that visits at least one vertex in each set once. Next, we describe our GTSP setup details. 

\subsection{Viewpoint Generation}
For our implementation, vertices are the center-points of voxels $v_F$ that the UAV can fly to and clusters are the set of all $v_F$ a specific $v_{BN}$ can be inspected from. Each vertex in $G$ corresponds to a candidate viewpoint. We check all pairs of $v_F \in V_F$ and $v_{BN} \in V_{BN}$ to see if $v_F$ lies within $View$ of one of the faces of $v_{BN}$. If so, we add a vertex in the graph $G$ corresponding to the pair $v_F$ and $v_{BN}$. 

\begin{figure}
    \centering
    \includegraphics[width = \columnwidth]{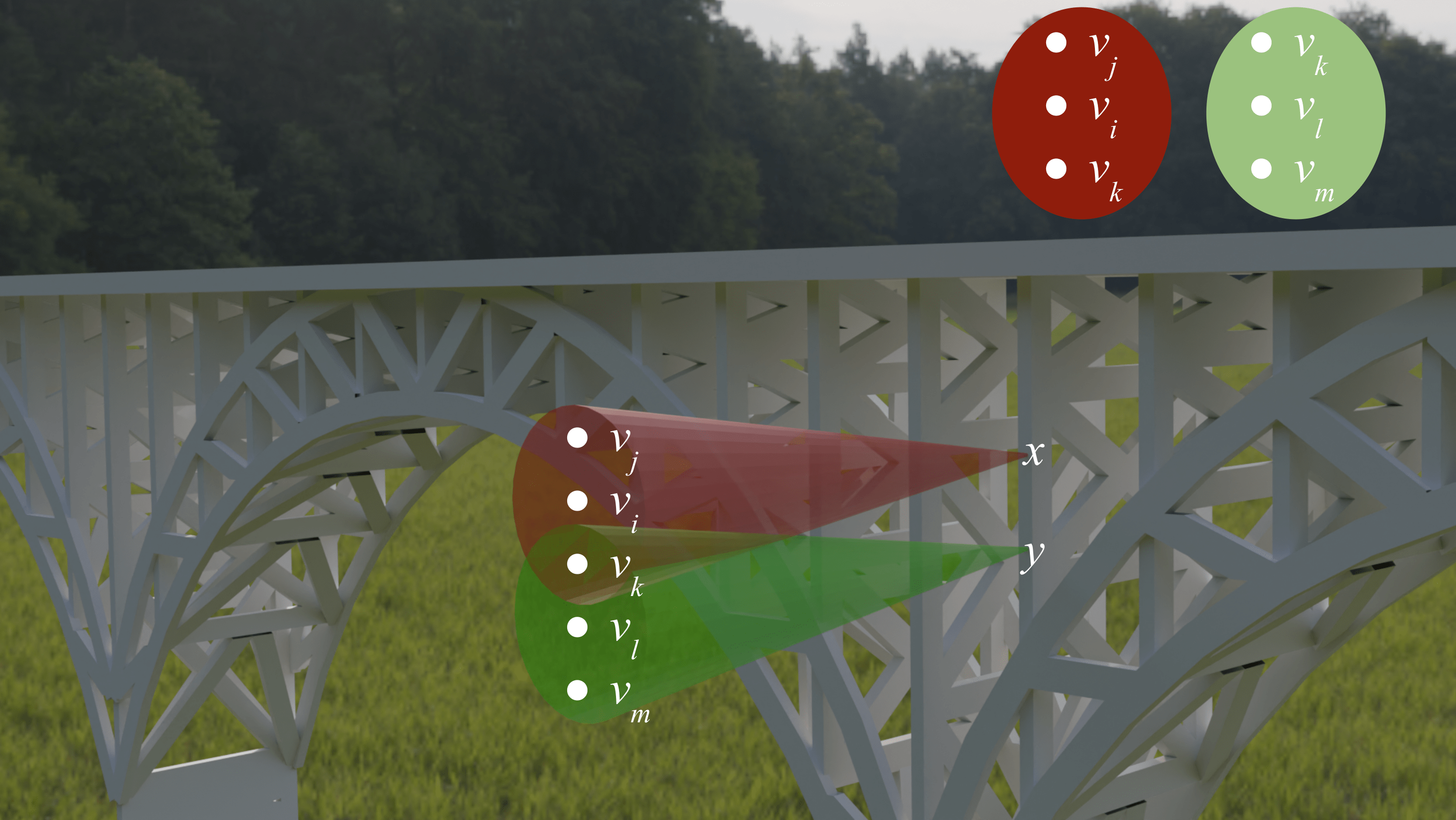}%
    \caption{Example GTSP setup for two infrastructure voxels. Each inspectable infrastructure voxel (x and y) can have multiple potential inspection viewpoints (vertices $v_i$ - $v_m$). All the vertices for a single infrastructure voxel are clustered together (red and green ovals). The edges between these vertices are initially their Euclidean distance.}%
    \label{fig:gtsp_example}%
\end{figure}

Each free voxel that can inspect the same $v_{BN}$ will add one vertex each to the cluster corresponding to $v_{BN}$. A simplified example of this graph setup is shown in Fig.~\ref{fig:gtsp_example}. The GTSP tour will ensure the UAV visits at least one viewpoint in each cluster.

Next, we create an edge between every pair of vertices in $G$. The cost for each of these edges is initially the Euclidean distance between the two vertices.  With the vertices, edges, and clusters, we create a GTSP instance and use the GTSP solver, GLNS~\cite{Smith2016GLNS}, to find a path for the UAV. We use GTSP as our planner because it will guarantee at least one point corresponding to every $V_{BN}$ will be visited.

\subsection{GTSP-Based Viewpoint Selection}
Before moving from one point to the next, we check the distance from the current location of the UAV to the next vertex in the GTSP path. Instead of Euclidean distance, we find the distance of the path between these points using an RRT* algorithm~\cite{rrtstar} with our environment occupancy grid. This ensures the path between these points is collision-free. If the difference between the RRT* distance and the Euclidean distance is greater than $DD$ (discrepancy distance), we update the edge costs in the GTSP instance and replan the GTSP path. We replan as needed to ensure that the first edge in the returned tour is within $DD$ of the Euclidean distance. We call this a lazy evaluation of edge costs. Computing the RRT* distance (which is a more accurate approximation of the actual travel distances) between every pair of vertices would be time-consuming. By checking the discrepancy lazily, we find a tour quickly while also not executing any edge where the actual distance is significantly larger than the expected distance. For our experiments, we set $DD$ to be 125\% of the Euclidean distance to account for some of the variances in paths generated using RRT* algorithms but still allow replanning when necessary.

\begin{algorithm}
\caption{Overview of single iteration of GATSBI}\label{alg:gatsbi}
\begin{algorithmic}[1]
\State{Update occupancy grids with latest localized pointcloud as described in Section~\ref{sec:perception}}
\State{Find all inspectable bridge voxels using occupancy grids and remove previously inspected bridge voxels, resulting in $V_{BN}$}
\If{$V_{BN} = \emptyset$}
\State{Terminate}
\EndIf
\State{Create GTSP instance G as described in Section~\ref{sec:GATSBI}}
\While{Difference in the RRT* distance of the first edge in the GTSP solution and its Euclidean distance is greater than a threshold}
\State{Update first edge cost in G with RRT* distance and re-solve GTSP}
\EndWhile
\State{Use RRT* as point-to-point planner for GTSP tour}
\State{Update $V_{BI}$ with latest inspected bridge voxels}
\end{algorithmic}
\end{algorithm}

We keep track of each newly visited cluster during the flight. Each of these newly visited clusters corresponds to a non-inspected infrastructure voxel. The camera is also used to take an image at each visited point in the path to obtain inspection images. Once inspected, GATSBI moves them from set $V_{BN}$ to $V_{BI}$. We execute the plan until one of two conditions is met: either a time limit ($RPT$) elapses, or we complete the path, whichever occurs first. We also record the raw sensor data during the flight. Once we complete navigation, we use the stored data to update the infrastructure inspected and non-inspected voxels and replan. Once $V_{BN}$ is empty, GATSBI considers the infrastructure inspected and terminates. An overview of a single iteration of the GATSBI algorithm can be viewed in Algorithm~\ref{alg:gatsbi}.

\section{Evaluation}\label{sec:eval_gatsbi}
In this section, we evaluate the algorithm in both simulation and hardware experiments. For the simulations, we compare it against SIP~\cite{BABOOMS_ICRA_15} and against a baseline frontier-exploration algorithm. We also evaluate parameter tuning. First, we describe the common setup that was used in both simulations and experiments.

\subsection{Setup}
For both simulation and experiments, we generated the occupancy grid using 3D pointclouds and input them into Octomap~\cite{hornung2013octomap}. We used the MoveIt~\cite{coleman2014reducing} software package based on the work done by Köse~\cite{tahsinko86:online} to implement the RRT* algorithm. MoveIt uses RRT* and the environmental 3D occupancy grid to find collision-free paths for point-to-point navigation. Finally, we used a viewing cone with an apex angle of 20\degree and a viewing distance between two to five meters. This restriction ensured the viewing cone was within the camera's field of view (FoV). We also set the viewing distance based on Dorafshan et al.~\cite{dorafshan2017fatigue}, where they suggest a minimum flight distance of two meters to allow for the safe flight of the UAV. For crack detection, we use YOLO-World~\cite{Cheng_2024_CVPR} to detect cracks in images captured during flight.

\subsection{Simulations}\label{sec:sim}
We present simulation results to evaluate the performance of our proposed unified inspection system. We discuss the setup used to perform simulations and then discuss the inspection environments. Next, we compare GATSBI with the frontier exploration baseline and with SIP~\cite{BABOOMS_ICRA_15} first quantitatively and then qualitatively.

\begin{figure*}
    \begin{minipage}{0.05\linewidth}\centering
        \vspace{1cm}
        \begin{subfigure}[t]{\textwidth}
            \rotatebox[origin=center]{90}{                }
        \end{subfigure}
        \vspace{1cm}
        \begin{subfigure}[t]{\textwidth}
            \rotatebox[origin=center]{90}{GATSBI}
        \end{subfigure}
        \begin{subfigure}[t]{\textwidth}
            \rotatebox[origin=center]{90}{SIP~\cite{BABOOMS_ICRA_15}}
        \end{subfigure}
    \end{minipage}
    \begin{minipage}{0.9\linewidth}\centering
    \begin{subfigure}[t]{0.13\textwidth}
        \begin{subfigure}[t]{\textwidth}
            \includegraphics[width=\textwidth]{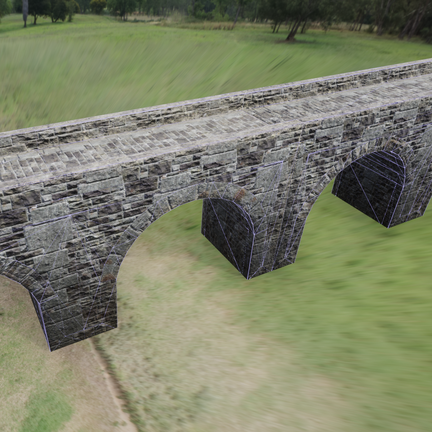}
        \end{subfigure}\vspace{.6ex}
        \begin{subfigure}[t]{\textwidth}
            \includegraphics[width=\textwidth]{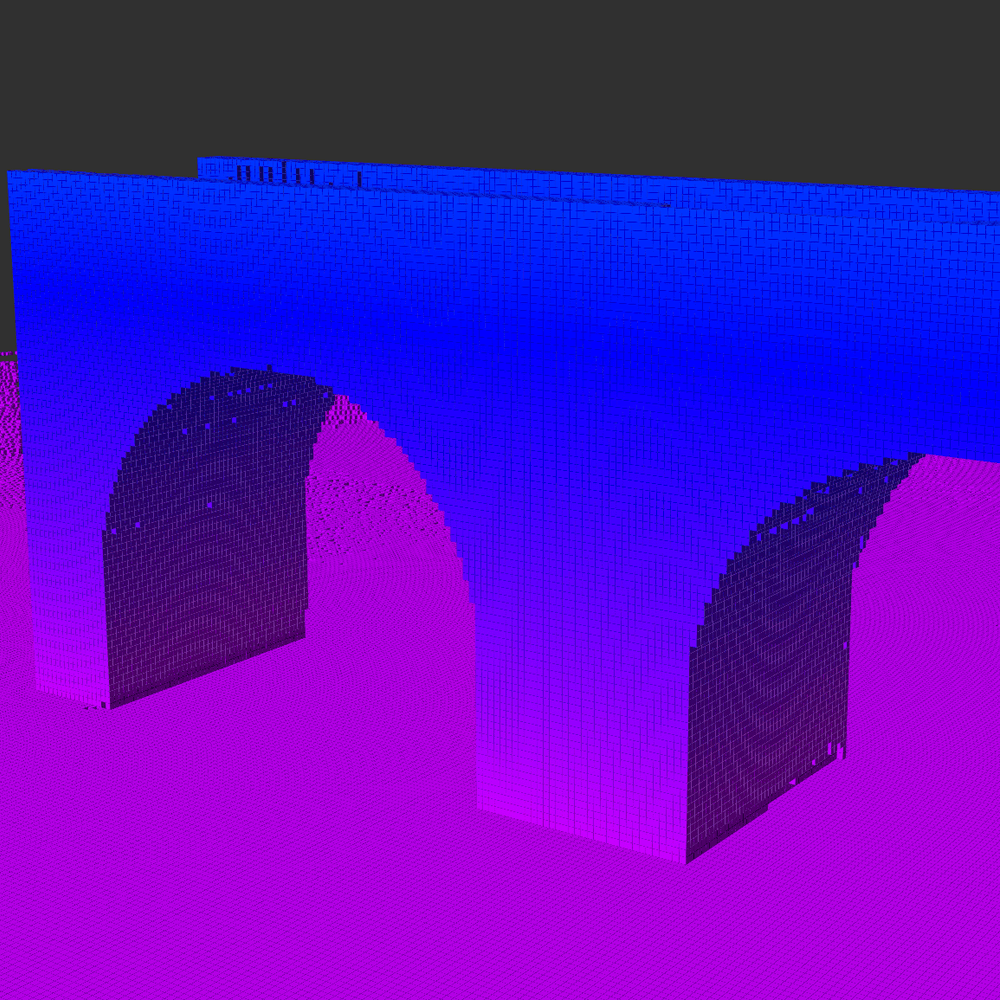}
        \end{subfigure}\vspace{.6ex}
        \begin{subfigure}[t]{\textwidth}
            \includegraphics[width=\textwidth]{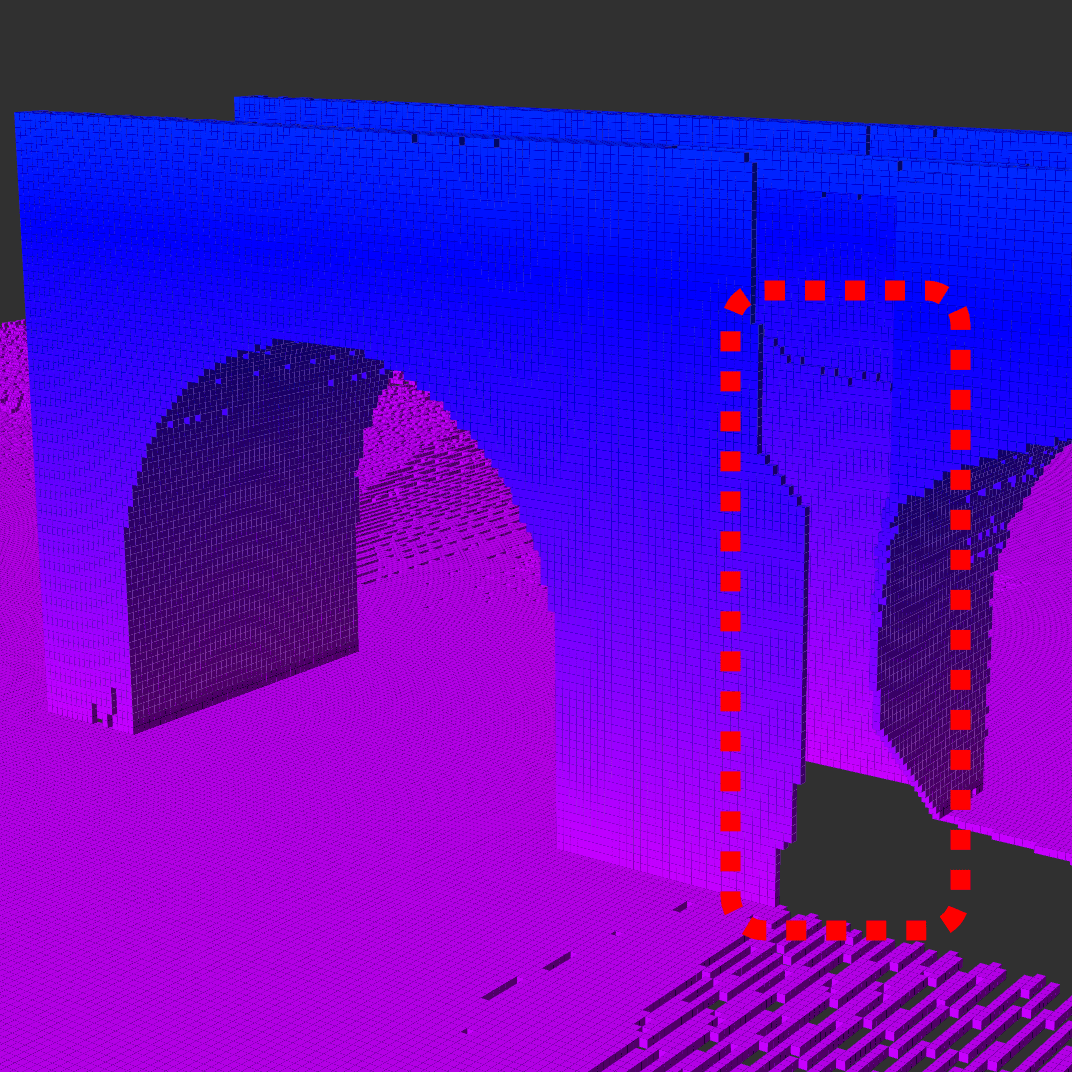}
        \end{subfigure}\vspace{.6ex}
        \caption{Arch}
    \end{subfigure}
    \hfill
    \begin{subfigure}[t]{0.13\textwidth}
        \begin{subfigure}[t]{\textwidth}
            \includegraphics[width=\textwidth]{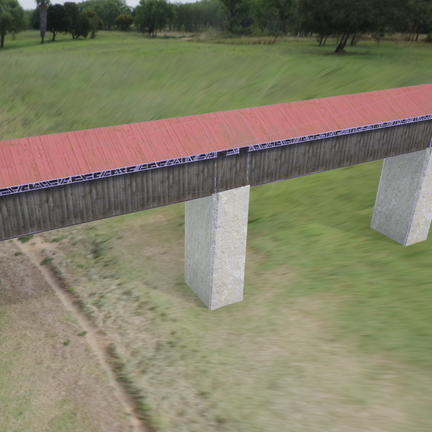}
        \end{subfigure}\vspace{.6ex}
        \begin{subfigure}[t]{\textwidth}
            \includegraphics[width=\textwidth]{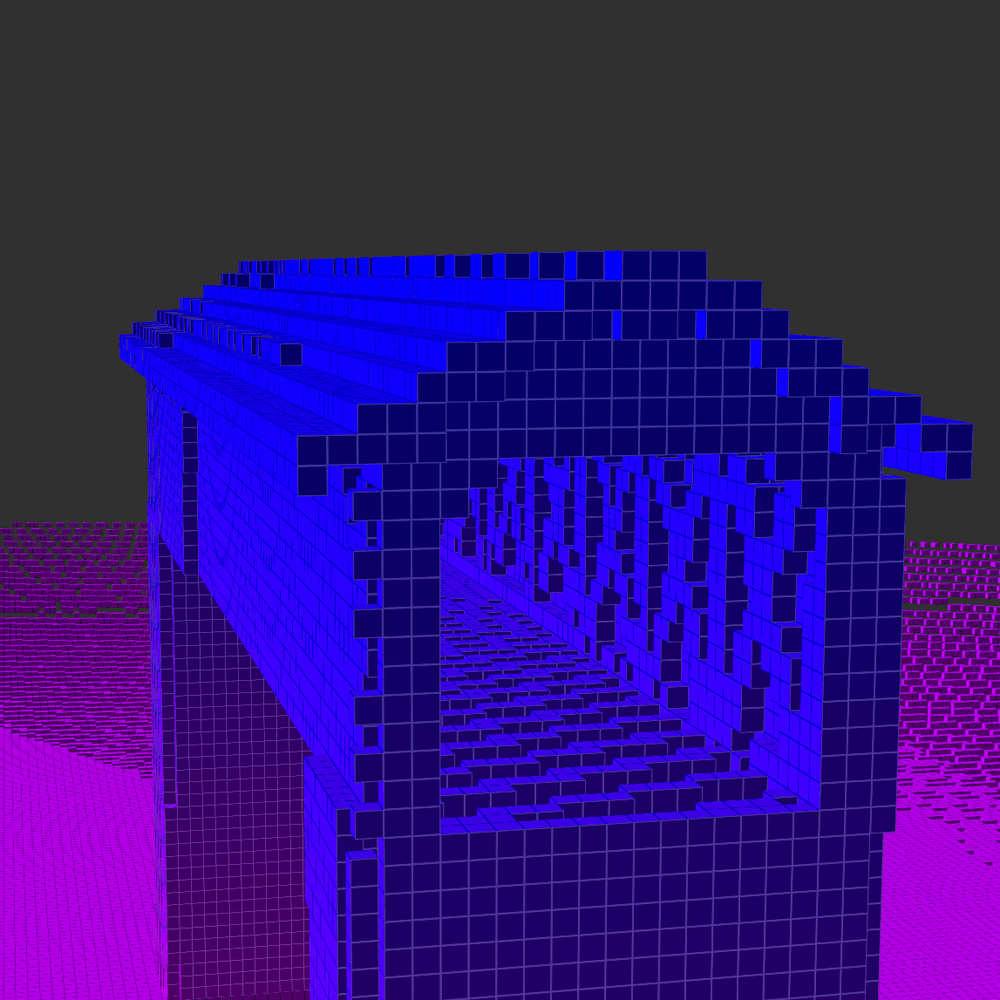}
        \end{subfigure}\vspace{.6ex}
        \begin{subfigure}[t]{\textwidth}
            \includegraphics[width=\textwidth]{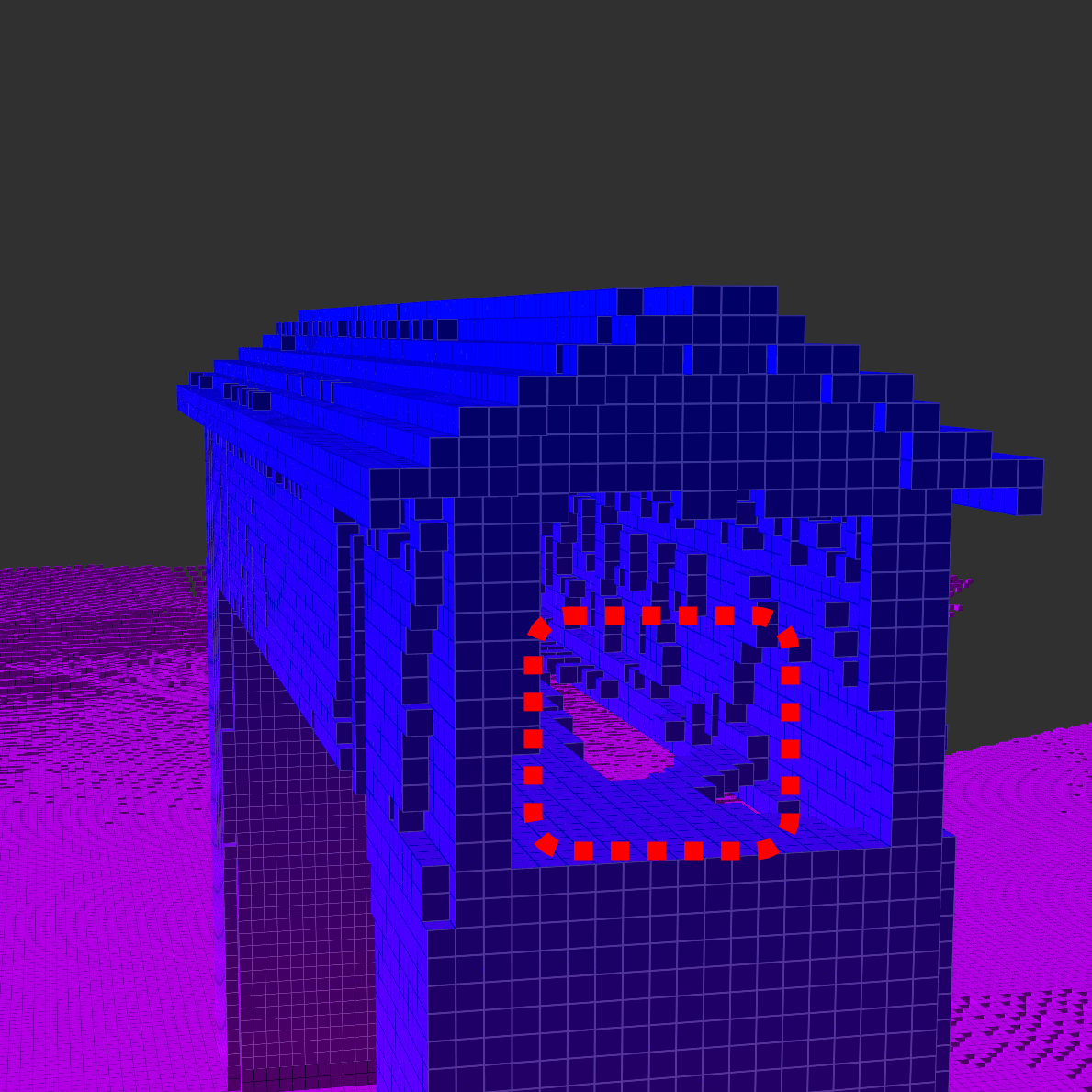}
        \end{subfigure}\vspace{.6ex}
        \caption{Covered}
    \end{subfigure}
    \hfill
    \begin{subfigure}[t]{0.13\textwidth}
        \begin{subfigure}[t]{\textwidth}
            \includegraphics[width=\textwidth]{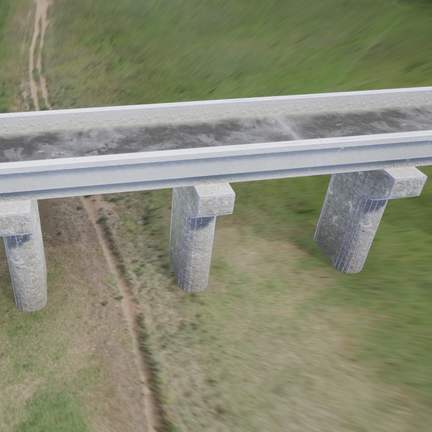}
        \end{subfigure}\vspace{.6ex}
        \begin{subfigure}[t]{\textwidth}
            \includegraphics[width=\textwidth]{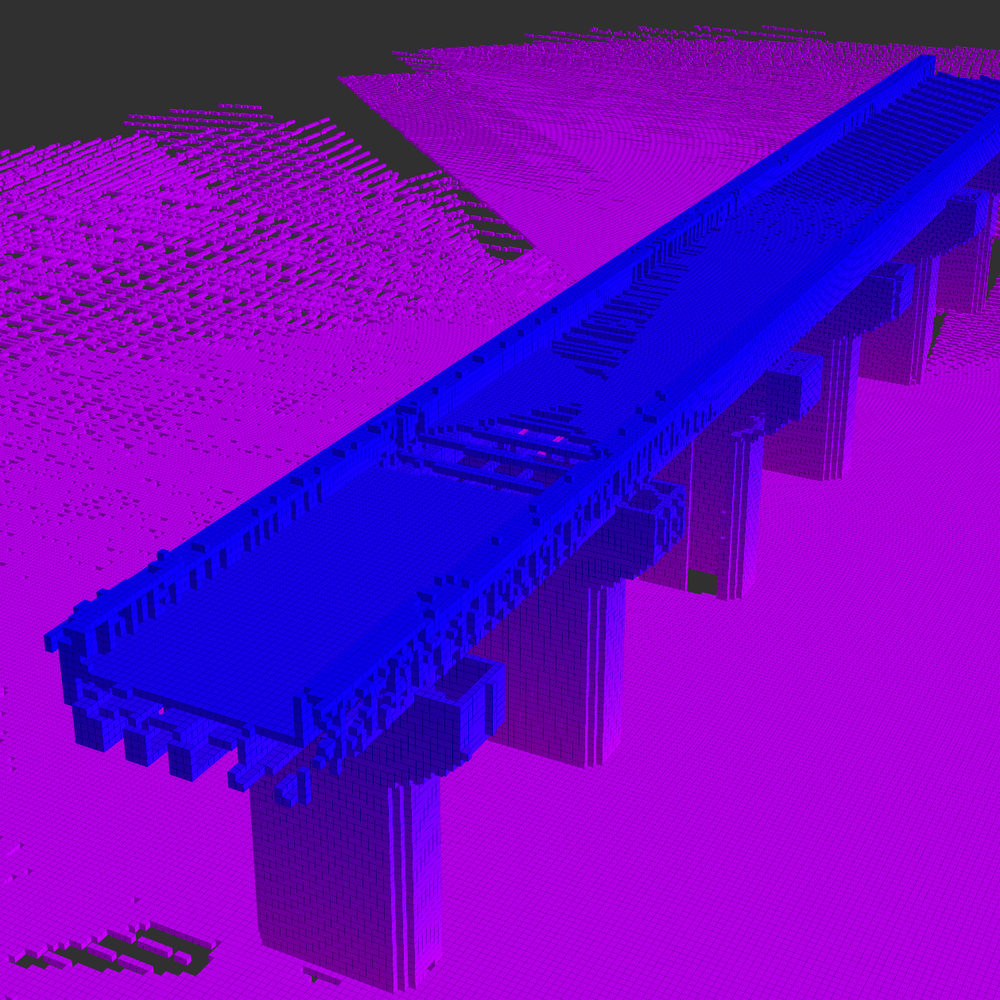}
        \end{subfigure}\vspace{.6ex}
        \begin{subfigure}[t]{\textwidth}
            \includegraphics[width=\textwidth]{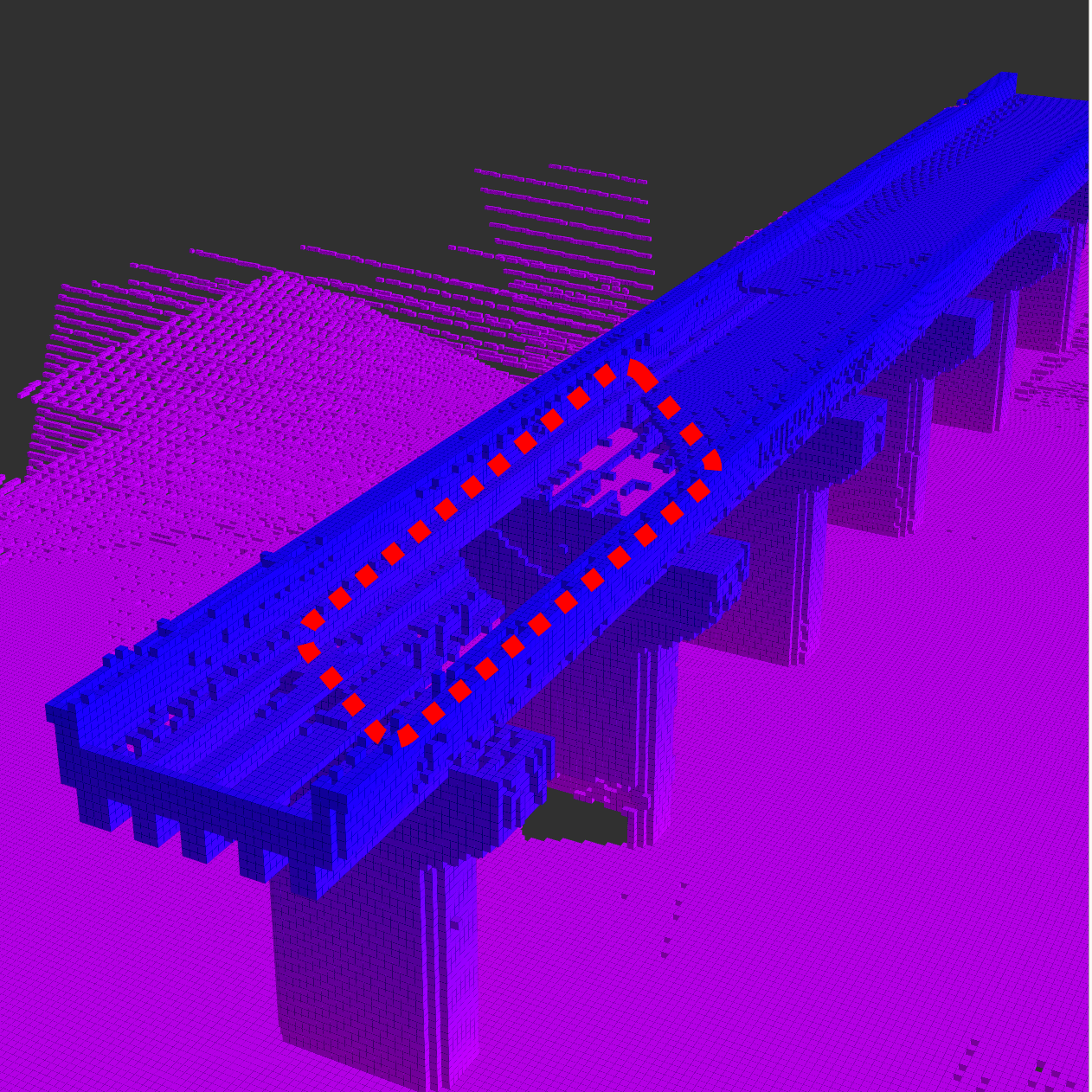}
        \end{subfigure}\vspace{.6ex}
        \caption{Girder}
    \end{subfigure}
    \hfill
    \begin{subfigure}[t]{0.13\textwidth}
        \begin{subfigure}[t]{\textwidth}
            \includegraphics[width=\textwidth]{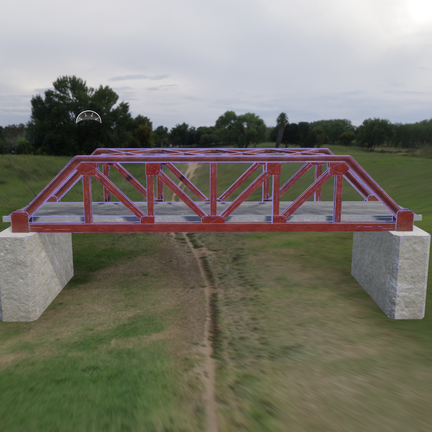}
        \end{subfigure}\vspace{.6ex}
        
        \begin{subfigure}[t]{\textwidth}
            \includegraphics[width=\textwidth]{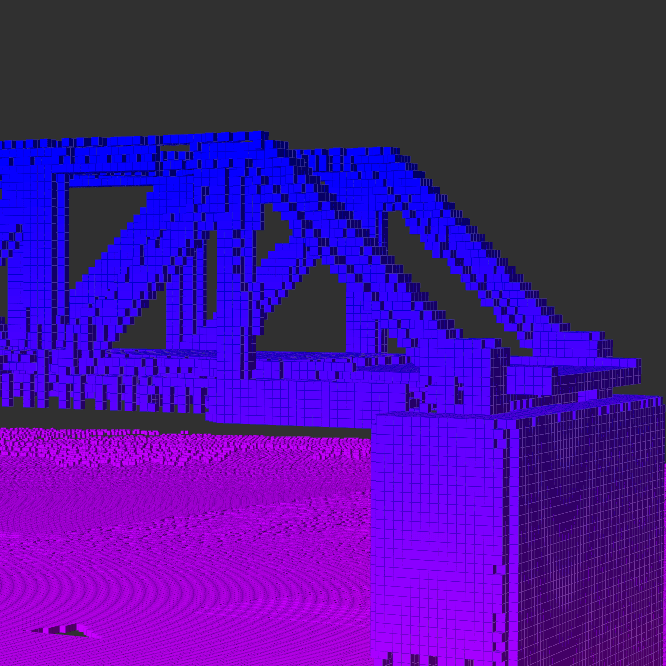}
        \end{subfigure}\vspace{.6ex}
        \begin{subfigure}[t]{\textwidth}
            \includegraphics[width=\textwidth]{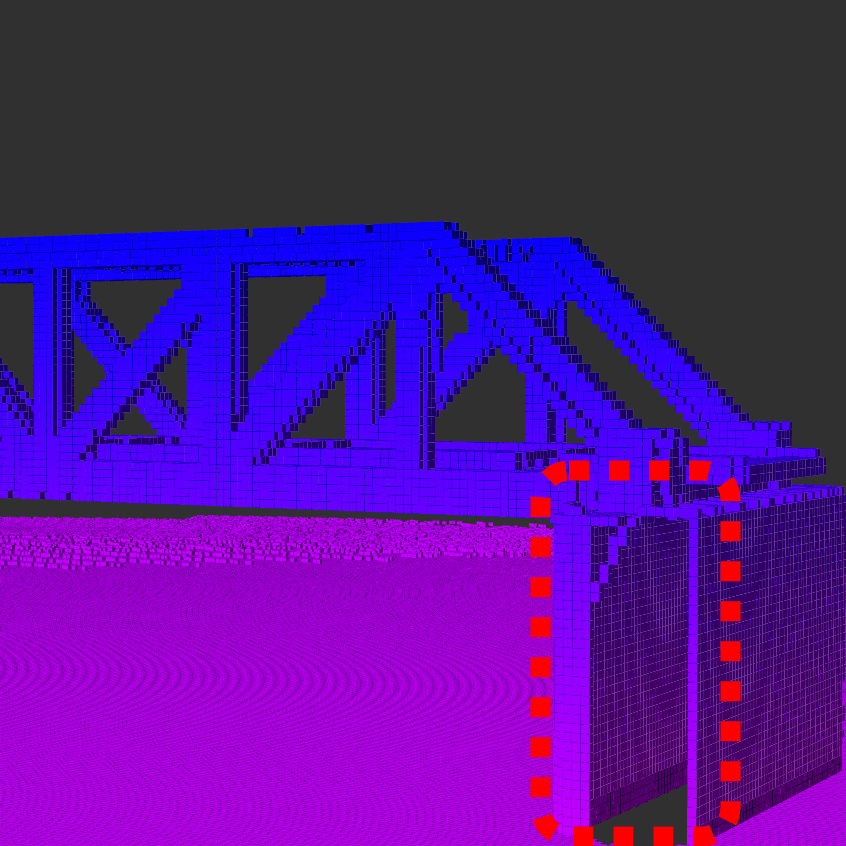}
        \end{subfigure}\vspace{.6ex}
        \caption{Iron}
    \end{subfigure}
    \hfill
    \begin{subfigure}[t]{0.13\textwidth}
        \begin{subfigure}[t]{\textwidth}
            \includegraphics[width=\textwidth]{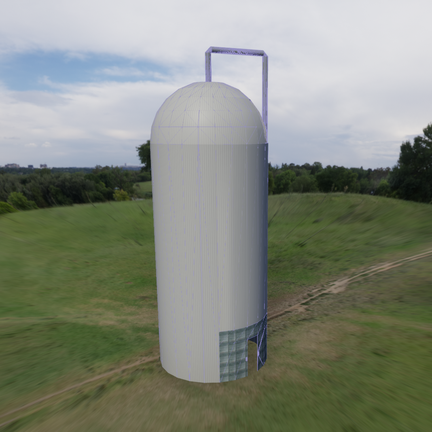}
        \end{subfigure}\vspace{.6ex}
        \begin{subfigure}[t]{\textwidth}
            \includegraphics[width=\textwidth]{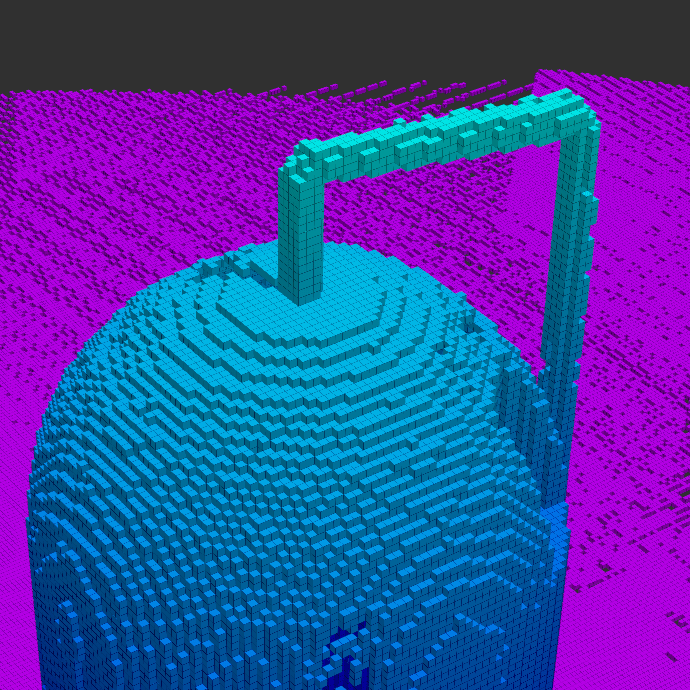}
        \end{subfigure}\vspace{.6ex}
        \begin{subfigure}[t]{\textwidth}
            \includegraphics[width=\textwidth]{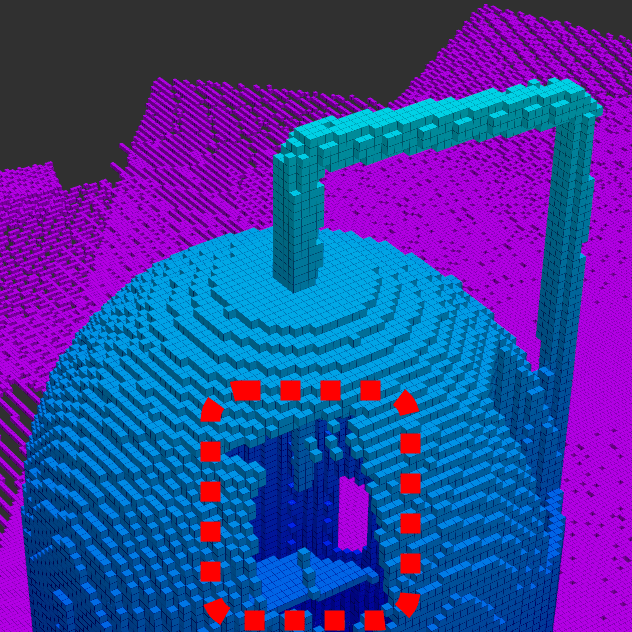}
        \end{subfigure}\vspace{.6ex}
        \caption{Silo}
    \end{subfigure}
    \hfill
    \begin{subfigure}[t]{0.13\textwidth}
        \begin{subfigure}[t]{\textwidth}
            \includegraphics[width=\textwidth]{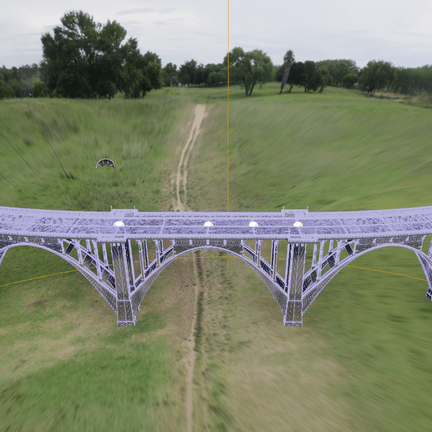}
        \end{subfigure}\vspace{.6ex}
        \begin{subfigure}[t]{\textwidth}
            \includegraphics[width=\textwidth]{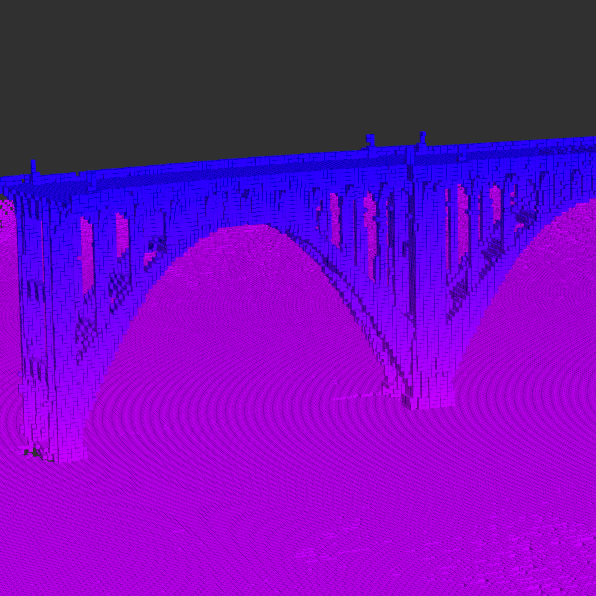}
        \end{subfigure}\vspace{.6ex}
        \begin{subfigure}[t]{\textwidth}
            \includegraphics[width=\textwidth]{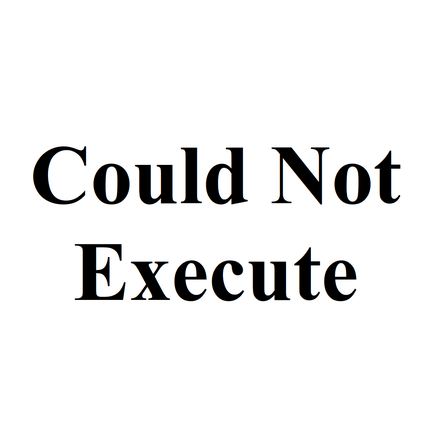}
        \end{subfigure}\vspace{.6ex}
        \caption{Steel}
    \end{subfigure}
    \hfill
    \begin{subfigure}[t]{0.13\textwidth}
        \begin{subfigure}[t]{\textwidth}
            \includegraphics[width=\textwidth]{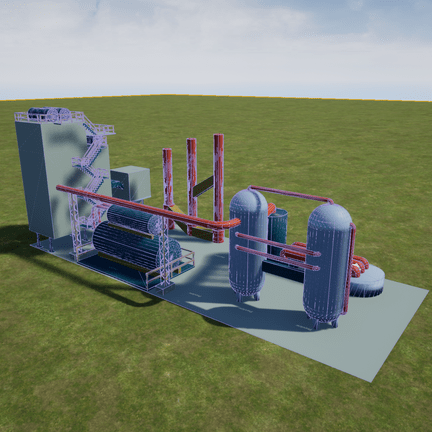}
        \end{subfigure}\vspace{.6ex}
        \begin{subfigure}[t]{\textwidth}
            \includegraphics[width=\textwidth]{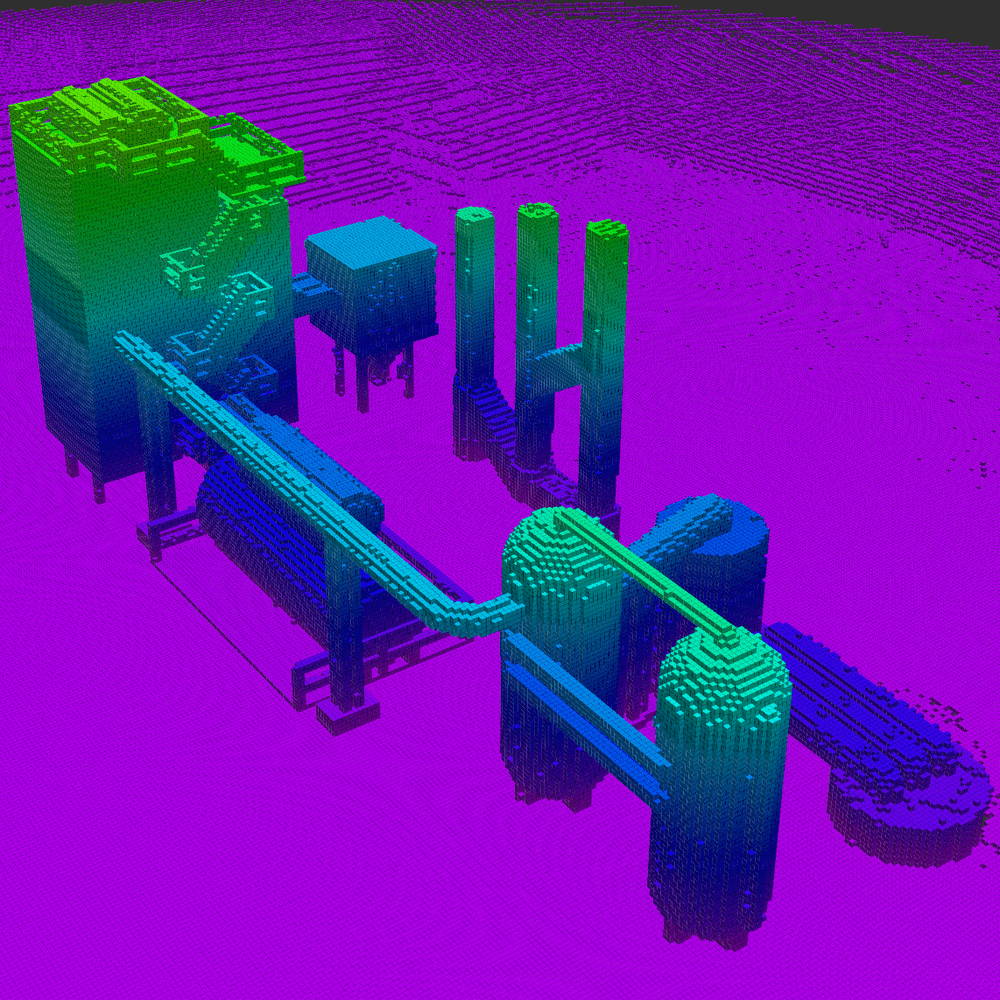}
        \end{subfigure}\vspace{.6ex}
        \begin{subfigure}[t]{\textwidth}
            \includegraphics[width=\textwidth]{figs/Qual_NA.png}
        \end{subfigure}\vspace{.6ex}
        \caption{Plant}
    \end{subfigure}
    \end{minipage}
    \caption{Qualitative comparison between GATSBI (top) and SIP~\cite{BABOOMS_ICRA_15} (bottom) in the seven simulation environments. The top row shows RGB images of the seven different environments. The middle row shows the high-resolution reconstruction after GATSBI's inspection flight. The bottom row shows the high-resolution reconstruction after SIP's reconstruction flight along with coverage gaps shown in the red-dotted boxes.}
    \label{fig:qual_results}
\end{figure*}

\subsubsection{Simulation Setup}
We ran the simulations on a laptop with an Intel Core i9-8950HK CPU, 32 GB of RAM, and an Nvidia RTX 2080 Max-Q GPU running Ubuntu 18.04. All simulated experiments were performed using the Robot Operating System (ROS) Melodic and AirSim~\cite{airsim2017fsr}. Our simulated inspection platform was an AirSim quadrotor. We equipped the UAV with a 512x512 depth camera and RGB camera. For simulations, pointcloud localization was provided by AirSim. We also used AirSim's built-in semantic segmentation to segment out the infrastructure in the RGB images. The segmented RGB images were aligned with the depth camera to generate a segmented depth image. Both the raw and segmented depth images are then converted into 3D pointclouds. The common setup described above was used for the rest of the pipeline.

\subsubsection{Environment Setup} 
\begin{figure}
    \centering
    \includegraphics[width = \columnwidth]{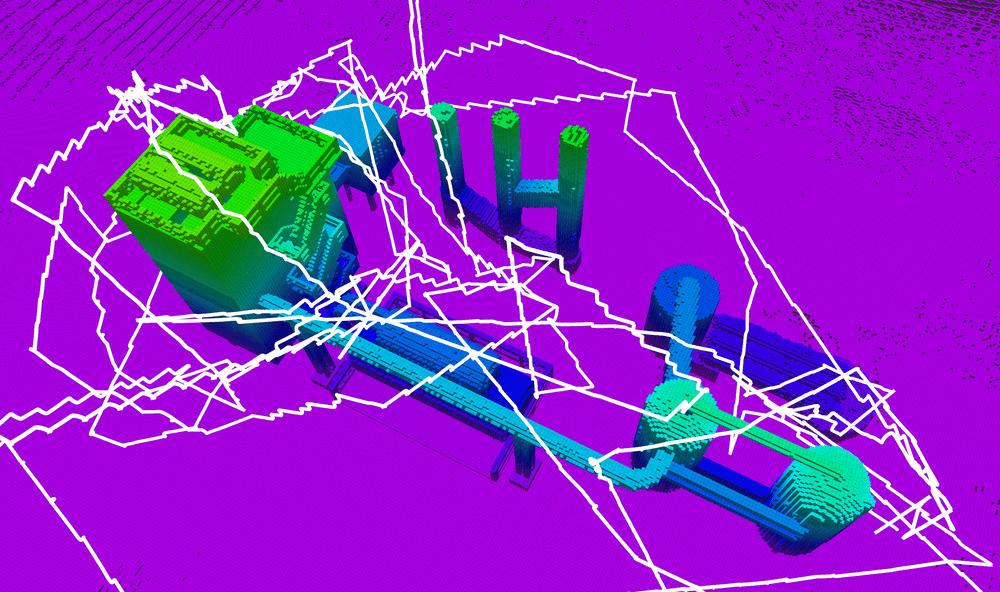}%
    \caption{UAV flight path during GATSBI in the chemical plant environment.}%
    \label{fig:flightPath_gatsbi}%
\end{figure}

We performed experiments using five bridge scenes, a silo scene, and a chemical plant scene (Fig.~\ref{fig:qual_results}). The steel bridge scene contained large hills on both ends of the bridge and trees far away from the bridge. However, the other scenes only contained the infrastructure itself. We chose these infrastructures (bridges and buildings) because they represent distinct types of infrastructure that are common to inspect. Figure~\ref{fig:flightPath_gatsbi} shows a path GATSBI generated and the UAV followed around the chemical plant scene.

\subsubsection{Computational Time}
\begin{figure}
    \centering
    \includegraphics[width = \columnwidth]{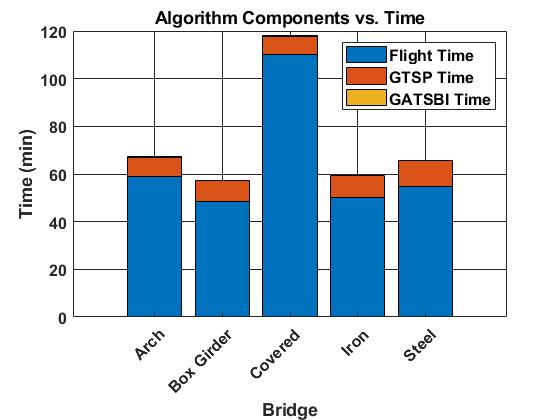}%
    \caption{We analyze how long different parts of the algorithm took to run. For all flights, the average flight time was 64.53 minutes. The average GTSP solver (GLNS) time was 8.82 minutes. Lastly, the average time for the remaining parts of GATSBI was 0.14 minutes.}%
    \label{fig:compTime}%
\end{figure}

We examine the time it takes to execute GATSBI. In Fig.~\ref{fig:compTime}, we report the average time for different components of the algorithm during all the simulations. We report three times: the time spent in the planner to create the GTSP instance (GATSBI), the time taken to solve the GTSP instance (GTSP), and the flight time before the algorithm calls the planner again (flight). We see that the time it takes GATSBI to perform segmentation and create a GTSP instance takes an average of 0.14 minutes. The GTSP solver takes an average of 8.82 minutes. Compared to the flight time (average of 64.5 minutes), the time taken by the planner is not significant. This suggests that GATSBI is not a bottleneck and is capable of running in real-time on UAVs that are executing 3D infrastructure inspection in unknown environments.

\subsubsection{Comparison with Frontier Exploration Baseline}
\begin{figure}
    \centering
    \begin{subfigure}[b]{0.75\columnwidth}%
        \includegraphics[width = \textwidth]{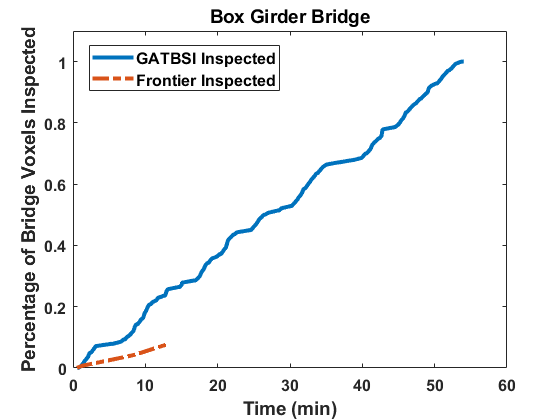}%
    \end{subfigure}%
    \hfill%
    \begin{subfigure}[b]{0.75\columnwidth}%
        \includegraphics[width = \textwidth]{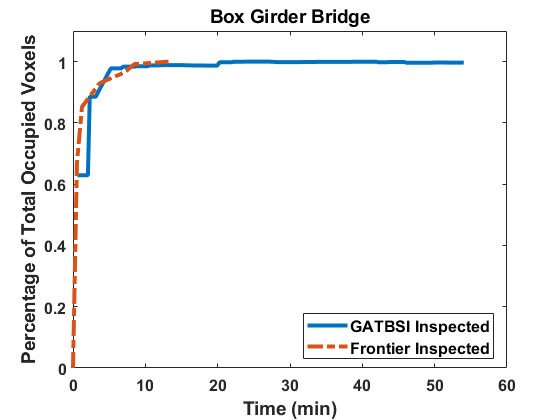}%
    \end{subfigure}%
    \caption{Top: The solid blue line is the percentage of inspectable voxels inspected by GATSBI over time. The dashed orange line represents the same for Frontier Exploration. Bottom: The solid blue line represents the percentage of environment voxels discovered over time. The dashed orange represents the same for Frontier Exploration. While Frontier Exploration is faster at finding the environment voxels as seen on the bottom, it is not good at inspecting the infrastructure voxels and only inspects a small percentage of them.}
    \label{fig:results}
\end{figure}

We compare the performance of GATSBI with a baseline algorithm that we developed that is based on frontier exploration. 
\revone{Since the baseline method does not directly count the number of inspected voxels, we implement a package on top of the baseline to count the inspected voxels. This way we compare only the inspected voxels, not the covered voxels.} 
We can see in Fig.~\ref{fig:results} and Table~\ref{tab:results} that our method does better than the baseline method when comparing the percentage of infrastructure voxels inspected. We obtain this value by dividing the inspected infrastructure voxels ($|V_{BI}|$) upon the termination of the algorithm by the total inspectable infrastructure voxels. Note, that obstructed infrastructure voxels that have no candidate viewpoints are uninspectable.

\begin{table}[ht!]
    \centering
    \begin{tabular}{|c|c|c|c|c|}
        \hline
        \textbf{Scene} & \textbf{Algorithm} & \textbf{Object Vox.} & \textbf{Inspected} & \textbf{Runtime} \\
        \hline
        \multirow{2}{*}{Arch} & Frontier & \multirow{2}{*}{124} & 10 & \textbf{15.2 min} \\
        \cline{2-2}
        \cline{4-5}
         & GATSBI & & \textbf{124} & 44.8 min \\
         \hline
        \multirow{2}{*}{Covered} & Frontier & \multirow{2}{*}{60} & 5 & \textbf{8.7 min} \\
        \cline{2-2}
        \cline{4-5}
         & GATSBI & & \textbf{60} & 69.3 min \\
        \hline
        \multirow{2}{*}{Girder} & Frontier & \multirow{2}{*}{140} & 11 & \textbf{13.2 min} \\
        \cline{2-2}
        \cline{4-5}
         & GATSBI & & \textbf{140} & 53.7 min \\
        \hline
        \multirow{2}{*}{Iron} & Frontier & \multirow{2}{*}{167} & 6 &\textbf{8.4 min} \\
        \cline{2-2}
        \cline{4-5}
         & GATSBI & & \textbf{167} & 66.9 min \\
        \hline
        \multirow{2}{*}{Steel} & Frontier & \multirow{2}{*}{214} & 12 & \textbf{24.5 min} \\
        \cline{2-2}
        \cline{4-5}
        & GATSBI & &\textbf{214} & 77.3 min \\
        \hline
    \end{tabular}
    \caption{Table showing the number of inspectable voxels that were inspected by GATSBI and Frontier for all 5 bridge scenes as well as the total algorithm runtime.}
    \label{tab:results}
\end{table}

Nevertheless, we observe that GATSBI achieves inspection of 100\% voxels while the baseline only achieves a maximum of 10\%. Frontier exploration does not explicitly take inspection into account. As shown in the right plot in Fig.~\ref{fig:results}, it performs as well as GATSBI at exploration. The environment is also simple; in a more complicated environment, it would be better at exploration than the GATSBI algorithm. This validates our claim that GATSBI targets the inspection of infrastructure surfaces instead of just covering the environment. We also see that GATSBI executes a more thorough inspection than the baseline. Therefore, we justify the claim that GATSBI is more efficient in inspection compared to a frontier exploration algorithm.

\subsubsection{Comparison with SIP}
Next, we compare the inspection performance of GATSBI with SIP~\cite{BABOOMS_ICRA_15}. SIP takes in as input a triangular mesh of the target infrastructure as well as viewing parameters (camera FoV and minimum/maximum viewing distance). For each triangle in the mesh, it generates a viewpoint that meets the viewing parameter constraints and a path that connects each viewpoint. Using AirSim, we flew this generated path and kept track of the number of infrastructure voxels that were inspected during the flight based on the viewing constraints. Similarly, we used GATSBI to generate a path in a receding horizon fashion as more and more of the infrastructure was observed and counted the number of voxels that were inspected during GATSBI's flight. For GATSBI, the infrastructure map was not known beforehand. The results from these flights are shown in Table~\ref{tab:sim_results}. Both algorithms were evaluated in five scenes: arch bridge, covered bridge, box girder bridge, iron bridge, and silo tower. At algorithm termination, on average GATSBI was able to inspect 38\% more voxels than SIP. However, SIP generally finished earlier than GATSBI while also flying a shorter distance. This is because the SIP knows the infrastructure mesh/map beforehand whereas GATSBI does not. 

\begin{table}[ht!]
    \vspace{2.5mm}
    \centering
    \begin{tabular}{|c|c|c|c|c|c|}
        \hline
        \multirow{2}{*}{\textbf{Scene}} & \multirow{2}{*}{\textbf{Algorithm}} & \textbf{Object} & \multirow{2}{*}{\textbf{Inspected}} & \textbf{Time} & \textbf{Distance} \\
        & & \textbf{Voxels} & & (min) & (meters) \\
        \hline
        \multirow{2}{*}{Arch} & SIP & \multirow{2}{*}{610} & 266 & \textbf{42.29} & \textbf{167.33} \\
        \cline{2-2}\cline{4-6}
         & GATSBI & & \textbf{610} & 91.32 & 295.88 \\
         \hline
        \multirow{2}{*}{Covered} & SIP & \multirow{2}{*}{79} & 70 & 32.22 & \textbf{89.07}  \\
        \cline{2-2}\cline{4-6}
        & GATSBI & & \textbf{79} & \textbf{21.36} & 89.12 \\
        \hline
        \multirow{2}{*}{Girder} & SIP & \multirow{2}{*}{529} & 250 & \textbf{61.02} & \textbf{267.87} \\
        \cline{2-2}\cline{4-6}
         & GATSBI & & \textbf{529} & 84.52 & 339.99 \\
        \hline
        \multirow{2}{*}{Iron} & SIP & \multirow{2}{*}{331} & 234 & 135.03 & 376.76 \\
        \cline{2-2}\cline{4-6}
         & GATSBI & & \textbf{331} & \textbf{58.90} & \textbf{225.63} \\
        \hline
        \multirow{2}{*}{Silo} & SIP & \multirow{2}{*}{481} & 288 & \textbf{83.41} & \textbf{206.53} \\
        \cline{2-2}\cline{4-6}
        & GATSBI & & \textbf{481} & 84.71 & 293.38 \\
        \hline
    \end{tabular}
    \caption{Table showing the number of inspectable voxels that were inspected by GATSBI and SIP for five scenes as well as the total algorithm runtime and flight distance.}
    \label{tab:sim_results}
\end{table}

For a more direct comparison, we compared the percentage of voxels inspected at the time the quickest method finished as well as at the shortest distance at termination, i.e., if SIP finished in 30 minutes, we compared the number of voxels inspected by GATSBI at 30 minutes and if SIP flew 200 meters, we compared the number of voxels GATSBI inspected at 200 meters flown. These results are shown in Table~\ref{tab:sim_results_equal}. When equalizing for time, GATSBI inspected 34\% more voxels on average than SIP. When equalizing for distance, GATSBI inspected 35\% more voxels than SIP on average. 

\begin{table}[ht!]
    \centering
    \begin{tabular}{|c|c|c|c|}
        \hline
        \multirow{2}{*}{\textbf{Scene}} & \multirow{2}{*}{\textbf{Algorithm}} & \textbf{\% Inspected} & \textbf{\% Inspected} \\
        & & \textbf{Equal Time} & \textbf{Equal Distance}\\
        \hline
        \multirow{2}{*}{Arch} & SIP & 43.61\% & 43.61\% \\
        \cline{2-4}
         & GATSBI & \textbf{54.56\%} & \textbf{75.14\%}  \\
         \hline
        \multirow{2}{*}{Covered} & SIP & 82.28\% & 88.61\% \\
        \cline{2-4}
         & GATSBI & \textbf{100.00\%} & \textbf{99.98\%} \\
        \hline
        \multirow{2}{*}{Girter} & SIP & 47.26\% & 47.26\%  \\
        \cline{2-4}
         & GATSBI & \textbf{88.48\%} & \textbf{91.12\%} \\
        \hline
        \multirow{2}{*}{Iron} & SIP & 36.25\% & 42.90\% \\
        \cline{2-4}
         & GATSBI & \textbf{100.00\%} & \textbf{100.00\%} \\
        \hline
        \multirow{2}{*}{Silo} & SIP & 59.88\% & 59.88\% \\
        \cline{2-4}
        & GATSBI & \textbf{100.00\%} & \textbf{92.96\%} \\
        \hline
    \end{tabular}
    \caption{Table comparing the percentage of inspectable voxels that were inspected by GATSBI and SIP for five scenes when equalizing for time and distance.}
    \label{tab:sim_results_equal}
\end{table}

Intuitively, it might make more sense for SIP to perform better since it has access to the infrastructure's mesh information beforehand; however, there are a few reasons why GATSBI performs better. Firstly, our use of voxels allows us to more easily define the "resolution" of inspection. If a mesh is larger than the voxel resolution, the mesh being inspected does not necessarily mean all voxels in the mesh have been inspected. Second, SIP does not consider collision-free navigation in its path generation. It only knows the infrastructure mesh beforehand and not other obstacles; therefore its generated path is not necessarily the most efficient. Lastly, SIP does not check whether the selected viewpoint can be navigated. It might be too close to other parts of the structure for the quadrotor to navigate. GATSBI handles collision-free path generation and viewpoint selection online based on the entire environment map. 

Qualitative comparisons between the two methods are shown in Figure~\ref{fig:qual_results}. While GATSBI observes the entire inspection surface, there are some coverage gaps, shown within the dotted-red boxes, with SIP. Also, GATSBI was able to inspect more complex structures as shown on the right side of the figure. SIP failed to generate paths for these structures due to the complexity of the triangular mesh. 

\subsubsection{Parameter Tuning}
\begin{figure}
    \centering
    \begin{subfigure}[b]{0.45\columnwidth}
        \includegraphics[width = \textwidth]{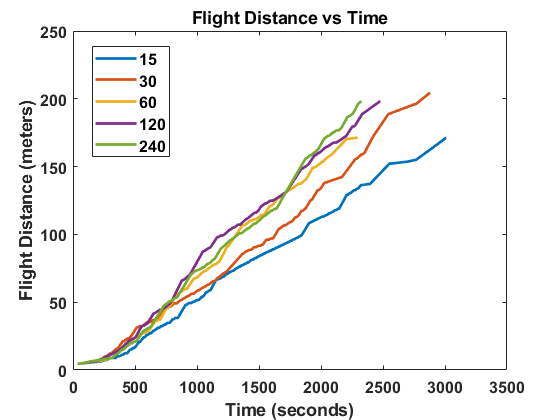}
        \caption{Replanning Time vs. Flight Distance and Total Runtime}
    \end{subfigure}
    \begin{subfigure}[b]{0.45\columnwidth}
        \includegraphics[width = \textwidth]{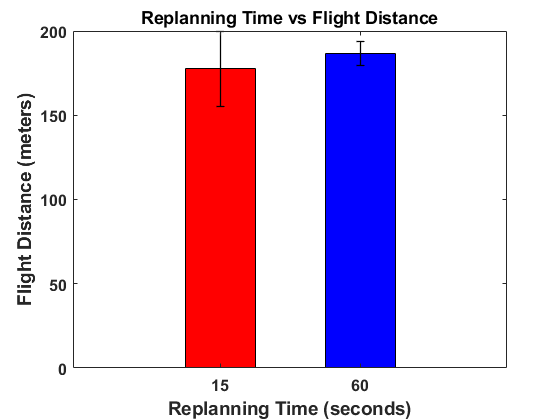}
        \caption{15 and 60 second Replanning Time vs. Flight Distance}
    \end{subfigure}
    \begin{subfigure}[b]{0.45\columnwidth}
        \includegraphics[width = \textwidth]{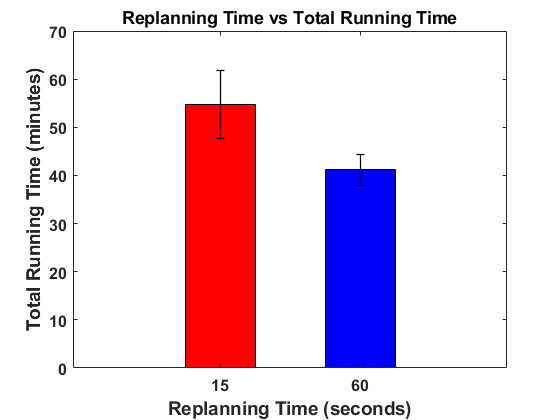}
        \caption{15 and 60 second Replanning Time vs. Total Runtime}
    \end{subfigure}
    \caption{Evaluation on different values for the replanning time parameter. In the first figure, we show how different values of replanning time affected the total flight distance of the UAV and the total runtime of the algorithm. For the second figure, we show the average and standard deviation of multiple runs at a replanning time of 15 and 60 seconds vs. the total flight distance. The last figure is the same but instead of total flight distance it is the total algorithm runtime.}
    \label{fig:replanning}
\end{figure}

One parameter used in the algorithm is replanning time, $RPT$. This time determines when to stop on the current GTSP tour if it has not been completed and replan with GATSBI using the most up-to-date environment and bridge information. Here, we discuss how we determined what to set $RPT$ to. Initially, we evaluated different values of $RPT$ for one of the infrastructure simulations. The results of this are shown in Fig.~\ref{fig:replanning}a. $RPT$ values of 15 and 60 seconds were then chosen for further evaluation due to them having the shortest flight distance. 5 runs each at these $RPT$ values were conducted using the simulation setup. Figure~\ref{fig:replanning}b shows the average flight distance during these 5 runs. On average, an $RPT$ value of 15 seconds had a slightly shorter total flight time (177.5 vs 186.5 meters) compared to an $RPT$ value of 60 seconds but with a higher standard deviation (22.4 vs 7.1). However, the average total runtime was longer using an $RPT$ value of 15 seconds (54.7 vs 41.2 min) while also having a higher standard deviation (7.0 vs 3.3) compared to an $RPT$ value of 60 seconds as shown in Fig.~\ref{fig:replanning}c. Because of this, an $RPT$ value of 60 seconds was used for the algorithm. 

\subsubsection{Simulated Crack Detection}

We also evaluated GATSBI on a simulated crack detection setup and compared its performance to frontier exploration. We randomly placed defects on 10\% of the infrastructure's surface voxels and determined how many of them were detected by both algorithms. Detected is defined as inspecting the voxel that the crack is on. This evaluation was done in an offline manner and each bridge was evaluated 10,000 times for both algorithms. The results are shown in Table~\ref{tab:cracks}. GATSBI performed at least 11.5x better at inspecting voxels with defects than frontier exploration. The reason that GATSBI's detection is not close to 100\% is that cracks were also placed on top of the infrastructure and below. For our bridge models, on average 40\% of the bridge surface is inspectable using a front-facing camera. Since the UAV is not equipped with an upwards and downwards-facing camera, the remaining 60\% cracks are not possible to be found. However, by adding these cameras and setting up the GTSP instance accordingly, these cracks would also be detected as long as they are not obstructed. Cracks could also be blocked by obstacles in the environment making them impossible to inspect.

\begin{table}[ht!]
    \centering
    \begin{tabular}{|c|c|c|c|}
        \hline
        \textbf{Bridge} & \textbf{Algorithm} & \textbf{\% Found} & \textbf{Std. Dev.} \\
        \hline
        \multirow{2}{*}{Arch} & Frontier & 2.6\% & 2.5\% \\
        \cline{2-4}
        & GATSBI & \textbf{32.5\%} & 7.2\% \\
        \hline
        \multirow{2}{*}{Covered} & Frontier & 3.0\% & 3.9\% \\
        \cline{2-4}
        & GATSBI & \textbf{34.5\%} & 10.9\% \\
        \hline
        \multirow{2}{*}{Girder} & Frontier & 3.2\% & 2.8\% \\
        \cline{2-4}
        & GATSBI & \textbf{49.4\%} & 8.0\% \\
        \hline
        \multirow{2}{*}{Iron} & Frontier & 1.6\% & 1.9\% \\
        \cline{2-4}
        & GATSBI & \textbf{44.1\%} & 7.6\% \\
        \hline
        \multirow{2}{*}{Steel} & Frontier & 2.4\% & 2.1\%\\
        \cline{2-4}
        & GATSBI & \textbf{42.8\%} & 6.7\% \\
        \hline
    \end{tabular}
    \caption{Table showing the results from the crack detection simulations on the multiple bridges. We compare the percentage of cracks found using GATSBI and Frontier Exploration.}
    \label{tab:cracks}
\end{table}
    
\subsection{Real-World Experiments}

This section presents real-world experiments that evaluate the performance of GATSBI. Here, we show results with only GATSBI since our simulated experiments demonstrated that GATSBI outperformed SIP. Below, we present quantitative results that compare the computation time, flight time, and the number of voxels inspected with GATBSI. 

\subsubsection{Experiment Setup} We performed these experiments using a DJI Matrice M600 Pro (see Fig.~\ref{fig:flight}). The M600 Pro was equipped with an NVIDIA Jetson TX2 (which ran Ubuntu 16.04 and ROS Kinetic), Velodyne VLP-16 3D LiDAR, and GPS. Due to energy limitations, a practical version of GATSBI was run. These limitations are addressed in Section~\ref{sec:conclusion}. For our experiments, we used the UAV's LiDAR to create a map of our bridge and then ran GATSBI in an offline manner to find the inspection path. Instead of using a pointcloud generated from a depth camera as we did in simulations, here we use pointclouds captured from a 3D LiDAR. For initial localization, we used DJI's GPS-based localization through their SDK. To combat real-world noise, we can use off-the-shelf LIO-SAM~\cite{liosam2020shan} along with GPS for localization. For segmentation, we manually used geographical segmentation (setting a bounding box around the target infrastructure). For our use case, geographical segmentation worked well but for more complicated experiments, color-based or learning-based segmentation networks can be used. The common setup described above was used for the rest of the pipeline. 

\subsubsection{UMD F3}

\begin{figure}[ht!]
\vspace{2.5mm}
    \centering
    \begin{subfigure}[b]{0.49\columnwidth}%
        \includegraphics[height = 6cm]{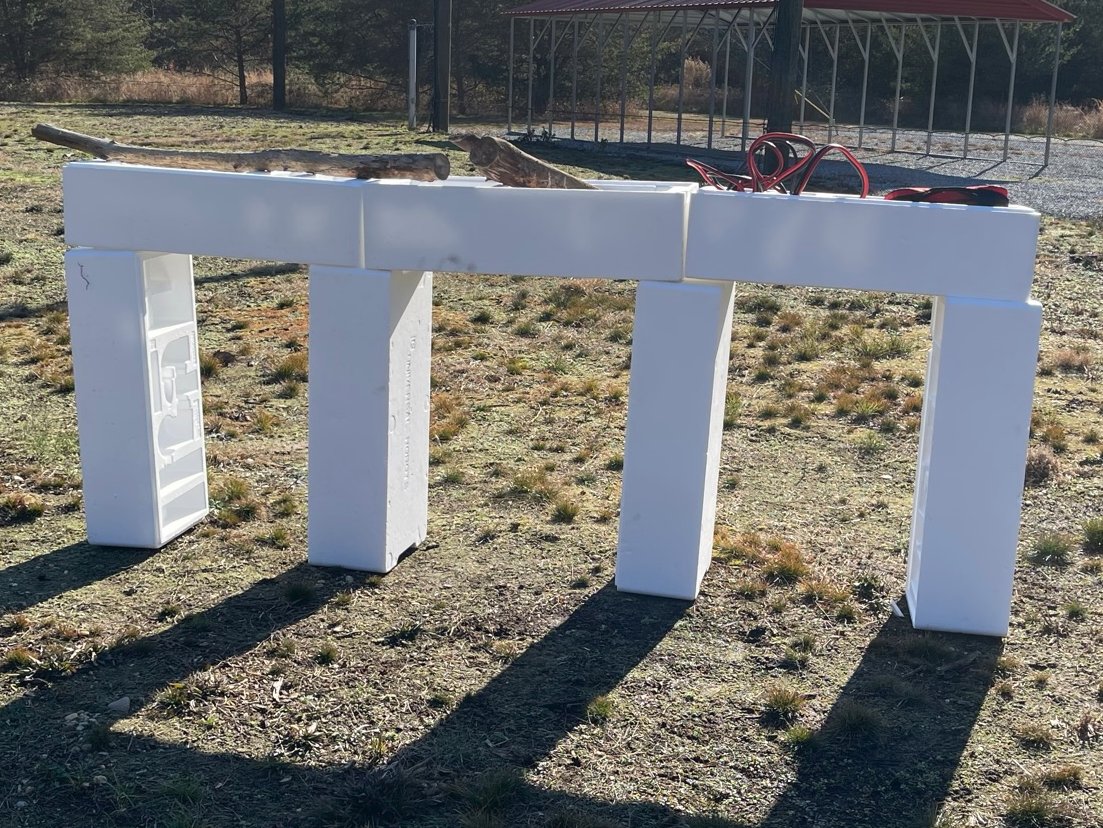}%
    \end{subfigure}%
    \hfill%
    \begin{subfigure}[b]{0.49\columnwidth}%
        \includegraphics[height = 6cm]{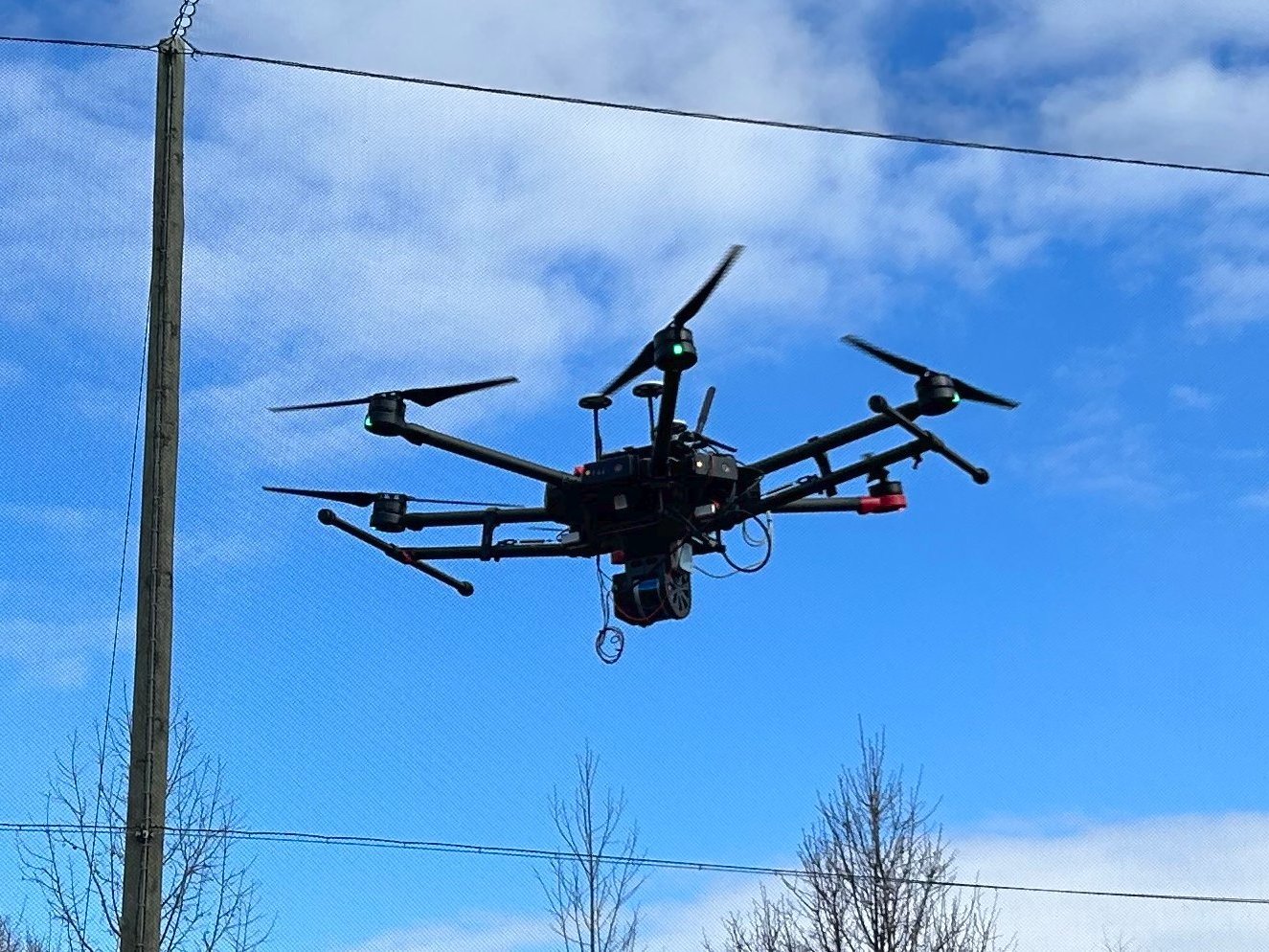}%
    \end{subfigure}%
    \caption{Left: The mock bridge used for the UMD F3 experiments. Right: DJI M600 Pro mid-flight during the inspection.}
    \label{fig:flight}
\end{figure}

\begin{figure}[ht!]
    \centering
    \includegraphics[width = \columnwidth]{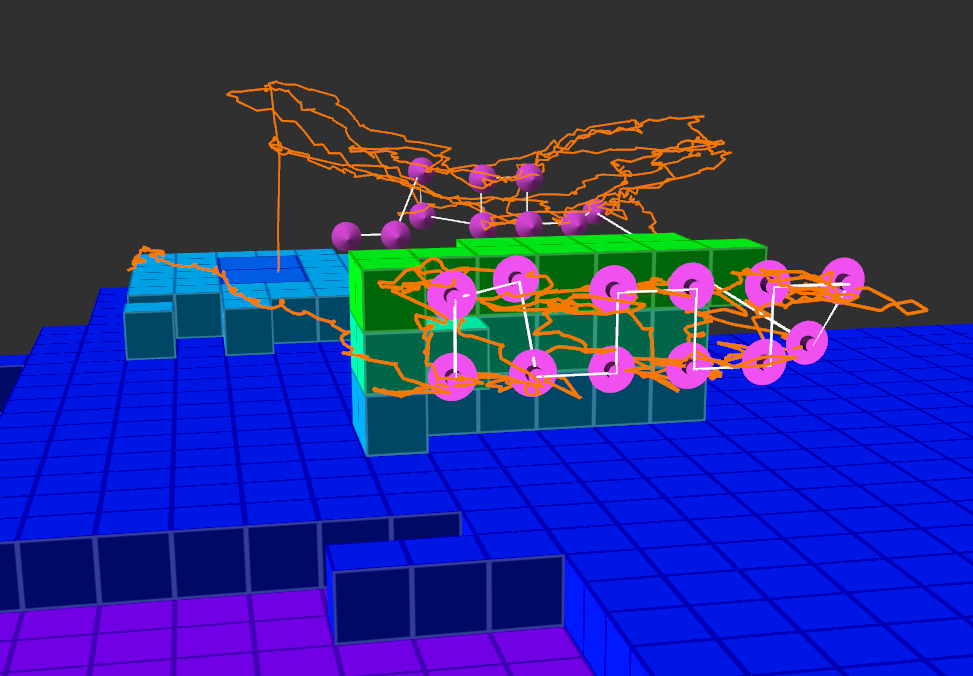}%
    \caption{Flight path and target inspection path on the real-world mock bridge.}%
    \label{fig:paths_hardware}%
\end{figure}

We constructed a mock bridge (Fig.~\ref{fig:flight}) and flew the UAV around it within an outdoor UAV cage called the Fearless Flight Facility (F3) at the University of Maryland, College Park. We implemented a practical version of the GATSBI algorithm on the real-world mock bridge. Due to the size of the mock bridge, our viewing cone angle was set to 0\degree with a minimum distance of two meters and a maximum distance of five meters. Our algorithm was able to inspect all 15 inspectable bridge voxels. The target inspection path and actual flown path are shown in Fig.~\ref{fig:paths_hardware}. The white line is the direct path between the target inspection points. The orange line is the actual flown path. The pink cones represent the inspection points. This experiment validates that GATSBI can be used for real-world infrastructure inspection. The computation time of the GATSBI algorithm was 0.32 seconds, GTSP time was 8.25 seconds, and flight time was 772 seconds. As shown in our conference~\cite{dhamiGATSBI}, the GATSBI algorithm time is not the bottleneck. Including the GTSP solver time, the total algorithm time is still much smaller than the flight time.

\subsubsection{College Park Bridge}

\begin{figure}[ht!]
    \centering
    \begin{subfigure}[b]{0.49\columnwidth}%
        \includegraphics[height = 6cm]{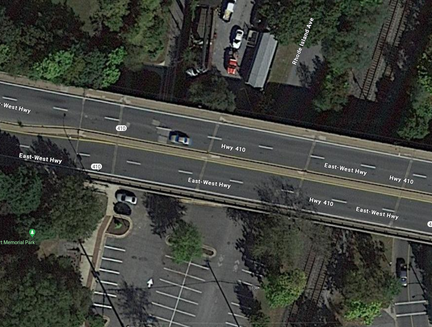}%
    \end{subfigure}%
    \hfill%
    \begin{subfigure}[b]{0.49\columnwidth}%
        \includegraphics[height = 6cm]{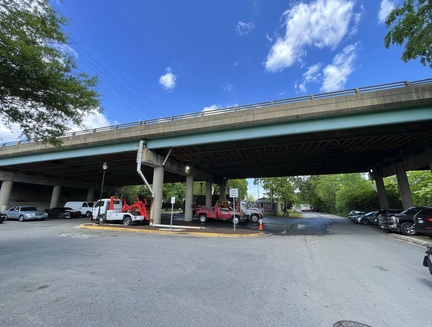}%
    \end{subfigure}%
    \caption{Left: Google Earth image of College Park Bridge. Right: Ground View of College Park Bridge.}
    \label{fig:college_park}
\end{figure}

\begin{figure}
\vspace{2.5mm}
    \centering
    \begin{subfigure}[t]{\columnwidth}
        \includegraphics[width=\textwidth]{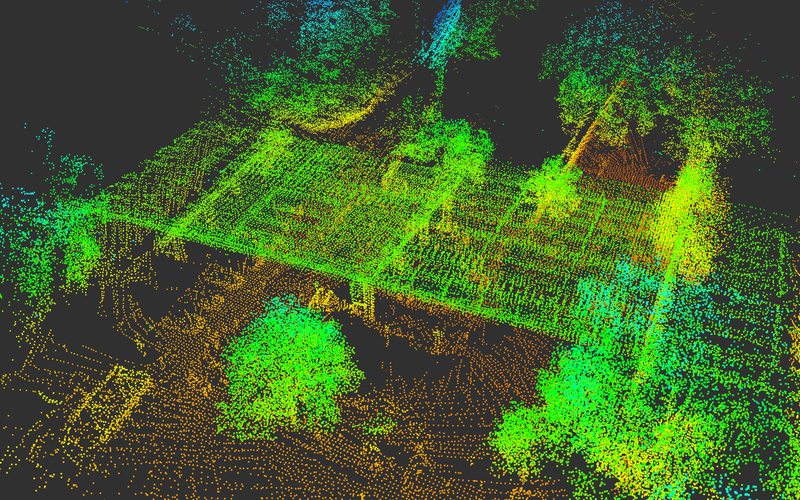}
        \caption{SLAM Pointcloud}
    \end{subfigure}
    \begin{subfigure}[t]{\columnwidth}
        \includegraphics[width=\textwidth]{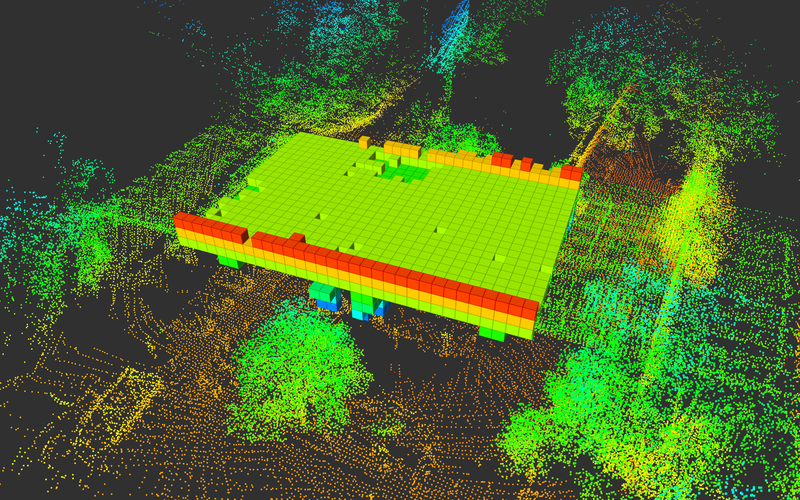}
        \caption{Segmented bridge Octomap}
    \end{subfigure}\vspace{.6ex}
    \caption{SLAM pointcloud and segmented bridge Octomap of real-world College Park Bridge.}
    \label{fig:college_park_points}
    \vspace{-2mm}
\end{figure}

\begin{figure}
\vspace{2.5mm}
    \centering
    \begin{subfigure}[t]{0.65\columnwidth}
        \includegraphics[width=\textwidth]{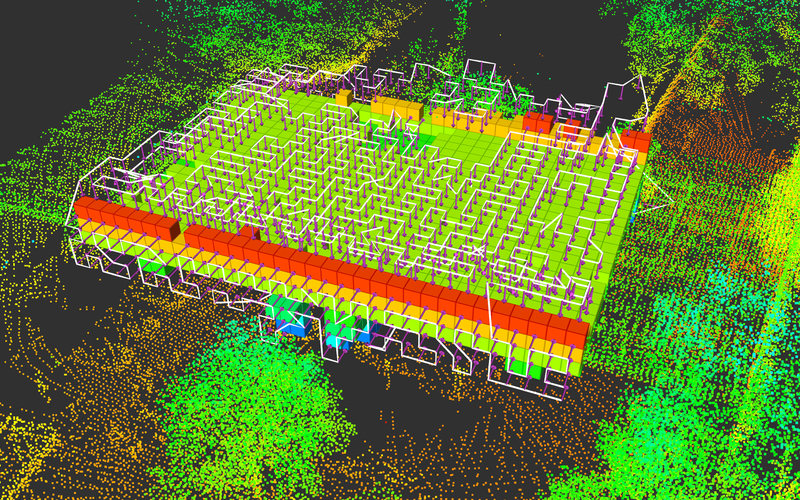}
        \caption{Inspection Path with Viewing Cone 0\degree and 2-5 meter distance}
    \end{subfigure}
    \begin{subfigure}[t]{0.65\columnwidth}
        \includegraphics[width=\textwidth]{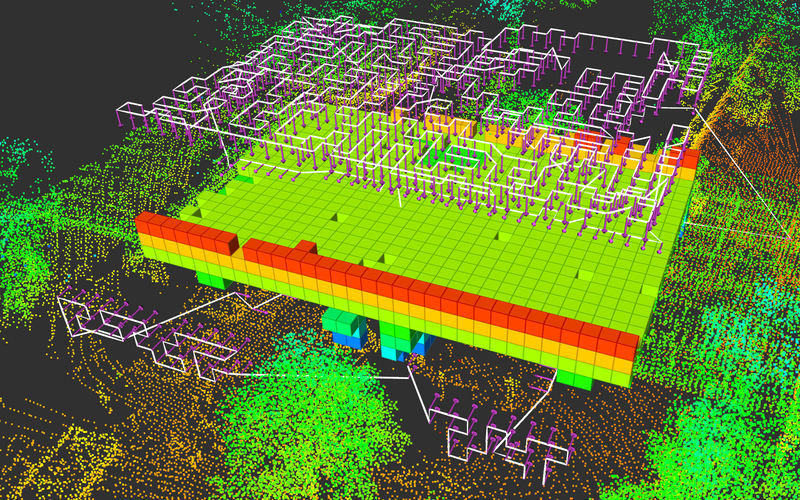}
        \caption{Inspection Path with Viewing Cone 0\degree and 8-10 meter distance}
    \end{subfigure}
    \begin{subfigure}[t]{0.65\columnwidth}
        \includegraphics[width=\textwidth]{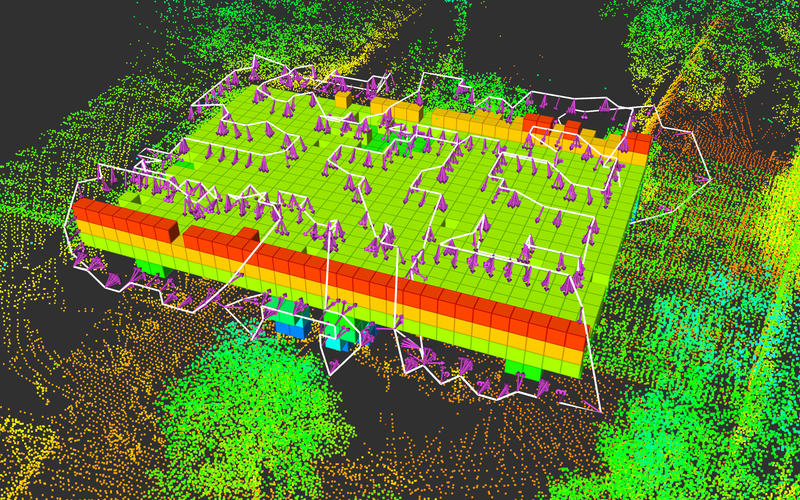}
        \caption{Inspection Path with Viewing Cone 20\degree and 2-5 meter distance}
    \end{subfigure}
    \caption{Output from simplified GATSBI algorithm on real-world College Park bridge in 3 different viewing cone scenarios.}
    \label{fig:college_park_gatsbi}
\end{figure}

Next, we tested the validity of our method on actual infrastructure. We used the same hardware setup as described above. However, due to being inside Washington DC's no-fly zone restriction, we had to capture data on the ground. We placed the DJI M600 Pro in a cart and walked it around the bottom of a real bridge in College Park, Maryland. The bridge and setup are shown in Figure~\ref{fig:college_park}. After collecting the data, we ran it through our pipeline. The SLAM pointcloud along with the segmented bridge octomap can be seen in Figure~\ref{fig:college_park_points}. These were used by our simplified GATSBI algorithm to generate our inspection paths. We also further modified GATSBI to work on voxel faces instead of just voxels. Previously, a voxel was marked as inspected if it any face of it was inspected. With the modified version, each of the six faces is separated. We show three example inspection paths in Figure~\ref{fig:college_park_gatsbi}. The first path is generated with a viewing cone angle of 0\degree with a minimum distance of two meters and maximum distance of five meters. The second path is generated with the same angle but from a distance of 8-10 meters. In this example, you can see the inspection path avoiding the tree in the foreground. Lastly, we show an example with a viewing cone angle of 20\degree and distance of 2-5 meters. Note here that each inspection point has multiple arrows coming out of it. This is because multiple faces are inspected at each point due to the viewing cone angle. 

%


\subsection{Crack Detection Results}
\begin{figure}
\vspace{2.5mm}
    \centering
    \begin{subfigure}[t]{0.49\columnwidth}
        \includegraphics[width=\textwidth]{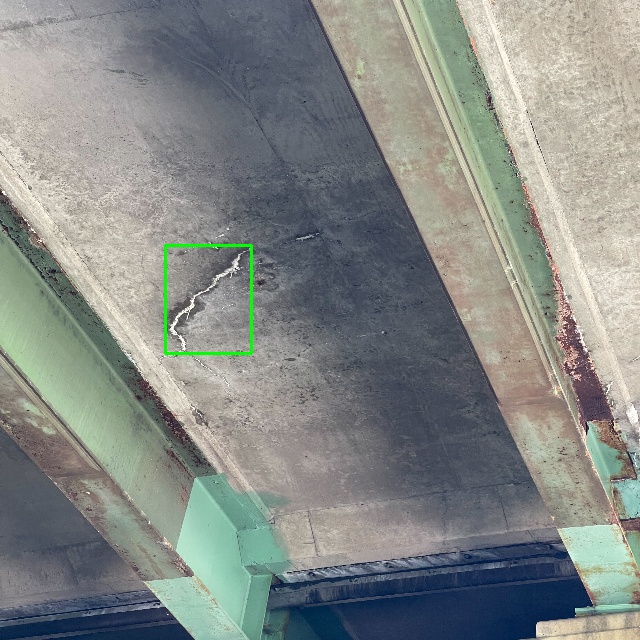}
        \caption{}
    \end{subfigure}
    \begin{subfigure}[t]{0.49\columnwidth}
        \includegraphics[width=\textwidth]{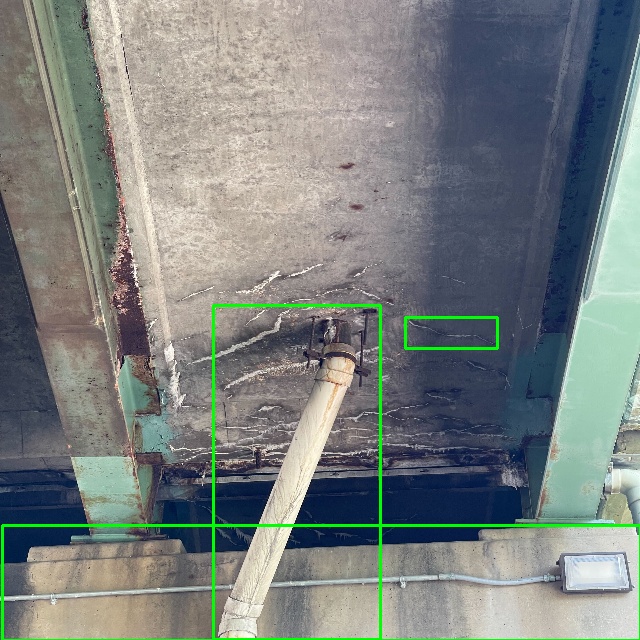}
        \caption{}
    \end{subfigure}
    \begin{subfigure}[t]{0.49\columnwidth}
        \includegraphics[width=\textwidth]{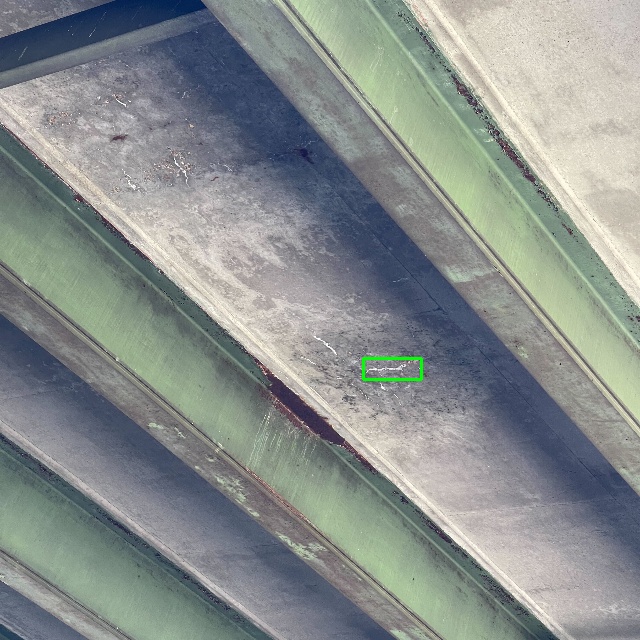}
        \caption{}
    \end{subfigure}
    \begin{subfigure}[t]{0.49\columnwidth}
        \includegraphics[width=\textwidth]{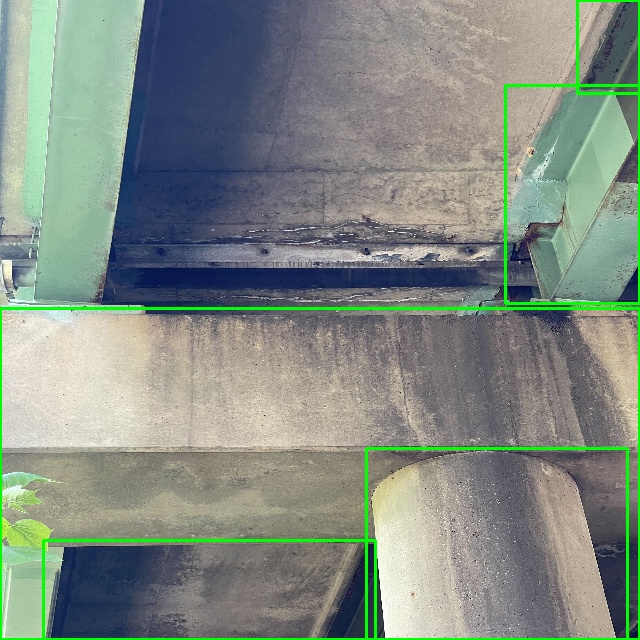}
        \caption{}
        \label{fig:cracks:pillar1}
    \end{subfigure}
    \caption{Crack detection results over real-world College Park bridge images.}
    \label{fig:college_park_cracks}
\end{figure}

For crack detection, we use pre-trained YOLO-World~\cite{Cheng_2024_CVPR} large model (\textit{YOLOw-l}) conditioned on the text prompt `crack'. This crack detection model effectively acts in a zero-shot manner for this task. We run this model on some images from the College Park Bridge, as shown in Figure~\ref{fig:college_park_cracks}. Here we show results with a confidence threshold of 0.008.

We found that YOLO-World can appreciably detect and localize the cracks in the bridge. In some cases, some false positives may occur which may have a high likelihood of containing cracks, such as the pillars in Figure~\ref{fig:cracks:pillar1}. Note that these results were obtained using a pre-trained model and were not fine-tuned for crack detection, highlighting its generalizability. 


\subsection{Code Release}
We provide the GATSBI code as well as two ROS packages that integrate MoveIt in our GitHub repository\footnote{\url{https://github.com/raaslab/GATSBI}}. The GATSBI code is used to run our inspection planner on the target infrastructure. One of the ROS packages integrates MoveIt with AirSim's ROS wrapper allowing the use of MoveIt in the AirSim simulation environment. The second ROS package integrates MoveIt with DJI's SDK allowing the use of MoveIt with DJI multirotors in real environments.

\section{Conclusion}\label{sec:conclusion}
We present GATSBI, a 3D infrastructure inspection planner. We evaluate the performance of the algorithm through AirSim simulations and real-world hardware experiments with a UAV equipped with a 3D LiDAR and an RGB camera. The simulations show that GATSBI outperforms SIP~\cite{BABOOMS_ICRA_15}. The hardware experiments show that GATSBI is a viable solution to real-world infrastructure inspection. In particular, we show that the algorithm is efficient in the sense that it targets inspectable voxels rather than simply exploring a volume. The simulations and experiments also demonstrate that the algorithm can run in real time. In future work, we intend to improve our real-world experiments. In particular, we are investigating implementing a multi-agent solution to account for the limited battery life of UAVs.

\renewcommand{\thechapter}{5}

\chapter{Environmental Monitoring: Change Detection}\label{chap:agri}

\section{Introduction}

In this chapter, we shift focus from monitoring a specific object in the environment to monitoring conditions within the entire environment. Here, if monitoring is visual, then we can simply use the algorithm from the previous chapter (~\textit{GATSBI}) without the semantic segmentation with perhaps a bounding box constraint. However, typically, in environmental monitoring applications, we are interested in detecting changes over time. ~\textit{GATSBI} was designed for only static environments. In this chapter, we examine how to detect changes in the environment in the context of precision agriculture. The goal of precision agriculture is to optimize the growth, maintenance, and harvesting of crops using data-driven technologies~\cite{rogers,tokekar2016sensor}. This will become especially important as the population grows, leading to a higher demand for efficiency from farms~\cite{bommarco_kleijn_potts_2013, franzluebbers_paine_winsten_krome_sanderson_ogles_thompson_2012, godfray_2011}. One way of achieving higher efficiency is through selective breeding with the aid of plant phenotyping. Phenotyping refers to collecting large-scale phenotypic information about plants, including their heights, which are used for selective breeding and associative mapping through, for example, genome-wide association studies.

A bottleneck in phenotyping is the data collection process. One important phenotypic trait to monitor during a plant's growth cycle is its height. Recording the plant growth allows agronomists to monitor and predict vital features of crops such as flowering time and yield~\cite{moles_leishman, moles_warton_warman_swenson_laffan_zanne_pitman_hemmings_leishman_2009}. Manual height measurements are labor-intensive and quickly become infeasible as the plot and farm size increases. Manual height measurements are also sometimes biased. Using wheat as one example, it is only feasible to measure a few selected plants in each plot where hundreds of wheat plants are grown. In this chapter, our goal is to alleviate this bottleneck by using a UAV equipped with a 3D LiDAR for plant height estimation.



\begin{figure}[htp]
\centering
\includegraphics[width=\linewidth]{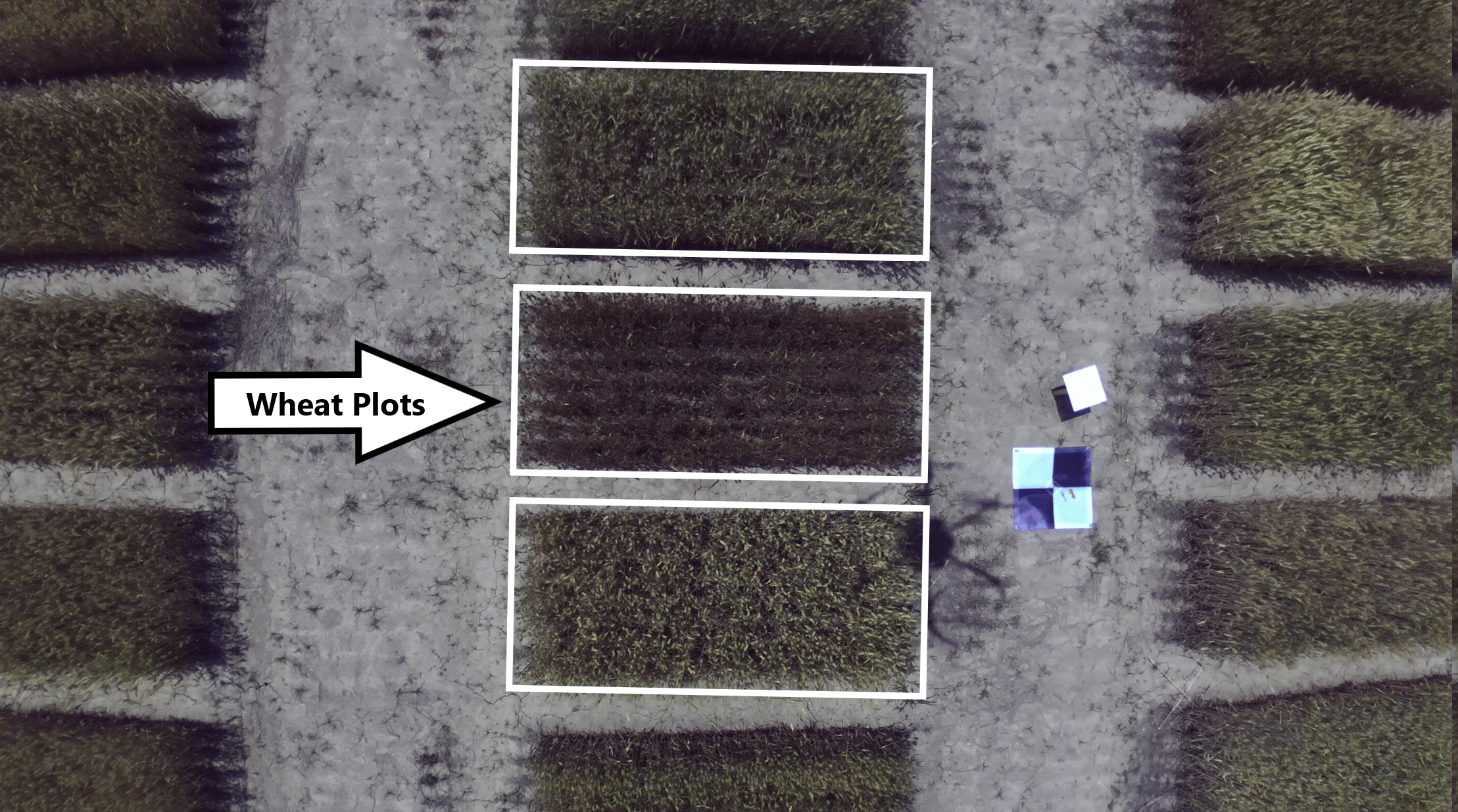}
\caption{This overhead picture was taken during a flight over wheat crops by our UAV. The farm is organized into plots, indicated by white boxes. The plots are organized as a grid. Our technique finds the plots and estimates the height of the crops within a plot.}
\label{fig:wheat_rgb_crop}
\end{figure}

We present a technique for determining crop heights using a 3D LiDAR mounted on a UAV. In particular, we focus on farms organized into smaller plots as shown in Figure~\ref{fig:wheat_rgb_crop}, as is typical in the case of phenotyping studies. Each plot (or a collection of plots) is a specific breed of plants being cultivated and monitored by the agronomist. The best cultivars are then selected for breeding. Traditionally, these plots are monitored by manual height measurements. Instead, we show how to use LiDAR to enable scalable and high-throughput phenotyping.

We collect raw 3D LiDAR scans from a UAV flown above the farm. We then present several data processing techniques that produce as output, bounding boxes around individual plots as well as height estimates for the plants within each plot. We do not make any assumptions on the prior knowledge about the size of each plot or the flatness of the terrain. We also present a toolchain to generate virtual phenotyping farms that can be used in simulations. In the real world, it is hard to find multiple farms to test the robustness of algorithms. This makes it difficult to compare algorithms or for robotics researchers to conduct precision agriculture research. Our goal in designing this toolchain is to make this research more accessible to the rest of the community. 

While there has been recent work on plant height estimation with LiDAR, as we describe in Section~\ref{chap:relWork}, this is the first such work using a 3D LiDAR mounted on a UAV that also detects individual plots. The two main contributions of this chapter are as follows:
\begin{itemize}
  \item A toolchain to generate virtual plot-based farms with varying ground terrains useful for phenotyping. The dataset released from this work has models for three representative environments, simulated 3D LiDAR scans, and ground truth information. This is released for the community at large to benchmark their algorithms against.
  \item A suite of point cloud processing tools for 3D LiDAR scans collected from a UAV on a wheat farm. These include open-loop mapping, data-filtering, plot clustering, and the height estimation tool. The datasets and the point cloud processing tools are also released along with this work. Our technique was able to estimate the plant heights of the real-life wheat plots with an RMSE of 6.1 cm.
\end{itemize}

We describe the hardware and software setup in Section~\ref{chap:systemDescription}. We describe our algorithm for plot and height estimation along with the underlying assumptions in Section~\ref{chap:rowAlgorithm}. The virtual farm generation toolchain is discussed in Section~\ref{chap:farmGen}. We conducted both hardware experiments and simulations with real wheat plots and virtual soybean plots. Lastly, we present the results and observations in Section~\ref{chap:experiments} followed by improvements we are working on in Section~\ref{chap:conclusionFutureWork}.





\section{Related Work}
\label{chap:relWork}

Automatically estimating the height of plants is an important problem and as such, there has been recent work on this (Table~\ref{table:comp}). However, as we will describe in this section, there are key differences between the prior work and the work presented in this chapter.
Prior methods include using a 2D LiDAR mounted on a UAV~\cite{anthony_elbaum_lorenz_detweiler_2014}, using RGB cameras mounted on fixed-wing~\cite{ziliani_parkes_hoteit_mccabe_2018} and multi-rotor UAVs~\cite{madec, yuan_li_bhatta_shi_baenziger_ge_2018}, and using a ground robot for navigation between rows of crops~\cite{kayacan_young_peschel_chowdhary_2018}.

The method described by Anthony et al. uses a 2D LiDAR mounted on the bottom of a UAV facing downwards to measure corn heights~\cite{anthony_elbaum_lorenz_detweiler_2014}. They estimate the ground and the crown of the crop plants for each scan. They reported the accuracy of height estimations within 5 cm. However, they do not produce a 3D map of the farm or find individual plots, as we do in this chapter.

Madec et al. used a similar method, except with a ground-based, 3D LiDAR on a row-based wheat farm along with a UAV equipped with an RGB camera~\cite{madec}. Using structure-from-motion, they extract a 3D dense point cloud from a camera on a UAV. 
They found a strong correlation between the estimated heights from the UAV and the ground-based LiDAR. The LiDAR estimations had an RMSE of 3.5 cm while the UAV estimations had an RMSE that ranged from 2.6 - 6.8 cm. It is expected for the ground-based method to be more accurate since a UGV does not have to constantly adjust its position to maintain stability. These adjustments add noise to the air-based measures. Unlike their method, we find each of the wheat plots as opposed to arbitrarily breaking up the 3D point cloud.

Yuan et al. used a similar setup with the 3D LiDAR mounted on a ground vehicle~\cite{yuan_li_bhatta_shi_baenziger_ge_2018}. 
The ground vehicle drove among the plots of the row-based wheat crops. To determine the height of the crops using the LiDAR, they relied on finding the ground in the areas between plots. The distance of the LiDAR off the ground was manually recorded and they used this distance to estimate the heights within the scans. Their ground-based LiDAR system had an RMSE of 5 cm; whereas the UAV approach had an RMSE of 9 cm. As opposed to manually determining the distance between the LiDAR and the ground, our method automates this process through the algorithm.

Ziliani et al. used a fixed-wing UAV with a downwards-facing camera ~\cite{ziliani_parkes_hoteit_mccabe_2018} to determine intra-field variability within a farm of maize crops. 
They used ground control points to help calibrate the image data and estimate the height. Also, to validate the estimations, they used a ground-based LiDAR mounted on a vehicle. There was a correlation of up to 0.99 between the LiDAR and structure-from-motion based approach for the RGB images. During the flowering of the crops, there was an increased variability and the correlation decreased to 0.65. However, they do not cover the entire farm with their 3D LiDAR as we do with our method.

There has also been some work done in regard to row detection by Li et al.~\cite{corn_row_detection:online}. They flew a UAV overhead a field of corn and captured RGB images from a camera. These RGB images were then stitched together to create a single image. They detect the rows of corn using computer vision techniques. Our method focuses on plot detection as opposed to row detection due to how our target farms are planted.

A detailed comparison between these prior works and ours is presented in Table~\ref{table:comp}. Our techniques can estimate the heights within an RMSE of 6.1 cm which is comparable to the other techniques. However, there are key differences between ours and the prior work: None of the previous works used a 3D LiDAR on the UAV to estimate the height. They either used 2D LiDAR or RGB cameras. If a 3D LiDAR was used, it was only from the ground (and therefore restricted to either the edges of the farm or between the rows) and for validating the RGB data. Unlike prior work, we estimate the ground plane using point-cloud analysis. Anthony et al.~\cite{anthony_elbaum_lorenz_detweiler_2014} did this for individual 2D line scans whereas we do this for local 3D patches. We also detect bounding boxes around each plot in the 3D point cloud map of the entire farm. This is particularly useful during phenotyping. 


\begin{table}[htp]
\centering
\includegraphics[width=\linewidth]{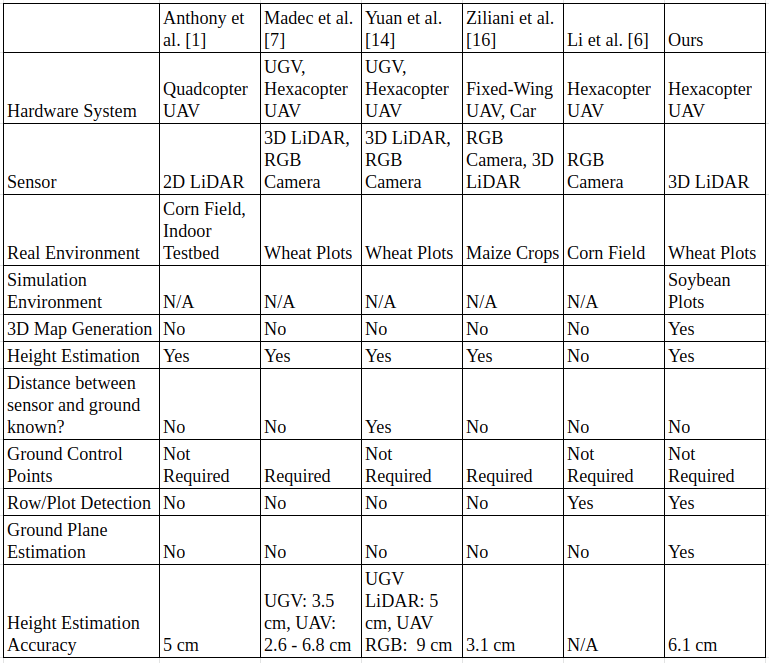}
\caption{Comparison between related work and our method.}
\label{table:comp}
\end{table}




\section{System Description} \label{chap:systemDescription}

\begin{figure}
    \centering
    \begin{subfigure}[b]{0.8\columnwidth}
        \includegraphics[width = \textwidth]{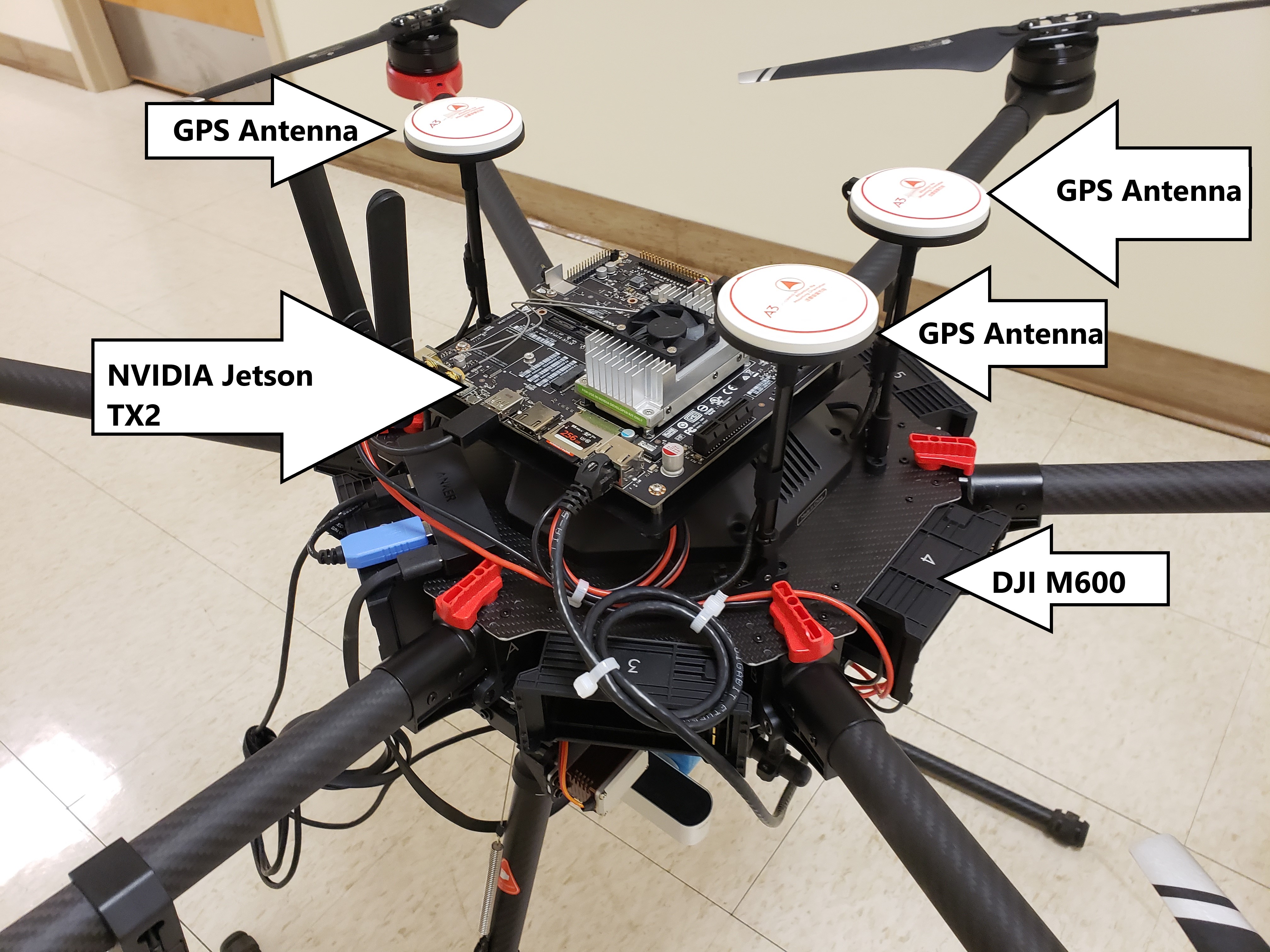}
        \caption{}
    \end{subfigure}
    \begin{subfigure}[b]{0.8\columnwidth}
        \includegraphics[width = \textwidth]{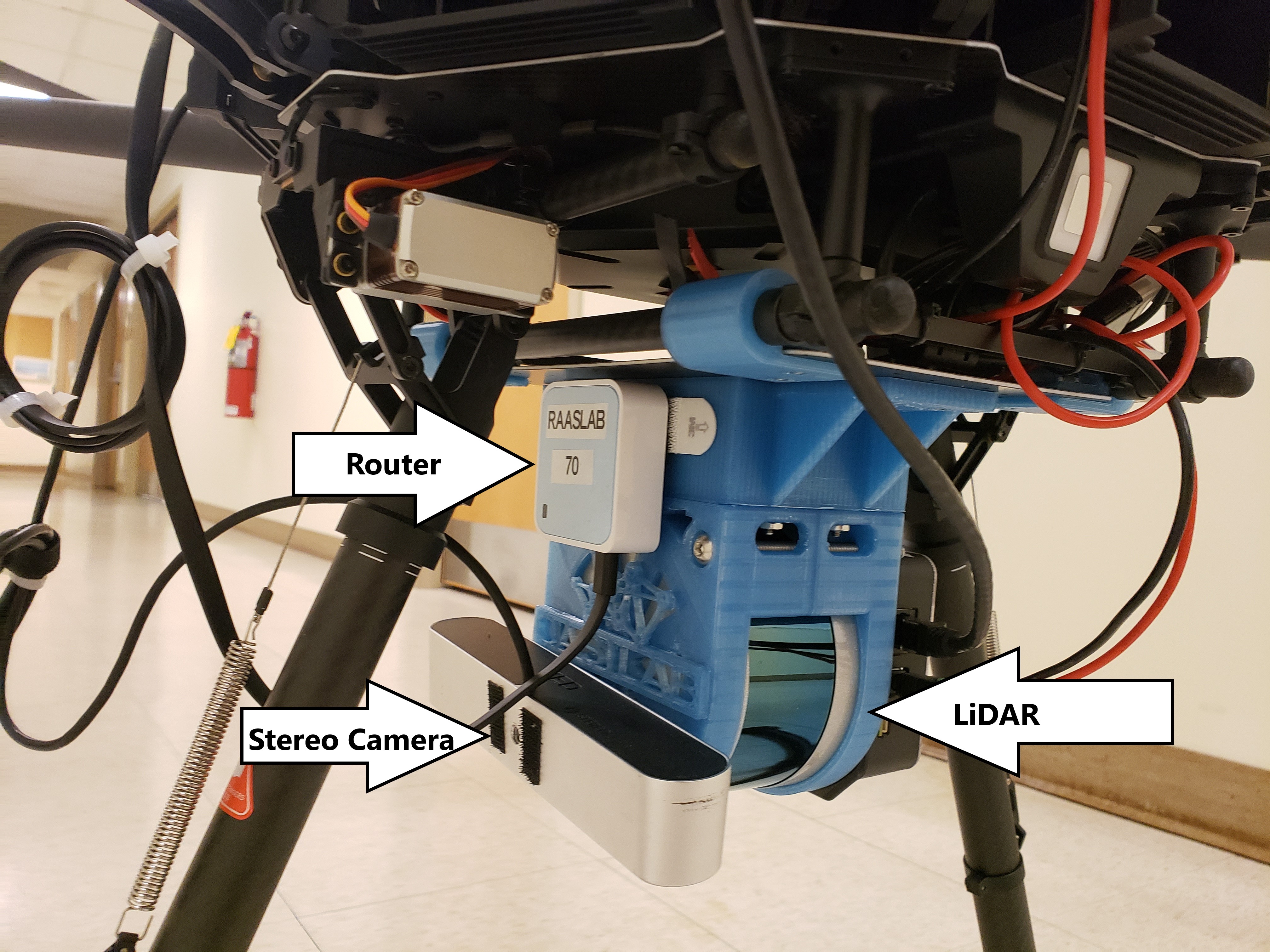}
        \caption{}
    \end{subfigure}
    \caption{We show the selected hardware mounted on top (a) and beneath (b) of our platform.}
    \label{fig:hardware_mounted}
\end{figure}

The UAV platform that we use is the DJI Matrice 600 Pro. The on-board computer is the NVIDIA Jetson TX2. The 3D LiDAR of choice is the Velodyne VLP 16. 
The DJI Matrice 600 Pro has a maximum takeoff weight of 15.5 kg. The platform has a maximum speed of 17.88 m/s if there is no wind and a hovering time of 16 minutes with a 6 kg payload. The VLP-16 is a 3D LiDAR consisting of 16 channels that can refresh at a rate of 5--20 Hz. It also has a range of 100 meters and an accuracy of $\pm3$ cm. The 16 channels of the 3D LiDAR, compared to a single channel for a 2D LiDAR, allow for the use of more information during the crop height estimation pipeline. Figure~\ref{fig:hardware_mounted} shows all the hardware mounted on the DJI M600 Pro platform. 
We used NVIDIA Jetpack 3.2.1 on the TX2 with Ubuntu 16.04 installed, as well as NVIDIA's CUDA 9.0 and OpenCV. We also used Robot Operating System (ROS) Kinetic along with the DJI SDK and the Point Cloud Library~\cite{Rusu_ICRA2011_PCL} (PCL).

\section{Data Products: Height Estimation, Plot Detection, and 3D Maps} \label{chap:rowAlgorithm}

\begin{figure}[htp]
    \centering
    \includegraphics[width=\linewidth]{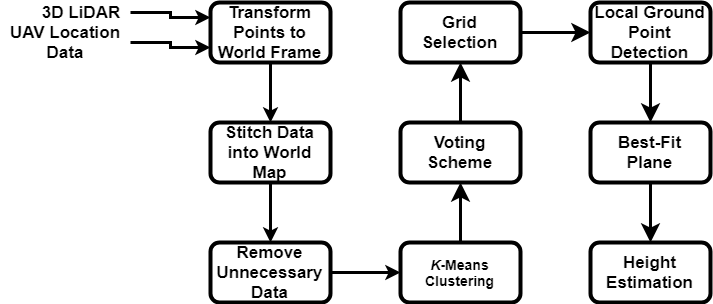}
    \caption{The system block diagram describing our height estimation algorithm.}
    \label{fig:block}
\end{figure}

Figure~\ref{fig:block} shows the system block diagram for the algorithm pipeline. We discuss each of the blocks step-by-step below.
We make the following assumptions based on the structure of typical phenotyping farms. We assume that the plots are aligned in a regularly ordered grid locally. We also know the number of plots. In principle, we only need to know the number of plots for a small subset of the environment. These are justified in phenotyping studies since the agronomists are the ones who decide the layout of the farms. Nevertheless, what the agronomists need for scalable and high-throughput phenotyping are 3D maps of the farm, to automatically find bounding boxes around individual plots, and to find the height relative to the ground for each plot. We describe the point cloud processing tools that we have developed to generate these data products in this section.

\subsection{Pre-processing and Map Building}

We transform each incoming scan into the world coordinate frame and merge them to produce a 3D map. 
Several open-source software packages can be used such as LOAM~\cite{Zhang-2014-7903} and BLAM~\cite{nelson_2016}. However, we found that in our setting this was not necessary since the pose obtained by the UAV was good enough to generate an open-loop map. 
The DJI M600 fuses information from three GPS receivers along with the IMU data to obtain its global pose. Since we are operating on a farm, the UAV has a clear line-of-sight with the GPS satellites which, along with the three GPS corrections, yields satisfactory mapping. Nevertheless, any of the open-source mapping tools can be used.

\begin{figure}[htp]
\centering
\includegraphics[width=\linewidth]{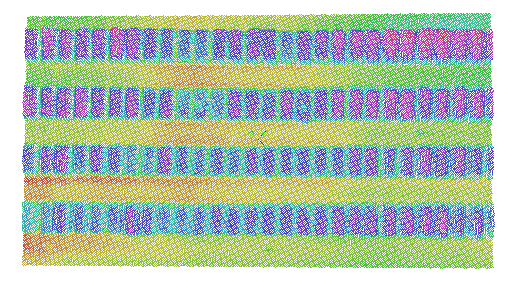}
\caption{This is an example output of the processing done on the point cloud scans. These scans were from the Kentland wheat farm. The colors represent the height of the points.}
\label{fig:wheat}
\end{figure}

After building a 3D map, we carry out several pre-processing steps to support the main techniques. 
As the UAV covers a larger area, the dataset quickly increases in size due to the immense amount of incoming data. It was necessary to down-sample the data to not only reduce the data size but also reduce the noise in the data. We do this using a voxel filter. The voxel filter allows down-sampling through averaging points in a set-size grid. Down-sampling also helps increase the speed at which we process the data for other algorithms. 
We also apply a crop box filter to crop the point cloud data and extract only the region of interest. 
Figure~\ref{fig:wheat} shows the cropped and down-sampled map obtained after these steps. Note that while we down-sample the data here, in a later step, we will use the full dataset for a more precise estimation of the heights. 

\subsection{Plot Detection}

After the pre-processing, the next step in the pipeline is to detect individual plots within the farm. Recall, that a ``plot'' refers to a cluster of plants as shown in Figure~\ref{fig:wheat_rgb_crop}. We need this to use the ground plane and height estimation algorithm on each plot given a point cloud file of the entire farm. We show an example dataset for this in Figure~\ref{fig:wheat}. 

\subsubsection{$K$--means Clustering}
We start with a down-sampled 3D point cloud. We first filter out points below a certain z-axis value (height). We then perform $k$--means clustering over the remaining points based on spatial distance. The k-value set for the $k$--means clustering algorithm is the number of clusters we expect, or in our case, the number of plots we are looking for. 


\begin{figure}
    \centering
    \begin{subfigure}[b]{0.8\columnwidth}
        \includegraphics[width = \textwidth]{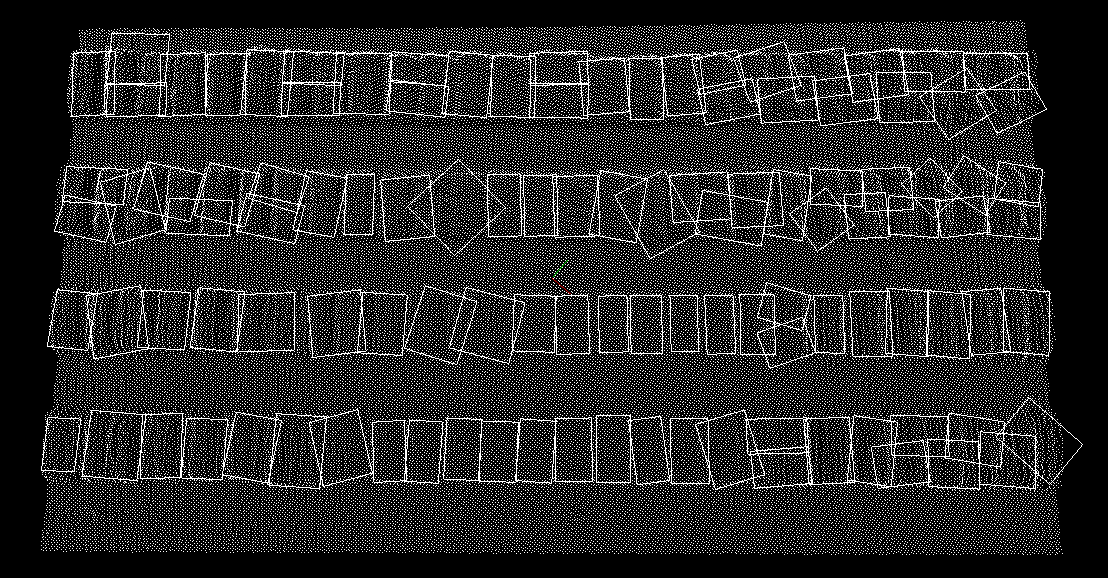}
        \caption{}
    \end{subfigure}
    \begin{subfigure}[b]{0.8\columnwidth}
        \includegraphics[width = \textwidth]{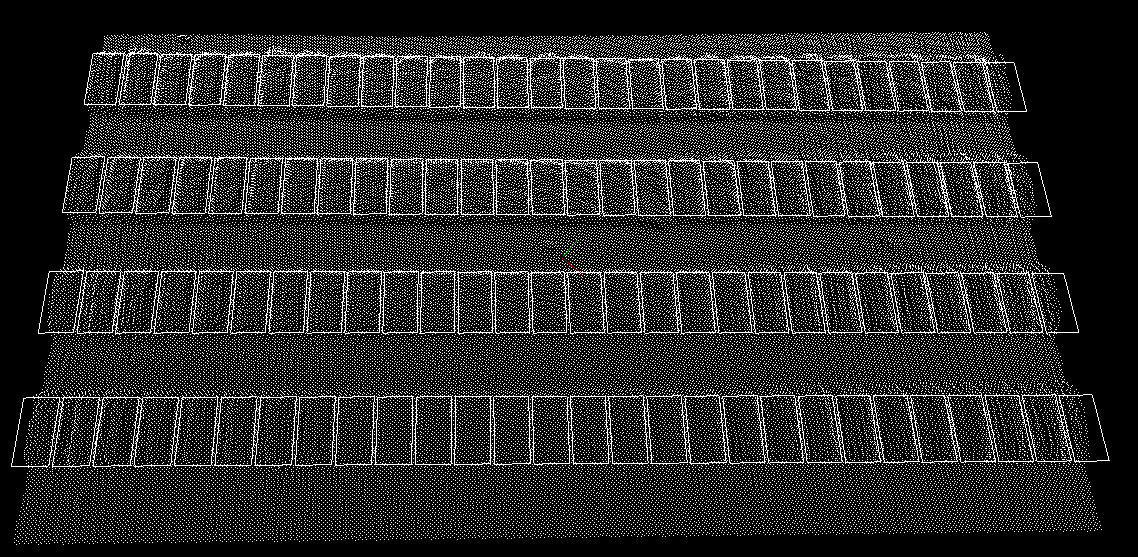}
        \caption{}
    \end{subfigure}
    \caption{(a) This is example data that we fed into the $k$--means clustering algorithm. The minimum-oriented boxes computed using the $k$--means clustering algorithm overlaid on top of the plots. (b) The estimated grid after the voting scheme overlaid on top of the plots. Note: While the figures were aligned with the $x$ and $y$ axes, the actual data and oriented boxes were not aligned. The algorithm also determines the rotation of the oriented boxes.}
    \label{fig:clusters}
\end{figure}


In Figure~\ref{fig:clusters}.a, we show the minimum-oriented bounding box of each cluster. A minimum-orientated bounding box is the smallest area box that fits all the points. As shown in this figure, the algorithm does not cluster all plots correctly, but it correctly clusters a large amount of them. Because of this, we used a voting scheme to help determine the plot size using the bounding boxes.


\subsubsection{Voting Scheme for Correcting Clusters}

For the voting scheme, we use the bounding box dimensions and orientations of each of the clusters determined in the previous step. These are binned, meaning they are all broken into separate ranges. We count the number of clusters that fit into each of the ranges. After, we take the range with the largest number of votes. Then, all the values within this range are averaged and used as the orientation that we set for the plots. We use a similar method for the length and width of the clusters to determine the length and width of the plots. 
\subsubsection{Grid Selection}

Using the estimated orientation, width, and length of the plots, we fill in a grid for the dataset. The distances between the plots are manually fit to help create the grid. We highlight that even though the grid is currently manually overlaid on top of the target dataset, we are working to automate this process. After we do this grid selection, the results look similar to what we show in Figure~\ref{fig:clusters}.b. We use these bounding boxes on the original dataset. We extend the lengths of the boxes so that they cover the ground between plots. All of the points in these boxes are then extracted one box at a time and then fed into the ground plane and height estimation algorithm which we discuss next.

We assume that we know the number of plots in the farm, i.e., we know the correct number of clusters for $k$--means. However, this is not a strict requirement. We do not need to know the total number of clusters in the entire farm. During pre-processing, we can set the crop box to a small area -- small enough to manually count the number of plots. Then, we can use the $k$--means clustering followed by the voting scheme to determine the size of the cluster. Once we determine the size of the cluster and the grid pattern for the smaller cropped area, we can extrapolate that grid to the rest of the (uncropped) dataset.


\subsection{Ground Plane and Height Estimation}
The purpose of this algorithm is to take a point cloud file, find the ground plane in it, and then determine height data for objects that are not part of the ground. We describe the process of this algorithm in this section.

\subsubsection{Detect Points in Ground Plane}
We detect the ground plane in a LiDAR scan by applying a sample plane model segmentation available in PCL~\cite{Rusu_ICRA2011_PCL}. This segmentation implementation uses a random sample consensus (RANSAC) algorithm to find a plane within the given dataset. 

\subsubsection{Best-Fit Plane of Inlier Points}
We find the inlier points from the planar model segmentation and use them to find the equation that describes a best-fit plane. We do this using the linear least-squares approximation on the inlier dataset. We take each of the inlier points and use them for this approximation to output the centroid of the data, as well as the normal vector of the best-fit plane.

\subsubsection{Height Estimation}

With a centroid and normal vector for the best-fit plane, we determine the height of each of the outlier points. We do this by finding the distance of each outlier point to the best-fit plane. Since the plane describes the ground, the distance from the outlier point to the plane is the height of that point over the ground. Doing this for all of the outlier points indicates each of their heights relative to the ground plane. As stated in the grid selection subsection, we extended the bounding box lengths. We did this so that the ground plane estimation works on a localized scale relative to each plot. This helps improve the accuracy of the algorithm by adjusting for the disparities in the ground over a large field. We purposely created a virtual farm environment with ground disparities shown in Figure~\ref{fig:sim_ground}.b. The output of the local ground plane estimation is shown in Figure~\ref{fig:sim_plots}. It was later discovered that using only voxel-filtered data led to underestimations. This is to be expected since voxel-filtering averages points within a set voxel resolution. To account for this, we concatenated the raw point cloud scans of the plots with the voxel-filtered data. More details are given in Section~\ref{chap:experiments}.

\section{Farm Generation Toolchain} \label{chap:farmGen}
In this section, we describe the algorithm that generates the virtual phenotyping farm world. The virtual environment is a Gazebo simulation environment. The hector quadrotor ROS package~\cite{2012simpar_meyer} is used as the virtual UAV with a Velodyne VLP-16 model attached to the bottom. We also use commercial 3D models of individual soybean plants. These are under the standard 3D Model License of \url{https://www.turbosquid.com}. We then create an automatically generated farm model from the individual plants. The algorithm is composed of two main parts, random ground creation and plot generation, discussed below.

\subsection{Random Ground Creation}
The first step of the farm generation algorithm is to create a ground model based on the following input parameters. We take as input the size of the square farm. We also take as input a parameter that gives the number of random vertices that are generated within the square. These vertices are then used to create a Delaunay triangulation model of the environment. The last parameter is the range within which the heights of each vertex are randomly varied. This creates a ground model that has a random variation in height and is represented as a triangulated terrain. 

\subsection{Plot Generation}
\begin{figure}[htp]
    \centering
    \includegraphics[width=\linewidth]{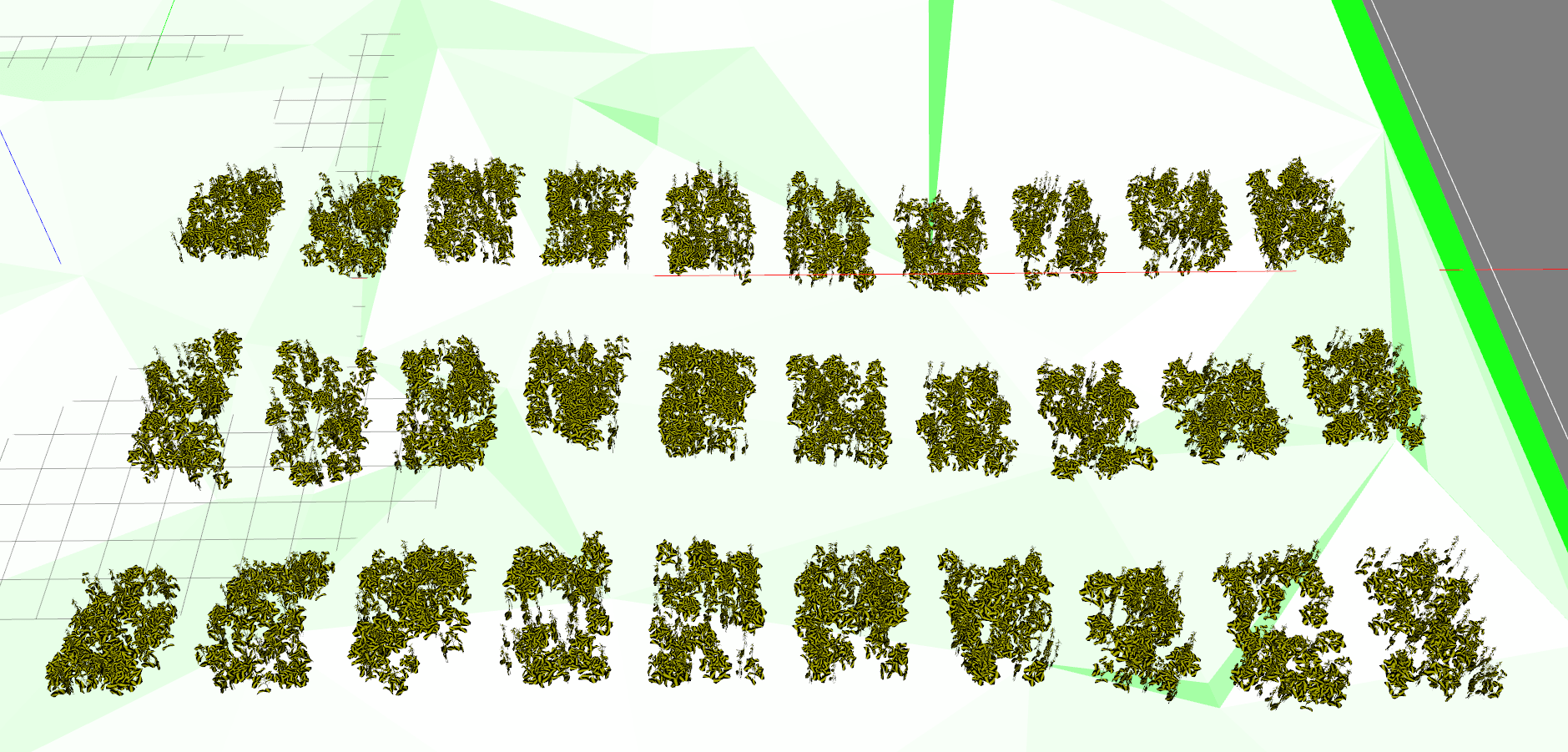}
    \caption{Screenshot taken of one of the generated phenotyping farms. This shows an overhead view of the entire farm. While it is hard to see in the figure, the ground model in this environment does vary between $\pm1$ m.}
    \label{fig:gazebo}
\end{figure}

Next, we create individual plots within this virtual farm. This step takes as input the following parameters: the width and length of the plots, the width and length between plots, the number of plants within each plot, the minimum and maximum scaling applied to individual plant models, and the row data. The row data is comprised of the number of plots within rows and the offset of where the row starts. Using these parameters, the individual plant models are randomly composed into the described environment. They are randomly scaled and rotated within the minimum and maximum range. Their heights are adjusted based on the height of the terrain at that point. An example of this is shown in Figure~\ref{fig:gazebo}.

\section{Experiments, Simulations, and Results} \label{chap:experiments}

We conducted experiments and data collection at Virginia Tech's Kentland Farm. Kentland Farm provided real-life examples of plotted crops that we could use to collect data. We describe the experiments in detail in this section, along with their results. We also describe simulations we conducted using the virtual farm generation tool that was developed and provide the results of those simulations.

\subsection{Kentland Experiments}

We performed real-world testing of our algorithm at Virginia Tech's Kentland Farm on a set of wheat plots maintained by Carl Griffey, Ph.D. for state variety testing in Spring 2019. Figure~\ref{fig:wheat} shows the processed point cloud data that we worked with and is comprised of 112 wheat plots. Flights were conducted during Spring and Summer 2019. However, for the algorithm pipeline, only a single flight is needed. For the wheat dataset provided, the flight was conducted in June 2019 during the peak height of the wheat crops.

The goal of our plot detection algorithm was to correctly find a bounding box for each of these plots. Figure~\ref{fig:clusters}.a highlights the bounding boxes when using just a $k$--means clustering algorithm to determine clusters within the point cloud data. As stated before, the algorithm correctly found many of the plots/clusters, but some did not fit at all. We fed the dimensions of the bounding boxes into the voting scheme step of the plot detection algorithm. Of the total 112 plots, the output of the voting scheme found 52 plots that fit within a certain orientation as well as width range, whereas, 54 were found to fit within a length range. We averaged all of the values within the range to determine the best-fit orientation, width, and length of the bounding boxes for the plots. The grid was then manually set to overlay on top of the point cloud data to extract the points within each plot. We showed this grid fitting previously in Figure~\ref{fig:clusters}.b.

\begin{figure}[htp]
\centering
\includegraphics[width=\linewidth]{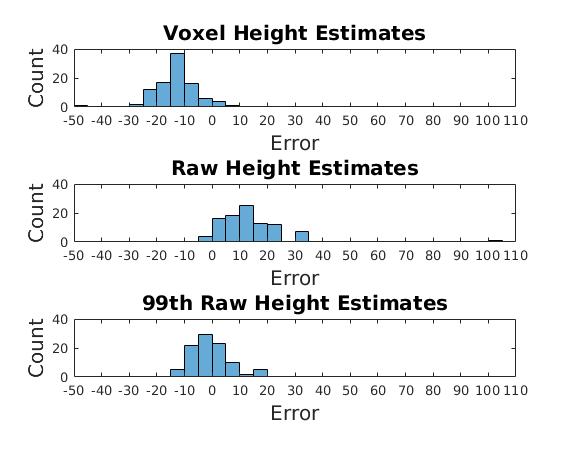}
\caption{The results of the Kentland wheat farm experiment. The top histogram shows the error between the manual measurements and the voxel-only estimations. The middle histogram shows the error using the raw data estimations. The bottom histogram shows the error using the 99th percentile of the raw data estimations.}
\label{fig:p_dif}
\end{figure}

The height estimation on the farm field gave promising results. We manually measured 3 plants within each plot of wheat. We averaged these 3 measurements to get a plot height. Figure~\ref{fig:p_dif} shows the error between the hand measurements and the height estimation. For the height estimation, we tried a few different methods. The first method was to use the voxel-filtered data set shown in Figure~\ref{fig:wheat}. We used the maximum height within each plot as the height estimate. We show this in the first histogram of Figure~\ref{fig:p_dif}. Since it was a down-sampled dataset, the majority of the output was underestimations and the average error between the manual measurements and height estimations was $\pm13.4$ \%. We show in the figure that the errors center around the -15 to -10 \% bin. The next method was to fit the raw plot data over the voxelized ground data. We cropped and recentered the raw data, but did not down-sample it using the voxel filter. Afterward, we used the plot detection bounding boxes to extract the points within each plot. These were then concatenated with the already voxel-filtered dataset in Figure~\ref{fig:wheat}. The maximum height within each plot was again used. We show the results of this methodology in the second histogram in Figure~\ref{fig:p_dif}. These gave less accurate results. Because we used the raw data, the new dataset was noisier and the error was $\pm13.8$ \%. The error centered around the +10 to +15 \% bin for this method, but there was also a single data point within the +100 to +105 \% bin due to noise. To get more accurate results, we extracted the 99th percentile height point from each plot. This gave the most accurate height estimations with the error shown in the bottom histogram of Figure~\ref{fig:p_dif}. The average RMSE across all plots was 6.1 cm using this method with an error of $\pm5.4$ \%. The error for this method centered around the -5 to 0 \% bin of the figure. The noisy outlier point in the raw-only methodology was also removed using this.

\begin{figure}[htp]
\centering
\includegraphics[width=\linewidth]{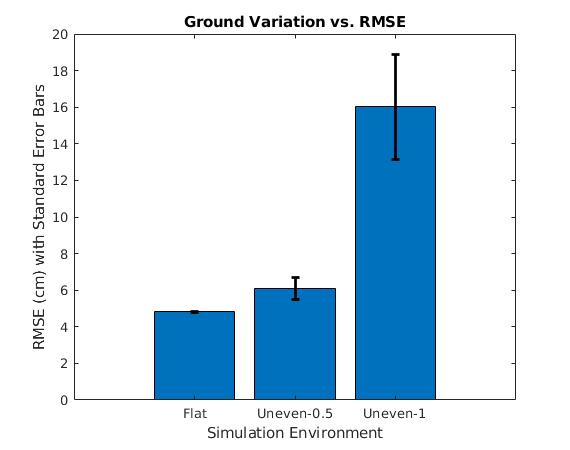}
\caption{The results of the simulated farm world experiments. The columns represent the 3 different simulation environments: one with a flat ground, one with a ground where the height varies $\pm0.5$ m, and one where the height varies $\pm1$ m. As seen in the figure, the more the height of the ground varies, the greater the RMSE and standard error is.}
\label{fig:sim_results_bar}
\end{figure}


\subsection{Soybean Plot Simulations}
\begin{figure}
    \centering
    \begin{subfigure}[b]{0.4\columnwidth}
        \includegraphics[width = \textwidth]{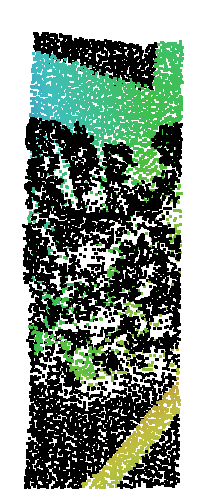} \label{fig:plot1} 
        \caption{}
    \end{subfigure}
    \begin{subfigure}[b]{0.4\columnwidth}
        \includegraphics[width = \textwidth]{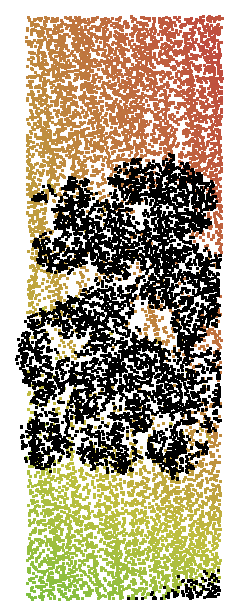} \label{fig:plot2} 
        \caption{}
    \end{subfigure}
    \caption{Zoomed in plots that are shown in Figure~\ref{fig:sim_ground} to highlight the ground plane estimation algorithm. These 2 plots are placed on a ground that varies. The colored points show points that lie within the ground plane while the black points are outliers. (a) The ground in this plot varies a lot compared to other plots and it can be seen that the colored points are not all of the ground points. (b) This ground varies also, but not as much. Almost all of the ground points in this are correctly colored.}
    \label{fig:sim_plots}
\end{figure}

\begin{figure}
    \centering
    \begin{subfigure}[b]{\columnwidth}
        \includegraphics[width = \textwidth]{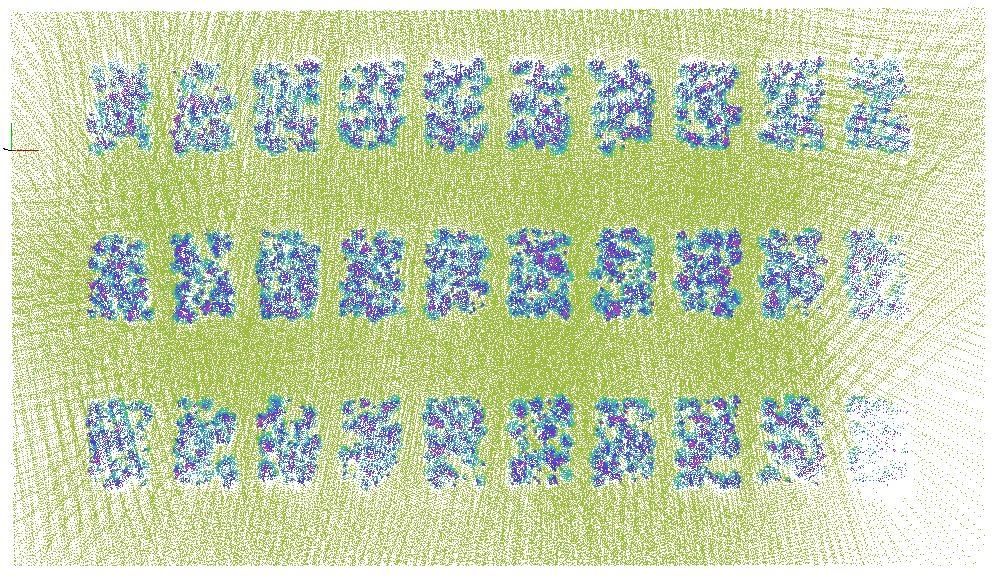}
        \caption{}
    \end{subfigure}
    \begin{subfigure}[b]{\columnwidth}
        \includegraphics[width = \textwidth]{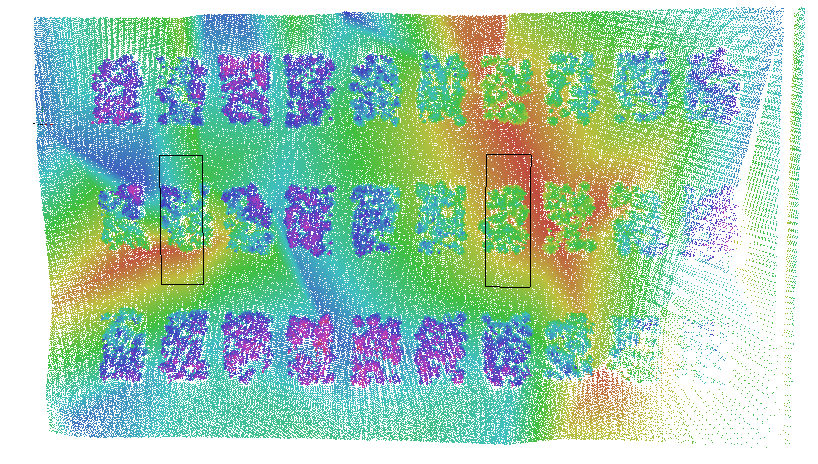}
        \caption{}
    \end{subfigure}
    \caption{2 different simulation environments. The different RGB values correspond to the height (z-value) of the points. (a) A simulated soybean farm world with a flat ground model. (b) A simulated soybean farm world with a ground model with 150 vertices that have $\pm1$ m height values.}
    \label{fig:sim_ground}
\end{figure}

We also performed simulations on 3 different environments generated using our farm generation toolchain. These environments consisted of the same number of plots. There were 3 rows of 10 plots, each consisting of 90 soybean plants. 
The difference between the 3 environments is the ground terrain. 2 of these environments are shown in Figure~\ref{fig:sim_ground}. 
Notably, Figure~\ref{fig:sim_plots} highlights the ground plane estimation algorithm. Figure~\ref{fig:sim_plots}.a consisted of a more extreme ground; however, the algorithm still estimates a good ground plane. Figure~\ref{fig:sim_plots}.b still had a ground model that varied, but not as intensely. This environment represents a more realistic scenario for a farming environment. The algorithm was able to find more of the ground points accurately.

The results of the simulations are shown in Figure~\ref{fig:sim_results_bar}. As expected, the environment with the flat ground model performed the best with an RMSE of 4.8 cm. The more realistic ground model is shown in the second column and resulted in an RMSE of 6.1 cm. Lastly, the environment with the more extreme ground model is shown in the third column and had an RMSE of 16.0 cm. As expected, as the ground varies more extremely, the RMSE of the height estimation algorithm increases. Also of note, the standard error of the mean increases correspondingly.

\section{Future Work} \label{chap:conclusionFutureWork}
In this chapter, we present several tools that are useful in high-throughput phenotyping of wheat and soybean plants. Specifically, we focus on easing the data processing pipeline of high-throughput phenotyping with a 3D LiDAR mounted on a UAV. We present point cloud processing techniques that produce as output a 3D model of the farm, semi-automatically find individual plots within a farm, and estimate the height of crops in each plot. In addition to hardware experiments, we also present a simulation toolchain to randomly produce virtual farms to test these algorithms. Our algorithms can estimate heights with an accuracy comparable to other methods. More importantly, our algorithms can produce other outputs (such as a 3D map and individual plots) that are needed for phenotyping that prior work has not focused on. An immediate avenue of future work is to test the algorithm for other types of farms besides the plot-based phenotyping farms considered in this chapter. The data processing software, real-world datasets, and simulation toolchain are released along with this work. We hope that this will make agriculture research more accessible to robotics research by reducing the hardware barrier to entry.

\renewcommand{\thechapter}{6}

\chapter{Environmental Monitoring: Change Validation}\label{chap:fire}

\section{Introduction}

In the previous chapter, we studied methods to detect changes happening in environments where changes occur slowly. However, there are many environments where changes happen fast. This rapid pace can cause high levels of noise generated by change detection methods leading to false positives. Due to this, there is a need to validate whether a change has occurred before deploying further resources to monitor the detected changes. One such example of a fast-changing environment is wildfires. Wildfire monitoring is a critical societal problem due to the pervasive threat of wildfires to our world~\cite{gill2013worldwide, wardle2003long}. Current wildfire detection practices incorporate reports from the general public, air patrols with pilots, satellite imagery, and data from sensor stations~\cite{canadaECC, ollero2006unmanned, xu2017real, allison2016airborne}. An ongoing challenge with these practices is that they are not always immediately available, preventing early fire detection. 
Instead, on-demand sensing with an Unmanned Aerial Vehicle (UAV) is a much more appealing solution. The recently announced XPrize Competition~\cite{XPRIZE} reflects this vision.
 
We envision a system where a network of sensor stations is deployed, that depending on their sensing range, makes it difficult to guarantee complete coverage of the target environment with only sensor stations. When focusing on early fire detection, their sensitivity to environmental phenomena is very high, leading to false positives. This is based on our experience working with N5 Sensors~\cite{N5Sensors_2023} where we have been developing machine-learning models for early wildfire detection using a network of sensor stations. Each station has multiple sensing modalities: chemical, particulate matter,  and infrared sensors. Due to false positives, there is a need to validate whether there is a fire. Once a fire is validated, it is also important to localize where it is actively burning to ensure quick fire management. In this chapter, we focus on the problem of wildfire validation and localization through the use of UAVs.

 \begin{figure}[ht!]
    \centering
    \includegraphics[width=\linewidth]{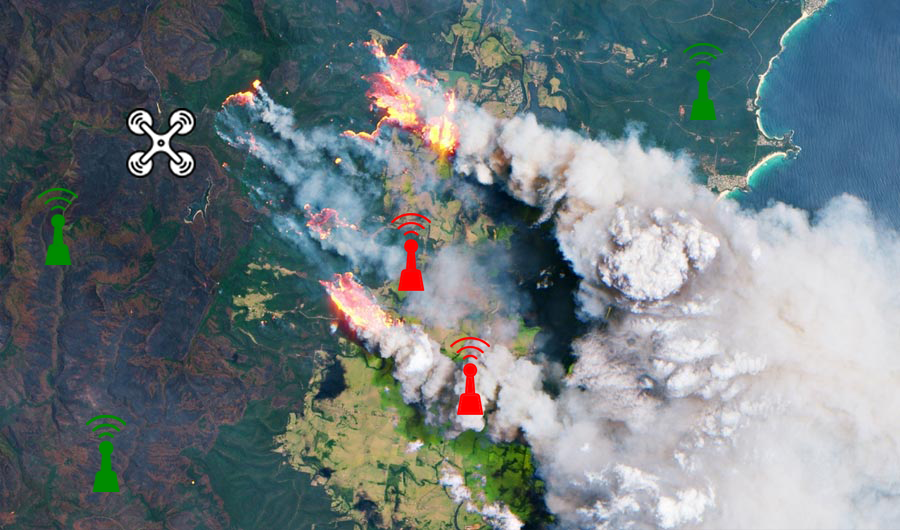}
    \caption{An example of sensor stations deployed out in the real world. Red stations detect the presence of a wildfire while green stations do not. UAV flies towards alerting stations for wildfire validation and localization.}
    \label{fig:overview_fire}
\end{figure}

While regular path planning typically aims to find the shortest or quickest path between two points, informative path planning focuses on learning the spatiotemporal distribution of an underlying field~\cite{binney2010informative,binney2012branch, Meliou2007}. 
For instance, a search-and-rescue UAV can utilize informative path planning to target its search in regions that have a higher likelihood of finding the entity being searched for~\cite{Lim2015}. This approach finds applications in areas such as precision agriculture~\cite{tokekar2016sensor} and environmental monitoring~\cite{hitz2017adaptive}. Standard informative path planning algorithms use metrics such as mutual information to drive the robot to reduce the uncertainty in the underlying field being estimated. See a recent comprehensive survey by Sung et al.~\cite{sung2023decision} for more applications of informative path planning. 

In our setup, we have the added challenge of validating whether there is a fire or not before localizing it. This is a difficult problem due to the complex dynamics of natural environments. Previous wildfire detection work already assumes a wildfire is taking place. We do not make this assumption. Specifically, the sensor stations will trigger the deployment of the UAV if one or more stations alert a positive detection. As we seek early wildfire detection, we expect the UAV to be deployed in cases where the stations have false positive alerts. Therefore, unlike standard informative path planning algorithms, our first objective is to plan paths for the UAV to determine if there is a fire or not. 

We first create a search area where the probability of detecting a fire, if there is one, is high. This is based on a model for how wildfire spreads. We then search in this area where the probabilities are updated based on real-time observations. If a fire is detected, we continue searching in the area for the location of the wildfire while also updating the probabilities based on observations.

We study the effectiveness of our proposed method, \textsc{Fire-GIPP}, against coverage-focused baselines. In summary, we make the following contributions:
\begin{itemize}
    \item Present \textsc{Fire-GIPP}, a greedy informative path planning algorithm that leverages probabilistic decision-making for wildfire validation. 
    \item Further, once a wildfire has been validated, \textsc{Fire-GIPP} is used for wildfire localization.
    \item We use thorough simulations to compare \textsc{Fire-GIPP} against coverage-focused baselines. \textsc{Fire-GIPP} is 53.89\% faster at localization than the baseline planners and 71.93\% faster over a comparable baseline in validation while being 90.78\% less computationally expensive. Similar results are seen in the high-fidelity AirSim simulations. 
\end{itemize}

\section{Related Work}\label{sec:related_fire}

\subsection{Wildfire Monitoring}
The use of UAVs as autonomous solutions has increased in recent years. One well-studied application is the use of UAVs for wildfire detection~\cite{merino2005cooperative}. In this particular work, the authors propose using a heterogeneous fleet of UAVs equipped with visual and infrared cameras to detect static fires and localize them. 

The authors extended this work for more general wildfire monitoring~\cite{merino2012unmanned}. In this extended work, they use the same setup as before, but instead of detection and localization of a static fire, they work on mapping the front of a dynamic fire. One difference is that there is no early alert system, the UAVs have to go on patrols for early fire detection. Also, all of the information from the fleet is sent and processed in a centralized manner. Another difference is that the control effort of the UAVs is not taken into consideration, they can fly to specific waypoints but the path selection is not optimized. 

Deep learning can also be utilized for autonomous fire detection~\cite{bouguettaya2022review, Lee2017}. Convolutional neural networks~\cite{Krizhevsky2012, Szegedy2014, simonyan2015deep} are used to make fire predictions from input images and are found to be highly accurate. The authors gathered images from online datasets for training and testing but did not deploy the CNNs on UAVs. Similar to before, they do not focus on the control effort of the UAV or provide any guarantees for early detection. Other works testing different machine-learning methods on the same task have also been studied~\cite{alexandrov2019analysis}. 

Remote sensing has been used for early fire detection~\cite{leblon2012use, allison2016airborne, bailon2020wildfire}. These works propose using aerial data, both from satellite imagery and UAVs, for wildfire monitoring. Similar to the work that was previously described, the drawback of using UAVs for early fire detection is the cost constraints. Specifically, limited battery life can limit the amount of patrol missions a UAV can go on. For satellite imagery, the location of focus can not always be guaranteed which can cause a delay in imagery and detection. Deployment of remote sensors in potential risk areas can help alleviate these concerns. 

\subsection{Informative Path Planning}
On the path planning side, one recent area of focus has been informative path planning~\cite{binney2010informative, Lim2015, binney2012branch, hitz2017adaptive}. In this area, observations from the environment are used to make informative path planning decisions to help create efficient paths. While it has been proposed to use informative path planning for wildfire management, these works do not study that problem. One example is by Lim et al.~\cite{Lim2015}, where they set up their planner as a Steiner tree problem~\cite{Kou1981} and apply it to other domains such as search-and-rescue. They also assume the object of interest is static which is an unrealistic assumption for wildfires.

Minimizing the spread of fire is another focus of research. This is called the firefighter problem~\cite{hartnell1995firefighter} where the goal is to minimize the spread of a fire in a graph by defending nodes in the graph. However, the focus of this body of work~\cite{Finbow2009} is to manage a spreading fire as opposed to detecting, validating, or locating it.

In summary, previous research focuses on a subset of the problems we solve in this chapter. They either focus on only wildfire detection or wildfire localization without an emphasis on planning. Previous work on informative path planning does not consider dynamic objects of interest or wildfires.

\section{Overview and Problem Formulation}\label{sec:prob_fire}
In this section, we give an overview of the actual system that we are developing and highlight the need for informative path planning with a UAV for wildfire validation.

\subsection{Motivating System Setup} We have a collection of $n$ sensor stations deployed in the environment. Each station has multi-modal sensors consisting of particulate, gas, infrared (IR) camera, CO2, temperature, and other sensors~\cite{N5Sensors_2023}. Each modality by itself is not sufficient to determine if there is a fire or not. However, we can combine the data from all modalities to detect fire. In our ongoing work, we use Long Short-Term Memory (LSTM) networks~\cite{lstm} since both spatial and temporal data are pivotal for wildfire detection. Formally, we use the LSTM-based predictor, \( \mathnormal{g} \), to predict the presence of a fire based on data from a subset of sensors, \( \mathnormal{F_D = g(S_n)} \). Each station is grouped up with its three closest neighbors to create a sub-network of four stations. The sensor readings from an entire sub-network are used as input to the LSTM for a total of 16 inputs (4 sensor modalities per sensor, 4 sensors per subnetwork). There is a single output from 0-1 for the LSTM that indicates the likelihood of a wildfire. 

\begin{figure}
    \centering
    \begin{subfigure}[b]{\columnwidth}%
        \includegraphics[width = \textwidth]{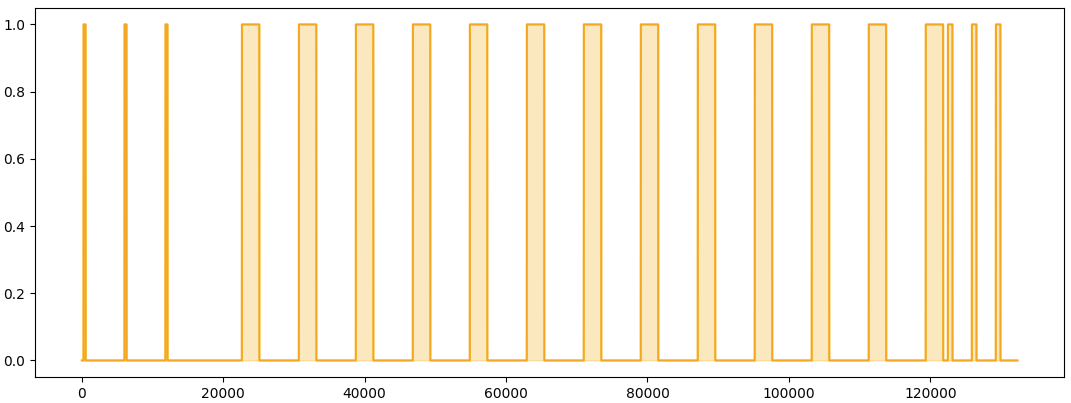}%
    \end{subfigure}%
    \hfill%
    \begin{subfigure}[b]{\columnwidth}%
        \includegraphics[width = \textwidth]{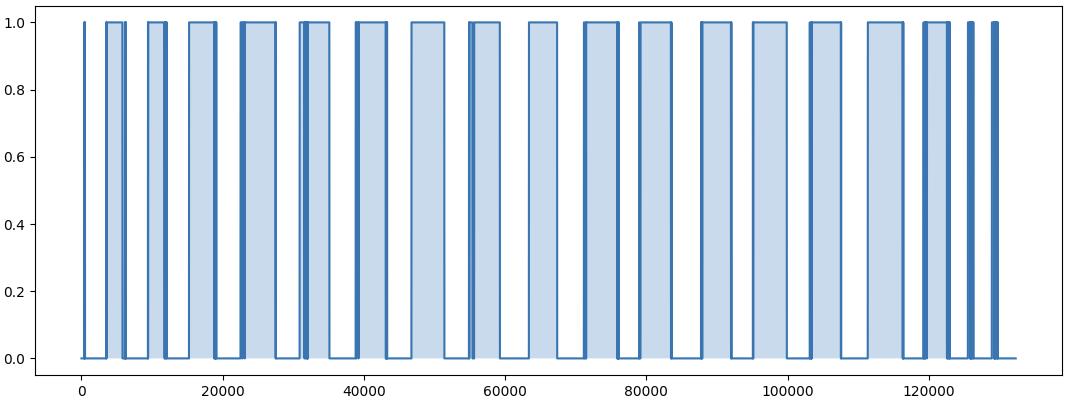}%
    \end{subfigure}%
    \caption{LSTM-based wildfire detection prediction (0: no-fire, 1: fire) over time for multiple real-world controlled burns. Ground truth in \textcolor{orange}{orange}. Predictions in \textcolor{blue}{blue}. }
    \label{fig:lstm}
\end{figure}

The results of our LSTM-based fire detection model can be seen in Figure~\ref{fig:lstm} with ground truth shown in orange and the model's prediction shown in blue. 10 fires at different locations were started for the data collected by the deployed ChemNodes. Half of these were used to train the model; whereas the other half were kept for testing. The model achieved an F-Score of 0.607 with an accuracy of 71.9\%, sensitivity of 46.4\%, and specificity of 81.1\% for the testing dataset. While the model does not achieve perfect accuracy, it is still able to detect every single fire that occurred. 

These preliminary findings show that the sensors are highly sensitive since they target early fire detection and this leads to false positive alarms. In this application, false negatives (undetected fire) are much more consequential than false positives (alerting when there is not a fire), since they could lead to disastrous consequences; therefore, high-sensitivity sensors are safer than low-sensitivity.  

Because of the false positives, once a sensor alerts, there is a need to validate whether it is a true or false positive. This motivates our work on using a UAV for wildfire validation and localization. 

\subsection{Informative Planning Problem Formulation} We assume the UAV has perfect sensing. The coverage area depends on the onboard sensor. It can be a camera FoV or a sensing range if it's a chemical sensor onboard. A fire is considered validated when one of three measurements are obtained by the UAV: fire is observed, smoke is observed, or a burned area is observed. A fire is invalidated if a false positive is detected. If the UAV observes the sensor location and neither fire, smoke, or a burned area is observed, then the alert is considered a false positive. To be complete, a planner must find a validation path if one exists; otherwise, it correctly reports that no solution is possible. Therefore, in our case, we seek a complete planner that validates the fire, if the fire exists, or reports that there is no fire. Amongst all complete planners, we seek the one that minimizes the time taken by the UAV. There can also be obstacles in the grid, such as a tall hill/mountain the UAV cannot fly over. The path must avoid collisions with these obstacles.

This is the first problem we focus on. 

\textbf{Problem 1} (\textit{Validation}) Given a grid \( \mathnormal{G} \), wind model \( \mathcal{W} \), and a set of sensors with positive alerts \( \mathnormal{s \subseteq S}\), find a path that validates a fire while minimizing time. 

Ideally, an active fire is observed in the same step as validation allowing for the deployment of resources to fight the fire. However, if one of the other observations is made; smoke or burned area, there is a need to locate the fire. We describe localization as observing a single actively burning cell. This is the second problem we focus on. 

\textbf{Problem 2} (\textit{Localization}) Given a grid \( \mathnormal{G} \), wind model \( \mathcal{W} \), and a set of sensors with positive alerts \( \mathnormal{s \subseteq S}\), find a path that localizes the fire while minimizing time. 

 Once an actively burning area is found, resources can be sent to map out the rest of the fire and extinguish it. This is beyond the scope of this problem and we refer to other work that focuses on firefront mapping~\cite{merino2012unmanned}. 

\section{Proposed Informative Path Planner} \label{sec:approach_fire}

In this section, we describe our proposed approach to solving the above two problems. Given our LSTM-based network is predicting a wildfire, we first create a search area based on the alerting sub-network of sensor stations. Then we deploy a UAV to that search area for wildfire validation while updating the probabilities of the search area with our observations. If a wildfire is validated, we continue this same process until it is located.

\subsection{Validation and Localization}
It is inefficient to search the entire grid to validate and locate the fire. Our planner first creates a subset of the total area to search in. Let $S$ be the set of sensors alerting the existence of a fire. A wind model $W$ of the environment and the locations of alerting sensors $s \subseteq S$ can be used to create a search area $A \subseteq G$.

\begin{figure*}[ht!]
    \centering
    \begin{subfigure}[c]{.48\textwidth}%
        \includegraphics[width = \textwidth]{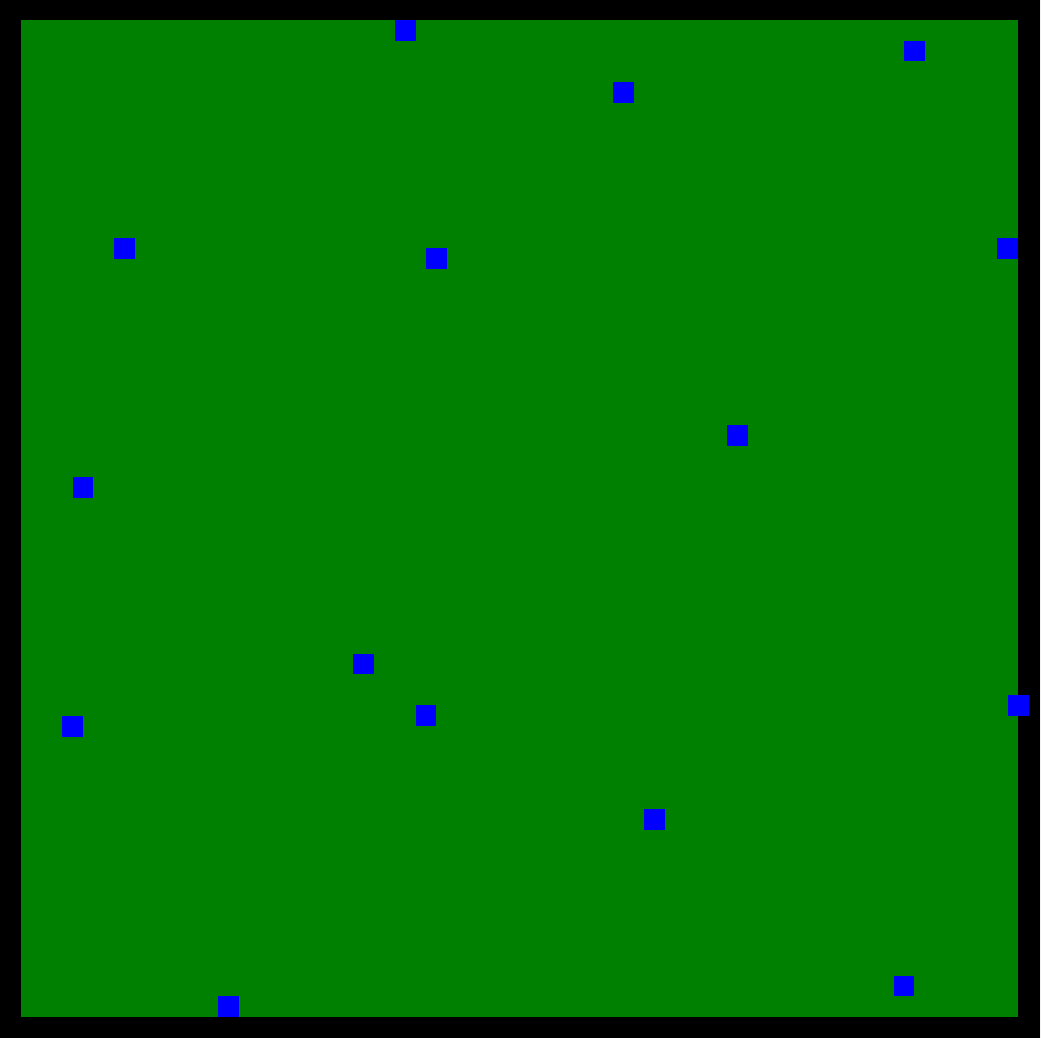}%
        \caption{}
    \end{subfigure}%
    \hfill
    \begin{subfigure}[c]{.48\textwidth}%
        \includegraphics[width = \textwidth]{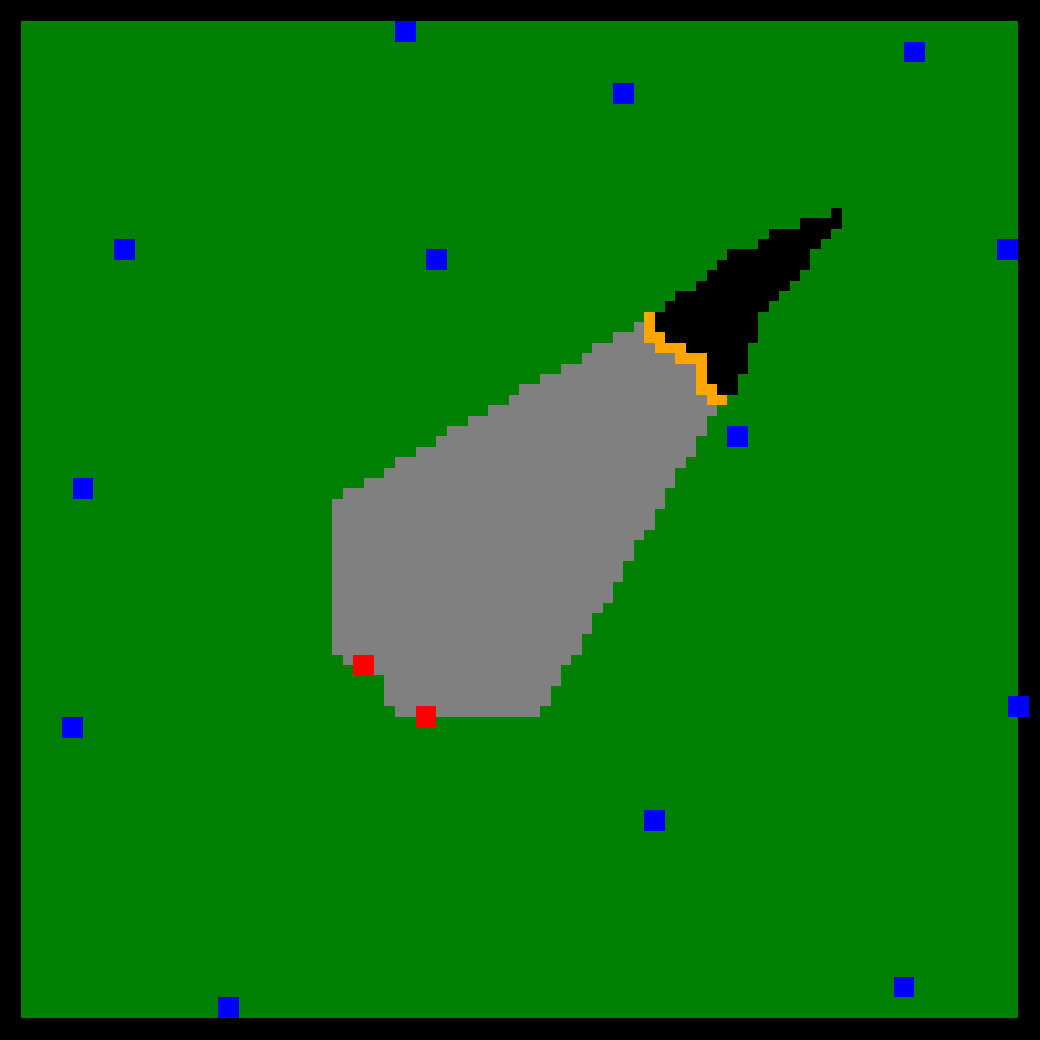}%
        \caption{}
    \end{subfigure}%
    \hfill
    \begin{subfigure}[c]{.48\textwidth}%
        \includegraphics[width = \textwidth]{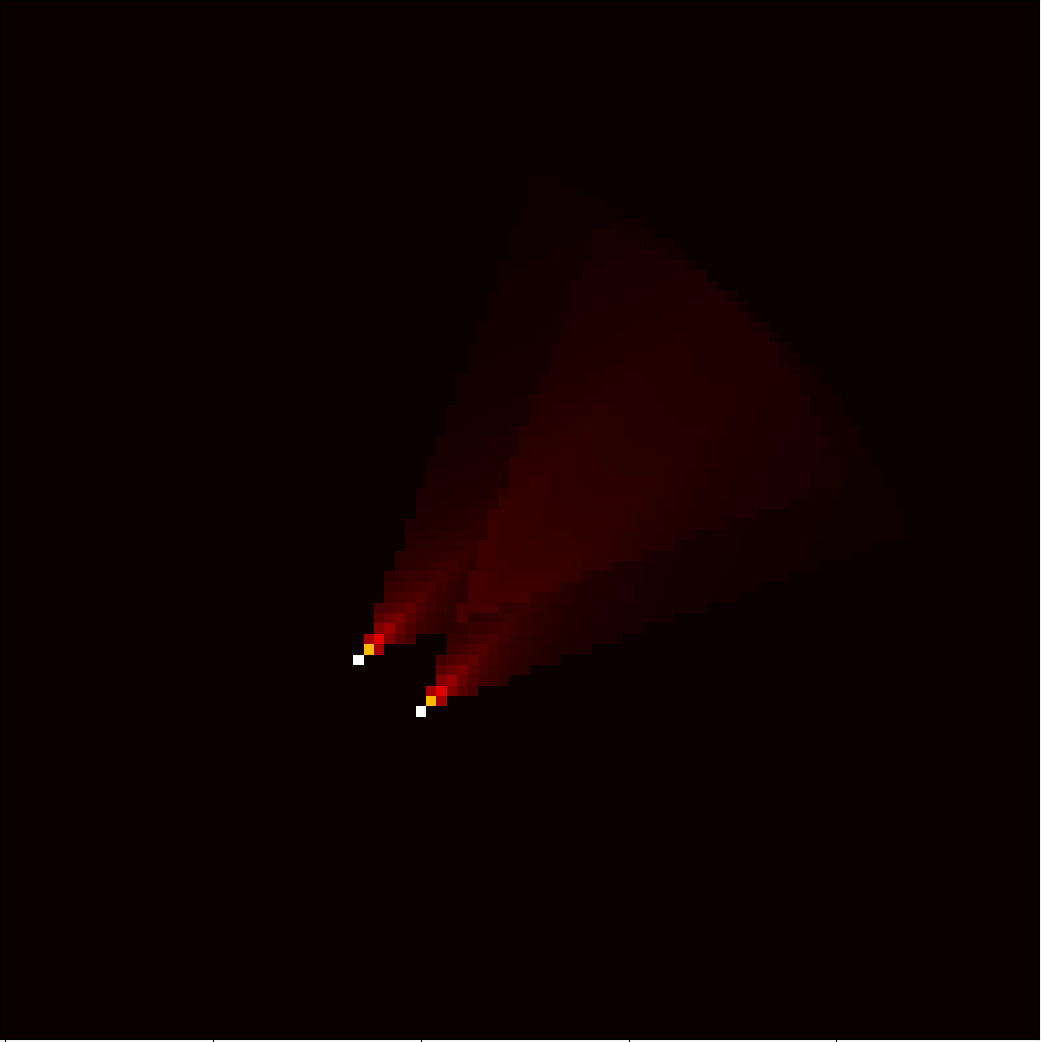}%
        \caption{}
    \end{subfigure}%
    \hfill
    \begin{subfigure}[c]{.48\textwidth}%
        \includegraphics[width = \textwidth]{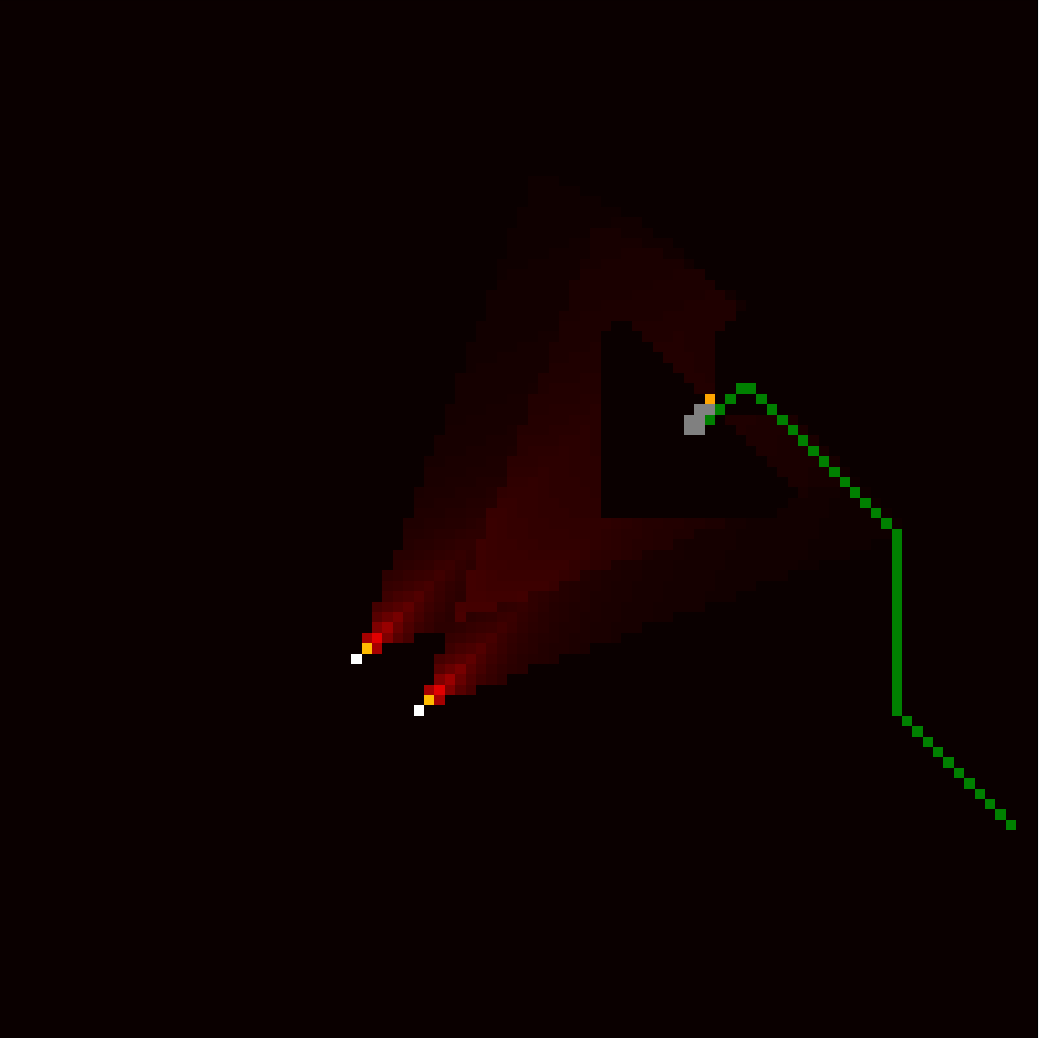}%
        \caption{}
    \end{subfigure}
    \caption{(a) An example of 15 sensors deployed in a 99x99 grid. The sensor locations are in blue. (b) A fire starts and spreads with smoke and 2 sensors (red) detect it. (c) Search area probability heatmap. (d) UAV path to validation and localization with observations.}
    \label{fig:sensor_search}
\end{figure*}

For each cell \( \mathnormal{c \in A}\), a validation probability can be assigned based on $W$ and the distance to the alerting sensor. Naturally, the further a cell is from the alerting sensor, the lower the validation probability. Also, the probability of the cell can be updated depending on the wind model $W$.

This search area probability field can be used for informative path-planning. In this situation with highly sensitive stationary sensors, a UAV can be used for fast fire validation. Assuming UAVs with high-resolution onboard sensors are located at a base station within \( \mathnormal{G} \), they can be deployed to the search area \( \mathnormal{A \subseteq G} \) to validate sensors that are detecting fires. A simple solution is to coverage plan for the entire search area \( \mathnormal{A} \). This is problematic for a few reasons. Firstly, UAVs have energy constraints that limit the time frame that they can actively monitor an area. Secondly, most wildfires are not static events. They continue to spread over time and if a UAV planner is too slow, the wildfire could spread past the search area \( \mathnormal{A} \). One way around this is to be less restrictive with the search area generation but UAV energy then becomes a bigger constraint. We can use the search area probability field to generate faster validation paths. Also, as the UAV moves from cell to cell in the grid, the high-resolution onboard sensors can be used to make observations and update the search area probability field depending on the observation made.

\subsection{Fire-GIPP}

\begin{algorithm}
    \caption{\textsc{Fire-GIPP}}\label{alg:ipp}
    \begin{algorithmic}
        \State $x_0 \gets$ UAV Initial Location
        \State $S_a \gets$ Alerting Sensors
        \State $A \gets$ Search Area
        \State $CP \gets$ \textproc{ClosestPoint}($x_0, A$)
        \State $path \gets$ \textproc{A*}($x_0, CP$)
        \For{$node$ in $path$}
            \State $o \gets$ \textproc{MoveToAndObserve}($node$)
            \If{$o$ is burning}
                \State Fire validated and located. Terminate.
            \ElsIf{$o$ is smoke or burned}
                \State Fire validated.
            \ElsIf{$node \in S_a$}
                \State False positive. Terminate.
            \EndIf
            \State \textproc{UpdateSearchArea}($A$)
        \EndFor
        \While{$A \neq \emptyset$}
            \State $node \gets$ \textproc{HighestProbableNeighbor}
            \State $o \gets$ \textproc{MoveToAndObserve}($node$)
            \If{$o$ is burning}
                \State Fire validated and located. Terminate.
            \ElsIf{$o$ is smoke or burned}
                \State Fire validated.
            \ElsIf{$node \in S_a$}
                \State False positive. Terminate.
            \EndIf
            \State \textproc{UpdateSearchArea}($A$)
        \EndWhile
    \end{algorithmic}
\end{algorithm}

With a probabilistic search area, we can now describe \textsc{Fire-GIPP}. We assume that initially the UAV is located in some cell within grid \( \mathnormal{G} \). First, we find the closest cell in the search area and navigate to it. We use the A* algorithm~\cite{Hart1968} to accomplish this to guarantee a collision-free path. As the UAV moves from point to point within the path, it makes observations of each cell $c_o \in G$. Each cell can be 1 of 4 values: \textbf{clear}, \textbf{burning}, \textbf{burned}, or \textbf{smoke}. If an actively burning cell is observed, we have validated and located the fire and can terminate. If a burned cell or cell with smoke is observed, the fire is validated but we still need to locate the fire. Based on the observation made, we can update the probabilities of every cell $c_a$ within the search area $A$. This probability update can be described using Bayes' theorem: 
\begin{equation}
P(c_a | c_o) = \frac{P(c_o | c_a) \cdot P(c_a)}{P(c_o)}
\end{equation}

Given we are modeling an accurate sensor, $P(c_o) = 1$:

\begin{equation}
P(c_a | c_o) = P(c_o | c_a) \cdot P(c_a)
\end{equation}

where $P(c_a)$ is the prior probability that we have assigned to $c_a$ in the search area $A$. $P(c_o|c_a)$ is the likelihood model and is determined from the observation and the wind model $W$.

Once the initial flight to the search area is complete and the search area probabilities have been updated, we select the neighbor cell of our current cell that has the highest probability. We move to that cell, observe it, and update the search area probabilities based on that observation. If the currently observed cell $c_o$ is where the alerting sensor is and we observe clear, we deem the event a false positive and terminate. Otherwise, we repeat the process of moving to the highest-probability neighboring cell, observing it, and updating the search area until the fire has been validated and located. \textsc{Fire-GIPP} is described in Algorithm~\ref{alg:ipp} and an example validation and localization flight is shown in Figure~\ref{fig:sensor_search}.

\section{Evaluation Setup}\label{sec:setup}

In this section, we discuss the specifics of our setup and implementation of the informative path planner. To start, we discuss the baseline coverage planners we compare \textsc{Fire-GIPP} against and then describe our simulation environment. 

\subsection{Baseline Coverage Planner}

For the baseline coverage planner, we follow the same initial A* path planner process as \textsc{Fire-GIPP}. Instead of updating the probabilities of the search area, we only use the observations to validate and locate the fire. Once the UAV gets to the search area, we use the search area to set up a traveling salesman problem~\cite{Laporte1992} (TSP). This ensures an approximate shortest-cost tour of the search area. The UAV flies along the tour until a validation and localization observation has been made. We call this closest point planner TSP-CP. We also implement a variant of this baseline planner where instead of going to the closest point in the search area, we go to the closest alerting sensor and call this TSP-Sensor. 

\subsection{Simulation Environments}

\begin{figure}[ht!]
    \centering
    \begin{subfigure}[b]{\columnwidth}%
        \includegraphics[width = \textwidth]{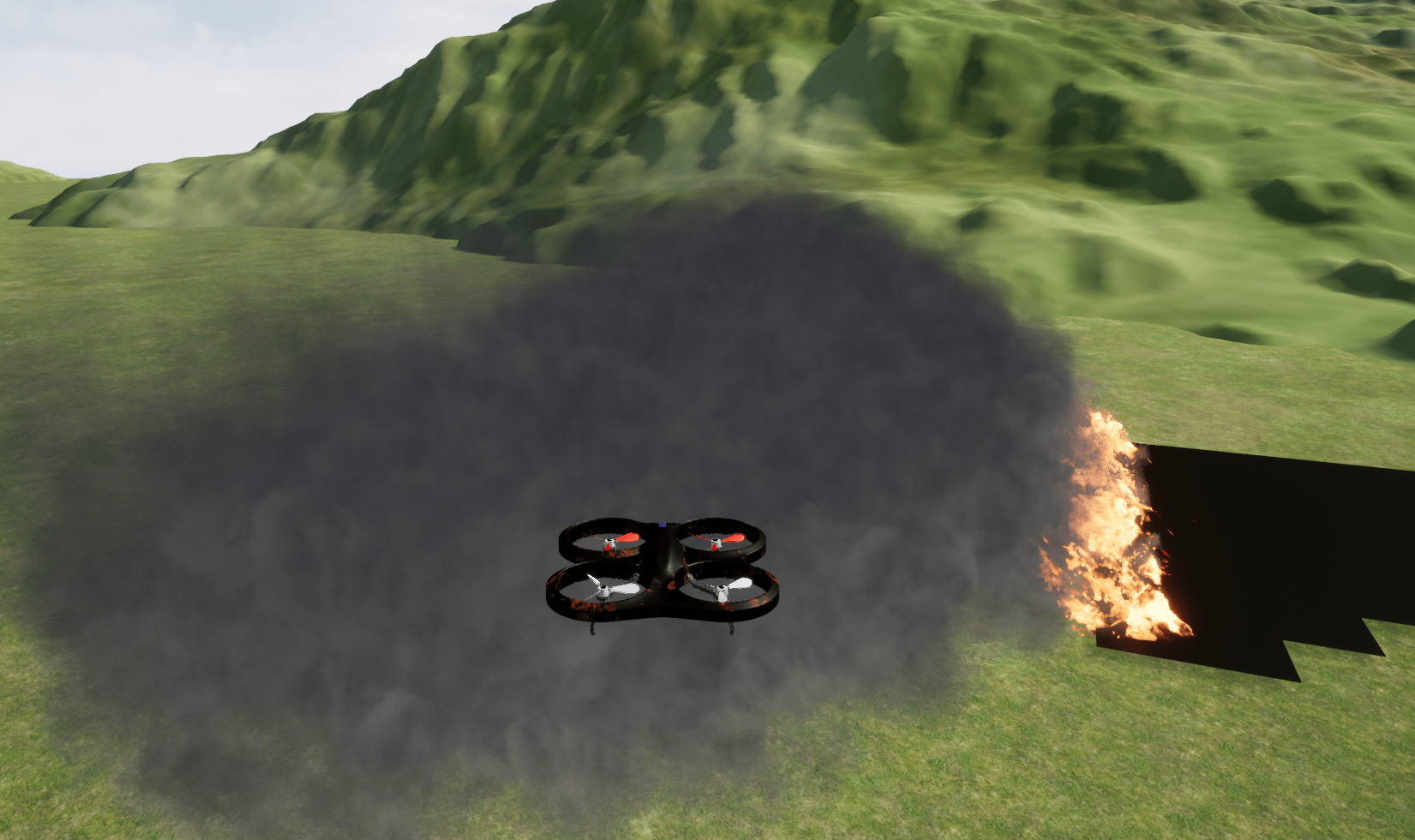}%
    \end{subfigure}%
    \hfill%
    \begin{subfigure}[b]{\columnwidth}%
        \includegraphics[width = \textwidth]{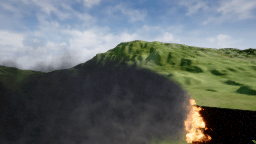}%
    \end{subfigure}%
    \caption{UAV and camera point of view in AirSim simulations.}
    \label{fig:observations_airsim_fire}
    \vspace{-5mm}
\end{figure}

We used 2 separate simulation environments. For more realistic planning and perception, we conducted high-fidelity simulations in AirSim~\cite{airsim2017fsr}. We equipped a virtual UAV with an RGB camera and spawned it into a realistic wildfire environment. Due to the limitations of AirSim, the simulated wildfires were static but the rest of the Fire-GIPP pipeline is the same. We also include non-AirSim simulations to evaluate Fire-GIPP in dynamic wildfire scenarios where the fire spreads over time. Further details about the dynamic simulation setup are discussed in the rest of this section. An example flight and camera data is shown in Figure~\ref{fig:observations_airsim_fire}.

\subsection{Wildfire Simulation Model}

The grid \( \mathnormal{G} \) is a 2-dimensional grid with dimensions $M$x$N$. Initially, every cell in this grid is marked as a clear cell. We assume a static wind model \( \mathnormal{W} \) comprised of 3 parameters. The wind direction \( \mathnormal{\vec{w}} \) is the direction the wind blows. The wind spread angle \( \mathnormal{\delta} \) is the offset in degrees from the wind direction  \( \mathnormal{\vec{w}} \) at which the fire and smoke in the simulation also spread. If this parameter is set to 0, the fire and smoke would only spread in a straight line in the grid. This is not realistic behavior of wildfires so \( \mathnormal{\delta} \) must have a greater than 0 value to allow realistic fire modeling. The third parameter is the smoke spread distance \( \mathnormal{\mu} \) which indicates how far from an actively burning cell the smoke spreads.

At time \( \mathnormal{t_0} \), a random cell \( \mathnormal{c_b} \) within the grid becomes a burning cell. At each subsequent timestamp, the fire spread can be modeled as follows: for every cell \( \mathnormal{c} \) that is burning, let \( \mathnormal{N} \) be the immediate neighborhood of \( \mathnormal{c} \). For every $c_n \in N$, if the angle between $\vec{c_nc}$ and $\vec{w}$ is less than $\delta$, change $c_n$ to a burning cell.

After this, each cell \( \mathnormal{c} \) that was burning during the previous timestamp becomes a burned cell. Once the fire has spread, we can model the smoke spread as well. It is similar to fire spread, except the neighborhood cells are different. Instead of looking at only immediate neighbor cells, for every cell \( \mathnormal{c} \) that is burning, let the smoke neighborhood \( \mathnormal{SN} \) include any cells within \( \mathnormal{\mu} \) distance of \( \mathnormal{c} \). For every $c_{sn} \in SN$, if the angle between $\vec{c_{sn}c}$ and $\vec{w}$ is less than $\delta$, change $c_{sn}$ to a smoke cell.

The model can be made more realistic if each cell is given a combustion probability based on fuel/dry vegetative level similar to how other models handle it~\cite{papadopoulos2011comparative, filippi2013representation}. Also, instead of assuming a static wind model \( \mathnormal{W} \), a wind vector field~\cite{schlipf2012model} can be used to give dynamic characteristics to the wind. However, this is not the focus of this work and a simpler model is sufficient for comparing \textsc{Fire-GIPP} with the baseline coverage planners.

\subsection{Sensor Deployment}
Next, we can discuss how we deploy sensors in the model. We base this deployment on how N5 Sensors~\cite{N5Sensors_2023} deploy their ChemNode sensors in the real world for fire detection. For every sensor \( \mathnormal{s \in S} \), we place it so that the distance between it and any other sensor must be greater than the parameter \( \mathnormal{d} \). This is to ensure that sensors are not too close to each other and that there is a decent amount of distribution and coverage of the grid \( \mathnormal{G} \). An example of the deployment is shown in Figure~\ref{fig:sensor_search}(a).

\subsection{Search Area Generation}

As stated before, these are based on multi-modality sensors consisting of particulate, IR camera, gas, and other sensors. They are highly sensitive to environmental changes and can sometimes alert when there are no fires. This is where the need for fire validation comes in. If there is a false positive, we do not want to waste resources to fight a fire that does not exist. Once a sensor detects a potential fire, we generate a search area for the UAV to go validate the fire. Each sensor also has a local wind sensor. We can use this measurement to help create the search area. Instead of the wind direction \( \mathnormal{\vec{w}} \), we define the search direction \( \mathnormal{\vec{\sigma} = -\vec{w}} \). We also define the search area distance threshold \( \mathnormal{\mu_{a}} \) which indicates how far from the alerting sensor the search area can be. This is also similar to the fire and smoke spread model, except we have a search area neighborhood \( \mathnormal{AN} \). For each sensor $s \in S$ detecting a fire,  let the search area neighborhood \( \mathnormal{AN} \) include any cells within \( \mathnormal{\mu_a} \) distance of \( \mathnormal{s} \). For every $c_{an} \in AN$, if the angle between $\vec{c_{an}s}$ and $\vec{\sigma}$ is less than $\delta$, $c_{an}$ becomes a cell in the search area $A$.

With a generated search area, we can now create a validation probability for each search area cell \( \mathnormal{c_{a}} \). We define this as follows:

\begin{equation}
\forall c_{a} \in A, P(c_{a}) = \frac{1}{(1 + \frac{\alpha|\theta_{a}|}{\delta})(\beta\norm{c_{a} - s})}
\end{equation}

The probability function \( \mathnormal{P} \) is based on two metrics, how far off from the wind direction a search cell is and how far from the sensor a search cell is. \( \mathnormal{\alpha} \) and \( \mathnormal{\beta} \) are parameters to control the effect of either of these metrics. The search area with probability values is shown in Figure~\ref{fig:sensor_search}(d) as a heat map.

\subsection{Search Area Update}

As stated, during a UAV flight we make observations $c_o$ and update the cells $c_a$ in the search area $A$. We update them as follows, if the observation made is a clear cell we know certain cells that the fire cannot be in due to the wind direction $\vec{w}$ and a wind noise parameter $\delta$, these are in the opposite wind direction $\vec{\sigma}$ from the current cell: 

\begin{equation}\label{eq:angle}
\forall c_a \in A, \theta_o = VectorAngle(\vec{c_ac_o}, \vec{\sigma})
\end{equation}

\begin{equation}\label{eq:prob}
P(c_a | c_o) = 0 \iff \theta_o < \delta
\end{equation}

$VectorAngle(\vec{a}, \vec{b})$ outputs the angle between the two input vectors. For a burning cell, we know the fire is past the current observation point in the wind direction $\vec{w}$. Therefore, we can update the probabilities of all cells in the opposite wind direction $\sigma$ to 0. We use equations \ref{eq:angle} and \ref{eq:prob} except $\delta$ is 90\degree. For a smoke cell, we know the fire has not yet reached this cell or cells perpendicular to it and the wind direction $\vec{w}$. Therefore, we can update the probability of all these cells to 0. We use the same equations as the burning cell, except we replace $\vec{\sigma}$ with $\vec{w}$.

The UAV may have a sensor with a specific FoV. For our purposes, we assume the UAV only observes a specific cell as opposed to having a larger FoV. This does not change the algorithm, we wanted to evaluate the planner under the highest information constraints. 

\section{Evaluation}\label{sec:eval_fire}
In this section, we compare the performance of \textsc{Fire-GIPP} to TSP-CP and TSP-Sensor. We compare results for fire localization and validation, including false positive validation. First, we describe the parameters implemented for testing.

\subsection{Simulation Parameters}

Table~\ref{tab:parameters} highlights the parameters we used for testing \textsc{Fire-GIPP} and the TSP baselines. We had a grid size of 99x99 where 50 sensors were randomly deployed as long as they were further than 5 cells away from any other sensor. 
The parameters were chosen to reflect the real data we observed from our pilot experiments used to train our LSTM detection network. 
We wanted the smoke to spread far enough for a sensor to detect a fire before it gets too close. This is similar to how the real-world N5 ChemNode~\cite{N5Sensors_2023} sensors measure particulate matter for fire detection. Similarly, we wanted the search area to be large enough so that the algorithms spent time validating and localizing the fire. Each test had a random value for $\vec{w}$, the wind direction because we wanted to test it under multiple wind directions. $\mu$ and $\mu_a$ were set to values of 7 and 8. These indicate how far the smoke spreads from a fire cell and the distance from the detected cell to include in the search area.

Lastly, we had a parameter called spread rate. This indicates how often the fire spreads relative to the UAV moving. We base this rate on a study on wildfire spread rates~\cite{Cruz2019} and the average UAV speed of a large consumer drone. Based on this information, the UAV moves every timestamp and the fire spreads every 20 timestamps. We also randomly placed an obstacle of varying size within the grid to add complexity to the environment and represent a hill/mountain the UAV cannot fly over.

\begin{table}[ht!]
    \centering
    \vspace{1.3mm}
    \caption{Testing Parameters}
    \begin{tabular}{lr}
        \toprule
        {Parameters} & Value \\
        \midrule 
           Grid Dimensions ($M$x$N$) & 99x99 \\
           Number of Sensors, $d$ & 50, 5  \\
           $w$, $\delta$ & Random, 60\degree  \\
           $\mu$, $\mu_a$ & 7, 8 \\
           $\alpha, \beta$ & 1, 1 \\
           UAV Speed & 1 cell/timestamp \\
           Spread Rate & 20 \\
        \bottomrule
    \end{tabular}
    \label{tab:parameters}
\end{table}

\subsection{Airsim Results}
We had 388 total runs per planner, one at each perimeter location of the grid excluding corners. The results of the simulations are shown in Table~\ref{tab:overall_results_airsim}. Fire-GIPP was 28.3\% faster to localize compared to the best baseline, TSP-CP; while only being 2.7\% slower to validate than TSP-Sensor, the validation-prioritized baseline.

\begin{table}[ht!]
    \centering
    \vspace{1.3mm}
    \caption{AirSim Validation and Localization Results}
    \begin{tabular}{lrrr}
    
        \toprule
        & \textsc{Fire-GIPP} & TSP-CP & TSP-Sensor \\
        \midrule
           Total Time-to-Localize\tablefootnote{Assuming UAV speed of 1 m/s.} (TtL) & \textbf{64.82} & 90.36 & 91.11  \\
           TtL (Non-Trivial) & \textbf{69.43} & 82.83 & 98.83  \\
           TtL (Search Area) & \textbf{6.81} & 20.21 & 28.27  \\
        \midrule
           Total Time-to-Validate\textsuperscript{2} (TtV) & 58.69 & 58.92 & \textbf{57.13}  \\
           TtV (Non-Trivial) & \textbf{48.53} & 52.18 & N/A  \\
           TtV (Search Area) & \textbf{5.27} & 8.92 & N/A  \\
        \bottomrule
    \end{tabular}
    \label{tab:overall_results_airsim}
\end{table}

\begin{figure*}[ht!]
    \centering
    \begin{subfigure}[c]{.48\textwidth}%
        \includegraphics[width = \textwidth]{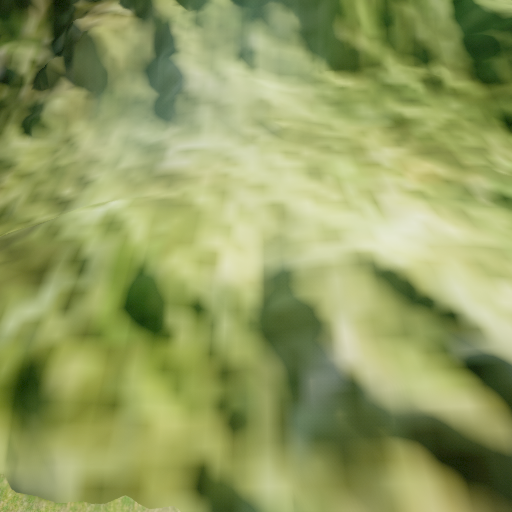}%
        \caption{}
    \end{subfigure}%
    \hfill
    \begin{subfigure}[c]{.48\textwidth}%
        \includegraphics[width = \textwidth]{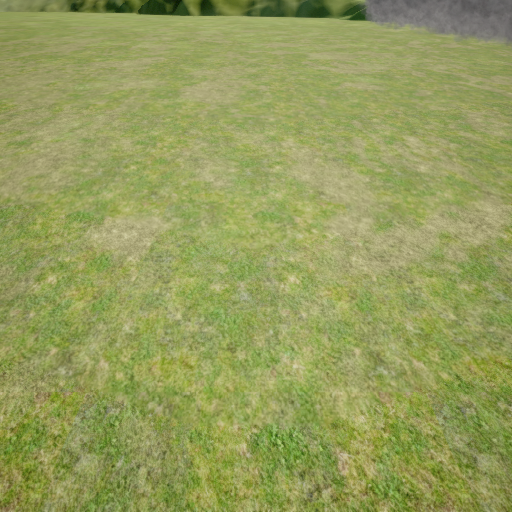}%
        \caption{}
    \end{subfigure}%
    \hfill
    \begin{subfigure}[c]{.48\textwidth}%
        \includegraphics[width = \textwidth]{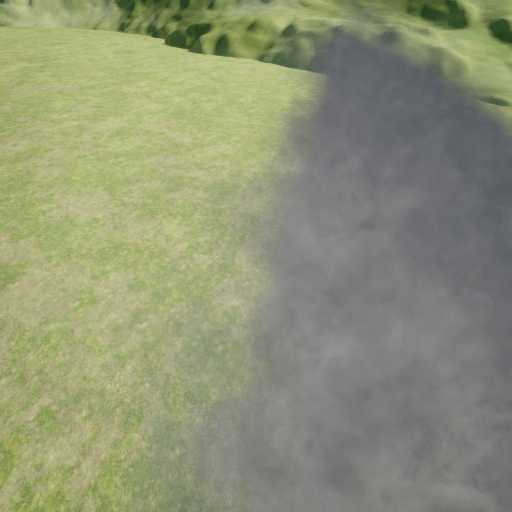}%
        \caption{}
    \end{subfigure}%
    \hfill
    \begin{subfigure}[c]{.48\textwidth}%
        \includegraphics[width = \textwidth]{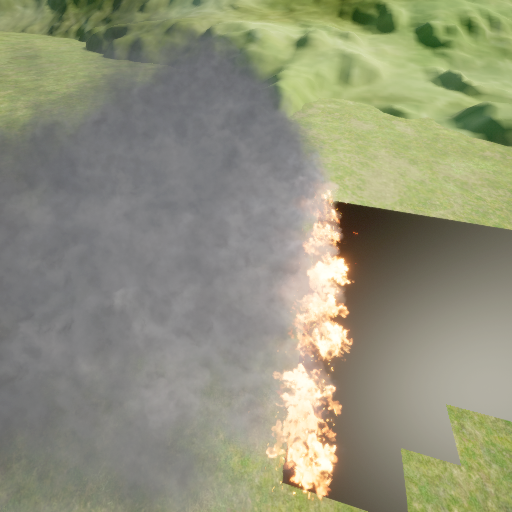}%
        \caption{}
    \end{subfigure}
    \caption{Images captured from the AirSim environment. (a) and (b) Example images where there is no fire, smoke, or burned area across different terrains. (c) An example image where smoke is detected, therefore the fire is validated. (d) An example image where the fire is detected, therefore the fire is localized.}
    \label{fig:airsim}
\end{figure*}

\subsection{Validation and Localization Results}
We evaluate the results of \textsc{Fire-GIPP}, TSP-CP, and TSP-Sensor in validation and localization in the non-AirSim simulations with dynamic wildfires. The results of these are shown in Table~\ref{tab:overall_results}. Localization is finding an actively burning cell. We had 16 unique simulated fires and initiated the UAV at each of the 388 border locations in the grid for all 3 planners for a total of 6208 test runs per planner. \textsc{Fire-GIPP} localized the fire at least 21.46\% faster than the best baseline, TSP-CP. Of the 6208 test runs, 5868 were non-trivial and were not localized during the initial flight to the search area for \textsc{Fire-GIPP} and TSP-CP, and 4301 were non-trivial for TSP-Sensor. \textsc{Fire-GIPP} and TSP-CP have identical initial paths since they both fly to the closest point first but TSP-Sensor flies to the sensor first, hence a different number of non-trivial test runs. \textsc{Fire-GIPP} localized the fire 16.35\% faster in these test runs. We also examined the time flown only in the search area, ignoring the initial time spent to get to the search area. This is done because the planning within the search area is where \textsc{Fire-GIPP} differs from the baselines so comparing this accurately shows the specific improvement of \textsc{Fire-GIPP}. In this metric, \textsc{Fire-GIPP} was at least 53.89\% faster. 

An additional 6208 test runs per planner had a random sensor giving a false positive alert. These, plus the 6208 true positive test runs were evaluated for validation time. The false positives test runs cannot be localized since there is no fire. For these 12416 test runs, TSP-Sensor was 3.31\% faster to validation. This is expected because observing the location of the sensor will 100\% validate the fire. We could make the initial point for \textsc{Fire-GIPP} the sensor location as well to close this gap, but chose not to prioritize validation over localization. TSP-CP and \textsc{Fire-GIPP} initially fly to the closest point in the search area where the fire potentially cannot be validated. For these 2 planners, 8666 test runs were non-trivial and were not validated during the initial flight. TSP-Sensor does not apply here since it always validates during its initial flight. Of these non-trivial flights, \textsc{Fire-GIPP} was 22.90\% faster. As we did for localization, we examined the time flown only in the search area. \textsc{Fire-GIPP} was 71.93\% faster in this metric.

Lastly, we consider the computational time of the algorithms. Since \textsc{Fire-GIPP} greedily selects neighbor cells, it is not computationally expensive compared to planning a TSP tour. \text{Fire-GIPP} had at least a 90.78\% lower computation time than the baselines. 

\begin{table}[ht!]
    \centering
    \vspace{1.3mm}
    \caption{Validation and Localization Results}
    \begin{tabular}{lrrr}
    
        \toprule
        & \textsc{Fire-GIPP} & TSP-CP & TSP-Sensor \\
        \midrule
           Total Time-to-Localize\textsuperscript{2} (TtL) & \textbf{69.91} & 89.01 & 94.50  \\
           TtL (Non-Trivial) & \textbf{68.63} & 83.75 & 82.04  \\
           TtL (Search Area) & \textbf{8.71} & 24.35 & 18.89  \\
        \midrule
           Total Time-to-Validate\textsuperscript{2} (TtV) & 67.29 & 80.70 & \textbf{65.06}  \\
           TtV (Non-Trivial) & \textbf{67.95} & 88.14 & N/A  \\
           TtV (Search Area) & \textbf{7.88} & 28.07 & N/A  \\
        \midrule
           Computational Time (ms) & \textbf{45.12} & 489.48 & 495.86 \\
        \bottomrule
    \end{tabular}
    \label{tab:overall_results}
\end{table}

\subsection{Validation: True Positives}
Here, we specifically investigate validation performance for true positives. The results of this are shown in Table~\ref{tab:val_results}. For these 6208 test runs, TSP-Sensor performed 4.66\% faster than \textsc{Fire-GIPP} as expected.  For \textsc{Fire-GIPP} and TSP-CP, 3649 test runs were non-trivial and were not validated during the initial flight to the search area.  \textsc{Fire-GIPP} validated 11.68\% faster than TSP-CP in this metric. For time in the search area, \textsc{Fire-GIPP} is 60.48\% faster than the TSP-CP.

\begin{table}[ht!]
    \centering
    \vspace{1.3mm}
    \caption{Validation Results: True Positives}
    \begin{tabular}{lrrr}
        \toprule
        & \textsc{Fire-GIPP} & TSP-CP & TSP-Sensor \\
        \midrule 
           Total Time-to-Validate\textsuperscript{2} (TtV) & 64.63 & 68.62 & \textbf{61.62}  \\
           TtV (Non-Trivial)  & \textbf{64.71} & 73.27 & N/A  \\
           TtV (Search Area) & \textbf{5.60} & 14.17 & N/A  \\
        \bottomrule
    \end{tabular}
    \label{tab:val_results}
\end{table}

\subsection{False Positives}
Lastly, we evaluated the other 6208 test runs which were false positives. In this scenario, a sensor is alerting when there is not a fire. These results are shown in Table~\ref{tab:fp_results}. As stated before, TSP-Sensor guarantees validation during its initial flight path. Even with this, it was only 2.16\% faster than \textsc{Fire-GIPP} at false positive validation. Further, for \textsc{Fire-GIPP} and TSP-CP, we study the non-trivial test runs where the false positive is not validated during the initial path, i.e., the closest point is not the sensor location. For these 5017 test runs, \textsc{Fire-GIPP} was 28.82\% faster. Investigating time within the search area, \textsc{Fire-GIPP} validates the false positives 74.99\% faster.

\begin{table}[ht!]
    \centering
    \vspace{1.3mm}
    \caption{Validation Results: False Positive}
    \begin{tabular}{lrrr}
        \toprule
        & \textsc{Fire-GIPP} & TSP-CP & TSP-Sensor \\
        \midrule 
           Total Time-to-Validate\textsuperscript{2} (TtV) & 70.23 & 93.91 & \textbf{68.71}  \\
           TtV (Non-Trivial) & \textbf{72.34} & 101.63 & N/A  \\
           TtV (Search Area) & \textbf{9.77} & 39.07 & N/A  \\
        \bottomrule
    \end{tabular}
    \label{tab:fp_results}
\end{table}

\section{Conclusion}\label{sec:con_fire}
We propose a greedy informative path planner for fast and efficient wildfire validation and localization. Our analysis shows that once in the search area, \textsc{Fire-GIPP} performs at least 53.89\% faster at localization than baseline coverage planners. For validation of fires and false positives, it performs 71.93\% better at validation within the search area than a comparable baseline while only being 3.31\% slower than a coverage planner that prioritizes immediate validation over balancing localization. Similar results are shown in the high-fidelity AirSim simulations. Further, \text{Fire-GIPP} was at least 90.78\% less computationally expensive than the baselines.

In future work, we would like to add more complexity to the evaluation model. Specifically, we would like to incorporate a more complicated wind model instead of having static wind characteristics. We would also like to test \textsc{Fire-GIPP} in real-world experiments.


\renewcommand{\thechapter}{7}

\chapter{Conclusion and Future Work}\label{chap:con}

\section{Summary of Contributions}

This dissertation solves planning and perception problems for UAVs in object and environmental monitoring applications. In Chapters~\ref{chap:prednbv} and~\ref{chap:mapnbv}, we focus on utilizing predictions to create efficient 3D reconstruction paths for UAVs targeting previously unknown objects. In Chapter~\ref{chap:insp}, we take it a step further and focus on generating paths for UAVs to inspect an unknown object in an unknown environment. In Chapter~\ref{chap:agri}, the focus shifts to monitoring the environment where we use the UAV to detect changes in crop height. Lastly, in Chapter~\ref{chap:fire}, data from deployed sensor stations is used to monitor for changes in the environment that indicate the presence of a wildfire. If detected, we create paths for UAVs for rapid validation and localization of the wildfires. The main contributions of this dissertation can be summarized as follows:
\begin{itemize}
    \item A next-best-view planner, \textit{Pred-NBV}, that performs object reconstruction without any prior information about the target object, using predictions to optimize information gain and control effort over a range of objects in a model-agnostic fashion.
    \item A decentralized, multi-agent extension of \textit{Pred-NBV}, named\textit{ MAP-NBV}, for active 3D reconstruction of various objects with a novel objective combining visual information gain and control effort.
    \item A receding horizon inspection planner, GATSBI, plans paths to efficiently inspect infrastructure when no prior information about the infrastructure or environment is present.
    \item A crop height estimation algorithm for a UAV equipped with a 3D LiDAR including crop-plot detection. We also present a simulated farm generation toolchain to help with future work.
    \item An LSTM model for wildfire detection based on environmental data collected from sensor stations. We also present a greedy informative path planner, \textsc{Fire-GIPP}, that leverages probabilistic decision-making for wildfire validation and localization. 
\end{itemize}

\section{Future Directions}

In this dissertation, we have solved a diverse set of planning and perception problems for UAVs in object and environmental monitoring settings. However, there are still many areas where further research would be beneficial. A few of these areas are discussed in more detail below. 

\subsection{Object Monitoring: Prediction-augmented Inspection}

As stated, in Chapter~\ref{chap:insp}, we focused on surface inspection of previously unknown objects in unknown environments. During the testing of our algorithm, we noticed the planner tended to overlap coverage as it planned inspection paths for newly explored surfaces. Currently, the planner creates a path to inspect what was discovered in all previous steps and then re-plans for inspection of the discovered areas in the next step. This leads to inefficient inspection plans where the UAV crosses over into sections it has already been in. In Chapters~\ref{chap:prednbv} and~\ref{chap:mapnbv}, we utilized predictions for 3D point cloud completion based on previous observations. An interesting avenue of research would be to combine the efforts of these chapters, i.e., using point cloud predictions to generate efficient inspection paths of unknown objects. People are very good at taking in a single observation of an object and generalizing (or predicting) what the rest of that object looks like. Currently, human inspectors can use this generalization ability to manually fly an efficient inspection path around an object of interest. Our point cloud prediction work mimics this generalization ability and combining it with our inspection work could help create similarly efficient inspection paths for manually-operated inspection UAVs. As the UAV makes more observations during its inspection flight, the rest of the object can be predicted. If a UAV is close to an area where points are predicted to be but currently have not been observed or inspected, the UAV should fly closer to the predictions to observe and then inspect these areas as opposed to coming back to them at a later point. 

\subsection{Object Monitoring: Co-Robotic Inspection}

Another task humans are great at but difficult to replicate is on-the-fly decision-making when presented with unexpected or previously unknown data. One example of this is in manual operation of UAVs in infrastructure inspection. An inspector might have a pre-planned route in mind when flying the UAV around infrastructure of interest. However, when a defect is detected, the inspector might want to fly closer to the defect to collect higher resolution data or investigate it with further scrutiny. How can this behavior be replicated in autonomous systems? The recent advent of Large Language Models (LLMs)~\cite{llm2024} provides one such method. To simplify the problem, one approach is to separate the high-level decision-making from the planning. We can use our work to generate autonomous inspection paths while a human inspector studies the incoming data from the UAV and decides if an area needs further scrutiny. If an area does need further scrutiny, the human inspector can indicate this to the planner which can utilize the zero-shot planning ability of LLMs to generate detour paths. This methodology would allow the interruption of the autonomous inspection path to allow the UAV to gather higher-resolution data of areas of interest and then resume the autonomous inspection flight afterward, similar to how a human inspector would manually fly a UAV for an inspection flight. 

\subsection{Environmental Monitoring: Change Prediction}

In Chapter~\ref{chap:fire}, we used environmental data with LSTMs to detect the presence of a wildfire. If a wildfire was detected, we planned for quick wildfire validation and localization. Currently, in the literature, wildfire front mapping is also a well-studied problem. Solutions to this problem target mapping every actively burning location of the fire. However, due to the dynamic nature of wildfires, any delay in response could lead to unnecessary destruction. Therefore one interesting potential area of research is using environmental data to predict exactly where a wildfire will spread to. These predictions can be used to rapidly deploy defensive measures to areas before the fire spreads there. This would allow UAVs to be used in a defensive approach as opposed to an offensive approach where they might have to keep up with the fire.
\titleformat{\chapter}
{\normalfont\large}{Appendix \thechapter:}{1em}{}

\renewcommand{\baselinestretch}{1}
\small\normalsize

\bibliographystyle{unsrt}
\bibliography{main}

\begin{thebibliography}{100}

\bibitem{aleotti2014global}
Jacopo Aleotti, Dario~Lodi Rizzini, Riccardo Monica, and Stefano Caselli.
\newblock Global registration of mid-range 3d observations and short range next best views.
\newblock In {\em 2014 IEEE/RSJ International Conference on Intelligent Robots and Systems (IROS)}, pages 3668--3675. IEEE, 2014.

\bibitem{shapenet2015}
Angel~X. Chang, Thomas Funkhouser, Leonidas Guibas, Pat Hanrahan, Qixing Huang, Zimo Li, Silvio Savarese, Manolis Savva, Shuran Song, Hao Su, Jianxiong Xiao, Li~Yi, and Fisher Yu.
\newblock {ShapeNet: An Information-Rich 3D Model Repository}.
\newblock Technical Report arXiv:1512.03012 [cs.GR], Stanford University --- Princeton University --- Toyota Technological Institute at Chicago, 2015.

\bibitem{BABOOMS_ICRA_15}
{A. Bircher, K. Alexis, M. Burri, P. Oettershagen, S. Omari, T. Mantel and R. Siegwart}.
\newblock Structural inspection path planning via iterative viewpoint resampling with application to aerial robotics.
\newblock In {\em Robotics and Automation (ICRA), 2015 IEEE International Conference on}, pages 6423--6430, May 2015.

\bibitem{tsouros2019review}
Dimosthenis~C Tsouros, Stamatia Bibi, and Panagiotis~G Sarigiannidis.
\newblock A review on uav-based applications for precision agriculture.
\newblock {\em Information}, 10(11):349, 2019.

\bibitem{villa2016overview}
Tommaso~Francesco Villa, Felipe Gonzalez, Branka Miljievic, Zoran~D Ristovski, and Lidia Morawska.
\newblock An overview of small unmanned aerial vehicles for air quality measurements: Present applications and future prospectives.
\newblock {\em Sensors}, 16(7):1072, 2016.

\bibitem{semsch2009autonomous}
Eduard Semsch, Michal Jakob, Du{\v{s}}an Pavlicek, and Michal Pechoucek.
\newblock Autonomous uav surveillance in complex urban environments.
\newblock In {\em 2009 IEEE/WIC/ACM International Joint Conference on Web Intelligence and Intelligent Agent Technology}, volume~2, pages 82--85. IEEE, 2009.

\bibitem{shakhatreh2019unmanned}
Hazim Shakhatreh, Ahmad~H Sawalmeh, Ala Al-Fuqaha, Zuochao Dou, Eyad Almaita, Issa Khalil, Noor~Shamsiah Othman, Abdallah Khreishah, and Mohsen Guizani.
\newblock Unmanned aerial vehicles (uavs): A survey on civil applications and key research challenges.
\newblock {\em Ieee Access}, 7:48572--48634, 2019.

\bibitem{jin2020research}
Wenbo Jin, Jixing Yang, Yudong Fang, and Wenchuan Feng.
\newblock Research on application and deployment of uav in emergency response.
\newblock In {\em 2020 IEEE 10th International Conference on Electronics Information and Emergency Communication (ICEIEC)}, pages 277--280. IEEE, 2020.

\bibitem{DebevecPhd1996}
Paul~E. Debevec.
\newblock {\em Modeling and Rendering Architecture from Photographs}.
\newblock PhD thesis, University of California at Berkeley, Computer Science Division, Berkeley CA, 1996.

\bibitem{pichat2018survey}
Jonas Pichat, Juan~Eugenio Iglesias, Tarek Yousry, S{\'e}bastien Ourselin, and Marc Modat.
\newblock A survey of methods for 3d histology reconstruction.
\newblock {\em Medical image analysis}, 46:73--105, 2018.

\bibitem{dong20174d}
Jing Dong, John~Gary Burnham, Byron Boots, Glen Rains, and Frank Dellaert.
\newblock 4d crop monitoring: Spatio-temporal reconstruction for agriculture.
\newblock In {\em 2017 IEEE International Conference on Robotics and Automation (ICRA)}, pages 3878--3885. IEEE, 2017.

\bibitem{bitzidou2013multi}
Malamati Bitzidou, Dimitrios Chrysostomou, and Antonios Gasteratos.
\newblock Multi-camera 3d object reconstruction for industrial automation.
\newblock In {\em Advances in Production Management Systems. Competitive Manufacturing for Innovative Products and Services: IFIP WG 5.7 International Conference, APMS 2012, Rhodes, Greece, September 24-26, 2012, Revised Selected Papers, Part I}, pages 526--533. Springer, 2013.

\bibitem{xu2021toward}
Yusheng Xu and Uwe Stilla.
\newblock Toward building and civil infrastructure reconstruction from point clouds: A review on data and key techniques.
\newblock {\em IEEE Journal of Selected Topics in Applied Earth Observations and Remote Sensing}, 14:2857--2885, 2021.

\bibitem{ariaratnam2001assessment}
Samuel~T Ariaratnam, Ashraf El-Assaly, and Yuqing Yang.
\newblock Assessment of infrastructure inspection needs using logistic models.
\newblock {\em Journal of infrastructure systems}, 7(4):160--165, 2001.

\bibitem{madanat1994optimal}
Samer Madanat and Moshe Ben-Akiva.
\newblock Optimal inspection and repair policies for infrastructure facilities.
\newblock {\em Transportation science}, 28(1):55--62, 1994.

\bibitem{chang2003health}
Peter~C Chang, Alison Flatau, and Shih-Chii Liu.
\newblock Health monitoring of civil infrastructure.
\newblock {\em Structural health monitoring}, 2(3):257--267, 2003.

\bibitem{ellingwood2005risk}
Bruce~R Ellingwood.
\newblock Risk-informed condition assessment of civil infrastructure: state of practice and research issues.
\newblock {\em Structure and infrastructure engineering}, 1(1):7--18, 2005.

\bibitem{gillins2016cost}
Matthew~N Gillins, Daniel~T Gillins, and Christopher Parrish.
\newblock Cost-effective bridge safety inspections using unmanned aircraft systems (uas).
\newblock In {\em Geotechnical and Structural Engineering Congress 2016}, pages 1931--1940, 2016.

\bibitem{eschmann2013high}
C~Eschmann, C-M Kuo, C-H Kuo, and C~Boller.
\newblock High-resolution multisensor infrastructure inspection with unmanned aircraft systems.
\newblock {\em The International Archives of the Photogrammetry, Remote Sensing and Spatial Information Sciences}, 40:125--129, 2013.

\bibitem{duque2018synthesis}
Luis Duque, Junwon Seo, and James Wacker.
\newblock Synthesis of unmanned aerial vehicle applications for infrastructures.
\newblock {\em Journal of Performance of Constructed Facilities}, 32(4):04018046, 2018.

\bibitem{ellenberg2015use}
A~Ellenberg, L~Branco, A~Krick, I~Bartoli, and A~Kontsos.
\newblock Use of unmanned aerial vehicle for quantitative infrastructure evaluation.
\newblock {\em Journal of Infrastructure Systems}, 21(3):04014054, 2015.

\bibitem{gopalakrishnan2018crack}
Kasthurirangan Gopalakrishnan, Hoda Gholami, Akash Vidyadharan, Alok Choudhary, Ankit Agrawal, et~al.
\newblock Crack damage detection in unmanned aerial vehicle images of civil infrastructure using pre-trained deep learning model.
\newblock {\em Int. J. Traffic Transp. Eng}, 8(1):1--14, 2018.

\bibitem{lemos2023automatic}
Rafael Lemos, Rafael Cabral, Diogo Ribeiro, Ricardo Santos, Vinicius Alves, and Andr{\'e} Dias.
\newblock Automatic detection of corrosion in large-scale industrial buildings based on artificial intelligence and unmanned aerial vehicles.
\newblock {\em Applied Sciences}, 13(3):1386, 2023.

\bibitem{maes2019perspectives}
Wouter~H Maes and Kathy Steppe.
\newblock Perspectives for remote sensing with unmanned aerial vehicles in precision agriculture.
\newblock {\em Trends in plant science}, 24(2):152--164, 2019.

\bibitem{gokool2023crop}
Shaeden Gokool, Maqsooda Mahomed, Richard Kunz, Alistair Clulow, Mbulisi Sibanda, Vivek Naiken, Kershani Chetty, and Tafadzwanashe Mabhaudhi.
\newblock Crop monitoring in smallholder farms using unmanned aerial vehicles to facilitate precision agriculture practices: a scoping review and bibliometric analysis.
\newblock {\em Sustainability}, 15(4):3557, 2023.

\bibitem{mohapatra2022early}
Ankita Mohapatra and Timothy Trinh.
\newblock Early wildfire detection technologies in practice—a review.
\newblock {\em Sustainability}, 14(19):12270, 2022.

\bibitem{kumar2018impact}
Subramania~Ananda Kumar and Paramasivam Ilango.
\newblock The impact of wireless sensor network in the field of precision agriculture: A review.
\newblock {\em Wireless Personal Communications}, 98:685--698, 2018.

\bibitem{murugan2017development}
Deepak Murugan, Akanksha Garg, and Dharmendra Singh.
\newblock Development of an adaptive approach for precision agriculture monitoring with drone and satellite data.
\newblock {\em IEEE Journal of Selected Topics in Applied Earth Observations and Remote Sensing}, 10(12):5322--5328, 2017.

\bibitem{casbeer2005forest}
David~W Casbeer, Randal~W Beard, Timothy~W McLain, Sai-Ming Li, and Raman~K Mehra.
\newblock Forest fire monitoring with multiple small uavs.
\newblock In {\em Proceedings of the 2005, American Control Conference, 2005.}, pages 3530--3535. IEEE, 2005.

\bibitem{Lim2015}
Zhan~Wei Lim, David Hsu, and Wee~Sun Lee.
\newblock {\em Adaptive Informative Path Planning in Metric Spaces}, pages 283--300.
\newblock Springer International Publishing, Cham, 2015.

\bibitem{bostrom2019informative}
Per Bostr{\"o}m-Rost.
\newblock {\em On informative path planning for tracking and surveillance}, volume 1838.
\newblock Link{\"o}ping University Electronic Press, 2019.

\bibitem{scott2003view}
William~R Scott, Gerhard Roth, and Jean-Fran{\c{c}}ois Rivest.
\newblock View planning for automated three-dimensional object reconstruction and inspection.
\newblock {\em ACM Computing Surveys (CSUR)}, 35(1):64--96, 2003.

\bibitem{delmerico2018comparison}
Jeffrey Delmerico, Stefan Isler, Reza Sabzevari, and Davide Scaramuzza.
\newblock A comparison of volumetric information gain metrics for active 3d object reconstruction.
\newblock {\em Autonomous Robots}, 42(2):197--208, 2018.

\bibitem{tarabanis1995survey}
Konstantinos~A Tarabanis, Peter~K Allen, and Roger~Y Tsai.
\newblock A survey of sensor planning in computer vision.
\newblock {\em IEEE transactions on Robotics and Automation}, 11(1):86--104, 1995.

\bibitem{kuipers1991robot}
Benjamin Kuipers and Yung-Tai Byun.
\newblock A robot exploration and mapping strategy based on a semantic hierarchy of spatial representations.
\newblock {\em Robotics and autonomous systems}, 8(1-2):47--63, 1991.

\bibitem{yamauchi1997frontier}
Brian Yamauchi.
\newblock A frontier-based approach for autonomous exploration.
\newblock In {\em Proceedings 1997 IEEE International Symposium on Computational Intelligence in Robotics and Automation CIRA'97.'Towards New Computational Principles for Robotics and Automation'}, pages 146--151. IEEE, 1997.

\bibitem{georgakis2022uncertainty}
Georgios Georgakis, Bernadette Bucher, Anton Arapin, Karl Schmeckpeper, Nikolai Matni, and Kostas Daniilidis.
\newblock Uncertainty-driven planner for exploration and navigation.
\newblock In {\em 2022 International Conference on Robotics and Automation (ICRA)}, pages 11295--11302, 2022.

\bibitem{Yang_2019}
Bo~Yang, Stefano Rosa, Andrew Markham, Niki Trigoni, and Hongkai Wen.
\newblock Dense 3d object reconstruction from a single depth view.
\newblock {\em {IEEE} Transactions on Pattern Analysis and Machine Intelligence}, 41(12):2820--2834, dec 2019.

\bibitem{yuan2018pcn}
Wentao Yuan, Tejas Khot, David Held, Christoph Mertz, and Martial Hebert.
\newblock Pcn: Point completion network.
\newblock In {\em 2018 international conference on 3D vision (3DV)}, pages 728--737. IEEE, 2018.

\bibitem{yu2021pointr}
Xumin Yu, Yongming Rao, Ziyi Wang, Zuyan Liu, Jiwen Lu, and Jie Zhou.
\newblock Pointr: Diverse point cloud completion with geometry-aware transformers.
\newblock In {\em Proceedings of the IEEE/CVF international conference on computer vision}, pages 12498--12507, 2021.

\bibitem{Peralta2020}
Daryl Peralta, Joel Casimiro, Aldrin~Michael Nilles, Justine~Aletta Aguilar, Rowel Atienza, and Rhandley Cajote.
\newblock Next-best view policy for 3d reconstruction.
\newblock In {\em Computer Vision – ECCV 2020 Workshops: Glasgow, UK, August 23–28, 2020, Proceedings, Part IV}, page 558–573, Berlin, Heidelberg, 2020. Springer-Verlag.

\bibitem{zeng2020pc}
Rui Zeng, Wang Zhao, and Yong-Jin Liu.
\newblock Pc-nbv: A point cloud based deep network for efficient next best view planning.
\newblock In {\em 2020 IEEE/RSJ International Conference on Intelligent Robots and Systems (IROS)}, pages 7050--7057, 2020.

\bibitem{burgard2005coordinated}
Wolfram Burgard, Mark Moors, Cyrill Stachniss, and Frank~E Schneider.
\newblock Coordinated multi-robot exploration.
\newblock {\em IEEE Transactions on robotics}, 21(3):376--386, 2005.

\bibitem{amanatiadis2013multi}
Angelos~A Amanatiadis, Savvas~A Chatzichristofis, Konstantinos Charalampous, Lefteris Doitsidis, Elias~B Kosmatopoulos, Phillipos Tsalides, Antonios Gasteratos, and Stergios~I Roumeliotis.
\newblock A multi-objective exploration strategy for mobile robots under operational constraints.
\newblock {\em IEEE Access}, 1:691--702, 2013.

\bibitem{hardouin2020next}
Guillaume Hardouin, Julien Moras, Fabio Morbidi, Julien Marzat, and El~Mustapha Mouaddib.
\newblock Next-best-view planning for surface reconstruction of large-scale 3d environments with multiple uavs.
\newblock In {\em 2020 IEEE/RSJ International Conference on Intelligent Robots and Systems (IROS)}, pages 1567--1574. IEEE, 2020.

\bibitem{almadhoun2021multi}
Randa Almadhoun, Tarek Taha, Lakmal Seneviratne, and Yahya Zweiri.
\newblock Multi-robot hybrid coverage path planning for 3d reconstruction of large structures.
\newblock {\em IEEE Access}, 10:2037--2050, 2021.

\bibitem{wu2019plant}
Chenming Wu, Rui Zeng, Jia Pan, Charlie~CL Wang, and Yong-Jin Liu.
\newblock Plant phenotyping by deep-learning-based planner for multi-robots.
\newblock {\em IEEE Robotics and Automation Letters}, 4(4):3113--3120, 2019.

\bibitem{zhu20153d}
Cheng Zhu, Rong Ding, Mengxiang Lin, and Yuanyuan Wu.
\newblock A 3d frontier-based exploration tool for mavs.
\newblock In {\em 2015 IEEE 27th International Conference on Tools with Artificial Intelligence (ICTAI)}, pages 348--352. IEEE, 2015.

\bibitem{da2020novel}
Daniel~Louback da~Silva~Lubanco, Markus Pichler-Scheder, and Thomas Schlechter.
\newblock A novel frontier-based exploration algorithm for mobile robots.
\newblock In {\em 2020 6th International Conference on Mechatronics and Robotics Engineering (ICMRE)}, pages 1--5. IEEE, 2020.

\bibitem{niroui2017robot}
Farzad Niroui, Ben Sprenger, and Goldie Nejat.
\newblock Robot exploration in unknown cluttered environments when dealing with uncertainty.
\newblock In {\em 2017 IEEE International Symposium on Robotics and Intelligent Sensors (IRIS)}, pages 224--229. IEEE, 2017.

\bibitem{dai2020fast}
Anna Dai, Sotiris Papatheodorou, Nils Funk, Dimos Tzoumanikas, and Stefan Leutenegger.
\newblock Fast frontier-based information-driven autonomous exploration with an mav.
\newblock {\em arXiv preprint arXiv:2002.04440}, 2020.

\bibitem{shen2012autonomous}
Shaojie Shen, Nathan Michael, and Vijay Kumar.
\newblock Autonomous indoor 3d exploration with a micro-aerial vehicle.
\newblock In {\em 2012 IEEE international conference on robotics and automation}, pages 9--15. IEEE, 2012.

\bibitem{corah2019communication}
Micah Corah, Cormac O’Meadhra, Kshitij Goel, and Nathan Michael.
\newblock Communication-efficient planning and mapping for multi-robot exploration in large environments.
\newblock {\em IEEE Robotics and Automation Letters}, 4(2):1715--1721, 2019.

\bibitem{premkumar2020combining}
Aravind~Preshant Premkumar, Kevin Yu, and Pratap Tokekar.
\newblock Combining geometric and information-theoretic approaches for multi-robot exploration.
\newblock {\em arXiv preprint arXiv:2004.06856}, 2020.

\bibitem{pito1999solution}
Richard Pito.
\newblock A solution to the next best view problem for automated surface acquisition.
\newblock {\em IEEE Transactions on pattern analysis and machine intelligence}, 21(10):1016--1030, 1999.

\bibitem{peng2019adaptive}
Cheng Peng and Volkan Isler.
\newblock Adaptive view planning for aerial 3d reconstruction.
\newblock In {\em 2019 International Conference on Robotics and Automation (ICRA)}, pages 2981--2987. IEEE, 2019.

\bibitem{roberts2017submodular}
Mike Roberts, Debadeepta Dey, Anh Truong, Sudipta Sinha, Shital Shah, Ashish Kapoor, Pat Hanrahan, and Neel Joshi.
\newblock Submodular trajectory optimization for aerial 3d scanning.
\newblock In {\em Proceedings of the IEEE International Conference on Computer Vision}, pages 5324--5333, 2017.

\bibitem{bircher2018receding}
Andreas Bircher, Mina Kamel, Kostas Alexis, Helen Oleynikova, and Roland Siegwart.
\newblock Receding horizon path planning for 3d exploration and surface inspection.
\newblock {\em Autonomous Robots}, 42(2):291--306, 2018.

\bibitem{LaValle1998RapidlyExploringRT}
Steven~M. LaValle.
\newblock Rapidly-exploring random trees : a new tool for path planning.
\newblock {\em The annual research report}, 1998.

\bibitem{song2020online}
Soohwan Song, Daekyum Kim, and Sungho Jo.
\newblock Online coverage and inspection planning for 3d modeling.
\newblock {\em Autonomous Robots}, pages 1--20, 2020.

\bibitem{anthony_elbaum_lorenz_detweiler_2014}
David Anthony, Sebastian Elbaum, Aaron Lorenz, and Carrick Detweiler.
\newblock On crop height estimation with uavs.
\newblock {\em 2014 IEEE/RSJ International Conference on Intelligent Robots and Systems}, 2014.

\bibitem{madec}
Madec, Simon, Baret, Fred, de~Solan, Thomas, Samuel, Dan, Jezequel, Hemmerlé, and et~al.
\newblock High-throughput phenotyping of plant height: Comparing unmanned aerial vehicles and ground lidar estimates, Nov 2017.

\bibitem{yuan_li_bhatta_shi_baenziger_ge_2018}
Wenan Yuan, Jiating Li, Madhav Bhatta, Yeyin Shi, P.~Baenziger, and Yufeng Ge.
\newblock Wheat height estimation using lidar in comparison to ultrasonic sensor and uas.
\newblock {\em Sensors}, 18(11):3731, 2018.

\bibitem{ziliani_parkes_hoteit_mccabe_2018}
Matteo Ziliani, Stephen Parkes, Ibrahim Hoteit, and Matthew Mccabe.
\newblock Intra-season crop height variability at commercial farm scales using a fixed-wing uav.
\newblock {\em Remote Sensing}, 10(12):2007, 2018.

\bibitem{Ollero2006}
Anibal Ollero, J.~Ramiro Martinez-de Dios, and Luis Merino.
\newblock Unmanned aerial vehicles as tools for forest-fire fighting.
\newblock {\em Forest Ecology and Management - FOREST ECOL MANAGE}, 234, 11 2006.

\bibitem{Merino2005}
L.~Merino, F.~Caballero, J.R. Martinez-de Dios, and A.~Ollero.
\newblock Cooperative fire detection using unmanned aerial vehicles.
\newblock In {\em Proceedings of the 2005 IEEE International Conference on Robotics and Automation}, pages 1884--1889, 2005.

\bibitem{Merino2012}
Luis Merino, Fernando Caballero, J.~Ramiro Martinez-de Dios, Ivan Maza, and Anibal Ollero.
\newblock An unmanned aircraft system for automatic forest fire monitoring and measurement.
\newblock {\em Journal of Intelligent and Robotic Systems}, 65:533--548, 01 2012.

\bibitem{Krizhevsky2012}
Alex Krizhevsky, Ilya Sutskever, and Geoffrey~E Hinton.
\newblock Imagenet classification with deep convolutional neural networks.
\newblock In F.~Pereira, C.J. Burges, L.~Bottou, and K.Q. Weinberger, editors, {\em Advances in Neural Information Processing Systems}, volume~25. Curran Associates, Inc., 2012.

\bibitem{Szegedy2014}
Christian Szegedy, Wei Liu, Yangqing Jia, Pierre Sermanet, Scott~E. Reed, Dragomir Anguelov, Dumitru Erhan, Vincent Vanhoucke, and Andrew Rabinovich.
\newblock Going deeper with convolutions.
\newblock {\em CoRR}, abs/1409.4842, 2014.

\bibitem{simonyan2015deep}
Karen Simonyan and Andrew Zisserman.
\newblock Very deep convolutional networks for large-scale image recognition, 2015.

\bibitem{Lee2017}
Wonjae Lee, Seonghyun Kim, Yong-Tae Lee, Hyun-Woo Lee, and Min Choi.
\newblock Deep neural networks for wild fire detection with unmanned aerial vehicle.
\newblock In {\em 2017 IEEE International Conference on Consumer Electronics (ICCE)}, pages 252--253, 2017.

\bibitem{dhami2023prednbv}
Harnaik Dhami, Vishnu~D. Sharma, and Pratap Tokekar.
\newblock Pred-nbv: Prediction-guided next-best-view for 3d object reconstruction, 2023.

\bibitem{airsim2017fsr}
Shital Shah, Debadeepta Dey, Chris Lovett, and Ashish Kapoor.
\newblock Airsim: High-fidelity visual and physical simulation for autonomous vehicles.
\newblock In Marco Hutter and Roland Siegwart, editors, {\em Field and Service Robotics}, pages 621--635, Cham, 2018. Springer International Publishing.

\bibitem{dhami2023mapnbv}
Harnaik Dhami, Vishnu~D. Sharma, and Pratap Tokekar.
\newblock Map-nbv: Multi-agent prediction-guided next-best-view planning for active 3d object reconstruction, 2023.

\bibitem{dhamiGATSBI}
Harnaik Dhami, Kevin Yu, Troi Williams, Vineeth Vajipey, and Pratap Tokekar.
\newblock {GATSBI}: An online {GTSP}-based algorithm for targeted surface bridge inspection.
\newblock In {\em 2023 International Conference on Unmanned Aircraft Systems ({ICUAS})}. {IEEE}, jun 2023.

\bibitem{henry1969record}
AL~Henry-Labordere.
\newblock The record balancing problem: A dynamic programming solution of a generalized traveling salesman problem rairo, vol.
\newblock {\em The record balancing problem: A dynamic programming solution of a generalized traveling salesman problem RAIRO vol}, 1969.

\bibitem{saskena1970mathematical}
JP~Saskena.
\newblock Mathematical model of scheduling clients through welfare agencies.
\newblock {\em Journal of the Canadian Operational Research Society}, 8:185--200, 1970.

\bibitem{srivastava1969generalized}
SS~Srivastava, Santosh Kumar, RC~Garg, and Prasenjit Sen.
\newblock Generalized traveling salesman problem through n sets of nodes.
\newblock {\em CORS journal}, 7(2):97, 1969.

\bibitem{Smith2017GLNS}
S.~L. Smith and F.~Imeson.
\newblock {GLNS}: An effective large neighborhood search heuristic for the generalized traveling salesman problem.
\newblock {\em Computers \& Operations Research}, 87:1--19, 2017.

\bibitem{dhami2024gatsbijournal}
Harnaik Dhami, Charith Reddy, Vishnu~D. Sharma, Troi Williams, and Pratap Tokekar.
\newblock Gatsbi: An online gtsp-based algorithm for targeted surface bridge inspection and defect detection, 2024.

\bibitem{dhami2020crop}
Harnaik Dhami, Kevin Yu, Tianshu Xu, Qian Zhu, Kshitiz Dhakal, James Friel, Song Li, and Pratap Tokekar.
\newblock Crop height and plot estimation for phenotyping from unmanned aerial vehicles using 3d lidar.
\newblock In {\em Proceedings of the IEEE/RSJ International Conference on Intelligent Robots and Systems (IROS)}, 2020.

\bibitem{Koenig2004Gazebo}
N.~Koenig and A.~Howard.
\newblock Design and use paradigms for gazebo, an open-source multi-robot simulator.
\newblock In {\em 2004 IEEE/RSJ International Conference on Intelligent Robots and Systems (IROS) (IEEE Cat. No.04CH37566)}, volume~3, pages 2149--2154 vol.3, 2004.

\bibitem{Liu2022}
Jun Liu, Murtaza Rangwala, Kulbir~Singh Ahluwalia, Shayan Ghajar, Harnaik Dhami, Pratap Tokekar, Benjamin Tracy, and Ryan~K. Williams.
\newblock Intermittent deployment for large-scale multi-robot forage perception: Data synthesis, prediction, and planning.
\newblock {\em IEEE Transactions on Automation Science and Engineering}, pages 1--21, 2022.

\bibitem{Rangwala2021}
Murtaza Rangwala, Jun Liu, Kulbir~Singh Ahluwalia, Shayan Ghajar, Harnaik~Singh Dhami, Benjamin~F. Tracy, Pratap Tokekar, and Ryan~K. Williams.
\newblock Deeppastl: Spatio-temporal deep learning methods for predicting long-term pasture terrains using synthetic datasets.
\newblock {\em Agronomy}, 11(11), 2021.

\bibitem{N5Sensors_2023}
N5~Sensors, Jun 2023.

\bibitem{bajcsy2018revisiting}
Ruzena Bajcsy, Yiannis Aloimonos, and John~K Tsotsos.
\newblock Revisiting active perception.
\newblock {\em Autonomous Robots}, 42:177--196, 2018.

\bibitem{shanthakumar2018view}
Prajwal Shanthakumar, Kevin Yu, Mandeep Singh, Jonah Orevillo, Eric Bianchi, Matthew Hebdon, and Pratap Tokekar.
\newblock View planning and navigation algorithms for autonomous bridge inspection with uavs.
\newblock In {\em International Symposium on Experimental Robotics (ISER)}, 2018.
\newblock Accepted.

\bibitem{kim_2009_IROS}
Ayoung Kim and Ryan Eustice.
\newblock Pose-graph visual slam with geometric model selection for autonomous underwater ship hull inspection.
\newblock In {\em 2009 IEEE/RSJ International Conference on Intelligent Robots and Systems (IROS)}, pages 1559--1565, 2009.

\bibitem{ropek_2021}
Lucas Ropek.
\newblock This startup is using drones to conduct aircraft inspections, Mar 2021.

\bibitem{46965}
Tinghui Zhou, Richard Tucker, John Flynn, Graham Fyffe, and Noah Snavely.
\newblock Stereo magnification: Learning view synthesis using multiplane images.
\newblock {\em ACM Trans. Graph. (Proc. SIGGRAPH)}, 37, 2018.

\bibitem{ramakrishnan2021hm3d}
Santhosh~Kumar Ramakrishnan, Aaron Gokaslan, Erik Wijmans, Oleksandr Maksymets, Alexander Clegg, John~M Turner, Eric Undersander, Wojciech Galuba, Andrew Westbury, Angel~X Chang, Manolis Savva, Yili Zhao, and Dhruv Batra.
\newblock Habitat-matterport 3d dataset ({HM}3d): 1000 large-scale 3d environments for embodied {AI}.
\newblock In {\em Thirty-fifth Conference on Neural Information Processing Systems Datasets and Benchmarks Track (Round 2)}, 2021.

\bibitem{debevec1996modeling}
Paul~E Debevec, Camillo~J Taylor, and Jitendra Malik.
\newblock Modeling and rendering architecture from photographs: A hybrid geometry-and image-based approach.
\newblock In {\em Proceedings of the 23rd annual conference on Computer graphics and interactive techniques}, pages 11--20, 1996.

\bibitem{breyer2022closed}
Michel Breyer, Lionel Ott, Roland Siegwart, and Jen~Jen Chung.
\newblock Closed-loop next-best-view planning for target-driven grasping.
\newblock In {\em 2022 IEEE/RSJ International Conference on Intelligent Robots and Systems (IROS)}, pages 1411--1416. IEEE, 2022.

\bibitem{ramakrishnan2020occupancy}
Santhosh~K Ramakrishnan, Ziad Al-Halah, and Kristen Grauman.
\newblock Occupancy anticipation for efficient exploration and navigation.
\newblock In {\em Computer Vision--ECCV 2020: 16th European Conference, Glasgow, UK, August 23--28, 2020, Proceedings, Part V 16}, pages 400--418. Springer, 2020.

\bibitem{Wei2021}
Minghan Wei, Daewon Lee, Volkan Isler, and Daniel Lee.
\newblock Occupancy map inpainting for online robot navigation.
\newblock In {\em 2021 IEEE International Conference on Robotics and Automation (ICRA)}, pages 8551--8557, 2021.

\bibitem{yan2019data}
Xinchen Yan, Mohi Khansari, Jasmine Hsu, Yuanzheng Gong, Yunfei Bai, S{\"o}ren Pirk, and Honglak Lee.
\newblock Data-efficient learning for sim-to-real robotic grasping using deep point cloud prediction networks.
\newblock {\em arXiv preprint arXiv:1906.08989}, 2019.

\bibitem{3drecgan}
Bo~Yang, Stefano Rosa, Andrew Markham, Niki Trigoni, and Hongkai Wen.
\newblock Dense 3d object reconstruction from a single depth view.
\newblock {\em IEEE Transactions on Pattern Analysis and Machine Intelligence}, 41(12):2820--2834, 2019.

\bibitem{wu20153d}
Zhirong Wu, Shuran Song, Aditya Khosla, Fisher Yu, Linguang Zhang, Xiaoou Tang, and Jianxiong Xiao.
\newblock 3d shapenets: A deep representation for volumetric shapes.
\newblock In {\em Proceedings of the IEEE conference on computer vision and pattern recognition}, pages 1912--1920, 2015.

\bibitem{bengio2009curriculum}
Yoshua Bengio, J{\'e}r{\^o}me Louradour, Ronan Collobert, and Jason Weston.
\newblock Curriculum learning.
\newblock In {\em Proceedings of the 26th annual international conference on machine learning}, pages 41--48, 2009.

\bibitem{vasquez2014volumetric}
J~Irving Vasquez-Gomez, L~Enrique Sucar, Rafael Murrieta-Cid, and Efrain Lopez-Damian.
\newblock Volumetric next-best-view planning for 3d object reconstruction with positioning error.
\newblock {\em International Journal of Advanced Robotic Systems}, 11(10):159, 2014.

\bibitem{gonzalez2002navigation}
H{\'e}ctor~H Gonz{\'a}lez-Banos and Jean-Claude Latombe.
\newblock Navigation strategies for exploring indoor environments.
\newblock {\em The International Journal of Robotics Research}, 21(10-11):829--848, 2002.

\bibitem{adler2014autonomous}
Benjamin Adler, Junhao Xiao, and Jianwei Zhang.
\newblock Autonomous exploration of urban environments using unmanned aerial vehicles.
\newblock {\em Journal of Field Robotics}, 31(6):912--939, 2014.

\bibitem{morooka1998next}
Ken'ichi Morooka, Hongbin Zha, and Tsutomu Hasegawa.
\newblock Next best viewpoint (nbv) planning for active object modeling based on a learning-by-showing approach.
\newblock In {\em Proceedings. Fourteenth International Conference on Pattern Recognition (Cat. No. 98EX170)}, volume~1, pages 677--681. IEEE, 1998.

\bibitem{vasquez2009view}
Juan~Irving V{\'a}squez-G{\'o}mez, Efra{\'\i}n L{\"o}pez-Damian, and Luis~Enrique Sucar.
\newblock View planning for 3d object reconstruction.
\newblock In {\em 2009 IEEE/RSJ International Conference on Intelligent Robots and Systems (IROS)}, pages 4015--4020. IEEE, 2009.

\bibitem{banta2000next}
Joseph~E Banta, LR~Wong, Christophe Dumont, and Mongi~A Abidi.
\newblock A next-best-view system for autonomous 3-d object reconstruction.
\newblock {\em IEEE Transactions on Systems, Man, and Cybernetics-Part A: Systems and Humans}, 30(5):589--598, 2000.

\bibitem{kriegel2013combining}
Simon Kriegel, Manuel Brucker, Zoltan-Csaba Marton, Tim Bodenmüller, and Michael Suppa.
\newblock Combining object modeling and recognition for active scene exploration.
\newblock In {\em 2013 IEEE/RSJ International Conference on Intelligent Robots and Systems (IROS)}, pages 2384--2391, 2013.

\bibitem{sharma2023proxmap}
Vishnu~Dutt Sharma, Jingxi Chen, and Pratap Tokekar.
\newblock Proxmap: Proximal occupancy map prediction for efficient indoor robot navigation.
\newblock In {\em 2023 IEEE/RSJ International Conference on Intelligent Robots and Systems (IROS)}. IEEE, 2023, (in press).

\bibitem{Katyal2021}
Kapil~D. Katyal, Adam Polevoy, Joseph Moore, Craig Knuth, and Katie~M. Popek.
\newblock High-speed robot navigation using predicted occupancy maps.
\newblock In {\em 2021 IEEE International Conference on Robotics and Automation (ICRA)}, pages 5476--5482, 2021.

\bibitem{yang2022real}
Bowen Yang, Qingwen Zhang, Ruoyu Geng, Lujia Wang, and Ming Liu.
\newblock Real-time neural dense elevation mapping for urban terrain with uncertainty estimations.
\newblock {\em IEEE Robotics and Automation Letters}, 8(2):696--703, 2022.

\bibitem{Hani2020}
Nicolai H\"{a}ni, Selim Engin, Jun-Jee Chao, and Volkan Isler.
\newblock Continuous object representation networks: Novel view synthesis without target view supervision.
\newblock In {\em Proceedings of the 34th International Conference on Neural Information Processing Systems}, NIPS'20, Red Hook, NY, USA, 2020. Curran Associates Inc.

\bibitem{Alidoost2019}
Fatemeh Alidoost, Hossein Arefi, and Federico Tombari.
\newblock 2d image-to-3d model: Knowledge-based 3d building reconstruction (3dbr) using single aerial images and convolutional neural networks (cnns).
\newblock {\em Remote Sensing}, 11(19), 2019.

\bibitem{xie2020grnet}
Haozhe Xie, Hongxun Yao, Shangchen Zhou, Jiageng Mao, Shengping Zhang, and Wenxiu Sun.
\newblock Grnet: Gridding residual network for dense point cloud completion.
\newblock In {\em Computer Vision--ECCV 2020: 16th European Conference, Glasgow, UK, August 23--28, 2020, Proceedings, Part IX}, pages 365--381. Springer, 2020.

\bibitem{POP2022160}
Alexandru Pop and Levente Tamas.
\newblock Next best view estimation for volumetric information gain.
\newblock {\em IFAC-PapersOnLine}, 55(15):160--165, 2022.
\newblock 6th IFAC Conference on Intelligent Control and Automation SciencesICONS 2022.

\bibitem{vaswani2017attention}
Ashish Vaswani, Noam Shazeer, Niki Parmar, Jakob Uszkoreit, Llion Jones, Aidan~N Gomez, {\L}ukasz Kaiser, and Illia Polosukhin.
\newblock Attention is all you need.
\newblock {\em Advances in neural information processing systems}, 30, 2017.

\bibitem{katz2007direct}
Sagi Katz, Ayellet Tal, and Ronen Basri.
\newblock Direct visibility of point sets.
\newblock {\em ACM Trans. Graph.}, 26(3):24–es, jul 2007.

\bibitem{kuffner2000rrt}
James~J Kuffner and Steven~M LaValle.
\newblock Rrt-connect: An efficient approach to single-query path planning.
\newblock In {\em Proceedings 2000 ICRA. Millennium Conference. IEEE International Conference on Robotics and Automation. Symposia Proceedings (Cat. No. 00CH37065)}, volume~2, pages 995--1001. IEEE, 2000.

\bibitem{fan2017point}
Haoqiang Fan, Hao Su, and Leonidas~J Guibas.
\newblock A point set generation network for 3d object reconstruction from a single image.
\newblock In {\em Proceedings of the IEEE conference on computer vision and pattern recognition}, pages 605--613, 2017.

\bibitem{coleman2014reducing}
David Coleman, Ioan Sucan, Sachin Chitta, and Nikolaus Correll.
\newblock Reducing the barrier to entry of complex robotic software: a moveit! case study.
\newblock {\em arXiv preprint arXiv:1404.3785}, 2014.

\bibitem{tahsinko86:online}
Tahsincan Köse.
\newblock tahsinkose/hector-moveit: Hector quadrotor with moveit! motion planning framework.
\newblock \url{https://github.com/tahsinkose/hector-moveit}, 2018.

\bibitem{ozaslaninspection}
Tolga Ozaslan, Shaojie Shen, Yash Mulgaonkar, Nathan Michael, and Vijay Kumar.
\newblock Inspection of penstocks and featureless tunnel-like environments using micro uavs.
\newblock In {\em International Conference on Field and Service Robotics}, 2013.

\bibitem{dunbabin2012environmental}
M.~Dunbabin and L.~Marques.
\newblock Robots for environmental monitoring: Significant advancements and applications.
\newblock {\em IEEE Robotics and Automation Magazine}, 19(1):24 --39, Mar 2012.

\bibitem{sung2019competitive}
Yoonchang Sung and Pratap Tokekar.
\newblock A competitive algorithm for online multi-robot exploration of a translating plume.
\newblock In {\em Proceedings of the IEEE International Conference on Robotics and Automation (ICRA)}. {IEEE}, may 2019.

\bibitem{ammirato2017dataset}
Phil Ammirato, Patrick Poirson, Eunbyung Park, Jana Ko{\v{s}}eck{\'a}, and Alexander~C Berg.
\newblock A dataset for developing and benchmarking active vision.
\newblock In {\em 2017 IEEE International Conference on Robotics and Automation (ICRA)}, pages 1378--1385. IEEE, 2017.

\bibitem{connolly1985determination}
Cl~Connolly.
\newblock The determination of next best views.
\newblock In {\em Proceedings. 1985 IEEE international conference on robotics and automation}, volume~2, pages 432--435. IEEE, 1985.

\bibitem{johns2016pairwise}
Edward Johns, Stefan Leutenegger, and Andrew~J Davison.
\newblock Pairwise decomposition of image sequences for active multi-view recognition.
\newblock In {\em Proceedings of the IEEE Conference on Computer Vision and Pattern Recognition}, pages 3813--3822, 2016.

\bibitem{mendoza2020supervised}
Miguel Mendoza, J~Irving Vasquez-Gomez, Hind Taud, L~Enrique Sucar, and Carolina Reta.
\newblock Supervised learning of the next-best-view for 3d object reconstruction.
\newblock {\em Pattern Recognition Letters}, 133:224--231, 2020.

\bibitem{huang2021comprehensive}
Xiaoshui Huang, Guofeng Mei, Jian Zhang, and Rana Abbas.
\newblock A comprehensive survey on point cloud registration, 2021.

\bibitem{kirillov2023segment}
Alexander Kirillov, Eric Mintun, Nikhila Ravi, Hanzi Mao, Chloe Rolland, Laura Gustafson, Tete Xiao, Spencer Whitehead, Alexander~C Berg, Wan-Yen Lo, et~al.
\newblock Segment anything.
\newblock {\em arXiv preprint arXiv:2304.02643}, 2023.

\bibitem{calinescu2011maximizing}
Gruia Calinescu, Chandra Chekuri, Martin Pal, and Jan Vondr{\'a}k.
\newblock Maximizing a monotone submodular function subject to a matroid constraint.
\newblock {\em SIAM Journal on Computing}, 40(6):1740--1766, 2011.

\bibitem{Zhou2018}
Qian-Yi Zhou, Jaesik Park, and Vladlen Koltun.
\newblock {Open3D}: {A} modern library for {3D} data processing.
\newblock {\em arXiv:1801.09847}, 2018.

\bibitem{skydio}
Skydio 2+™ and x2™ -- skydio inc.
\newblock \url{https://www.skydio.com/}.
\newblock Accessed: 2023-2-1.

\bibitem{Exyn_Technologies_undated-gq}
{Exyn Technologies}.
\newblock Industrial drone technology.
\newblock \url{https://www.exyn.com/}.
\newblock Accessed: 2023-2-1.

\bibitem{9048979}
B.~{Kakillioglu}, J.~{Wang}, S.~{Velipasalar}, A.~{Janani}, and E.~{Koch}.
\newblock 3d sensor-based uav localization for bridge inspection.
\newblock In {\em 2019 53rd Asilomar Conference on Signals, Systems, and Computers}, pages 1926--1930, 2019.

\bibitem{liosam2020shan}
Tixiao Shan, Brendan Englot, Drew Meyers, Wei Wang, Carlo Ratti, and Rus Daniela.
\newblock Lio-sam: Tightly-coupled lidar inertial odometry via smoothing and mapping.
\newblock In {\em IEEE/RSJ International Conference on Intelligent Robots and Systems (IROS)}, pages 5135--5142. IEEE, 2020.

\bibitem{hornung2013octomap}
Armin Hornung, Kai~M. Wurm, Maren Bennewitz, Cyrill Stachniss, and Wolfram Burgard.
\newblock {OctoMap}: An efficient probabilistic {3D} mapping framework based on octrees.
\newblock {\em Autonomous Robots}, 2013.
\newblock Software available at \url{http://octomap.github.com}.

\bibitem{Smith2016GLNS}
S.~L. Smith and F.~Imeson.
\newblock {{GLNS}: An Effective Large Neighborhood Search Heuristic for the Generalized Traveling Salesman Problem}.
\newblock {\em Computers \& Operations Research}, 87:1--19, 2017.

\bibitem{Cheng_2024_CVPR}
Tianheng Cheng, Lin Song, Yixiao Ge, Wenyu Liu, Xinggang Wang, and Ying Shan.
\newblock Yolo-world: Real-time open-vocabulary object detection.
\newblock In {\em Proceedings of the IEEE/CVF Conference on Computer Vision and Pattern Recognition (CVPR)}, pages 16901--16911, June 2024.

\bibitem{li2020exploration}
Alberto~Quattrini Li.
\newblock Exploration and mapping with groups of robots: Recent trends.
\newblock {\em Current Robotics Reports}, pages 1--11, 2020.

\bibitem{tokekar2016algorithms}
Pratap Tokekar, Ashish~Kumar Budhiraja, and Vijay Kumar.
\newblock Algorithms for visibility-based monitoring with robot teams, 2016.

\bibitem{hollinger2013active}
Geoffrey~A Hollinger, Brendan Englot, Franz~S Hover, Urbashi Mitra, and Gaurav~S Sukhatme.
\newblock Active planning for underwater inspection and the benefit of adaptivity.
\newblock {\em The International Journal of Robotics Research}, 32(1):3--18, 2013.

\bibitem{alfarrarjeh2018deep}
Abdullah Alfarrarjeh, Dweep Trivedi, Seon~Ho Kim, and Cyrus Shahabi.
\newblock A deep learning approach for road damage detection from smartphone images.
\newblock In {\em 2018 IEEE International Conference on Big Data (Big Data)}, pages 5201--5204. IEEE, 2018.

\bibitem{faramarzi2020road}
Masoud Faramarzi.
\newblock Road damage detection and classification using deep neural networks (yolov4) with smartphone images.
\newblock {\em Available at SSRN 3627382}, 2020.

\bibitem{kuang2020computer}
Zijian Kuang, Xinran Tie, Lihang Ying, and Shi Jin.
\newblock Computer vision and normalizing flow-based defect detection.
\newblock {\em arXiv preprint arXiv:2012.06737}, 2020.

\bibitem{cha2017deep}
Young-Jin Cha, Wooram Choi, and Oral B{\"u}y{\"u}k{\"o}zt{\"u}rk.
\newblock Deep learning-based crack damage detection using convolutional neural networks.
\newblock {\em Computer-Aided Civil and Infrastructure Engineering}, 32(5):361--378, 2017.

\bibitem{maeda2018road}
Hiroya Maeda, Yoshihide Sekimoto, Toshikazu Seto, Takehiro Kashiyama, and Hiroshi Omata.
\newblock Road damage detection and classification using deep neural networks with smartphone images.
\newblock {\em Computer-Aided Civil and Infrastructure Engineering}, 33(12):1127--1141, 2018.

\bibitem{shang2022superpixel}
Hongbing Shang, Qixiu Yang, Chuang Sun, Xuefeng Chen, and Ruqiang Yan.
\newblock Superpixel perception graph neural network for intelligent defect detection.
\newblock {\em arXiv preprint arXiv:2210.07539}, 2022.

\bibitem{liu2021crackformer}
Huajun Liu, Xiangyu Miao, Christoph Mertz, Chengzhong Xu, and Hui Kong.
\newblock Crackformer: Transformer network for fine-grained crack detection.
\newblock In {\em Proceedings of the IEEE/CVF International Conference on Computer Vision}, pages 3783--3792, 2021.

\bibitem{zou2012cracktree}
Qin Zou, Yu~Cao, Qingquan Li, Qingzhou Mao, and Song Wang.
\newblock Cracktree: Automatic crack detection from pavement images.
\newblock {\em Pattern Recognition Letters}, 33(3):227--238, 2012.

\bibitem{zou2018deepcrack}
Qin Zou, Zheng Zhang, Qingquan Li, Xianbiao Qi, Qian Wang, and Song Wang.
\newblock Deepcrack: Learning hierarchical convolutional features for crack detection.
\newblock {\em IEEE Transactions on Image Processing}, 28(3):1498--1512, 2018.

\bibitem{konig2021optimized}
Jacob K{\"o}nig, Mark~David Jenkins, Mike Mannion, Peter Barrie, and Gordon Morison.
\newblock Optimized deep encoder-decoder methods for crack segmentation.
\newblock {\em Digital Signal Processing}, 108:102907, 2021.

\bibitem{nath2021s2d2net}
Vikanksh Nath and Chiranjoy Chattopadhyay.
\newblock S2d2net: An improved approach for robust steel surface defects diagnosis with small sample learning.
\newblock In {\em 2021 IEEE International Conference on Image Processing (ICIP)}, pages 1199--1203. IEEE, 2021.

\bibitem{zavrtanik2021draem}
Vitjan Zavrtanik, Matej Kristan, and Danijel Sko{\v{c}}aj.
\newblock Draem-a discriminatively trained reconstruction embedding for surface anomaly detection.
\newblock In {\em Proceedings of the IEEE/CVF International Conference on Computer Vision}, pages 8330--8339, 2021.

\bibitem{bovzivc2021mixed}
Jakob Bo{\v{z}}i{\v{c}}, Domen Tabernik, and Danijel Sko{\v{c}}aj.
\newblock Mixed supervision for surface-defect detection: From weakly to fully supervised learning.
\newblock {\em Computers in Industry}, 129:103459, 2021.

\bibitem{li2021cutpaste}
Chun-Liang Li, Kihyuk Sohn, Jinsung Yoon, and Tomas Pfister.
\newblock Cutpaste: Self-supervised learning for anomaly detection and localization.
\newblock In {\em Proceedings of the IEEE/CVF Conference on Computer Vision and Pattern Recognition}, pages 9664--9674, 2021.

\bibitem{bovzivc2021end}
Jakob Bo{\v{z}}i{\v{c}}, Domen Tabernik, and Danijel Sko{\v{c}}aj.
\newblock End-to-end training of a two-stage neural network for defect detection.
\newblock In {\em 2020 25th International Conference on Pattern Recognition (ICPR)}, pages 5619--5626. IEEE, 2021.

\bibitem{tabernik2020segmentation}
Domen Tabernik, Samo {\v{S}}ela, Jure Skvar{\v{c}}, and Danijel Sko{\v{c}}aj.
\newblock Segmentation-based deep-learning approach for surface-defect detection.
\newblock {\em Journal of Intelligent Manufacturing}, 31(3):759--776, 2020.

\bibitem{dougan2022new}
G{\"u}rkan Do{\u{g}}an and Burhan Ergen.
\newblock A new mobile convolutional neural network-based approach for pixel-wise road surface crack detection.
\newblock {\em Measurement}, 195:111119, 2022.

\bibitem{li2021automatic}
Gang Li, Qiangwei Liu, Wei Ren, Wenting Qiao, Biao Ma, and Jian Wan.
\newblock Automatic recognition and analysis system of asphalt pavement cracks using interleaved low-rank group convolution hybrid deep network and segnet fusing dense condition random field.
\newblock {\em Measurement}, 170:108693, 2021.

\bibitem{yang2019feature}
Fan Yang, Lei Zhang, Sijia Yu, Danil Prokhorov, Xue Mei, and Haibin Ling.
\newblock Feature pyramid and hierarchical boosting network for pavement crack detection.
\newblock {\em IEEE Transactions on Intelligent Transportation Systems}, 21(4):1525--1535, 2019.

\bibitem{inoue2021crack}
Yuki Inoue and Hiroto Nagayoshi.
\newblock Crack detection as a weakly-supervised problem: towards achieving less annotation-intensive crack detectors.
\newblock In {\em 2020 25th International Conference on Pattern Recognition (ICPR)}, pages 65--72. IEEE, 2021.

\bibitem{sun2022new}
Chen Sun, Liang Gao, Xinyu Li, and Yiping Gao.
\newblock A new knowledge distillation network for incremental few-shot surface defect detection.
\newblock {\em arXiv preprint arXiv:2209.00519}, 2022.

\bibitem{joshi2022automatic}
Deepa Joshi, Thipendra~P Singh, and Gargeya Sharma.
\newblock Automatic surface crack detection using segmentation-based deep-learning approach.
\newblock {\em Engineering Fracture Mechanics}, 268:108467, 2022.

\bibitem{rrtstar}
Sertac Karaman and Emilio Frazzoli.
\newblock Sampling-based algorithms for optimal motion planning, 2011.

\bibitem{dorafshan2017fatigue}
Sattar Dorafshan, Robert Thomas, and Marc Maguire.
\newblock Fatigue crack detection using unmanned aerial systems in fracture critical inspection of steel bridges.
\newblock {\em Journal of Bridge Engineering}, 23:04018078, 10 2018.

\bibitem{rogers}
Nicole Rogers.
\newblock What is precision agriculture?

\bibitem{tokekar2016sensor}
Pratap Tokekar, Joshua {Vander Hook}, David Mulla, and Volkan Isler.
\newblock Sensor planning for a symbiotic {UAV} and {UGV} system for precision agriculture.
\newblock {\em IEEE Transactions on Robotics}, 2016.

\bibitem{bommarco_kleijn_potts_2013}
Riccardo Bommarco, David Kleijn, and Simon~G. Potts.
\newblock Ecological intensification: harnessing ecosystem services for food security.
\newblock {\em Trends in Ecology and Evolution}, 28(4):230--–238, 2013.

\bibitem{franzluebbers_paine_winsten_krome_sanderson_ogles_thompson_2012}
A.~J. Franzluebbers, L.~K. Paine, J.~R. Winsten, M.~Krome, M.~A. Sanderson, K.~Ogles, and D.~Thompson.
\newblock Well-managed grazing systems: A forgotten hero of conservation.
\newblock {\em Journal of Soil and Water Conservation}, 67(4), 2012.

\bibitem{godfray_2011}
H.~C.~J. Godfray.
\newblock Food and biodiversity.
\newblock {\em Science}, 333(6047):1231--–1232, 2011.

\bibitem{moles_leishman}
Angela~T. Moles and Michelle~R. Leishman.
\newblock The seedling as part of a plants life history strategy.
\newblock {\em Seedling Ecology and Evolution}, pages 217–--238, 2008.

\bibitem{moles_warton_warman_swenson_laffan_zanne_pitman_hemmings_leishman_2009}
Angela~T. Moles, David~I. Warton, Laura Warman, Nathan~G. Swenson, Shawn~W. Laffan, Amy~E. Zanne, Andy Pitman, Frank~A. Hemmings, and Michelle~R. Leishman.
\newblock Global patterns in plant height.
\newblock {\em Journal of Ecology}, 97(5):923–--932, 2009.

\bibitem{kayacan_young_peschel_chowdhary_2018}
Erkan Kayacan, Sierra~N. Young, Joshua~M. Peschel, and Girish Chowdhary.
\newblock High-precision control of tracked field robots in the presence of unknown traction coefficients.
\newblock {\em Journal of Field Robotics}, 35(7):1050--–1062, 2018.

\bibitem{corn_row_detection:online}
Zhengqi Li, Nikolaos Stefas, Haluk Bayram, and Volkan Isler.
\newblock Mapping corn fields with uavs.

\bibitem{Rusu_ICRA2011_PCL}
Radu~Bogdan Rusu and Steve Cousins.
\newblock {3D is here: Point Cloud Library (PCL)}.
\newblock In {\em {IEEE International Conference on Robotics and Automation (ICRA)}}, Shanghai, China, May 9-13 2011.

\bibitem{Zhang-2014-7903}
Ji~Zhang and Sanjiv Singh.
\newblock Loam: Lidar odometry and mapping in real-time.
\newblock In {\em Proceedings of Robotics: Science and Systems Conference}, July 2014.

\bibitem{nelson_2016}
Erik Nelson.
\newblock erik-nelson/blam, Aug 2016.

\bibitem{2012simpar_meyer}
Johannes Meyer, Alexander Sendobry, Stefan Kohlbrecher, Uwe Klingauf, and Oskar von Stryk.
\newblock Comprehensive simulation of quadrotor uavs using ros and gazebo.
\newblock In {\em 3rd Int. Conf. on Simulation, Modeling and Programming for Autonomous Robots (SIMPAR)}, page to appear, 2012.

\bibitem{gill2013worldwide}
A~Malcolm Gill, Scott~L Stephens, and Geoffrey~J Cary.
\newblock The worldwide “wildfire” problem.
\newblock {\em Ecological applications}, 23(2):438--454, 2013.

\bibitem{wardle2003long}
David~A Wardle, Greger Hornberg, Olle Zackrisson, Maarit Kalela-Brundin, and David~A Coomes.
\newblock Long-term effects of wildfire on ecosystem properties across an island area gradient.
\newblock {\em Science}, 300(5621):972--975, 2003.

\bibitem{canadaECC}
Wildfire operations: Detecting wildfire.
\newblock {\em Government of Canada: Environment and Climate Change}.

\bibitem{ollero2006unmanned}
Anibal Ollero and L~Merino.
\newblock Unmanned aerial vehicles as tools for forest-fire fighting.
\newblock {\em Forest Ecology and Management}, 234(1):S263, 2006.

\bibitem{xu2017real}
Guang Xu and Xu~Zhong.
\newblock Real-time wildfire detection and tracking in australia using geostationary satellite: Himawari-8.
\newblock {\em Remote Sensing Letters}, 8(11):1052--1061, 2017.

\bibitem{allison2016airborne}
Robert~S Allison, Joshua~M Johnston, Gregory Craig, and Sion Jennings.
\newblock Airborne optical and thermal remote sensing for wildfire detection and monitoring.
\newblock {\em Sensors}, 16(8):1310, 2016.

\bibitem{XPRIZE}
Xprize wildfire.
\newblock {\em XPRIZE}, Mar 2024.

\bibitem{binney2010informative}
Jonathan Binney, Andreas Krause, and Gaurav~S Sukhatme.
\newblock Informative path planning for an autonomous underwater vehicle.
\newblock In {\em 2010 IEEE International Conference on Robotics and Automation}, pages 4791--4796. IEEE, 2010.

\bibitem{binney2012branch}
Jonathan Binney and Gaurav~S Sukhatme.
\newblock Branch and bound for informative path planning.
\newblock In {\em 2012 IEEE international conference on robotics and automation}, pages 2147--2154. IEEE, 2012.

\bibitem{Meliou2007}
Alexandra Meliou, Andreas Krause, Carlos Guestrin, and Joseph~M. Hellerstein.
\newblock Nonmyopic informative path planning in spatio-temporal models.
\newblock Technical Report UCB/EECS-2007-44, EECS Department, University of California, Berkeley, Apr 2007.

\bibitem{hitz2017adaptive}
Gregory Hitz, Enric Galceran, Marie-{\`E}ve Garneau, Fran{\c{c}}ois Pomerleau, and Roland Siegwart.
\newblock Adaptive continuous-space informative path planning for online environmental monitoring.
\newblock {\em Journal of Field Robotics}, 34(8):1427--1449, 2017.

\bibitem{sung2023decision}
Yoonchang Sung, Zhiang Chen, Jnaneshwar Das, and Pratap Tokekar.
\newblock A survey of decision-theoretic approaches for robotic environmental monitoring, 2023.

\bibitem{merino2005cooperative}
Luis Merino, Fernando Caballero, JR~Martinez-de Dios, and An{\'\i}bal Ollero.
\newblock Cooperative fire detection using unmanned aerial vehicles.
\newblock In {\em Proceedings of the 2005 IEEE international conference on robotics and automation}, pages 1884--1889. IEEE, 2005.

\bibitem{merino2012unmanned}
Luis Merino, Fernando Caballero, J~Ramiro Mart{\'\i}nez-de Dios, Iv{\'a}n Maza, and An{\'\i}bal Ollero.
\newblock An unmanned aircraft system for automatic forest fire monitoring and measurement.
\newblock {\em Journal of Intelligent \& Robotic Systems}, 65:533--548, 2012.

\bibitem{bouguettaya2022review}
Abdelmalek Bouguettaya, Hafed Zarzour, Amine~Mohammed Taberkit, and Ahmed Kechida.
\newblock A review on early wildfire detection from unmanned aerial vehicles using deep learning-based computer vision algorithms.
\newblock {\em Signal Processing}, 190:108309, 2022.

\bibitem{alexandrov2019analysis}
Dmitriy Alexandrov, Elizaveta Pertseva, Ivan Berman, Igor Pantiukhin, and Aleksandr Kapitonov.
\newblock Analysis of machine learning methods for wildfire security monitoring with an unmanned aerial vehicles.
\newblock In {\em 2019 24th conference of open innovations association (FRUCT)}, pages 3--9. IEEE, 2019.

\bibitem{leblon2012use}
Brigitte Leblon, Laura Bourgeau-Chavez, and Jes{\'u}s San-Miguel-Ayanz.
\newblock Use of remote sensing in wildfire management.
\newblock {\em Sustainable development-authoritative and leading edge content for environmental management}, pages 55--82, 2012.

\bibitem{bailon2020wildfire}
Rafael Bailon-Ruiz and Simon Lacroix.
\newblock Wildfire remote sensing with uavs: A review from the autonomy point of view.
\newblock In {\em 2020 international conference on unmanned aircraft systems (ICUAS)}, pages 412--420. IEEE, 2020.

\bibitem{Kou1981}
L.~Kou, G.~Markowsky, and L.~Berman.
\newblock A fast algorithm for steiner trees.
\newblock {\em Acta Informatica}, 15(2):141--145, Jun 1981.

\bibitem{hartnell1995firefighter}
Bert Hartnell.
\newblock Firefighter! an application of domination.
\newblock In {\em the 24th Manitoba Conference on Combinatorial Mathematics and Computing, University of Minitoba, Winnipeg, Cadada, 1995}, 1995.

\bibitem{Finbow2009}
Stephen Finbow and Gary Macgillivray.
\newblock The firefighter problem: A survey of results, directions and questions.
\newblock {\em The Australasian Journal of Combinatorics [electronic only]}, 43, 02 2009.

\bibitem{lstm}
Sepp Hochreiter and J\"{u}rgen Schmidhuber.
\newblock Long short-term memory.
\newblock {\em Neural Comput.}, 9(8):1735–1780, nov 1997.

\bibitem{Hart1968}
Peter~E. Hart, Nils~J. Nilsson, and Bertram Raphael.
\newblock A formal basis for the heuristic determination of minimum cost paths.
\newblock {\em IEEE Transactions on Systems Science and Cybernetics}, 4(2):100--107, 1968.

\bibitem{Laporte1992}
Gilbert Laporte.
\newblock The traveling salesman problem: An overview of exact and approximate algorithms.
\newblock {\em European Journal of Operational Research}, 59(2):231--247, 1992.

\bibitem{papadopoulos2011comparative}
George~D Papadopoulos and Fotini-Niovi Pavlidou.
\newblock A comparative review on wildfire simulators.
\newblock {\em IEEE systems Journal}, 5(2):233--243, 2011.

\bibitem{filippi2013representation}
Jean-Baptiste Filippi, Vivien Mallet, and Bahaa Nader.
\newblock Representation and evaluation of wildfire propagation simulations.
\newblock {\em International Journal of Wildland Fire}, 23(1):46--57, 2013.

\bibitem{schlipf2012model}
David Schlipf, Andreas Rettenmeier, Florian Haizmann, Martin Hofs{\"a}{\ss}, Mike Courtney, and Po~Wen Cheng.
\newblock Model based wind vector field reconstruction from lidar data.
\newblock 2012.

\bibitem{Cruz2019}
Miguel~G. Cruz and Martin~E. Alexander.
\newblock The 10{\%} wind speed rule of thumb for estimating a wildfire's forward rate of spread in forests and shrublands.
\newblock {\em Annals of Forest Science}, 76(2):44, Apr 2019.

\bibitem{llm2024}
Humza Naveed, Asad~Ullah Khan, Shi Qiu, Muhammad Saqib, Saeed Anwar, Muhammad Usman, Naveed Akhtar, Nick Barnes, and Ajmal Mian.
\newblock A comprehensive overview of large language models, 2024.

\end{thebibliography}

\end{document}